\documentclass[10pt,twocolumn,letterpaper]{article}

\usepackage[pagenumbers]{cvpr} 

\usepackage[dvipsnames]{xcolor}
\usepackage{colortbl}
\usepackage{tabularx}
\usepackage{multicol}
\usepackage{multirow}

\usepackage{graphicx}
\usepackage{adjustbox}
\usepackage{amsmath}
\usepackage{amssymb}
\usepackage{booktabs}
\usepackage{comment}

\definecolor{cvprblue}{rgb}{0.21,0.49,0.74}
\definecolor{forestgreen}{rgb}{0.133,0.545,0.133}

\definecolor{m3net_blue}{RGB}{99, 113, 250}
\definecolor{m3net_red}{RGB}{239, 99, 75}
\definecolor{m3net_green}{RGB}{0, 180, 139}

\definecolor{correct}{RGB}{173, 173, 173}
\definecolor{incorrect}{RGB}{234, 59, 46}

\usepackage[pagebackref=true,breaklinks,colorlinks,citecolor=cvprblue,linkcolor=cvprblue]{hyperref}

\usepackage{indentfirst}
\usepackage{titletoc}

\usepackage[accsupp]{axessibility} 

\def\confName{CVPR}
\def\confYear{2024}

\title{Multi-Space Alignments Towards Universal LiDAR Segmentation}

\author{
Youquan Liu$^{*,1}$ \quad Lingdong Kong$^{*,2,3}$ \quad Xiaoyang Wu$^{4}$ \quad Runnan Chen$^{4}$ \quad Xin Li$^{5}$\\Liang Pan$^{2}$ \quad Ziwei Liu$^{6}$ \quad Yuexin Ma$^{1}$\\
{$^1$ShanghaiTech University, $^2$Shanghai AI Laboratory, $^3$National University of Singapore}\\
{$^4$University of Hong Kong, $^5$East China Normal University, $^6$S-Lab, Nanyang Technological University}\\
\url{https://github.com/youquanl/M3Net}
}

\begin{document}

\maketitle

\begin{abstract}
    A unified and versatile LiDAR segmentation model with strong robustness and generalizability is desirable for safe autonomous driving perception. This work presents \textbf{M3Net}, a one-of-a-kind framework for fulfilling \textbf{m}ulti-task, \textbf{m}ulti-dataset, \textbf{m}ulti-modality LiDAR segmentation in a \textbf{universal} manner using just a \textbf{single} set of parameters. To better exploit data volume and diversity, we first combine large-scale driving datasets acquired by different types of sensors from diverse scenes and then conduct alignments in three spaces, namely data, feature, and label spaces, during the training. As a result, M3Net is capable of taming heterogeneous data for training state-of-the-art LiDAR segmentation models. Extensive experiments on twelve LiDAR segmentation datasets verify our effectiveness. Notably, using a \textbf{shared} set of parameters, M3Net achieves 75.1\%, 83.1\%, and 72.4\% mIoU scores, respectively, on the official benchmarks of SemanticKITTI, nuScenes, and Waymo Open.
\vspace{-0.4cm}
\end{abstract}

\newcommand\blfootnote[1]{%
\begingroup
\renewcommand\thefootnote{}{}\footnote{#1}%
\addtocounter{footnote}{-1}%
\endgroup
}

\blfootnote{$^{*}$~The first two authors contributed equally to this work.}

\section{Introduction}
\label{sec:intro}

Dense and structural 3D surrounding scene understanding provides crucial information for autonomous vehicles to make proper decisions \cite{mao2023survey}. With the recent advancements in sensing technologies, especially the Light Detection and Ranging (LiDAR) sensor, a holistic scene perception can be achieved by segmenting the acquired sensor data \cite{rizzoli2022survey,gao2021survey}.

\begin{figure}[t]
    \begin{center}
    \includegraphics[width=1.0\linewidth]{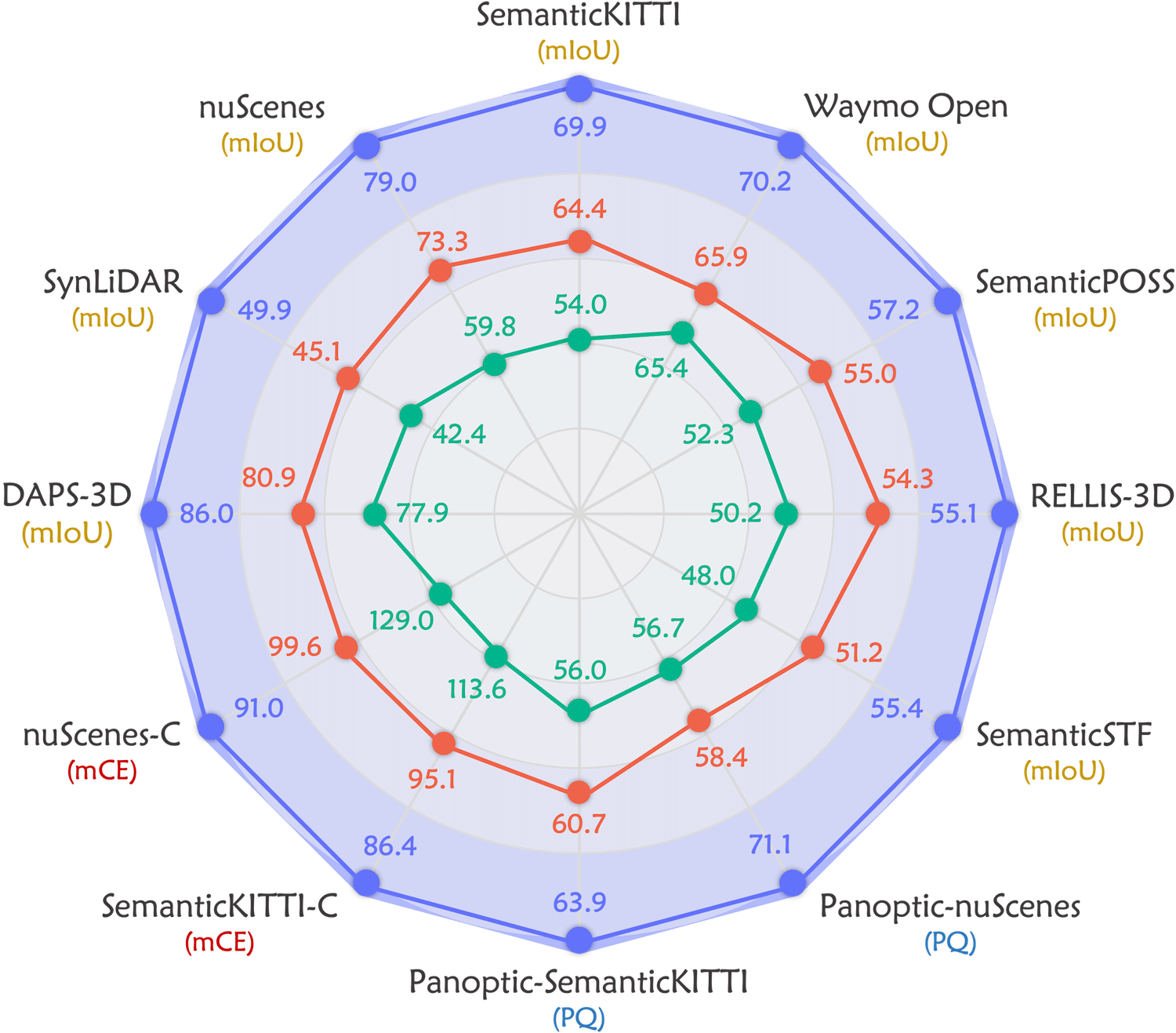}
    \end{center}
    \vspace{-0.4cm}
    \caption{Performance comparisons among \textbf{M3Net} [\textcolor{m3net_blue}{$\bullet$}], \textit{Single-Dataset Training} [\textcolor{m3net_red}{$\bullet$}], and \textit{Na\"{\i}ve Joint Training} [\textcolor{m3net_green}{$\bullet$}] across \textbf{twelve} LiDAR segmentation datasets. For better comparisons, the radius is normalized based on M3Net's scores. The larger the area coverage, the higher the overall performance. Best viewed in colors.}
    \label{fig:teaser}
     \vspace{-1ex}
\end{figure}

Most existing LiDAR segmentation models \cite{hu2021sensatUrban,xu2020squeezesegv3,zhu2021cylindrical,zhou2020polarNet,ando2023rangevit} are trained and tested in a \textit{single-task, single-dataset, single-modality} manner. Despite achieving commendable results in the single domain, there is a significant performance drop when transitioning to new domains \cite{kong2023conDA,jaritz2023xMUDA}. The limited generalization capability hinders their facilitation of real-world applications \cite{seppanen2022snowyKITTI,kong2023robo3D,kong2023robodepth}. In reality, LiDAR datasets are marred by significant variances, encompassing variations in data patterns due to different sensor types and weather conditions, diverse class distributions arising from varying capture scenarios, and distinct label spaces shaped by specific annotation protocols. These factors collectively pose a formidable challenge in harmonizing disparate LiDAR point clouds and jointly optimizing model parameters to effectively address multiple tasks across a range of sensor modalities \cite{zhang2023uni3d,fontez2023mdt3d}. Empirical evidence in Fig. \ref{fig:study} further reveals that na\"{\i}vely combining heterogeneous data to train a LiDAR segmentation model -- without strategic alignments -- often leads to sub-opt results.

Recent works \cite{triess2021survey,xiao2022synLiDAR,boulch2023also,peng2023sam,seppanen2022snowyKITTI,jaritz2020xMUDA,kong2023conDA} resort to unsupervised domain adaptation (UDA) for utilizing training data from both source and target domains to optimize one parameter set. Nevertheless, they either focus on only the sharing mapping between two domains (by ignoring disjoint classes) or directly merge source domain labels to align with the target domain \cite{jaritz2023xMUDA,xu2024visual}. The overlook of the performance degradation on the source dataset and the destruction of original label mappings inevitably constrains such a learning paradigm. Furthermore, there have been efforts~\cite{tsai2023ms3d++,zhang2023uni3d,wu2023ppt,sanchez2022cola} to employ multi-dataset learning strategies to bolster the generalization prowess of 3D perception models. However, they either necessitate dataset-specific fine-tuning, deviating from a truly universal learning approach, or converge label spaces to a coarser set, resulting in the dilution of fine-grained segmentation capabilities across diverse semantic categories.

\begin{figure*}[t]
    \begin{center}
    \includegraphics[width=1.0\linewidth]{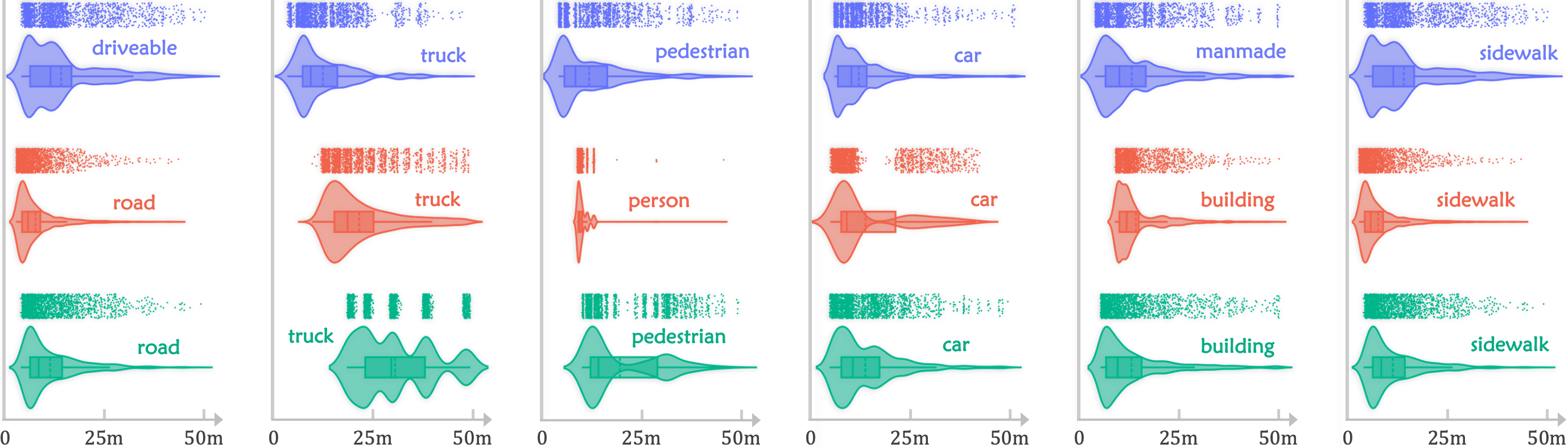}
    \end{center}
    \vspace{-0.4cm}
    \caption{\textbf{Statistical analysis} of six sharing semantic classes in the \textit{nuScenes} [\textcolor{m3net_blue}{$\bullet$}], \textit{SemanticKITTI} [\textcolor{m3net_red}{$\bullet$}], and \textit{Waymo Open} [\textcolor{m3net_green}{$\bullet$}] datasets. Each violin plot shows the class distribution across LiDAR scenes spanning $50$ meters, centered around the ego-vehicle. Best viewed in colors.}
    \label{fig:stats}
     \vspace{-1ex}
\end{figure*}

In this work, we define a novel paradigm towards leveraging LiDAR point clouds from different datasets to tame \textit{a single set of parameters} for multi-task LiDAR segmentation. Sibling to image segmentation communities \cite{lambert2020mseg,zhou2022lmseg,tsai2022learning}, we call this paradigm \textbf{universal LiDAR segmentation}. The ultimate goal of such a synergistic way of learning is to build a powerful segmentation model that can absorb rich cross-domain knowledge and, in return, achieve strong resilience and generalizability for practical usage. Given the substantial differences among datasets in terms of data characteristics, feature distributions, and labeling conventions, we introduce a comprehensive \textit{multi-space alignment} approach that encompasses data-, feature-, and label-level alignments, to effectively pave the path for efficient and universally applicable LiDAR segmentation. In particular, the multi-modal data, including images and texts, is fully exploited to assist the alignment process with the guidance of more general knowledge. Through aforementioned processes, we propose \textbf{M3Net} to learn common knowledge across datasets, modalities, and tasks, thereby significantly enhancing its applicability in practical scenarios.

To substantiate the efficacy of M3Net and the utility of each module developed, 
we have carried out a series of thorough comparative and ablation studies across an extensive array of driving datasets, as shown in \cref{fig:teaser}. Notably, our best model achieves state-of-the-art LiDAR segmentation performance with $75.1\%$, $83.1\%$, $72.4\%$ mIoU scores on \textit{SemanticKITTI} \cite{behley2019semanticKITTI}, \textit{nuScenes} \cite{fong2022panoptic-nuScenes}, \textit{Waymo Open} \cite{sun2020waymoOpen}, respectively, using a \textit{shared} set of parameters. Moreover, our approach also performs well for direct knowledge transfer and out-of-distribution adaptations, further underscoring its robust capability for effective knowledge transfer.




\section{Related Work}
\label{sec:related}

\noindent\textbf{LiDAR Segmentation.}
A holistic perception of 3D scenes is crucial for safe autonomous driving \cite{behley2021semanticKITTI,caesar2020nuScenes,hong20224dDSNet,kong2024calib3d,li2024place3d}. Various LiDAR segmentation models have been proposed, with distinct focuses on aspects include LiDAR representations \cite{milioto2019rangenet++,tang2020searching,zhu2021cylindrical,zhou2020polarNet,choy2019minkowski,thomas2019kpconv,triess2020scan,wu2022ptv2}, model architectures \cite{cheng2022cenet,kong2023rethinking,ando2023rangevit,lai2023sphereformer,puy23waffleiron,xu2021rpvnet,hu2020randla,cortinhal2020salsanext}, sensor fusion \cite{zhuang2021pmf,liong2020amvNet,liu2023uniseg,cheng2021af2S3Net,xu2024visual}, post-processing \cite{xu2020squeezesegv3,zhao2021fidnet}, data augmentations \cite{nekrasov2021mix3d,xiao2022polarmix,saltori2022cosmix}, \etc. Most recently, researchers started to explore data efficiency \cite{kong2022laserMix,li2023lim3d}, annotation efficiency \cite{liu2022less,liu2023segment,sautier2022slidr,unal2022scribbleKITTI,li2022coarse3D}, annotation-free learning \cite{chen2023clip2Scene,zhang2023growsp,chen2023towards}, zero-shot learning \cite{chen2023bridging,lu2023see}, domain adaptation \cite{jaritz2020xMUDA,kong2023conDA,boulch2023also,xiao2022synLiDAR,michele2023saluda,li2023adversarially,peng2023sam}, and robustness \cite{kong2023robo3D} in LiDAR segmentation, shedding lights for practitioners. Existing pursues, however, learn \textit{separate} parameter sets for each dataset, impeding the scalability. This motivates us to explore LiDAR segmentation in a multi-task, multi-dataset, multi-modality manner with just a \textit{single} set of parameters.

\noindent\textbf{Multi-Task Learning.}
A proper pipeline design could enable the model to generate suitable predictions to fulfill multiple tasks simultaneously \cite{he2017mask-rcnn,cheng2022maskformer}. The current research endeavors mainly focus on building image or video segmentation models to handle semantic, instance, and panoptic segmentation tasks \cite{wang2021max-deeplab,zhang2021k-net,li2022video-k-net,zou2023seem,zhang2023openSeeD,jain2023oneformer,wang2023hipie}. Recently, several attempts have been made to enable multi-task segmentation on LiDAR point clouds. MaskRange \cite{gu2022maskrange} and MaskPLS \cite{marcuzzi2023maskpls} extend the mask classification paradigm \cite{cheng2021maskformer} for joint semantic and panoptic LiDAR segmentation. LidarMultiNet \cite{ye2023lidarmultinet} uses global context pooling and task-specific heads to handle LiDAR-based detection and segmentation. P3Former \cite{xiao2023p3former} proposed a specialized positional embedding to handle the geometry ambiguity in panoptic LiDAR segmentation. Our framework also supports multi-task learning. Different from existing approaches, the proposed M3Net stands out by combining knowledge from different sensor data across multiple data sources, which achieves superior performance on each task.

\noindent\textbf{Multi-Dataset Learning.} 
Leveraging data samples from different sources for training has been proven effective in enhancing robustness and generalizability \cite{meletis2018training}. Various approaches have been proposed to merge image datasets for object detection \cite{zhou2022simple,zhou2022detecting,chen2023scaledet,wang2019towards,li2022homogeneous,li2023logonet}, image segmentation \cite{kalluri2019universal,lambert2020mseg,zhou2022lmseg,tsai2022learning,gu2023dataseg}, depth estimation \cite{rene2020towards,chen2020improving}, \etc. Due to large domain gaps, the image-based methods are often hard to be transferred to 3D. To combine multiple LiDAR datasets for 3D object detection, MDT3D \cite{fontez2023mdt3d} defines a coarse label set to handle the label space conflicts in different point cloud datasets. MS3D++ \cite{tsai2023ms3d++,tsai2023ms3d} ensembles pre-trained detectors from different source datasets for multi-domain adaptation. Uni3D \cite{zhang2023uni3d} resorts to dataset-specific detection heads and feature re-coupling for training a unified 3D object detector. Recently, PPT \cite{wu2023ppt} proposed to pre-train a point cloud segmentation network using data from multiple datasets. However, the pre-trained weights are then fine-tuned on each specific dataset, which breaks the universal learning manner. The closest work to us is COLA \cite{sanchez2022cola}, which trains a single model across multiple sources by converting dataset-specific labels to a common coarse set. Such a conversion, however, leads to the loss of fine-grained segmentation across the various semantic categories. Differently, our M3Net is tailored to tame a \textit{single} parameter set to fulfill multi-task prediction across multiple datasets while still maintaining the original label mappings.

\noindent\textbf{Multi-Modality Learning.} Recent trend favors synergistic learning from data of different modalities, such as vision, language, and speech \cite{baevski2022data2vec,radford2021clip,chowdhery2022palm,feichtenhofer2022mae,caron2021dino,oquab2023dinov2,wang2023segGPT}. For LiDAR segmentation, several works \cite{jaritz2020xMUDA,jaritz2023xMUDA,yan20202dpass,liu2023bevfusion,cen2023cmdfusion} explored the distillation of image features to point clouds. Recently, OpenScene \cite{peng2023openscene} and CLIP2Scene \cite{chen2023clip2Scene} proposed to leverage point clouds along with multi-view images and language for open-vocabulary learning. PPKT \cite{liu2021ppkt}, SLidR \cite{sautier2022slidr}, and Seal \cite{liu2023segment} form cross-sensor contrastive learning objectives to pre-train the LiDAR segmentation models. The advantages of sensor fusion have been consistently proven. In this work, to pursue universal LiDAR segmentation, we propose to align multi-space point clouds via images and texts.

\section{Approach}
\label{sec:approach}

Our study serves as an early attempt at combining \textit{multi-task, multi-dataset, multi-modality} knowledge into a \textit{single} set of parameters to fulfill \textbf{universal LiDAR segmentation}. We start with a pilot study to unveil the difficulties in merging heterogeneous LiDAR point clouds (\cf \cref{sec:study}). We then present M3Net, a versatile LiDAR segmentation network tailored to pursue \textit{i)} statistical consistency in the data space (\cf \cref{sec:align_data}), \textit{ii)} cross-modality-assisted alignment in the feature space (\cf \cref{sec:align_feature}), and \textit{iii)} language-guided unification in the label space (\cf \cref{sec:align_label}).

\subsection{Pilot Study}
\label{sec:study}
The current de facto of training a LiDAR segmentation network adopts a \textit{task-by-task} and \textit{dataset-by-dataset} pipeline. Despite the superior performance achieved under such standalone settings, the trained parameter sets cannot be shared to satisfy out-of-domain requirements and, therefore, limits their use cases for practical applications.

\begin{figure}[t]
    \begin{center}
    \includegraphics[width=1.0\linewidth]{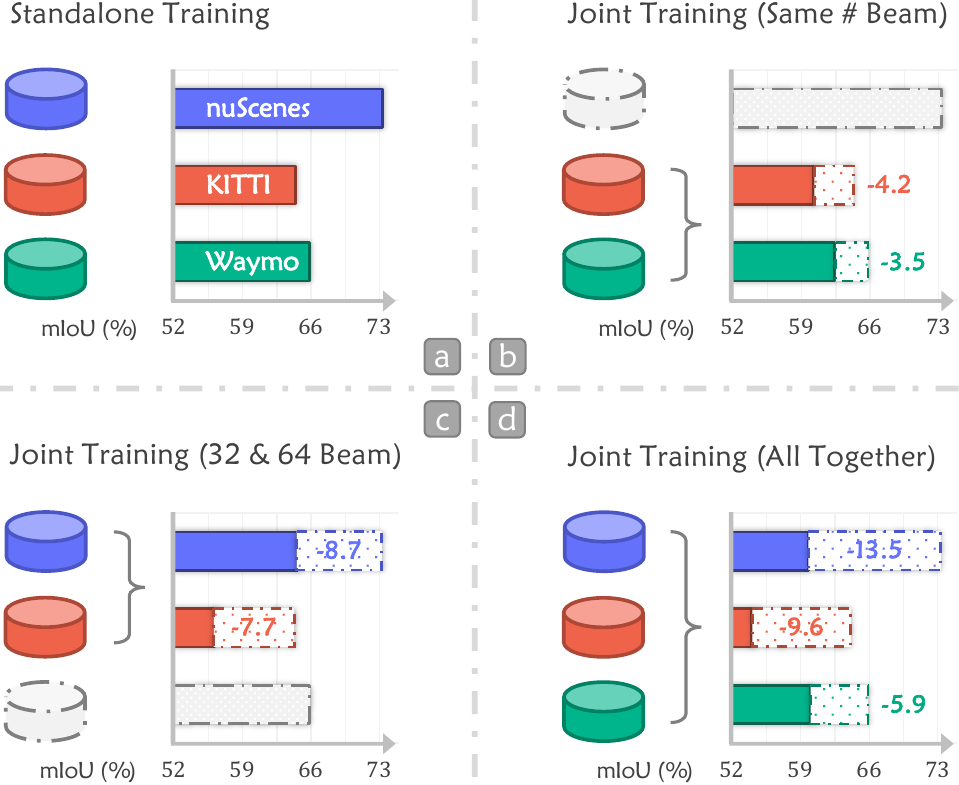}
    \end{center}
    \vspace{-0.4cm}
    \caption{\textbf{A pilot study} of na\"{\i}vely merging different datasets for training the MinkUNet \cite{choy2019minkowski} model. Compared to the standalone training in \textbf{(a)}, either jointly training with \textbf{(b)} the same, \textbf{(c)} different, or \textbf{(d)} all sensor-acquired data will cause severe degradation.}
    \label{fig:study}
     \vspace{-1ex}
\end{figure}

\noindent\textbf{Na\"{\i}ve Joint Training.}
A natural alternative to breaking the above constraint is to jointly train a network across multiple datasets for better generalizability. However, as depicted in \cref{fig:stats}, it is often non-trivial to na\"{\i}vely combine heterogeneous data with large data distribution gaps to train a universal LiDAR segmentation model without proper alignments. To testify this, we conducted a pilot study using the prior art MinkUNet \cite{choy2019minkowski} for both standalone and joint training on three large-scale datasets \cite{fong2022panoptic-nuScenes,behley2019semanticKITTI,sun2020waymoOpen}. As shown in \cref{fig:study}~\textcolor{red}{(a)} and \textcolor{red}{(d)}, a brutal combination undermines the segmentation performance. Due to large discrepancies in aspects like sensor configurations, data acquisitions, label mappings, and domain shifts, the jointly trained representations tend to be disruptive instead of being more general.

\noindent\textbf{LiDAR Sensor Discrepancy.}
To understand the root cause of performance degradation, we conducted another study that controls point cloud density discrepancies when merging datasets. As shown in \cref{fig:study}~\textcolor{red}{(b)} and \textcolor{red}{(c)}, joint training on data collected by sensors with different beam numbers tends to suffer more severely than merging less density variant data. We hypothesize that this is mainly caused by the data statistical variations. In light of these observations, we propose a bag of suitable operations in the following sections to alleviate the large domain gaps among different LiDAR segmentation datasets \cite{behley2019semanticKITTI,behley2021semanticKITTI,caesar2020nuScenes,fong2022panoptic-nuScenes,sun2020waymoOpen}.

\subsection{Data-Space Alignment}
\label{sec:align_data}
Given a total of $S$ datasets $D^s = \{(x^s, y^s) | 1 \leq s \leq S\}$, where $(x^s, y^s)$ denotes the data-label pairs constituting a dataset. For the LiDAR segmentation task, $x^s$ often encompasses the LiDAR point cloud $P^s = \{p_x,p_y,p_z\}^s \in \mathbb{R}^{N \times 3}$ and synchronized multi-view camera images $V^s = \{ \mathcal{I}_1,...,\mathcal{I}_l\}|l=1,...,L\}$, where $\mathcal{I}_t \in \mathbb{R}^{H \times W \times 3}$, $N$ is the number of points, $L$ denotes the number of camera sensors, $H$ and $W$ are the height and width of the image, respectively. $y^s \in \mathbb{R}^{N}$ denotes point cloud labels in the label space $\mathbb{Y}^s$, we unify the label space as $\mathbb{Y}^u = \mathbb{Y}^1 \cup \mathbb{Y}^2 ...  \cup \mathbb{Y}^S$. 

\noindent\textbf{Cross-Modality Data Alignment.}
As a multi-sensing system, the information encoded in $P_i^s$ and $V_i^s$ are intuitively complementary to each other \cite{behley2019semanticKITTI,caesar2020nuScenes,sun2020waymoOpen}. To leverage such an advantage, we resort to the correspondences embedded in camera calibration matrices to bridge the LiDAR points and camera image pixels. Specifically, for each point $\mathbf{p} = (p^x, p^y, p^z)$ in $P^s$, the corresponding pixel $(u, v)$ can be found by the following transformations:
\begin{equation}
\label{eqn:voxel_pixel_pair}
\begin{split}
[u, v, 1]^\text{T} = \frac{1}{p^z} \cdot T_s \cdot T \cdot [p^x, p^y, 1]^\text{T},
\end{split}
\end{equation}
where $T \in \mathbb{R}^{4\times4}$ is the camera extrinsic matrix that consists of a rotation matrix and a translation matrix, and $T_s \in \mathbb{R}^{3\times4}$ is the camera intrinsic matrix. As we will show in the following sections, such a cross-sensor data alignment serves as the foundation for alignments in other spaces.

\noindent\textbf{Cross-Sensor Statistical Alignment.} 
To mitigate the discrepancies in sensor installations across different datasets, we incorporate a point coordinate alignment operation. Specifically, drawing upon insights from prior domain adaptation approaches \cite{wei2022lidar,yang2021st3d}, we adjust the coordinate origins of point clouds from different datasets by introducing an offset $\sigma \in \mathbb{R}^{1 \times 3}$ to the ground plane. We find empirically that such an alignment can largely reduce the degradation caused by the variations in different sensor setups.

\noindent\textbf{Dataset-Specific Rasterization.}
It is conventional to rasterize LiDAR point clouds $P^s$ using unified rasterization parameters, \eg, voxel size \cite{zhu2021cylindrical,tang2020searching} or horizontal range view resolution \cite{milioto2019rangenet++,xu2020squeezesegv3}. However, the point clouds acquired in different LiDAR datasets naturally differ in density, range, intensity, \etc, which tends to favor different rasterization parameters. To meet such a requirement, we select dataset-specific parameters for rasterization on each dataset through empirical experiments and analyses.

\noindent\textbf{Decoupled BN.}
Another challenge in training across multiple datasets is the presence of domain gaps, which can result in significant statistical shifts of feature learning among datasets. Such shifts can hinder the convergence and affect the model's ability to generalize well across diverse datasets. We adopt a decoupled batch norm (BN) for point cloud features in each dataset. Instead of using the traditional BN, which calculates mean and variance across all samples in a mini-batch, the decoupled BN tends to adapt each dataset's specific characteristics independently.

\subsection{Feature-Space Alignment} 
\label{sec:align_feature}

We aim to acquire a generalized feature representation for downstream tasks. Compared to point clouds, images contribute stronger visual, textural, and semantic information. Thus, the collaboration between pixels and points could enrich the overall representation. Previous research \cite{li2023logonet,chen2023clip2Scene, liu2023uniseg} has consistently demonstrated that such a combination results often leads to improved performance.

\begin{figure}[t]
    \begin{center}
    \includegraphics[width=1.0\linewidth]{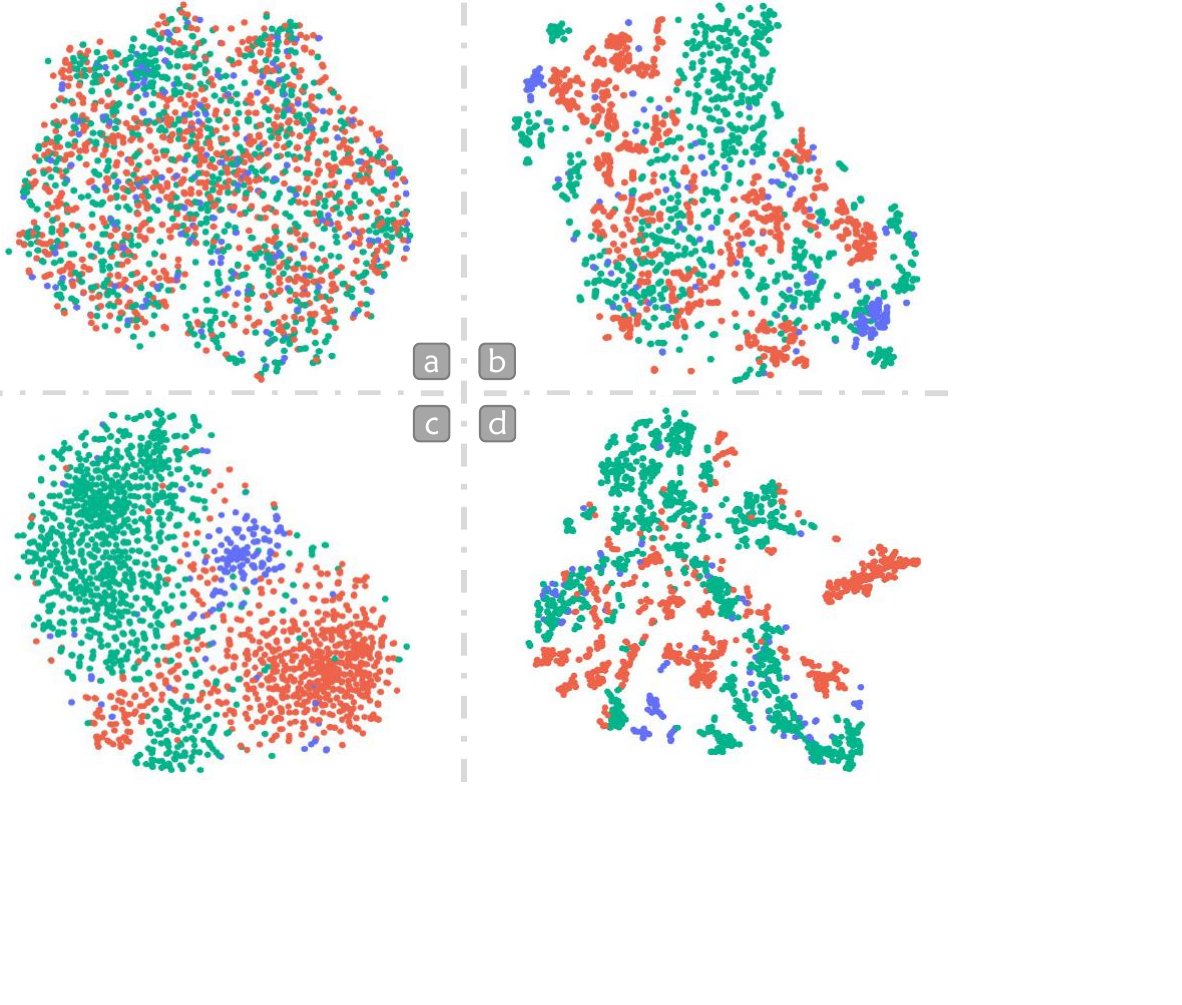}
    \end{center}
    \vspace{-0.5cm}
    \caption{\textbf{The t-SNE plots of learned features} before and after the feature-space alignment in merging the \textit{nuScenes} [\textcolor{m3net_blue}{$\bullet$}], \textit{SemanticKITTI} [\textcolor{m3net_red}{$\bullet$}], and \textit{Waymo Open} [\textcolor{m3net_green}{$\bullet$}] datasets. We show image features from \textbf{(a)} standalone networks; \textbf{(b)} SAM \cite{kirillov2023sam}, and point cloud features \textbf{(c)} before and \textbf{(d)} after the feature-space alignment.}
    \label{fig:tsne}
    \vspace{1ex}
\end{figure}

\noindent\textbf{Cross-Modality Assisted Alignment.} In the context of multi-dataset joint training, our objective is to establish a unified feature space by leveraging image features to assist point cloud features. Acknowledging that images used in training lack ground truth labels \cite{fong2022panoptic-nuScenes,behley2019semanticKITTI}, we utilize image features from a pre-trained model as an alternative, facilitating a more universally applicable representation. We feed camera images $V^s$ into a pre-trained DeepLab~\cite{wang2021max-deeplab} and a vision-language model (VLM) and visualize the output image features by t-SNE~\cite{maaten2008t-sne}. As shown in \cref{fig:tsne}, we observe that image features from DeepLab appear disorderly and lack semantics. In contrast, features from VLM share a more unified feature space. Motivated by this, we propose a cross-modality assisted alignment that uses VLM to help align the feature space. Specifically, the camera images $V^s$ are fed to the frozen image encoder from VLM to obtain image features $F_v = \{\mathcal{F}_v^1, \mathcal{F}_v^2, ..., \mathcal{F}_v^s\}$, where $\mathcal{F}_v^s \in \mathbb{R}^{ c \times h \times w}$. The LiDAR point clouds $P^s$, on the other hand, are fed to the point encoder followed by a projection layer to generate the point features $F_p = \{\mathcal{F}_p^1, \mathcal{F}_p^2, ..., \mathcal{F}_p^s\}$, where $\mathcal{F}_p^s \in \mathbb{R}^{m \times c}$; $m$ denotes the number of non-empty grids. We then leverage the paired image features $\hat{F}_v \in \mathbb{R}^{m_p \times c}$ and point feature $\hat{F}_p \in \mathbb{R}^{m_p \times c}$ for alignment, where $m_p$ is the number of point-pixel pairs. After obtaining $\hat{F}_p$ and $\hat{F}_v$, the cross-modality alignment is expressed as follows:
\begin{equation}
\mathcal{L}_{\text{cma}}(\hat{F}_v, \hat{F}_p) = 1 - \frac{\hat{F}_v \cdot \hat{F}_p}{\|\hat{F}_v\| \cdot \|\hat{F}_p\|}~.
\label{eqn:cos-loss1}
\end{equation}

\begin{figure}[t]
    \begin{center}
    \includegraphics[width=1.0\linewidth]{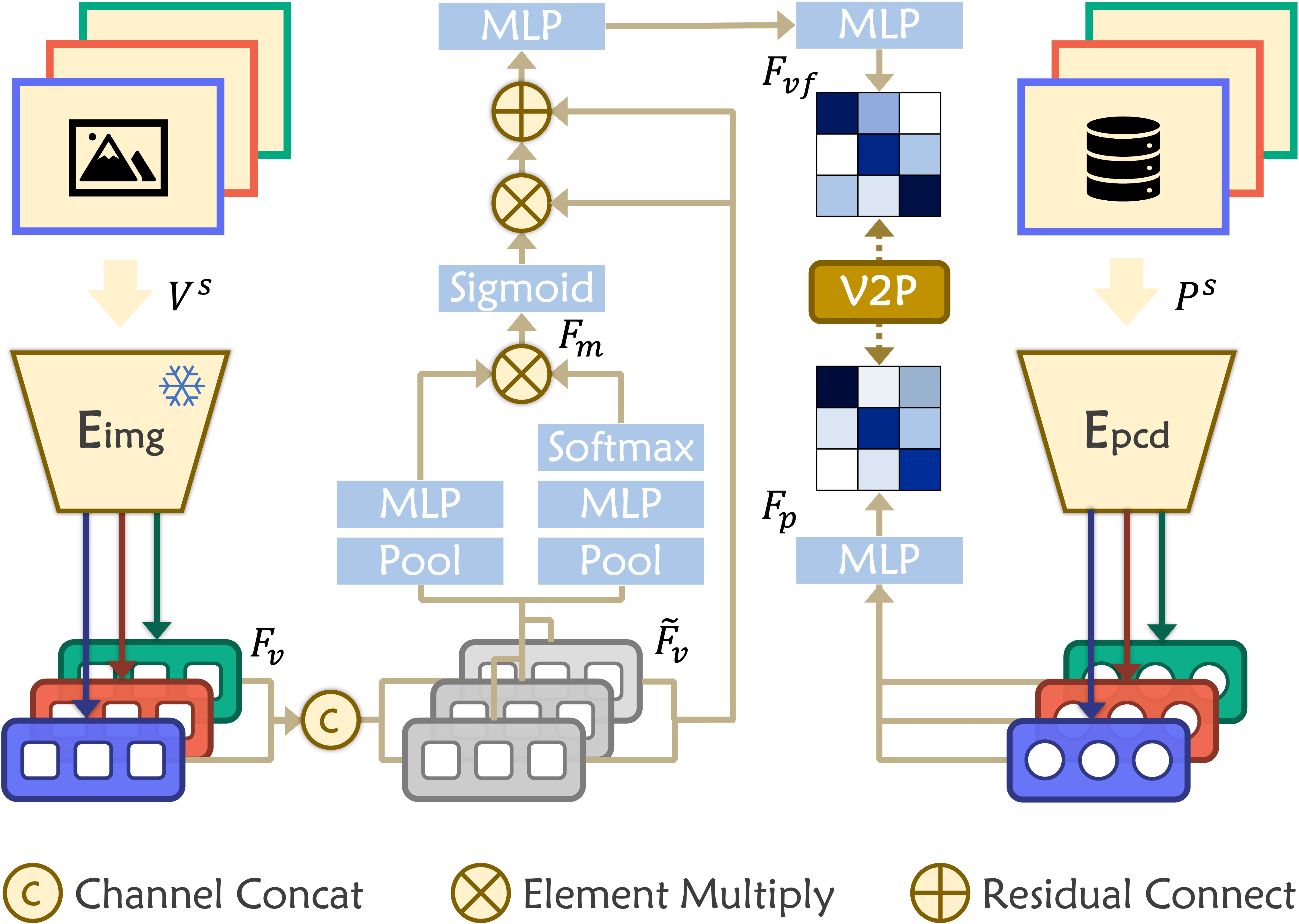}
    \end{center}
    \vspace{-0.45cm}
    \caption{\textbf{Feature-space alignment} in M3Net. We leverage both image features $F_v$ and LiDAR point cloud features $F_p$ extracted from image encoder $E_{img}$ and point encoder $E_{pcd}$ to employ the regularization via V2P loss and achieve feature-space alignment.}
    \label{fig:align_feature}
    \vspace{-1ex}
\end{figure}

\noindent\textbf{Domain-Aware Cross-Modality Alignment.}
With cross-modality alignment, we transfer the knowledge of VLM to the point encoder, enabling the point features to gain a more comprehensive representation. However, during the execution of the above alignment, we have narrowed it exclusively to image and point features from the same dataset. In this mode, point features solely learn from matching image features, restricting their knowledge acquisition. Ideally, we aim to ensure that image features encompass not only scenes identical to those represented in point clouds but also scenes from other datasets. To address this, we propose a domain-aware cross-modality guided alignment, as illustrated in Fig.~\ref{fig:align_feature}. Specifically, we first extract, for each dataset, $F_v$ and $F_p$ from the same image encoder $E_{img}$ and point encoder $E_{pcd}$ during the cross-modality assisted alignment. The sets of features from all datasets are concatenated along the channel dimension to form $\widetilde{F}_v \in \mathbb{R}^{ c_v \times h \times w}$. Subsequently, we sequentially feed $\widetilde{F}_v$ through a branch that consists of a global average pooling and an MLP. Simultaneously, $\widetilde{F}_v$ is fed to an auxiliary branch that undergoes the same processing flow and generates an output after the softmax function $\mathcal{G}(\cdot)$. The outputs from both branches are multiplied to obtain $F_m \in \mathbb{R}^{c_v \times 1 \times 1 }$. The overall process can be described as follows: 
\begin{equation}
F_m = MLP(Pool(\widetilde{F}_v)) \cdot \mathcal{G}( MLP(Pool(\widetilde{F}_v)))~.
\label{eqn:fea-forward1}
\end{equation}
Next, we forward $F_m$ to a sigmoid activation function $\mathcal{H}(\cdot)$ and multiply it with input image features $\widetilde{F}_v$. The resulting output is added to $\widetilde{F}_v$ and passed through the MLP layers to obtain the final image features $F_{vf}  \in \mathbb{R}^{ c \times h \times w}$. The forward process of this operation is depicted as follows:
\begin{equation}
F_{vf} = MLP((\mathcal{H}(F_m) \cdot \widetilde{F}_v) + \widetilde{F}_v)~.
\label{eqn:fea-forward2}
\end{equation}
Finally, we leverage the cross-modality data alignment to acquire paired image features $\hat{F}_{vf} \in \mathbb{R}^{m_p \times c}$ and paired point feature $\hat{F}_p$. The overall objective function is:
\begin{equation}
\mathcal{L}_{v2p}(\hat{F}_{vf}, \hat{F}_p) = 1 - \frac{\hat{F}_{vf} \cdot \hat{F}_p}{\|\hat{F}_{vf}\| \cdot \|\hat{F}_p\|}~.
\label{eqn:cos-loss2}
\end{equation}

\subsection{Label-Space Alignment} 
\label{sec:align_label}

\noindent\textbf{Label Conflict.} In multi-dataset joint training settings, label conflicts emerge as a significant challenge. This often refers to the inconsistencies in class labels across different datasets involved in the training process. The discrepancy can arise due to variations in annotation conventions, labeling errors, or even differences in the underlying semantics of classes between datasets. In our baseline, we unionize the different label spaces across datasets into $\mathbb{Y}^u$, where all datasets share a single LiDAR segmentation head. However, this may introduce several potential drawbacks: 
\begin{itemize}
    \item \textit{Loss of granularity:} Unified label spaces could lose semantic granularity, particularly when dealing with subtle category differences in between different datasets.
    \item \textit{Information loss:} During label space consolidation, details unique to each dataset may be obscured or lost, especially for those related to domain-specific categories.
    \item \textit{Increased complexity:} Handling a unified label space may necessitate more complex model architectures or training strategies, thereby increasing overall complexity.
\end{itemize}

To address these issues, we introduce a language-guided label-space alignment to facilitate a more holistic semantic correlation across datasets. Given the natural correspondence between images and texts and the strong correlation between images and point clouds, we aim to strategically utilize the image modality as a bridge to establish language-guided alignments. Such a process consists of a text-driven point alignment, a text-driven image alignment, and a cross-modality-assisted label alignment.

\begin{figure}[t]
    \begin{center}
    \includegraphics[width=1.0\linewidth]{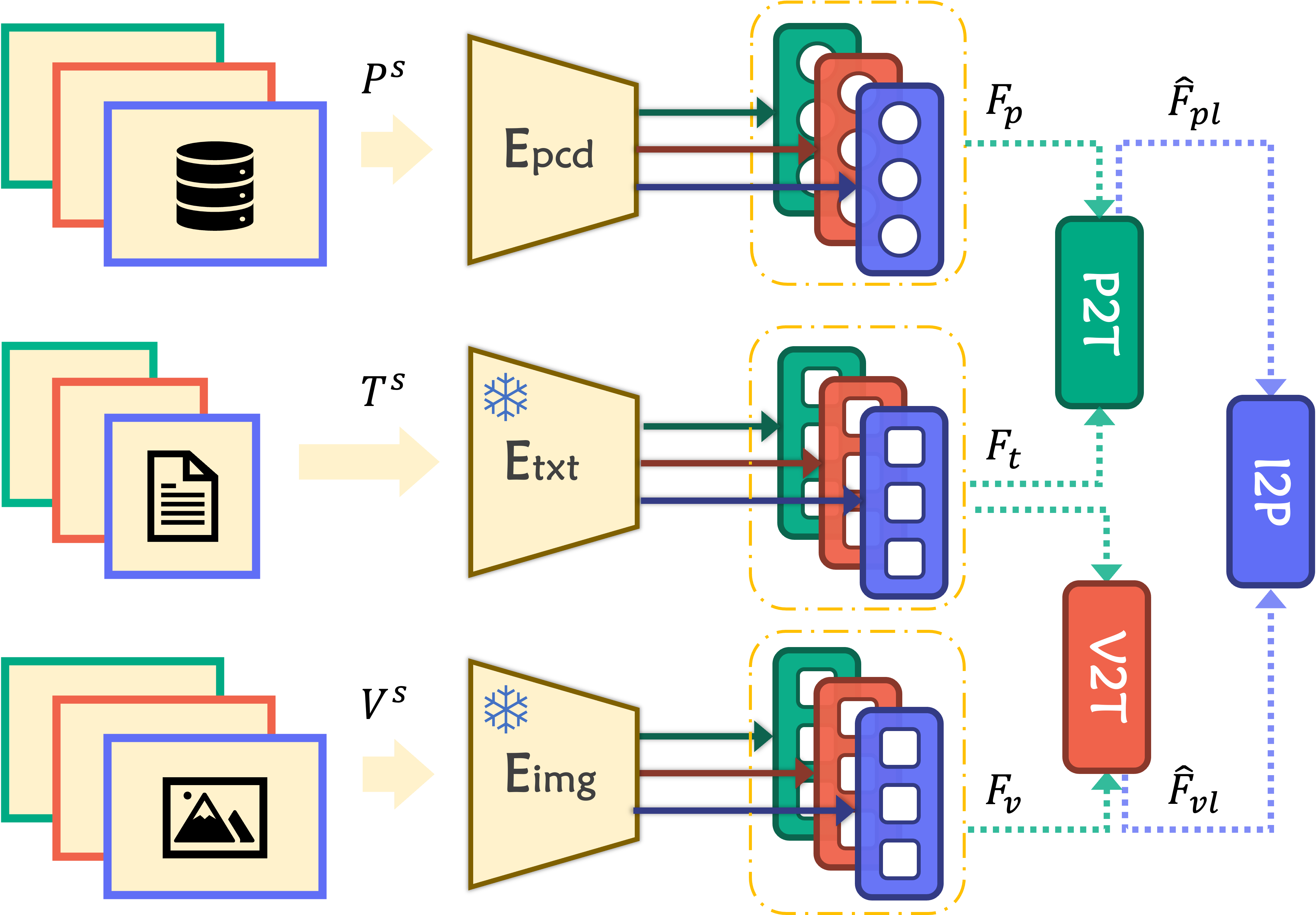}
    \end{center}
    \vspace{-0.4cm}
    \caption{\textbf{Label-space alignment} in M3Net. We leverage image features $F_v$, point cloud features $F_p$, and text embedding $F_t$ extracted from $E_{img}$, $E_{pcd}$, and $E_{txt}$, respectively, for  regularization via the I2P, P2T, and V2T losses in the label-space alignment.}
    \label{fig:align_label}
    \vspace{-1ex}
\end{figure}

\noindent\textbf{Text-Driven Alignments.} As depicted in Fig.~\ref{fig:align_label}, images $V^s$ are fed into the frozen image encoder $E_{img}$ to extract the image features $F_v$. Concurrently, the LiDAR point clouds $P^s$ are processed by the point encoder $E_{pcd}$ to generate the point features $F_p$. Additionally, given the text input $T^s$, text embedding features $F_t \in \mathbb{R}^{Q \times c}$ are obtained from a frozen text encoder $E_{txt}$, where $Q$ represents the number of categories across datasets. The text is composed of class names from unified label space $\mathbb{Y}_u$ placed into pre-defined templates, and the text embedding captures semantic information of the corresponding classes. Subsequently, pixel-text pairs $\{v_k,t_k\}_{k=1}^M$ and point-text pairs $\{p_k,t_k\}_{k=1}^M$ are generated, where $M$ represents the number of pairs. Leveraging the semantic information contained in the text, we selectively choose positive and negative samples for both images and points for contrastive learning. It is noteworthy that negative samples are confined to the specific dataset category space. The overall objective of the text-driven point alignment function is shown as follows:
\begin{equation}
\label{equ:point-text-info-loss}
\mathcal{L}_{p2t} = - \sum_{q=1}^{Q} \log( \frac{\sum_{t_{k} \in q, p_{k}}\exp(<t_{k}, p_{k}>/\tau)}{\sum_{t_{k} \in q, t_{k} \notin q, p_{j}}\exp(<t_{k}, p_{k}>/\tau)} )~,
\end{equation}
where $t_{k} \in q$ indicates that $t_{k}$ is generated by the $q$-th classes name, and $Q$ is the number of classes. Symbol $<,>$ denotes the scalar product operation and $\tau$ is a temperature term ($\tau > 0$). Similarly, the objective of the text-driven image alignment function is illustrated as follows:
\begin{equation}
\label{equ:image-text-info-loss}
\mathcal{L}_{v2t} = - \sum_{q=1}^{Q} \log( \frac{\sum_{t_{k} \in q, v_{k}}\exp(<t_{k}, v_{k}>/\tau)}{\sum_{t_{k} \in q, t_{k} \notin q, v_{j}}\exp(<t_{k}, v_{k}>/\tau)} )~.
\end{equation}

\noindent\textbf{Cross-Modality-Assisted Label Alignment.} After text-driven alignments, the subsequent crucial step entails aligning the point and image modalities within the label space. We first obtain image logits $F_{vl} = \{\mathcal{F}_{vl}^1, \mathcal{F}_{vl}^2, ..., \mathcal{F}_{vl}^s\}$ and point logits $F_{pl} =\{\mathcal{F}_{pl}^1, \mathcal{F}_{pl}^2, ..., \mathcal{F}_{pl}^s\}$ from text-driven alignments, where $\mathcal{F}_{vl}^s \in \mathbb{R}^{Q \times H \times W}$, $\mathcal{F}_{pl}^s \in \mathbb{R}^{N \times Q}$. Subsequently, we conduct cross-modality alignment to obtain paired image logits $\hat{F}_{vl} \in \mathbb{R}^{m_p \times Q}$ and paired point logits $\hat{F}_{pl} \in \mathbb{R}^{m_p \times Q}$. Formally, the cross-modality-assisted alignment in the label space is formulated as follows:
\begin{equation}
\mathcal{L}_{i2p}(\hat{F}_{vl}, \hat{F}_{pl}) = 1 - \frac{\hat{F}_{vl} \cdot \hat{F}_{pl}}{\|\hat{F}_{vl}\| \cdot \|\hat{F}_{pl}\|}~.
\label{eqn:i2p}
\end{equation}
Finally, the complete objective function for the language-guided label-space alignment is expressed as follows:
\begin{equation}
\mathcal{L}_{label} = \mathcal{L}_{p2t} + \mathcal{L}_{i2p} + \mathcal{L}_{v2t}~.
\label{eqn:label-align}
\end{equation}

\subsection{Universal LiDAR Segmentation}

We enhance the versatility of M3Net via multi-tasking learning. This integration involves an instance extractor to enable joint semantic and panoptic LiDAR segmentation.

\noindent\textbf{Panoptic LiDAR Segmentation.} Motivated by DSNet~\cite{hong2021dsnet,hong20224dDSNet}, our instance extractor comprises an instance head and a clustering step. The instance head encompasses several MLPs designed to predict the offsets between instance centers. The clustering step uses semantic predictions to filter out \textit{stuff} points, thereby retaining only those associated with \textit{thing} points. The remaining points undergo a mean-shift clustering \cite{marcuzzi2002MeanShift}, utilizing features from the instance head to discern distinct instances. Lastly, we employ the L1 loss $\mathcal{L}_{l1}$ to optimize the \textit{thing} point regression process. 

\noindent\textbf{Overall Objectives.}
Putting everything together, the overall objective of M3Net is to minimize the following losses: 
\begin{equation}
\mathcal{L} = \mathcal{L}_{v2p} + \mathcal{L}_{label} + \mathcal{L}_{ce} +
\mathcal{L}_{lovasz} + \mathcal{L}_{l1}~,
\label{eqn:overall-loss}
\end{equation}
where $\mathcal{L}_{ce}$ and $\mathcal{L}_{lovasz}$ denote the cross-entropy loss and the Lovasz-softmax \cite{berman2018lovasz} loss, respectively.

\section{Experiments}
\label{sec:experiments}

\begin{table*}[t]
\caption{\textbf{Ablation study} on the M3Net alignments happen in the \textcolor{m3net_blue}{\textbf{Data}}, \textcolor{m3net_red}{\textbf{Feature}}, and \textcolor{m3net_green}{\textbf{Label}} spaces, respectively, when combining the \textit{SemanticKITTI} \cite{behley2019semanticKITTI}, \textit{nuScenes} \cite{fong2022panoptic-nuScenes}, and \textit{Waymo Open} \cite{sun2020waymoOpen} datasets. The mAcc and mIoU scores are in percentage. Best scores are in \textbf{bold}.}
\vspace{-0.1cm}
\centering\scalebox{0.79}{
\begin{tabular}{c|p{32pt}<{\centering}p{32pt}<{\centering}p{32pt}<{\centering}|p{25.2pt}<{\centering}p{25.2pt}<{\centering}|p{25.2pt}<{\centering}p{25.2pt}<{\centering}|p{25.2pt}<{\centering}p{25.2pt}<{\centering}|p{25.2pt}<{\centering}p{25.2pt}<{\centering}|p{25.2pt}<{\centering}p{25.2pt}<{\centering}|p{25.2pt}<{\centering}p{25.2pt}<{\centering}}
    \toprule
    \multirow{3}{*}{-} & \multicolumn{3}{c|}{\multirow{3}{*}{\textcolor{darkgray}{\textbf{Configurations}}}} & \multicolumn{6}{c|}{\textcolor{darkgray}{\textbf{MinkUNet}} \cite{choy2019minkowski}} & \multicolumn{6}{c}{\textcolor{darkgray}{\textbf{PTv2+}} \cite{wu2022ptv2}}
    \\
    & & & & \multicolumn{2}{c|}{\textcolor{darkgray}{\textbf{SemKITTI}}} & \multicolumn{2}{c|}{\textcolor{darkgray}{\textbf{nuScenes}}} & \multicolumn{2}{c|}{\textcolor{darkgray}{\textbf{Waymo}}} & \multicolumn{2}{c|}{\textcolor{darkgray}{\textbf{SemKITTI}}} & \multicolumn{2}{c|}{\textcolor{darkgray}{\textbf{nuScenes}}} & \multicolumn{2}{c}{\textcolor{darkgray}{\textbf{Waymo}}}
    \\
    & & & & mAcc & mIoU & mAcc & mIoU & mAcc & mIoU & mAcc & mIoU & mAcc & mIoU & mAcc & mIoU
    \\\midrule\midrule
    \textcolor{gray}{Baseline} & \multicolumn{3}{c|}{\textcolor{gray}{Na\"{\i}ve Joint Training}} & \textcolor{gray}{$62.43$} & \textcolor{gray}{$54.03$} & \textcolor{gray}{$65.05$} & \textcolor{gray}{$59.84$} & \textcolor{gray}{$73.76$} & \textcolor{gray}{$65.39$} & \textcolor{gray}{$67.96$} & \textcolor{gray}{$61.59$} & \textcolor{gray}{$76.53$} & \textcolor{gray}{$69.65$} & \textcolor{gray}{$75.68$} & \textcolor{gray}{$67.00$}
    \\\midrule
    \multirow{5}{*}{\textbf{\parbox{1cm}{M3Net\\ (Ours)}}} & \textcolor{m3net_blue}{\textbf{Data}} & \textcolor{m3net_red}{\textbf{Feature}} & \textcolor{m3net_green}{\textbf{Label}} & \multicolumn{2}{c|}{-} & \multicolumn{2}{c|}{-} & \multicolumn{2}{c|}{-} & \multicolumn{2}{c|}{-} & \multicolumn{2}{c|}{-} & \multicolumn{2}{c}{-}
    \\
    & \textcolor{m3net_blue}{\textbf{\checkmark}} & & & $73.82$ & $69.01$ & $83.66$ & $76.89$ & $77.88$ & $69.37$ & $78.55$ & $69.95$ & $86.22$ & $79.13$ & $80.96$ & $72.15$
    \\
    & \textcolor{m3net_blue}{\textbf{\checkmark}} & \textcolor{m3net_red}{\textbf{\checkmark}} & & $74.36$ & $69.64$ & $85.17$ & $78.88$ & $78.31$ & $69.70$ & $79.43$ & $70.87$ & $87.10$ & $80.26$ & $80.74$ & $72.33$
    \\
    & \textcolor{m3net_blue}{\textbf{\checkmark}} & & \textcolor{m3net_green}{\textbf{\checkmark}} & $73.85$ & $69.34$ & $85.20$ & $78.90$ & $78.04$ & $69.55$ & $80.30$ & $71.13$ & $87.44$ & $80.45$  &  $80.69$ & $72.30$
    \\
    & \textcolor{m3net_blue}{\textbf{\checkmark}} & \textcolor{m3net_red}{\textbf{\checkmark}} & \textcolor{m3net_green}{\textbf{\checkmark}} & $\mathbf{74.40}$ & $\mathbf{69.85}$ & $\mathbf{85.30}$ & $\mathbf{79.00}$ & $\mathbf{78.66}$ & $\mathbf{70.15}$ & $\mathbf{80.00}$ & $\mathbf{72.00}$ & $\mathbf{87.91}$ & $\mathbf{80.90}$ & $\mathbf{81.11}$ & $\mathbf{72.40}$ 
    \\\bottomrule
\end{tabular}}
\label{tab:ablation_semantic}
\vspace{0.1cm}
\end{table*}
\begin{table*}[t]
\caption{\textbf{Panoptic LiDAR segmentation} results on the \textit{val} sets of the \textit{Panoptic-SemanticKITTI} \cite{behley2019semanticKITTI} and \textit{Panoptic-nuScenes} \cite{fong2022panoptic-nuScenes} datasets. All scores are given in percentage. The best and second-best scores are highlighted in \textbf{bold} and \underline{underline}, respectively.}
\vspace{-0.1cm}
\centering\scalebox{0.79}{
\begin{tabular}{r|p{33pt}<{\centering}p{33pt}<{\centering}p{33pt}<{\centering}|p{25.8pt}<{\centering}p{25.8pt}<{\centering}p{25.8pt}<{\centering}p{25.8pt}<{\centering}p{25.8pt}<{\centering}|p{25.8pt}<{\centering}p{25.8pt}<{\centering}p{25.8pt}<{\centering}p{25.8pt}<{\centering}p{25.8pt}<{\centering}}
    \toprule
    \multirow{2}{*}{\textcolor{darkgray}{\textbf{Method}}} & \multicolumn{3}{c|}{\multirow{2}{*}{\textcolor{darkgray}{\textbf{Configurations}}}} & \multicolumn{5}{c|}{\textcolor{darkgray}{\textbf{Panoptic-SemanticKITTI}}} & \multicolumn{5}{c}{\textcolor{darkgray}{\textbf{Panoptic-nuScenes}}}
    \\
    & & & & PQ & PQ$^\dagger$ & RQ & SQ & mIoU & PQ & PQ$^\dagger$ & RQ & SQ & mIoU
    \\\midrule\midrule
    Panoptic-TrackNet \cite{hurtado2020mopt} & \multicolumn{3}{c|}{\multirow{5}{*}{Single-Dataset Training}} & $40.0$ & - & $48.3$ & $73.0$ & $53.8$ & $51.4$ & $56.2$ & $63.3$ & $80.2$ & $58.0$ 
    \\
    Panoptic-PolarNet \cite{zhou2021panoptic} & & & & $59.1$ & $64.1$ & $70.2$ & $78.3$ & $64.5$ & $63.4$ & $67.2$ & $75.3$ & $83.9$ & $66.9$
    \\
    EfficientLPS \cite{sirohi2021efficientlps} & & & & $59.2$ & $65.1$ & $69.8$ & $75.0$ & $64.9$ & $59.2$ & $62.8$ & $82.9$ & $70.7$ & $69.4$
    \\
    DSNet \cite{hong2021dsnet} & & & & $61.4$ & $65.2$ & $72.7$ & $79.0$ & $69.6$ & $64.7$ & $67.6$ & $76.1$ & $83.5$ & $76.3$
    \\
    Panoptic-PHNet \cite{li2022panoptic} & & & & $61.7$ & - & - & - & $65.7$ & $\mathbf{74.7}$ & $\mathbf{77.7}$ & $\mathbf{84.2}$ & $\mathbf{88.2}$ & $79.7$
    \\\midrule
    \textcolor{gray}{Baseline}~~~~~~~~~~~~ & \multicolumn{3}{c|}{\textcolor{gray}{Na\"{\i}ve Joint Training}} & \textcolor{gray}{$56.03$} & \textcolor{gray}{$59.64$} & \textcolor{gray}{$65.78$} & \textcolor{gray}{$73.72$} & \textcolor{gray}{$61.59$} & \textcolor{gray}{$56.67$} & \textcolor{gray}{$60.61$} & \textcolor{gray}{$66.75$} & \textcolor{gray}{$83.49$} & \textcolor{gray}{$69.65$}
    \\\midrule
    \multirow{5}{*}{\textbf{\parbox{2cm}{M3Net\\ (Ours)}}}~~ & \textcolor{m3net_blue}{\textbf{Data}} & \textcolor{m3net_red}{\textbf{Feature}} & \textcolor{m3net_green}{\textbf{Label}} & \multicolumn{5}{c|}{-} & \multicolumn{5}{c}{-}
    \\
    & \textcolor{m3net_blue}{\textbf{\checkmark}} & & & $62.34$ & $65.17$ & $72.60$ & $74.67$ & $69.95$ & $68.49$ & $71.11$ & $79.13$ & $85.49$ & $79.13$
    \\
    & \textcolor{m3net_blue}{\textbf{\checkmark}} & \textcolor{m3net_red}{\textbf{\checkmark}} & & $62.91$ & $65.73$ & \underline{$73.32$} & $75.47$ & $70.87$ & $71.47$ & $73.86$ & $81.53$ & $86.71$ & $80.26$
    \\
    & \textcolor{m3net_blue}{\textbf{\checkmark}} & & \textcolor{m3net_green}{\textbf{\checkmark}} & \underline{$63.23$} & \underline{$67.89$} & $\mathbf{73.61}$ & \underline{$81.66$} & \underline{$71.13$} & $71.53$ & $73.91$ & $81.80$ & \underline{$86.92$} & \underline{$80.45$}
    \\
    & \textcolor{m3net_blue}{\textbf{\checkmark}} & \textcolor{m3net_red}{\textbf{\checkmark}} & \textcolor{m3net_green}{\textbf{\checkmark}} & $\mathbf{63.87}$ & $\mathbf{68.66}$ & $73.10$ & $\mathbf{82.35}$ & $\mathbf{72.00}$ & \underline{$71.70$} & \underline{$74.01$} & \underline{$82.20$} & $86.47$ & $\mathbf{80.90}$
    \\\bottomrule
\end{tabular}}
\label{tab:ablation_panoptic}
\vspace{0.1cm}
\end{table*}

\subsection{Experimental Setups}

\noindent\textbf{Datasets.}
Our M3Net framework and baselines are trained on a combination of \textbf{\textit{nuScenes}} \cite{fong2022panoptic-nuScenes}, \textbf{\textit{SemanticKITTI}} \cite{behley2019semanticKITTI}, and \textbf{\textit{Waymo Open}} \cite{sun2020waymoOpen}. Meanwhile, we resort to another five LiDAR-based perception datasets \cite{jiang2021rellis3D,pan2020semanticPOSS,xiao2023semanticSTF,xiao2022synLiDAR,klokov2023daps3D} and two 3D robustness evaluation datasets \cite{kong2023robo3D} to verify the strong generalizability of M3Net. Due to space limits, additional details regarding the datasets are in the Appendix.

\noindent\textbf{Implementation Details.}
M3Net is implemented based on Pointcept \cite{pointcept2023} and MMDetection3D \cite{mmdet3d2020}. We use two backbones in our experiments, \ie, MinkUNet~\cite{choy2019minkowski} and PTv2+~\cite{wu2022ptv2}. We trained M3Net on four A$100$ GPUs for $50$ epochs with a batch size of $6$ for each GPU. The initial learning rate is set to $0.002$. We adopt the AdamW optimizer~\cite{loshchilov2019adamw} with a weight decay of $0.005$ and cosine decay learning rate scheduler. For the dataset-specific rasterization, we set voxel sizes to $0.05$m, $0.1$m, and $0.05$m for \textit{SemanticKITTI} \cite{behley2019semanticKITTI}, \textit{nuScenes} \cite{fong2022panoptic-nuScenes}, and \textit{Waymo Open} \cite{sun2020waymoOpen}, respectively. For the data augmentation, we employ random flipping, jittering, scaling, rotation, and Mix3D~\cite{nekrasov2021mix3d}. Due to space limits, kindly refer to Appendix for additional details.

\noindent\textbf{Evaluation Metrics.}
We adopt conventional reportings of \textit{mAcc} and \textit{mIoU} for LiDAR semantic segmentation, \textit{PQ}, \textit{PQ}$^\dagger$, \textit{SQ}, and \textit{RQ} for panoptic segmentation, and \textit{mCE} and \textit{mRR} for 3D robustness evaluation. Due to the space limit, kindly refer to our Appendix for more detailed definitions.

\subsection{Ablation Study}

\noindent\textbf{Multi-Space Alignments.}
The effectiveness of three proposed alignments over the joint training baselines is shown in Tab.~\ref{tab:ablation_semantic}. We observe that the data-space alignment plays the most crucial role in improving the universal LiDAR segmentation performance. Without proper data alignments, joint training with either MinkUNet \cite{choy2019minkowski} or the stronger PTv2+ \cite{wu2022ptv2} will suffer severe degradation, especially on sparser point clouds \cite{fong2022panoptic-nuScenes}. On top of the data-space alignment, the combinations of multi-view images at the feature space and the language-guided knowledge at the label space further enhance the learned feature representations. The results show that they work synergistically in merging knowledge from heterogeneous domains during joint training.

\noindent\textbf{Panoptic LiDAR Segmentation.}
In \cref{tab:ablation_panoptic}, we present another ablation study focusing on panoptic LiDAR segmentation. All three alignments incorporated in M3Net demonstrate significant improvements over the baselines. This highlights the pronounced efficacy of our multi-space alignments. Moreover, our approach outperforms the single-dataset state-of-the-art method Panaptic-PHNet~\cite{li2022panoptic} by a notable $2.17$\% PQ on \textit{Panoptic-SemanticKITTI} \cite{behley2019semanticKITTI} and achieves compelling results on \textit{Panoptic-nuScenes} \cite{fong2022panoptic-nuScenes}.

\begin{table*}[t]
\caption{\textbf{Knowledge transfer and generalization} analyses across five LiDAR segmentation datasets and two 3D robustness evaluation datasets. All scores are given in percentage. The best and second-best scores are highlighted in \textbf{bold} and \underline{underline}, respectively.}
\vspace{-0.1cm}
\centering\scalebox{0.79}{
\begin{tabular}{r|p{27.4pt}<{\centering}p{27.4pt}<{\centering}|p{27.4pt}<{\centering}p{27.4pt}<{\centering}|p{27.4pt}<{\centering}p{27.4pt}<{\centering}|p{27.4pt}<{\centering}p{27.4pt}<{\centering}|p{27.4pt}<{\centering}p{27.4pt}<{\centering}|p{27.4pt}<{\centering}p{27.4pt}<{\centering}|p{27.4pt}<{\centering}p{27.4pt}<{\centering}}
    \toprule
    \multirow{2}{*}{\textbf{Method}} & \multicolumn{2}{c|}{\textcolor{darkgray}{\textbf{RELLIS-3D}}} & \multicolumn{2}{c|}{\textcolor{darkgray}{\textbf{SemanticPOSS}}} & \multicolumn{2}{c|}{\textcolor{darkgray}{\textbf{SemanticSTF}}} & \multicolumn{2}{c|}{\textcolor{darkgray}{\textbf{SynLiDAR}}} & \multicolumn{2}{c|}{\textcolor{darkgray}{\textbf{DAPS-3D}}} & \multicolumn{2}{c|}{\textcolor{darkgray}{\textbf{SemKITTI-C}}} & \multicolumn{2}{c}{\textcolor{darkgray}{\textbf{nuScenes-C}}}
    \\
    & $1\%$ & $10\%$ & Half & Full & Half & Full & $1\%$ & $10\%$ & Half & Full & mCE & mRR & mCE & mRR
    \\\midrule\midrule
    PPKT \cite{liu2021ppkt} & $49.71$ & $54.33$ & $50.18$ & $56.00$ & $50.92$ & $54.69$ & $37.57$ & $46.48$ & $78.90$ & $84.00$ & - & - & $105.64$ & $76.06$
    \\
    SLidR \cite{sautier2022slidr} & $49.75$ & $54.57$ & $51.56$ & $55.36$ & $52.01$ & $54.35$ & $42.05$ & $47.84$ & $81.00$  & $85.40$ & - & - & $106.08$ & $75.99$
    \\
    Seal \cite{liu2023segment} & \underline{$51.09$} & \underline{$55.03$} & \underline{$53.26$} & \underline{$56.89$} & \underline{$53.46$} & \underline{$55.36$} & \underline{$43.58$} & \underline{$49.26$} & \underline{$81.88$} & \underline{$85.90$} & - & - & \underline{$92.63$} & $\mathbf{83.08}$
    \\\midrule
    \textcolor{gray}{Na\"{\i}ve Joint} & \textcolor{gray}{$37.77$} & \textcolor{gray}{$50.23$} & \textcolor{gray}{$42.19$} & \textcolor{gray}{$52.31$} & \textcolor{gray}{$46.70$} & \textcolor{gray}{$48.00$} & \textcolor{gray}{$18.56$} & \textcolor{gray}{$42.37$} & \textcolor{gray}{$73.91$} & \textcolor{gray}{$77.89$} & \textcolor{gray}{$113.65$} & \textcolor{gray}{$84.73$} & \textcolor{gray}{$128.97$} & \underline{\textcolor{gray}{$81.45$}}
    \\
    Single-Dataset & $40.17$ & $54.25$ & $47.69$ & $55.00$ & $50.33$ & $51.19$ & $23.17$ & $45.08$ & $75.10$ & $80.87$ & \underline{$95.11$} & \underline{$84.95$} & $99.63$ & $79.06$
    \\
    \textbf{M3Net (Ours)} & $\mathbf{51.27}$ & $\mathbf{55.05}$ & $\mathbf{53.60}$ & $\mathbf{57.17}$ & $\mathbf{53.78}$ & $\mathbf{55.42}$ & $\mathbf{44.10}$ & $\mathbf{49.93}$ & $\mathbf{82.08}$ & $\mathbf{86.00}$ & $\mathbf{86.43}$ & $\mathbf{85.77}$ & $\mathbf{91.03}$ & $79.15$
    \\\bottomrule
\end{tabular}}
\label{tab:generalization}
\vspace{-0.1cm}
\end{table*}

\begin{table}[t]
\caption{\textbf{LiDAR semantic segmentation} results on the \textit{val} and \textit{test} sets of \textit{SemanticKITTI} \cite{behley2019semanticKITTI} and \textit{nuScenes} \cite{fong2022panoptic-nuScenes}, and the \textit{val} set of \textit{Waymo Open} \cite{sun2020waymoOpen}. All scores are in percentage. The best and second-best scores are highlighted in \textbf{bold} and \underline{underline}.}
\vspace{-0.1cm}
\centering\scalebox{0.72}{
\begin{tabular}{r|p{26.5pt}<{\centering}p{26.5pt}<{\centering}|p{26.5pt}<{\centering}p{26.5pt}<{\centering}|p{26.5pt}<{\centering}p{26.5pt}<{\centering}}
    \toprule
    \multirow{2}{*}{\textbf{Method}} & \multicolumn{2}{c|}{\textcolor{darkgray}{\textbf{SemKITTI}}} & \multicolumn{2}{c|}{\textcolor{darkgray}{\textbf{nuScenes}}} & \multicolumn{2}{c}{\textcolor{darkgray}{\textbf{Waymo}}}
    \\
    & Val & Test & Val & Test & mIoU & mAcc
    \\\midrule\midrule
    RangeNet++~\cite{milioto2019rangenet++} & - & $52.2$ & - & $65.5$ & - & -
    \\
    PolarNet~\cite{zhou2020polarNet} & $57.2$ & $54.3$ & $71.0$ & $69.8$ & - & -
    \\
    SalsaNext~\cite{cortinhal2020salsanext} & - & $59.5$ & - & $72.2$ & - & -
    \\
    RangeViT~\cite{ando2023rangevit} & $60.7$ & $64.0$ & $75.2$ & - & - & -
    \\
    MinkUNet~\cite{choy2019minkowski} & $63.8$ & $63.7$ & $73.3$ & - & $65.9$ & $76.6$
    \\
    SPVNAS~\cite{tang2020searching} & $64.7$ & $66.4$ & - & $77.4$ & $67.4$ & -
    \\
    AMVNet~\cite{liong2020amvNet} & $65.2$ & $65.3$ & $77.2$ & $77.3$ & - & -
    \\
    RPVNet~\cite{xu2021rpvnet} & $65.5$ & $70.3$ & $77.6$ & - & - & -
    \\
    (AF)$^2$-S3Net~\cite{cheng2021af2S3Net} & - & $69.7$ & - & $78.3$ & - & -
    \\
    Cylinder3D~\cite{zhu2021cylindrical} & $65.9$ & $67.8$ & $76.1$ & $77.9$ & $66.0$ & -
    \\
    PVKD~\cite{hou2022pvkd} & $66.4$ & $71.2$ & $76.0$ & - & - & -
    \\
    WaffleIron~\cite{puy23waffleiron} & $66.8$ & $70.8$ & $79.1$ & - & - & -
    \\
    RangeFormer~\cite{kong2023rethinking} & $67.6$ & $73.3$ & $78.1$ & $80.1$ & - & -
    \\
    SphereFormer~\cite{lai2023sphereformer} & $67.8$ & \underline{$74.8$} & $78.4$ & $81.9$ & $69.9$ & -
    \\
    FRNet~\cite{xu2023frnet} & $68.7$ & $73.3$ & $79.0$ & $82.5$ & - & -
    \\
    PTv2+~\cite{wu2022ptv2} & \underline{$70.3$} & $70.6$ & \underline{$80.2$} & \underline{$82.6$} & $70.6$ & \underline{$80.2$}
    \\
    LidarMultiNet~\cite{ye2023lidarmultinet} & - & - & - & $81.4$ & $\mathbf{73.8}$ & -
    \\\midrule
    \textbf{M3Net (Ours)} & $\mathbf{72.0}$ & $\mathbf{75.1}$ & $\mathbf{80.9}$ & $\mathbf{83.1}$ &  \underline{$72.4$}  & $\mathbf{81.1}$ 
    \\\bottomrule
\end{tabular}}
\vspace{-0.1cm}
\label{tab:comparative}
\end{table}

\noindent\textbf{Visual Feature Alignments.} 
We conduct a qualitative analysis of the learned visual feature distributions in the form of t-SNE \cite{maaten2008t-sne}. \cref{fig:tsne}~\textcolor{red}{(a)} and \textcolor{red}{(b)} represent the distributions of learned visual features among three datasets from DeepLab and VLM backbones, respectively. The features obtained by the latter exhibit more distinct semantics in feature space. The concentrated distribution space is advantageous for achieving feature alignments across multiple datasets. Additionally, \cref{fig:tsne}~\textcolor{red}{(c)} and \textcolor{red}{(d)} illustrate the distribution of point cloud features before and after feature-space alignment. As can be seen, the feature distribution distances between the three datasets have been largely reduced, providing evidence of the alignment effectiveness.

\subsection{Comparative Study}

\noindent\textbf{Comparisons to State of the Arts.}
In Tab.~\ref{tab:comparative}, we compare M3Net with current best-performing models on the benchmarks of \textit{SemanticKITTI} \cite{behley2019semanticKITTI}, \textit{nuScenes} \cite{fong2022panoptic-nuScenes}, and \textit{Waymo Open} \cite{sun2020waymoOpen}.
Remarkably, M3Net consistently outperforms existing approaches across all three datasets. Specifically, on \textit{SemanticKITTI} \cite{behley2019semanticKITTI}, M3Net achieves a $72.0\%$ mIoU on the validation set, surpassing the closest method by a notable margin of $1.7\%$ mIoU. Similarly, on \textit{nuScenes} \cite{fong2022panoptic-nuScenes}, M3Net achieves $80.9\%$ mIoU and $83.1\%$ mIoU on the validation and test sets, demonstrating its robustness and generalization capabilities. Additionally, the performance of M3Net on \textit{Waymo Open} \cite{sun2020waymoOpen} is competitive with prior arts. We achieve a mIoU of $72.4\%$ and a mAcc of $81.1\%$. These results highlight again the superiority of M3Net in handling complex diverse LiDAR segmentation tasks.

\noindent\textbf{Direct Knowledge Transfer.}
To further validate the strong knowledge transfer capability of M3Net, we conduct extensive experiments on five different LiDAR-based perception datasets \cite{jiang2021rellis3D,pan2020semanticPOSS,xiao2023semanticSTF,xiao2022synLiDAR,klokov2023daps3D}. These datasets have unique data collection protocols and data distributions. As shown in \cref{fig:teaser} and the first ten columns in \cref{tab:generalization}, our framework constantly outperforms the prior arts, the na\"{\i}ve joint training, and the single-dataset baselines across all five datasets. This concretely supports the strong knowledge transfer efficacy brought by multi-space alignments in M3Net.

\noindent\textbf{Out-of-Distribution Generalization.}
Evaluating the generalization ability of models on out-of-training-distribution data is crucial, particularly in safety-critical fields like autonomous driving \cite{kong2023robodepth,xie2024robobev,xie2023robobev}. In this context, we resort to the two corruption datasets from the Robo3D~\cite{kong2023robo3D} benchmark, \ie, \textit{SemanticKITTI-C} and \textit{nuScenes-C}, to conduct our assessment. From the last four columns of \cref{tab:generalization}, we observe that M3Net achieves better results than the na\"{\i}ve joint training and other single-dataset approaches, proving the strong generalizability of the learned representations.
\vspace{0.09cm}
\section{Conclusion}
\label{sec:conclusion}
In this work, we presented M3Net, a universal framework capable of fulfilling multi-task, multi-dataset, multi-modality LiDAR segmentation using a single set of parameters. Through extensive analyses, we validated the effectiveness of applying data-, feature-, and label-space alignments to handle such a challenging task. In addition, our comprehensive analysis and discourse have delved into the fundamental challenges of acquiring the general knowledge for scalable 3D perception, which holds substantial potential to propel further research in this domain. Our future strides focus on combining more data resources to further enhance the alignments and adaptations in our framework.

\vspace{0.2cm}
{\small\noindent\textbf{Acknowledgements}.
This work was partially supported by NSFC (No.62206173) and MoE Key Laboratory of Intelligent Perception and Human-Machine Collaboration (ShanghaiTech University). This work was also supported by the Ministry of Education, Singapore, under its MOE AcRF Tier 2 (MOET2EP20221- 0012), NTU NAP, and under the RIE2020 Industry Alignment Fund – Industry Collaboration Projects (IAF-ICP) Funding Initiative, as well as cash and in-kind contribution from the industry partner(s).}

\clearpage
\appendix
\section*{Appendix}
\startcontents[appendices]
\printcontents[appendices]{l}{1}{\setcounter{tocdepth}{3}}

\begin{table*}[t]
\caption{\textbf{Summary of the datasets} used in this work. We split different datasets into three categories: \textbf{i)} The \textit{nuScenes} \cite{fong2022panoptic-nuScenes}, \textit{SemanticKITTI} \cite{behley2019semanticKITTI}, and \textit{Waymo Open} \cite{sun2020waymoOpen} datasets are used for multi-dataset training and evaluations. \textbf{ii)} The \textit{RELLIS-3D} \cite{jiang2021rellis3D}, \textit{SemanticPOSS} \cite{pan2020semanticPOSS}, SemanticSTF \cite{xiao2023semanticSTF}, SynLiDAR \cite{xiao2022synLiDAR}, and DAPS-3D \cite{klokov2023daps3D} datasets are used for knowledge transfer and generalization (\textit{w/} fine-tuning). \textbf{iii)} The SemanticKITTI-C \cite{kong2023robo3D} and nuScenes-C \cite{kong2023robo3D} datasets are used for out-of-distribution generalization (\textit{w/o} fine-tuning).}
\vspace{-0.1cm}
\centering
\begin{adjustbox}{width=\textwidth}
\begin{tabular}{c|c|c|c|c}
\toprule
\multicolumn{5}{c}{\textbf{Dataset Summary}}
\\\midrule\midrule
\begin{minipage}[b]
{0.4\columnwidth}\centering\raisebox{-.3\height}{\includegraphics[width=\linewidth]{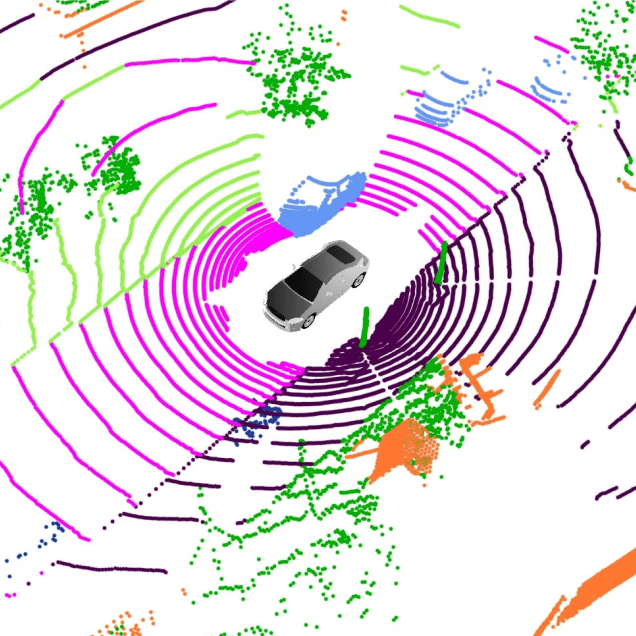}}\end{minipage} & \begin{minipage}[b]
{0.4\columnwidth}\centering\raisebox{-.3\height}{\includegraphics[width=\linewidth]{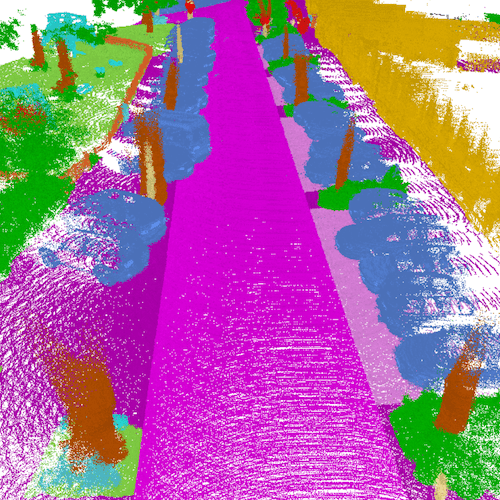}}\end{minipage} & \begin{minipage}[b]
{0.4\columnwidth}\centering\raisebox{-.3\height}{\includegraphics[width=\linewidth]{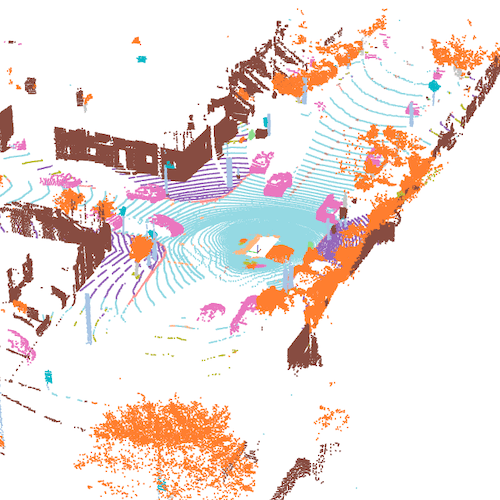}}\end{minipage} & \begin{minipage}[b]
{0.4\columnwidth}\centering\raisebox{-.3\height}{\includegraphics[width=\linewidth]{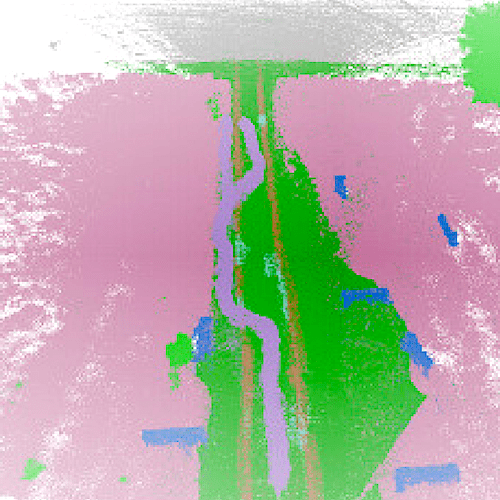}}\end{minipage} & \begin{minipage}[b]
{0.4\columnwidth}\centering\raisebox{-.3\height}{\includegraphics[width=\linewidth]{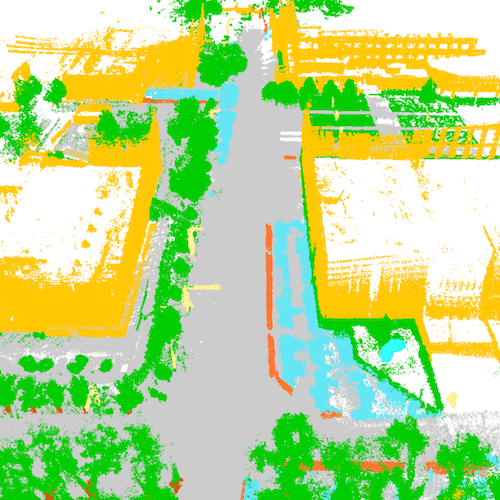}}\end{minipage}
\\\midrule
nuScenes \cite{fong2022panoptic-nuScenes} & SemanticKITTI \cite{behley2019semanticKITTI} & Waymo Open \cite{sun2020waymoOpen} & RELLIS-3D \cite{jiang2021rellis3D} & SemanticPOSS \cite{pan2020semanticPOSS}
\\
$[$\href{https://www.nuscenes.org/nuscenes}{Link}$]$ & $[$\href{http://semantic-kitti.org/}{Link}$]$ & $[$\href{https://waymo.com/open}{Link}$]$ & $[$\href{http://www.unmannedlab.org/research/RELLIS-3D}{Link}$]$ & $[$\href{http://www.poss.pku.edu.cn/semanticposs.html}{Link}$]$
\\\midrule\midrule
\begin{minipage}[b]
{0.4\columnwidth}\centering\raisebox{-.3\height}{\includegraphics[width=\linewidth]{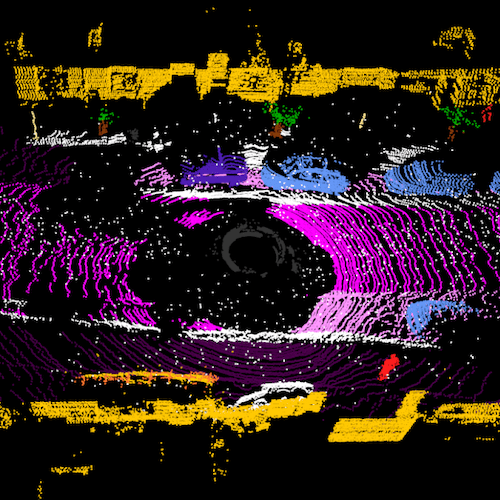}}\end{minipage} & \begin{minipage}[b]
{0.4\columnwidth}\centering\raisebox{-.3\height}{\includegraphics[width=\linewidth]{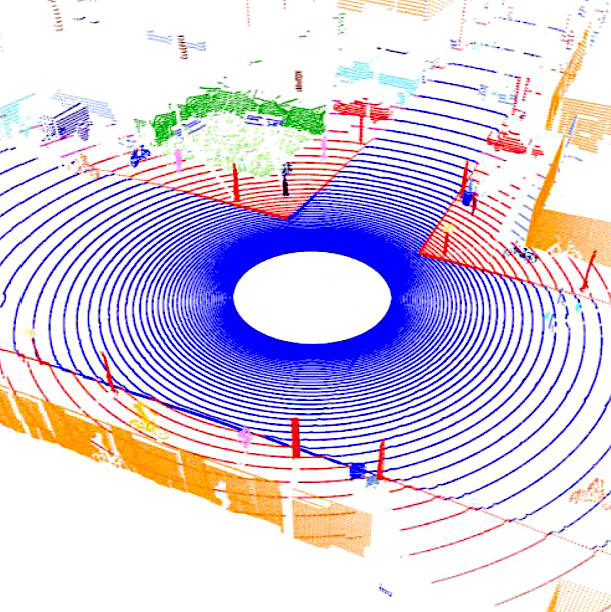}}\end{minipage} & \begin{minipage}[b]
{0.4\columnwidth}\centering\raisebox{-.3\height}{\includegraphics[width=\linewidth]{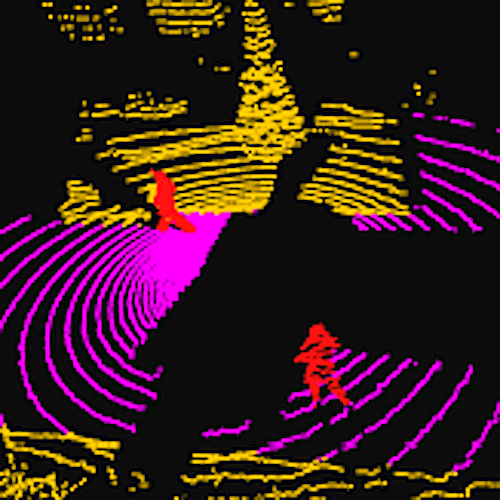}}\end{minipage} & \begin{minipage}[b]
{0.4\columnwidth}\centering\raisebox{-.3\height}{\includegraphics[width=\linewidth]{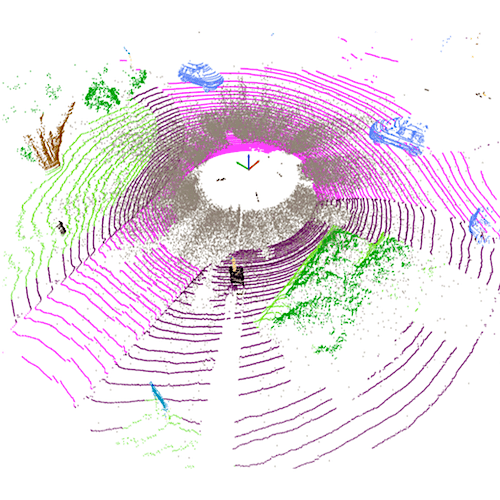}}\end{minipage} & \begin{minipage}[b]
{0.4\columnwidth}\centering\raisebox{-.3\height}{\includegraphics[width=\linewidth]{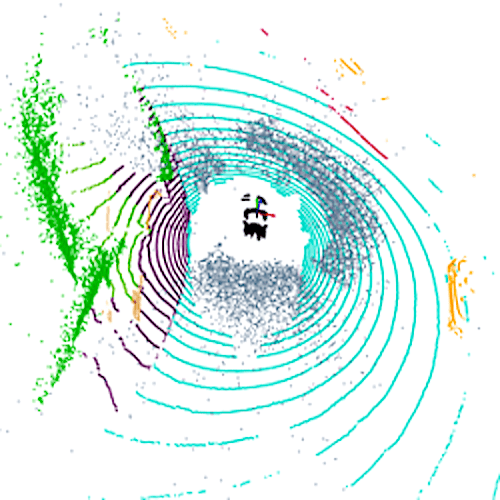}}\end{minipage}
\\\midrule
SemanticSTF \cite{xiao2023semanticSTF} & SynLiDAR \cite{xiao2022synLiDAR} & DAPS-3D \cite{klokov2023daps3D} & SemanticKITTI-C \cite{kong2023robo3D} & nuScenes-C \cite{kong2023robo3D}
\\
$[$\href{https://github.com/xiaoaoran/SemanticSTF}{Link}$]$ & $[$\href{https://github.com/xiaoaoran/SynLiDAR}{Link}$]$ & $[$\href{https://github.com/subake/DAPS3D}{Link}$]$ & $[$\href{https://github.com/ldkong1205/Robo3D}{Link}$]$ & $[$\href{https://github.com/ldkong1205/Robo3D}{Link}$]$
\\\bottomrule
\end{tabular}
\end{adjustbox}
\label{tab:dataset_summary}
\end{table*}

\vspace{0.2cm}

\section{Multi-Dataset Configuration}

In this section, we elaborate on the details of combining multiple heterogeneous LiDAR segmentation datasets to train a universal LiDAR segmentation model. 

\subsection{Overview}

In this work, we resort to ten driving datasets for achieving \textit{i)} multi-dataset training and evaluations, \textit{ii)} knowledge transfer and generalization, and \textit{iii)} out-of-distribution generalization. A summary of the datasets used in this work is shown in \cref{tab:dataset_summary}.
For multi-dataset training and evaluations, we use the LiDAR and camera data from the \textit{{nuScenes}} \cite{caesar2020nuScenes,fong2022panoptic-nuScenes}, \textit{{SemanticKITTI}} \cite{behley2019semanticKITTI}, and \textit{{Waymo Open}} \cite{sun2020waymoOpen} datasets.

\begin{itemize}
    \item \textbf{nuScenes} is a large-scale public dataset for autonomous driving, created by Motional (formerly nuTonomy). It is widely used in the research and development of autonomous vehicles and related technologies. The dataset includes a comprehensive range of sensor data crucial for autonomous driving. It typically contains data from multiple cameras, LiDAR, RADAR, GPS, IMU, and other sensors. This multimodal data collection is essential for developing and testing algorithms for perception, prediction, and motion planning in autonomous vehicles. One of the strengths of the nuScenes dataset is its diversity. The data encompasses various driving conditions, including different times of day, weather conditions, and urban environments. This diversity is crucial for training robust algorithms that can handle real-world driving scenarios. In this work, we use the LiDAR semantic and panoptic segmentation data from the \textit{lidarseg}\footnote{\url{https://www.nuscenes.org/lidar-segmentation}.} subset in the nuScenes dataset, which includes segmentation labels for the entire nuScenes dataset, encompassing thousands of scenes, each a $20$-second clip captured from a driving vehicle in various urban settings. $32$ classes are manually labeled, covering a wide range of objects and elements in urban scenes, where $16$ of them are typically adopted in evaluating the segmentation performance.  More details of this dataset can be found at \url{https://www.nuscenes.org/nuscenes}.
    
    \item \textbf{SemanticKITTI} is a well-known dataset in the field of autonomous driving and robotics, specifically tailored for the task of semantic and panoptic segmentation using LiDAR point clouds. It is an extension of the original KITTI Vision Benchmark Suite\footnote{\url{https://www.cvlibs.net/datasets/kitti}.} \cite{geiger2012kitti}, with annotations for over $20$ sequences of driving scenarios, each containing tens of thousands of LiDAR scans. The dataset covers a variety of urban and rural scenes. This includes city streets, residential areas, highways, and country roads, providing a diverse set of environments for testing algorithms. The dataset provides labels for $28$ different semantic classes, including cars, pedestrians, bicycles, various types of vegetation, buildings, roads, and so on. $19$ classes are typically adopted for evaluation. In total, around $4549$ million points are annotated, and such extensive labeling provides a dense coverage for each LiDAR scan. More details of this dataset can be found at \url{http://semantic-kitti.org}.

    \item \textbf{Waymo Open} is a large dataset for autonomous driving, provided by Waymo LLC, a company that specializes in the development of self-driving technology. This dataset is particularly notable for its comprehensive coverage of various scenarios encountered in autonomous driving. The data is collected using Waymo's self-driving vehicles, which are equipped with an array of sensors, including high-resolution LiDARs, cameras, and radars. This multimodal data collection allows for comprehensive perception modeling. The dataset includes a wide range of driving environments and conditions, such as city streets, highways, and suburban areas, captured at different times of day and in various weather conditions. This variety is crucial for developing robust autonomous driving systems. In this work, we use its 3D Semantic Segmentation subset, which specifically provides point-level annotations for 3D point clouds generated by LiDAR sensors. $22$ semantic classes are used during evaluation, encompassing a wide range of object classes, such as vehicles, pedestrians, and cyclists, as well as static objects like road signs, buildings, and vegetation. More details of this dataset can be found at \url{https://waymo.com/open}.
\end{itemize}

To validate that the learned features from our multi-dataset training setup are superior to that of the singe-dataset training in knowledge transfer and generalization, we conduct fine-tuning experiments on the following five datasets: \textit{{RELLIS-3D}} \cite{jiang2021rellis3D}, \textit{{SemanticPOSS}} \cite{pan2020semanticPOSS}, \textit{{SemanticSTF}} \cite{xiao2023semanticSTF}, \textit{{SynLiDAR}} \cite{xiao2022synLiDAR}, and \textit{{DAPS-3D}} \cite{klokov2023daps3D}.

\begin{itemize}
    \item \textbf{RELLIS-3D} is a dataset focusing on off-road environments for autonomous navigation and perception, developed by Texas A\&M University. It contains multimodal sensor data, including high-resolution LiDAR, RGB imagery, and GPS/IMU data, providing a comprehensive set for developing and evaluating algorithms for off-road autonomous driving. The dataset features diverse terrain types, such as grasslands, forests, and trails, offering unique challenges compared to urban scenarios. RELLIS-3D includes annotations for $13$ semantic classes, including natural elements and man-made objects, crucial for navigation in off-road settings. More details of this dataset can be found at \url{http://www.unmannedlab.org/research/RELLIS-3D}.

    \item \textbf{SemanticPOSS} focuses on panoramic LiDAR scans, which include urban scenes, highways, and rural areas. The dataset contains annotations for $14$ semantic classes, covering vehicles, pedestrians, cyclists, and various road elements. Its panoramic view provides a $360$-degree understanding of the vehicle's surroundings, which is beneficial for comprehensive scene analysis. More details of this dataset can be found at \url{https://www.poss.pku.edu.cn/semanticposs}.

    \item \textbf{SemanticSTF} studies the 3D semantic segmentation of LiDAR point clouds under adverse weather conditions, including snow, rain, and fog. It is built from the real-world STF \cite{bijelic2020stf} dataset with point-wise annotations of $21$ semantic categories. The original LiDAR data in STF was captured by a Velodyne HDL64 S3D LiDAR sensor. In total, SemanticSTF selected $2076$ scans for dense annotations, including $694$ snowy, $637$ dense-foggy, $631$ light-foggy, and $114$ rainy scans. More details of this dataset can be found at \url{https://github.com/xiaoaoran/SemanticSTF}.

    \item \textbf{SynLiDAR} is a synthetic dataset for LiDAR-based semantic segmentation. It is generated using advanced simulation techniques to create realistic urban, suburban, and rural environments. SynLiDAR offers an extensive range of annotations for a variety of classes, including dynamic objects like vehicles and pedestrians, as well as static objects like buildings and trees. This dataset is useful for algorithm development and testing in simulated environments where real-world data collection is challenging. More details of this dataset can be found at \url{https://github.com/xiaoaoran/SynLiDAR}.

    \item \textbf{DAPS-3D} is a dataset focusing on dynamic and static point cloud segmentation. It includes LiDAR scans from diverse urban environments, providing detailed annotations for dynamic objects such as vehicles, pedestrians, and cyclists, as well as static objects like buildings, roads, and vegetation. DAPS-3D is designed to advance research in dynamic scene understanding and prediction in autonomous driving, addressing the challenges posed by moving objects in complex urban settings. More details of this dataset can be found at \url{https://github.com/subake/DAPS3D}.
\end{itemize}

\begin{table*}[t]
\caption{\textbf{The statistical analysis} of the $16$ semantic classes in the \textit{\textbf{nuScenes}}~\cite{fong2022panoptic-nuScenes} dataset. Statistics are calculated from the \textit{training} split of the dataset. Each violin plot shows the LiDAR point cloud density distribution in a $50$ meters range. Best viewed in colors.}
\vspace{-0.1cm}
\centering
\begin{adjustbox}{width=\textwidth}
\begin{tabular}{c|c|c|c}
\toprule
\multicolumn{4}{c}{\textbf{nuScenes (16 classes)}}
\\\midrule\midrule
\begin{minipage}[b]
{0.49\columnwidth}\centering\raisebox{-.4\height}{\includegraphics[width=\linewidth]{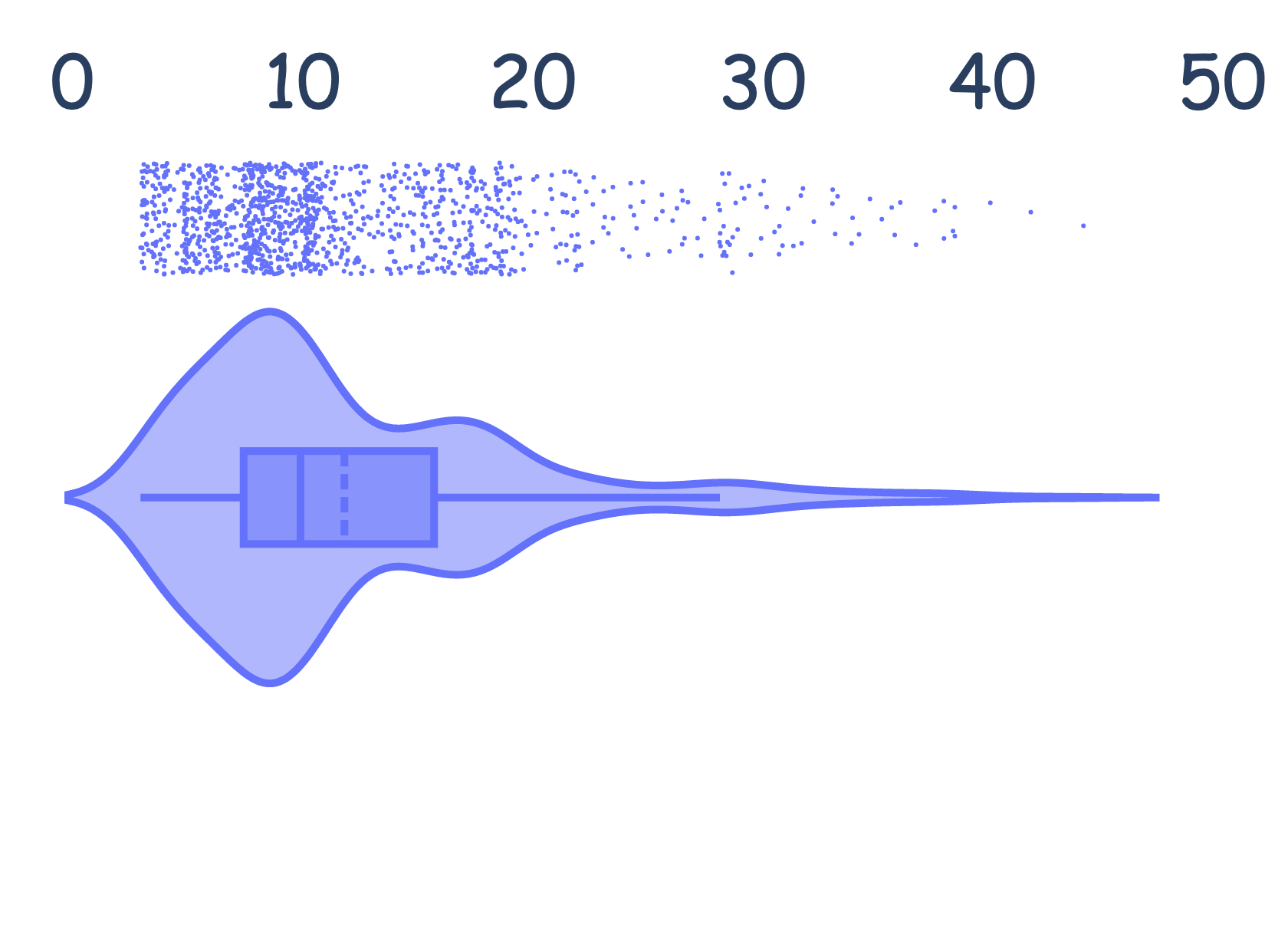}}\end{minipage} & \begin{minipage}[b]
{0.49\columnwidth}\centering\raisebox{-.4\height}{\includegraphics[width=\linewidth]{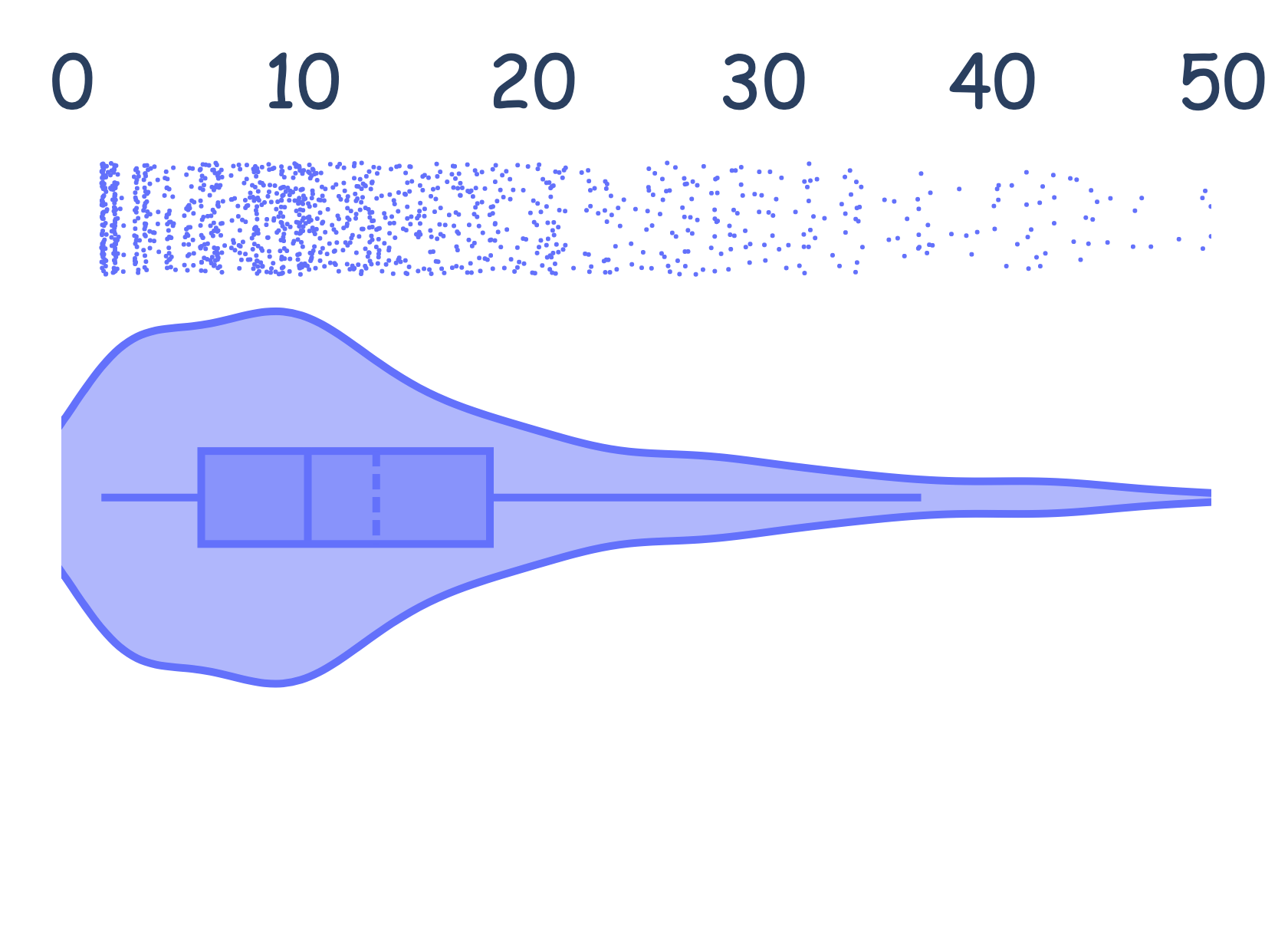}}\end{minipage} & \begin{minipage}[b]
{0.49\columnwidth}\centering\raisebox{-.4\height}{\includegraphics[width=\linewidth]{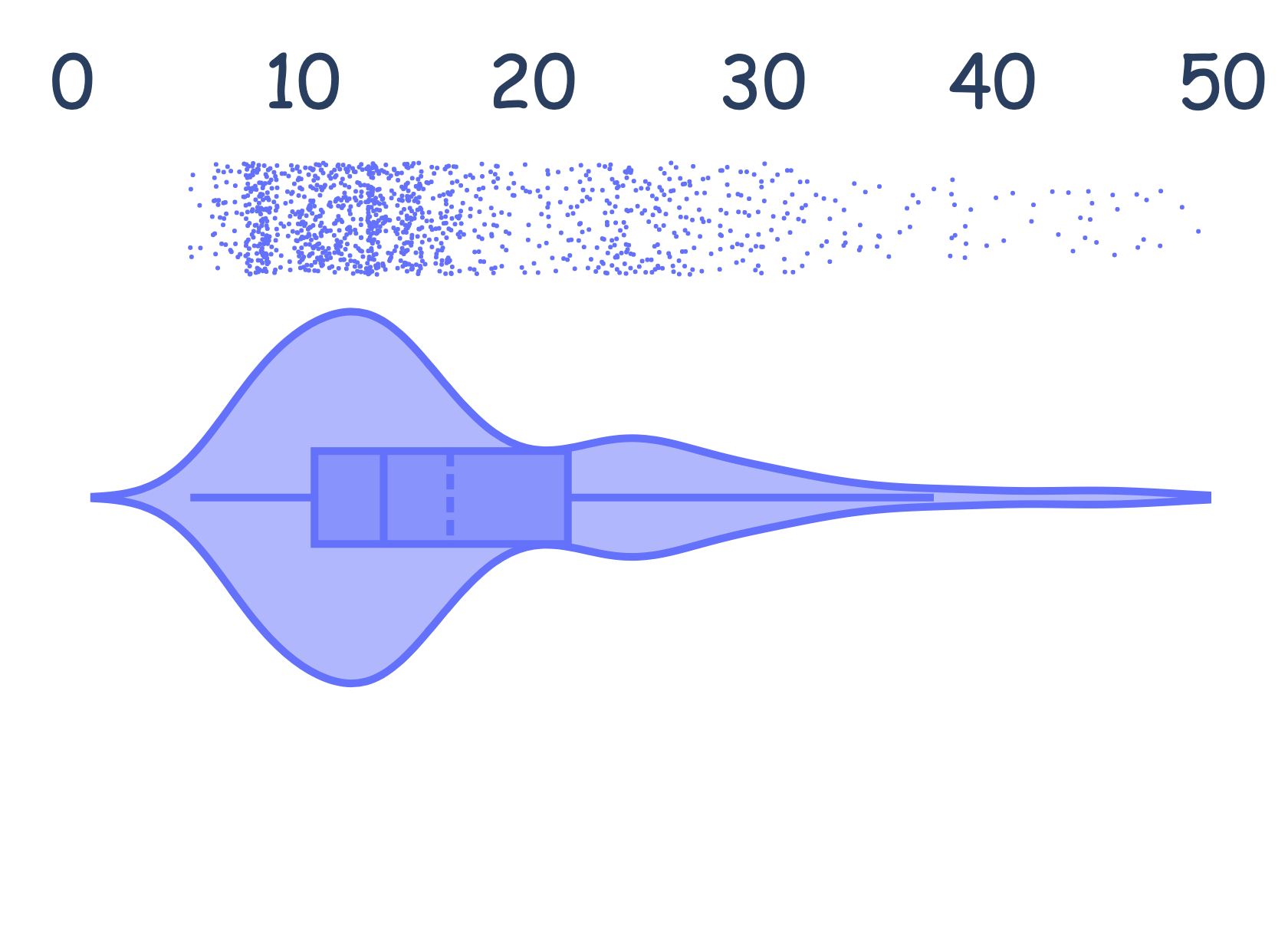}}\end{minipage} & \begin{minipage}[b]
{0.49\columnwidth}\centering\raisebox{-.4\height}{\includegraphics[width=\linewidth]{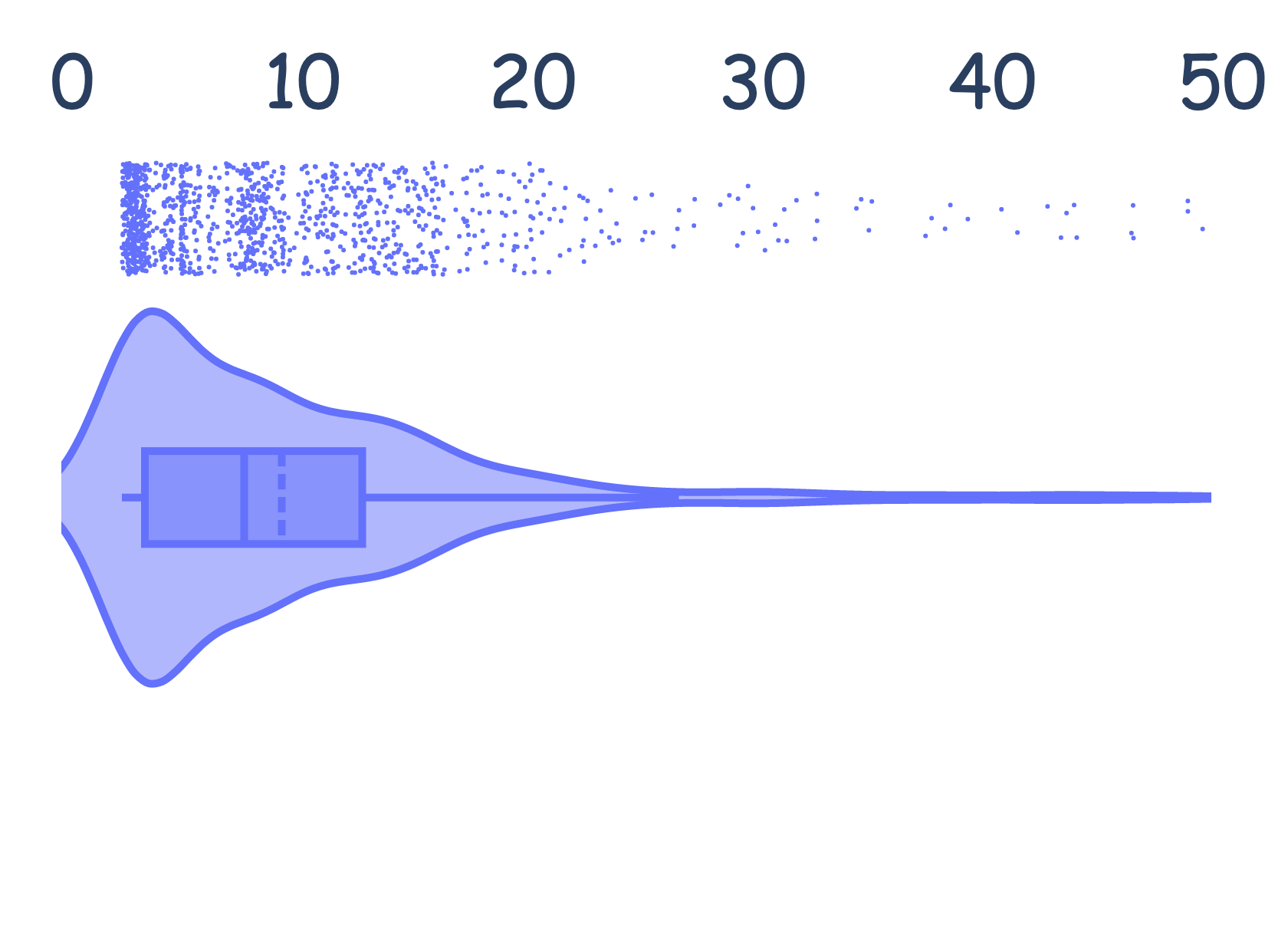}}\end{minipage}
\\
barrier & bicycle & bus & car 
\\\midrule
\begin{minipage}[b]
{0.49\columnwidth}\centering\raisebox{-.4\height}{\includegraphics[width=\linewidth]{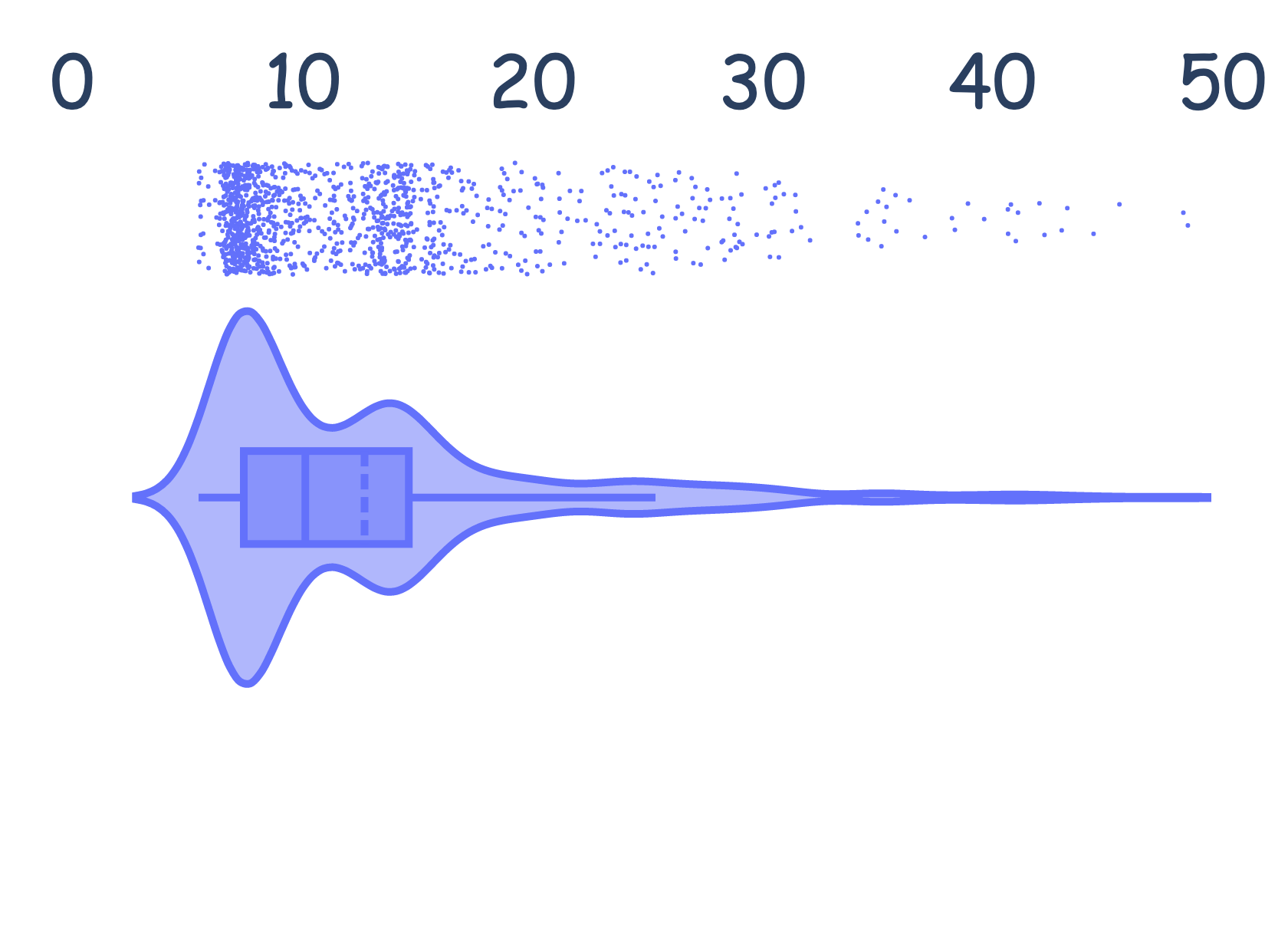}}\end{minipage} & \begin{minipage}[b]
{0.49\columnwidth}\centering\raisebox{-.4\height}{\includegraphics[width=\linewidth]{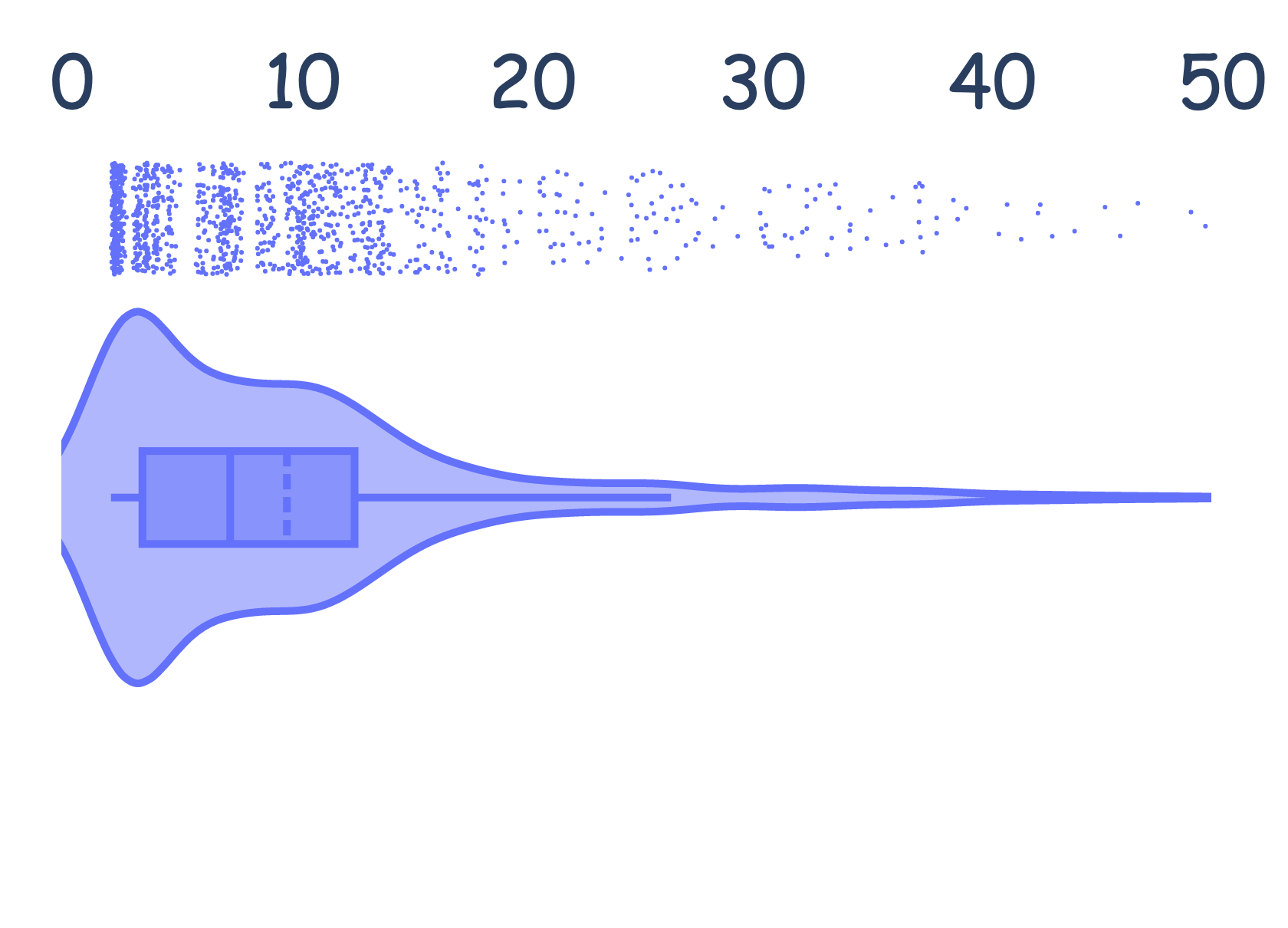}}\end{minipage} & \begin{minipage}[b]
{0.49\columnwidth}\centering\raisebox{-.4\height}{\includegraphics[width=\linewidth]{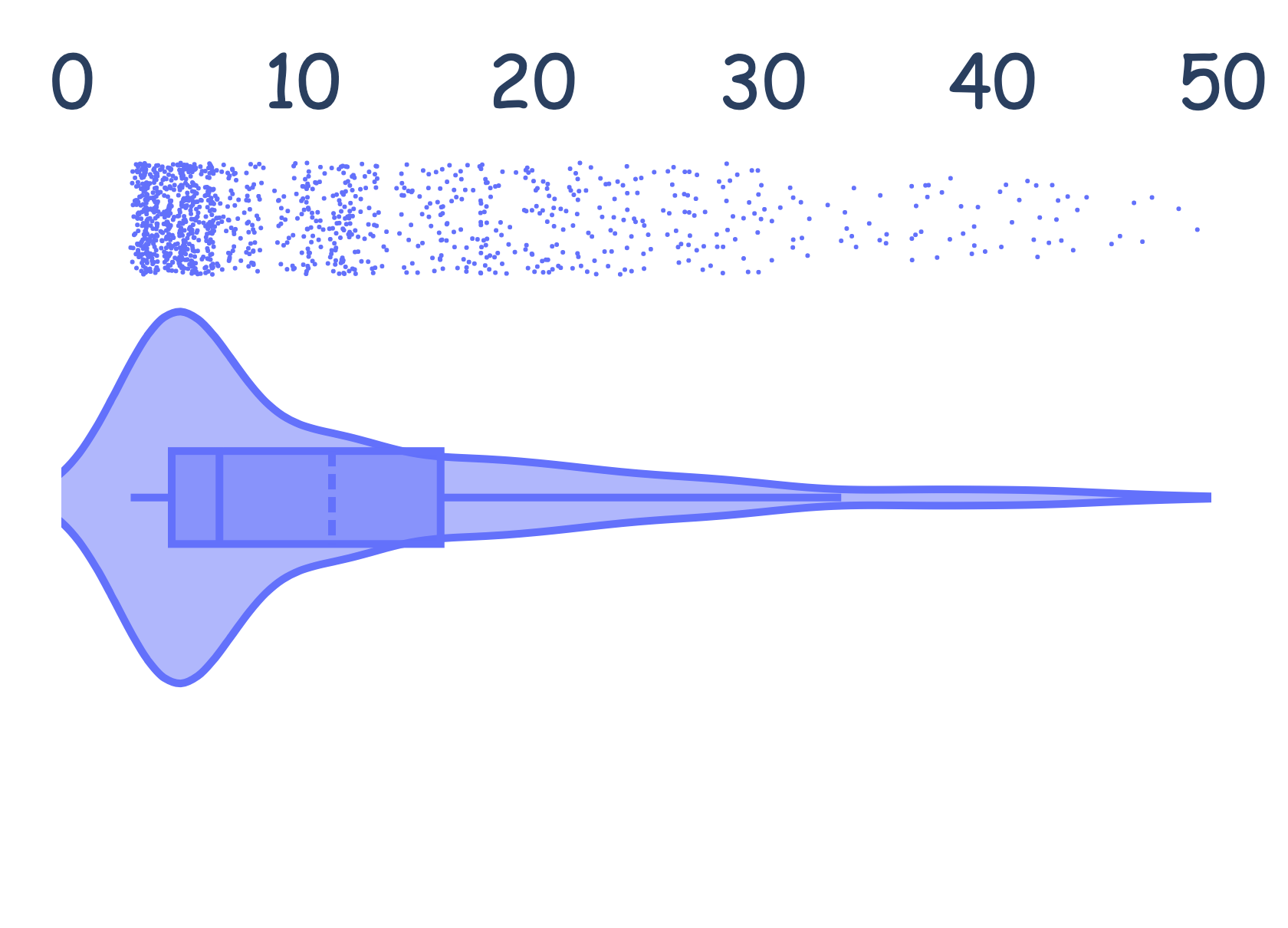}}\end{minipage} & \begin{minipage}[b]
{0.49\columnwidth}\centering\raisebox{-.4\height}{\includegraphics[width=\linewidth]{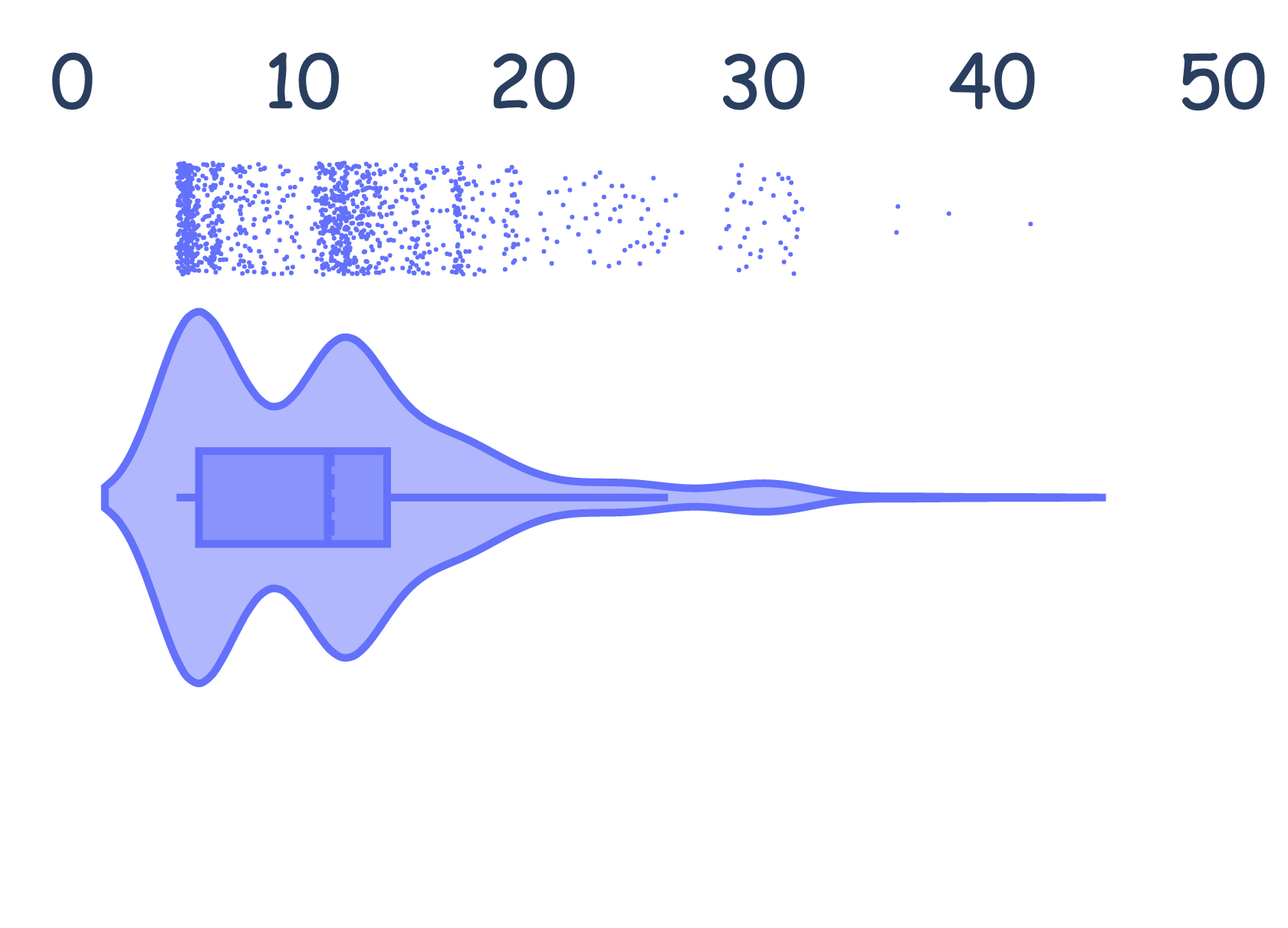}}\end{minipage}
\\
construction-vehicle & motorcycle & pedestrian & traffic-cone
\\\midrule
\begin{minipage}[b]
{0.49\columnwidth}\centering\raisebox{-.4\height}{\includegraphics[width=\linewidth]{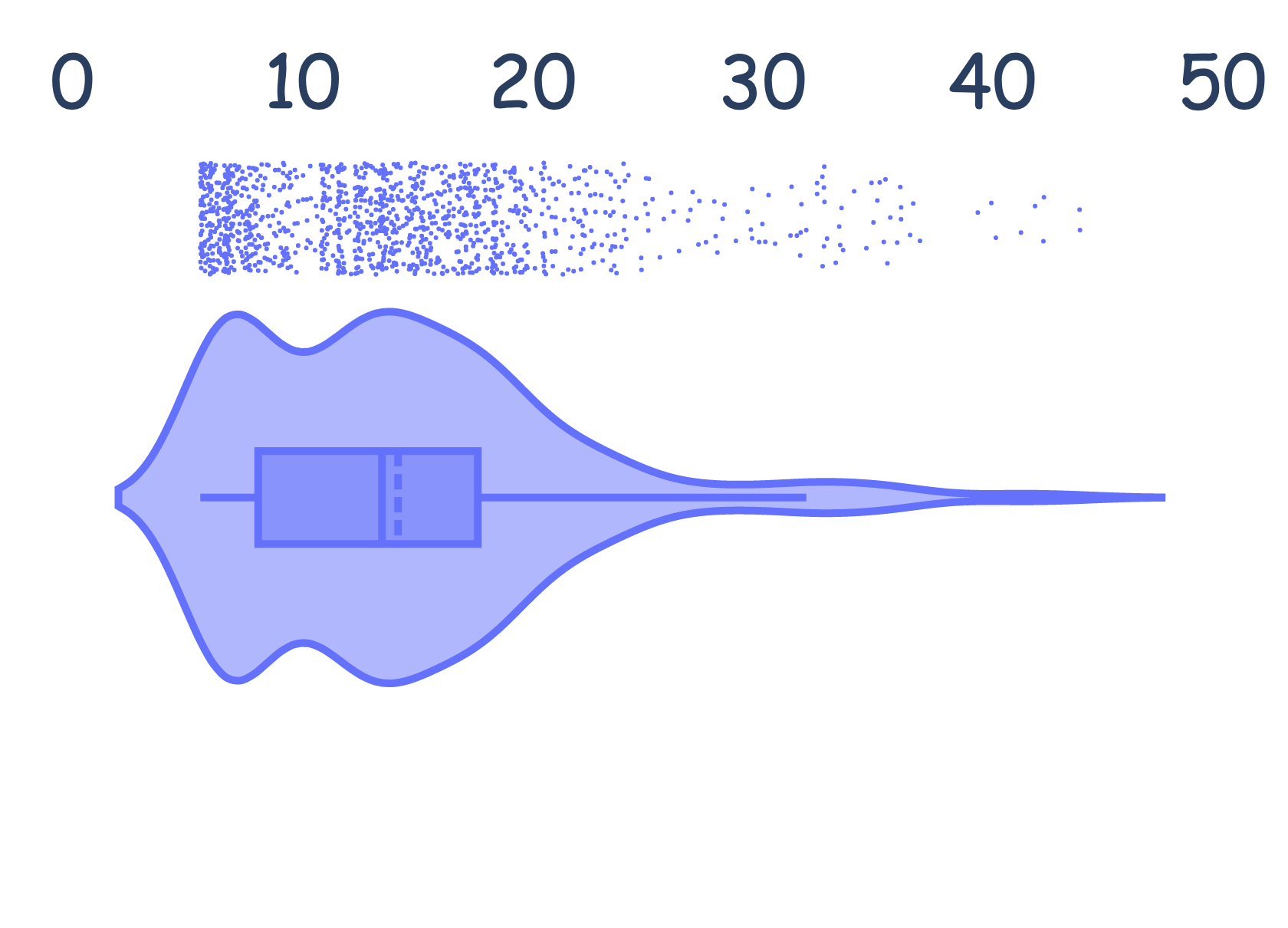}}\end{minipage} & \begin{minipage}[b]
{0.49\columnwidth}\centering\raisebox{-.4\height}{\includegraphics[width=\linewidth]{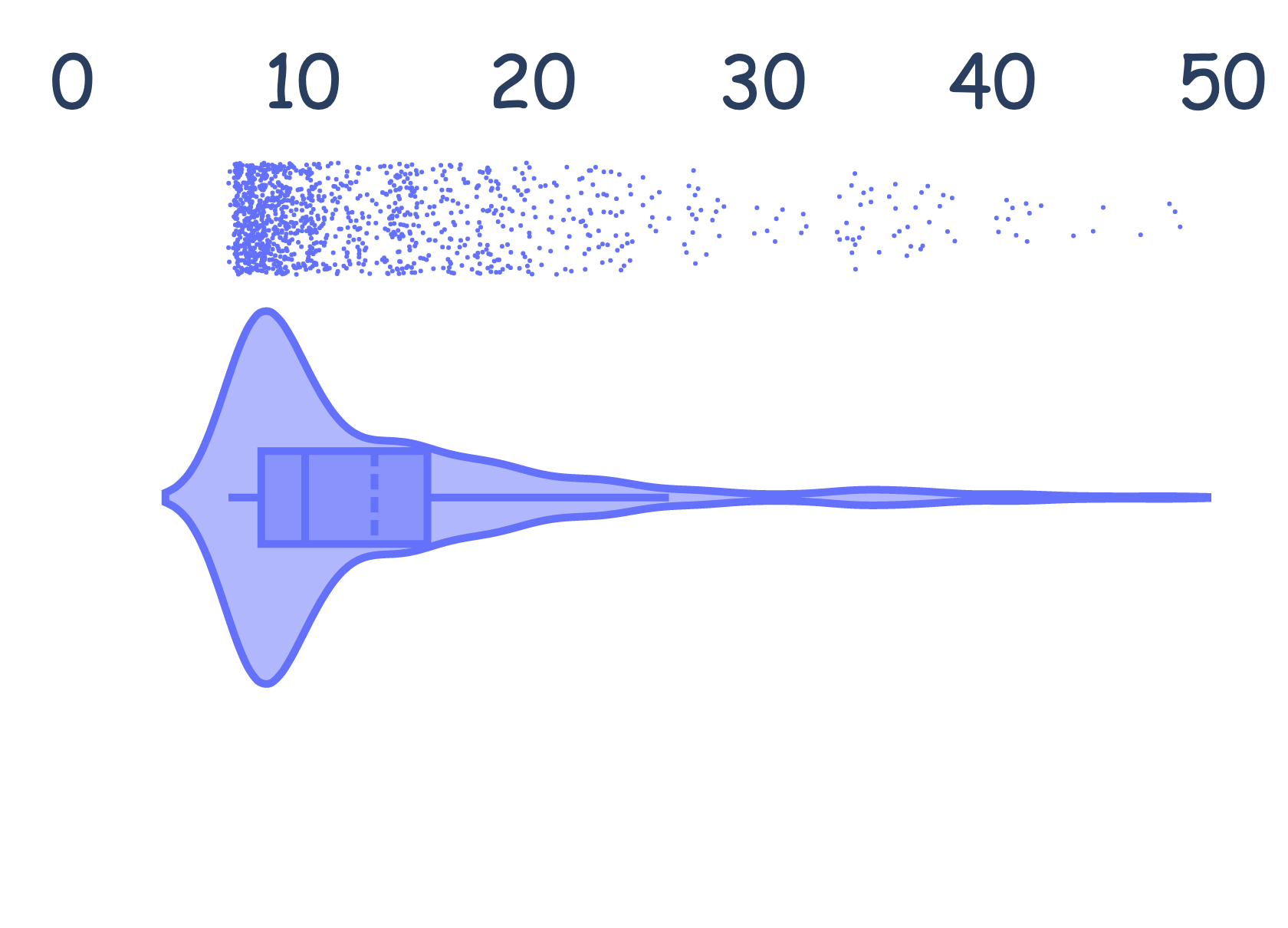}}\end{minipage} & \begin{minipage}[b]
{0.49\columnwidth}\centering\raisebox{-.4\height}{\includegraphics[width=\linewidth]{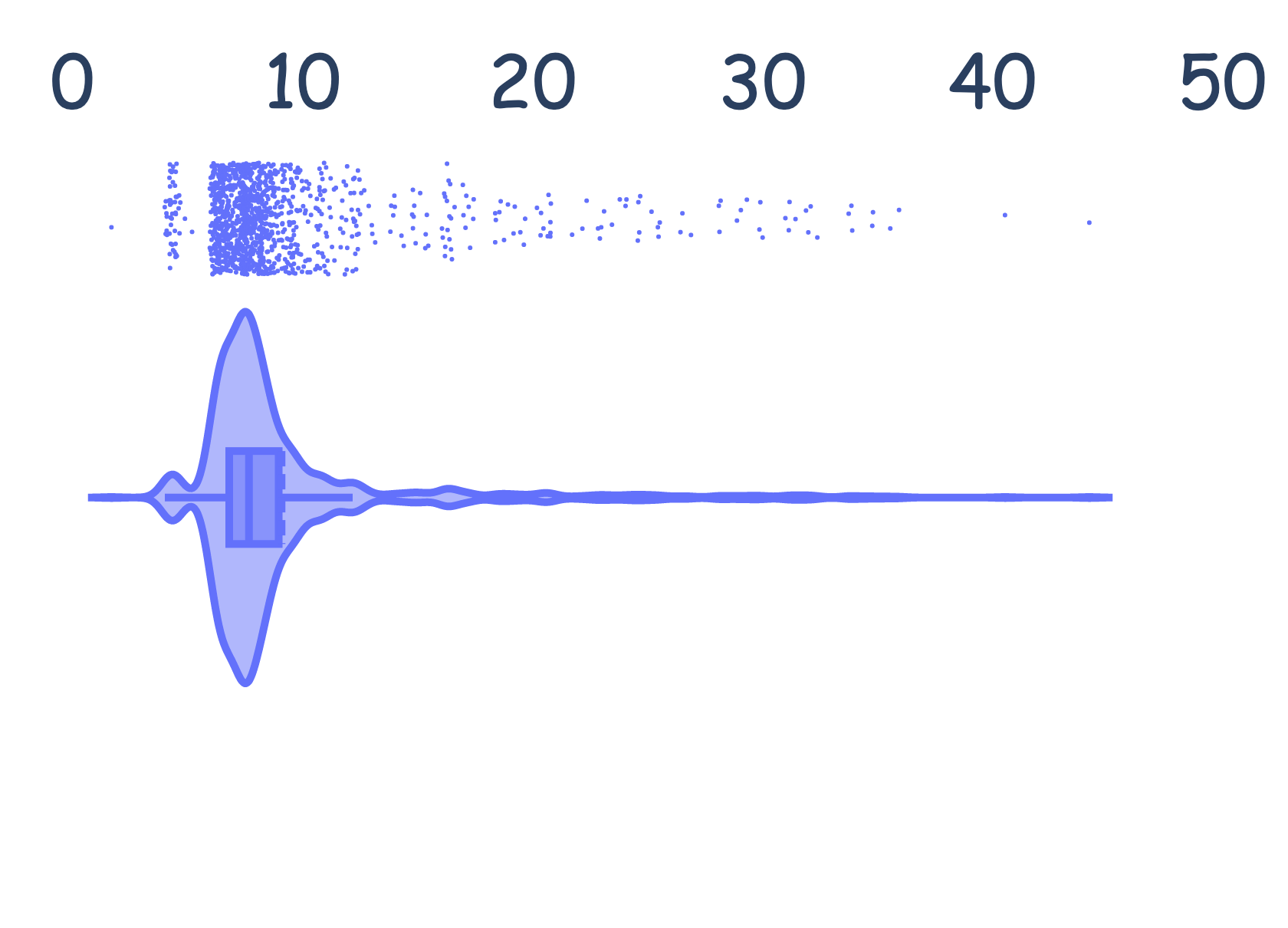}}\end{minipage} & \begin{minipage}[b]
{0.49\columnwidth}\centering\raisebox{-.4\height}{\includegraphics[width=\linewidth]{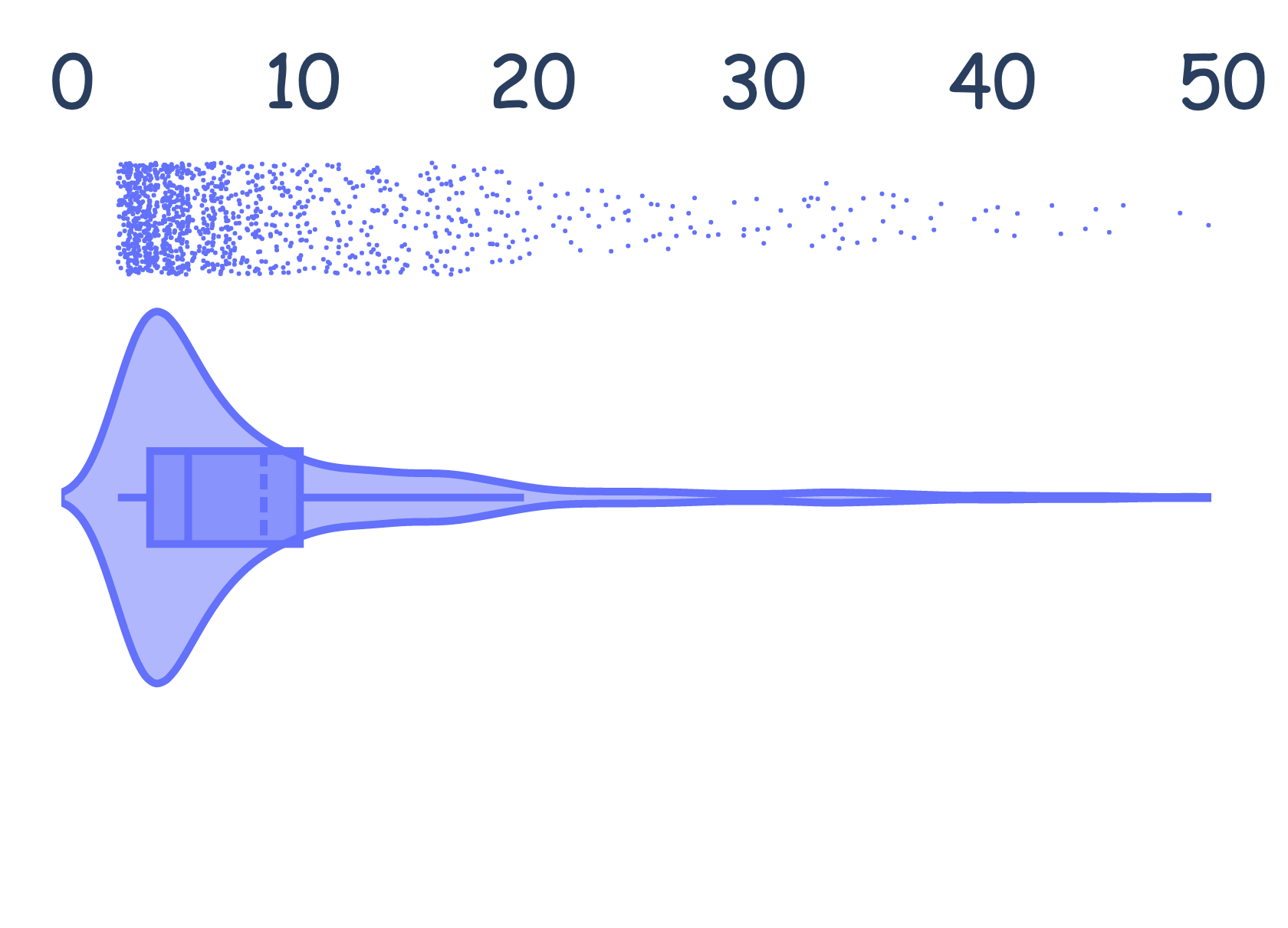}}\end{minipage}
\\
trailer & truck & driveable-surface & other-flat
\\\midrule
\begin{minipage}[b]
{0.49\columnwidth}\centering\raisebox{-.4\height}{\includegraphics[width=\linewidth]{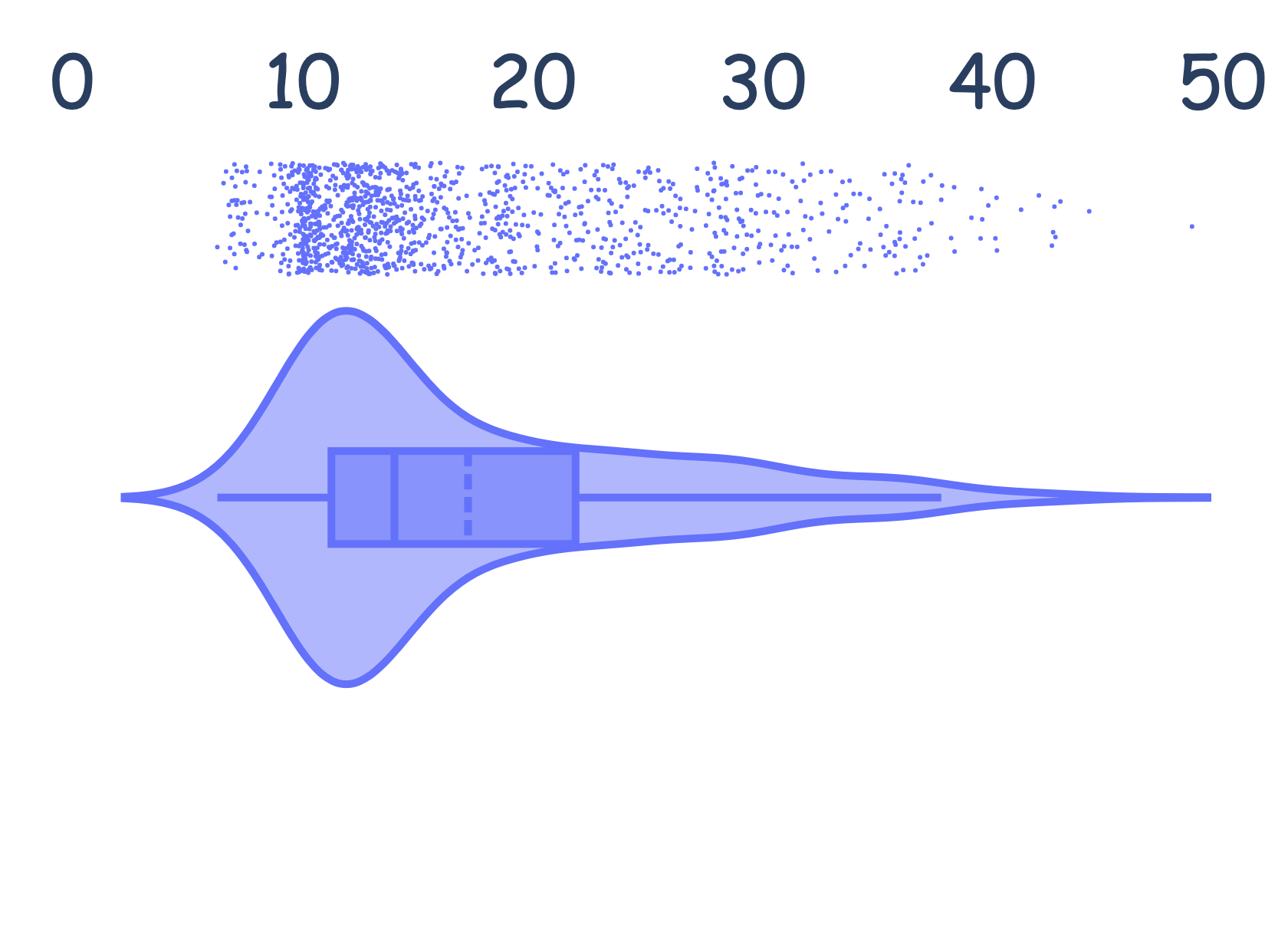}}\end{minipage} & \begin{minipage}[b]
{0.49\columnwidth}\centering\raisebox{-.4\height}{\includegraphics[width=\linewidth]{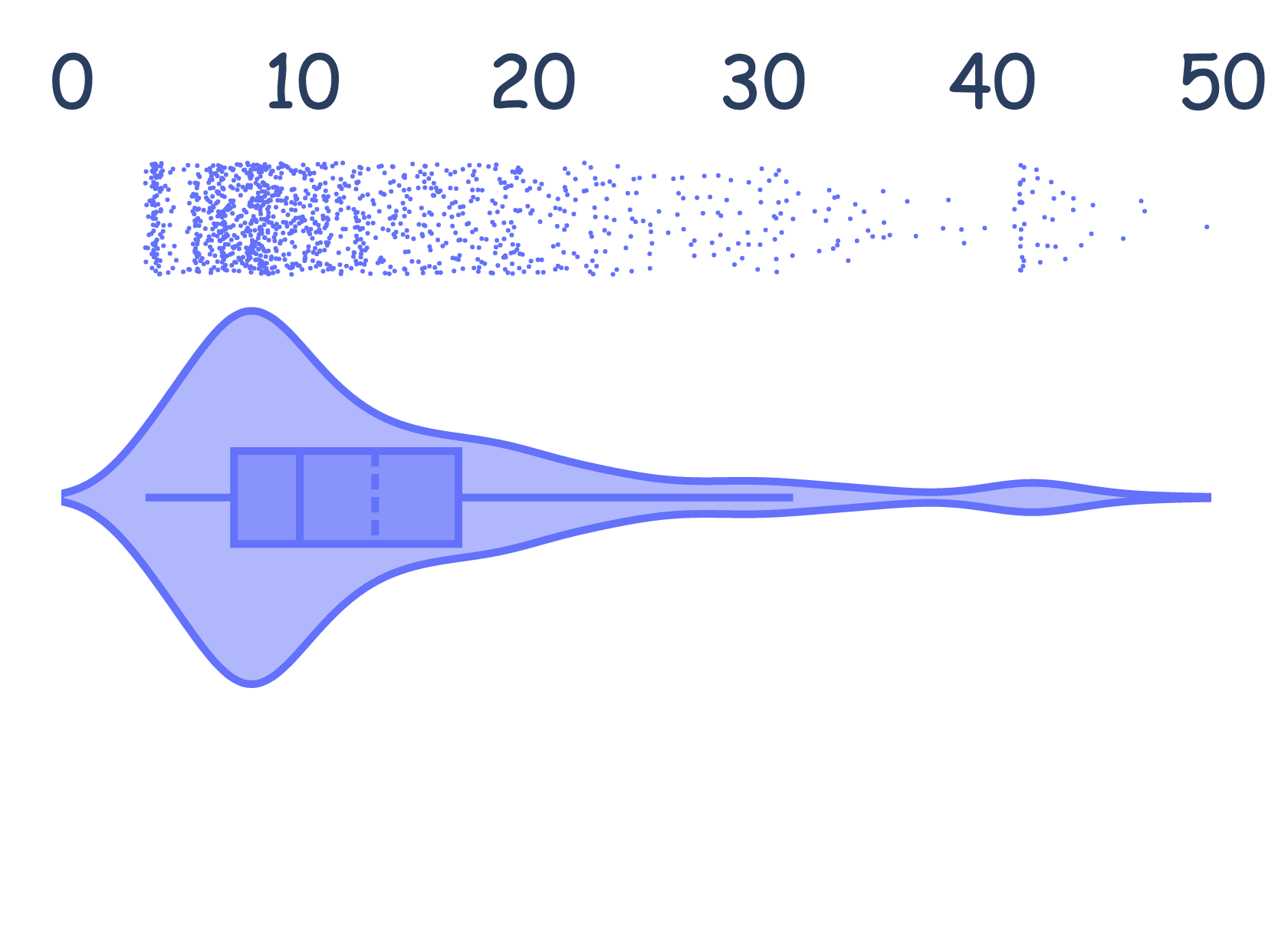}}\end{minipage} & \begin{minipage}[b]
{0.49\columnwidth}\centering\raisebox{-.4\height}{\includegraphics[width=\linewidth]{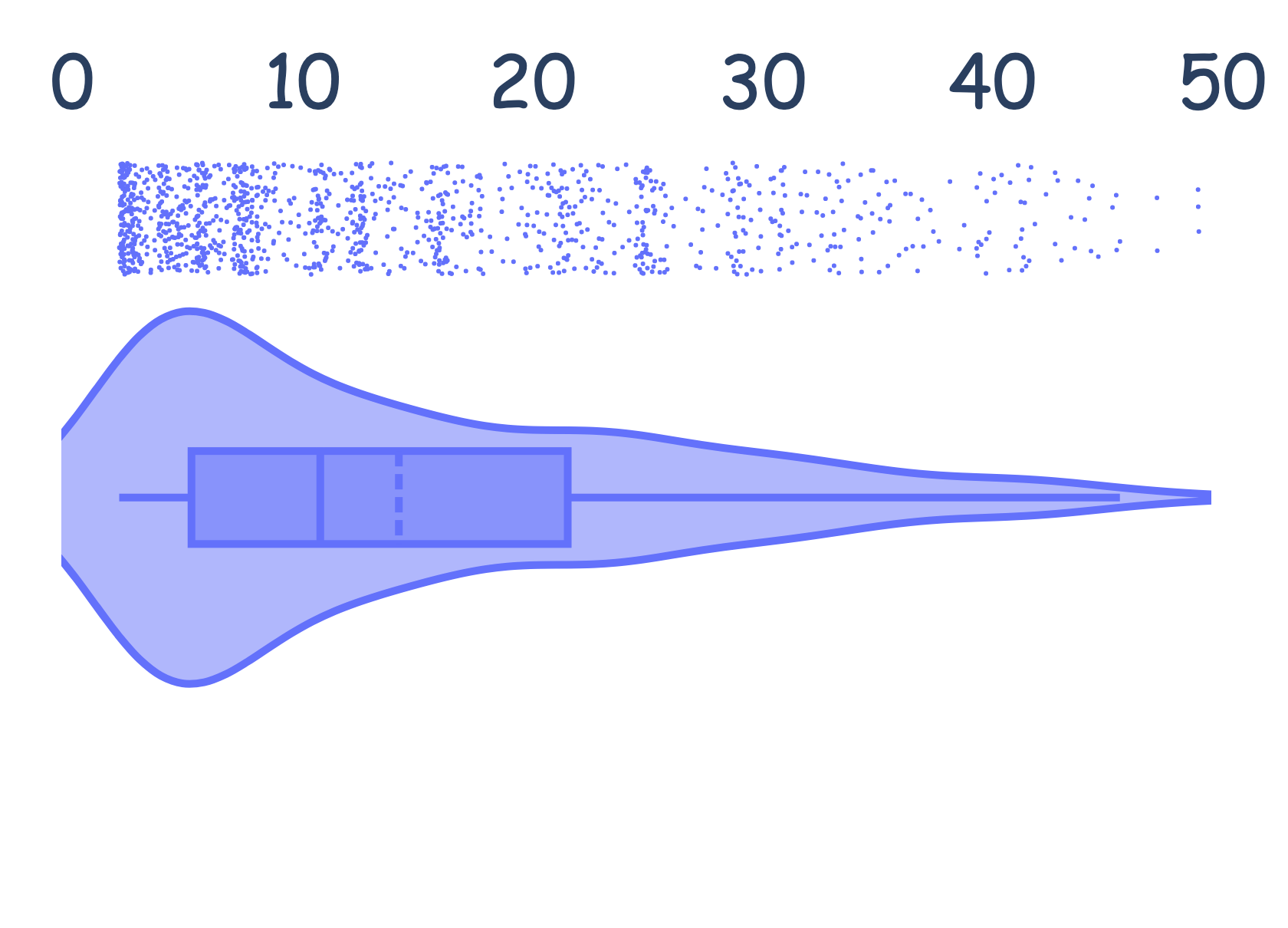}}\end{minipage} & \begin{minipage}[b]
{0.49\columnwidth}\centering\raisebox{-.4\height}{\includegraphics[width=\linewidth]{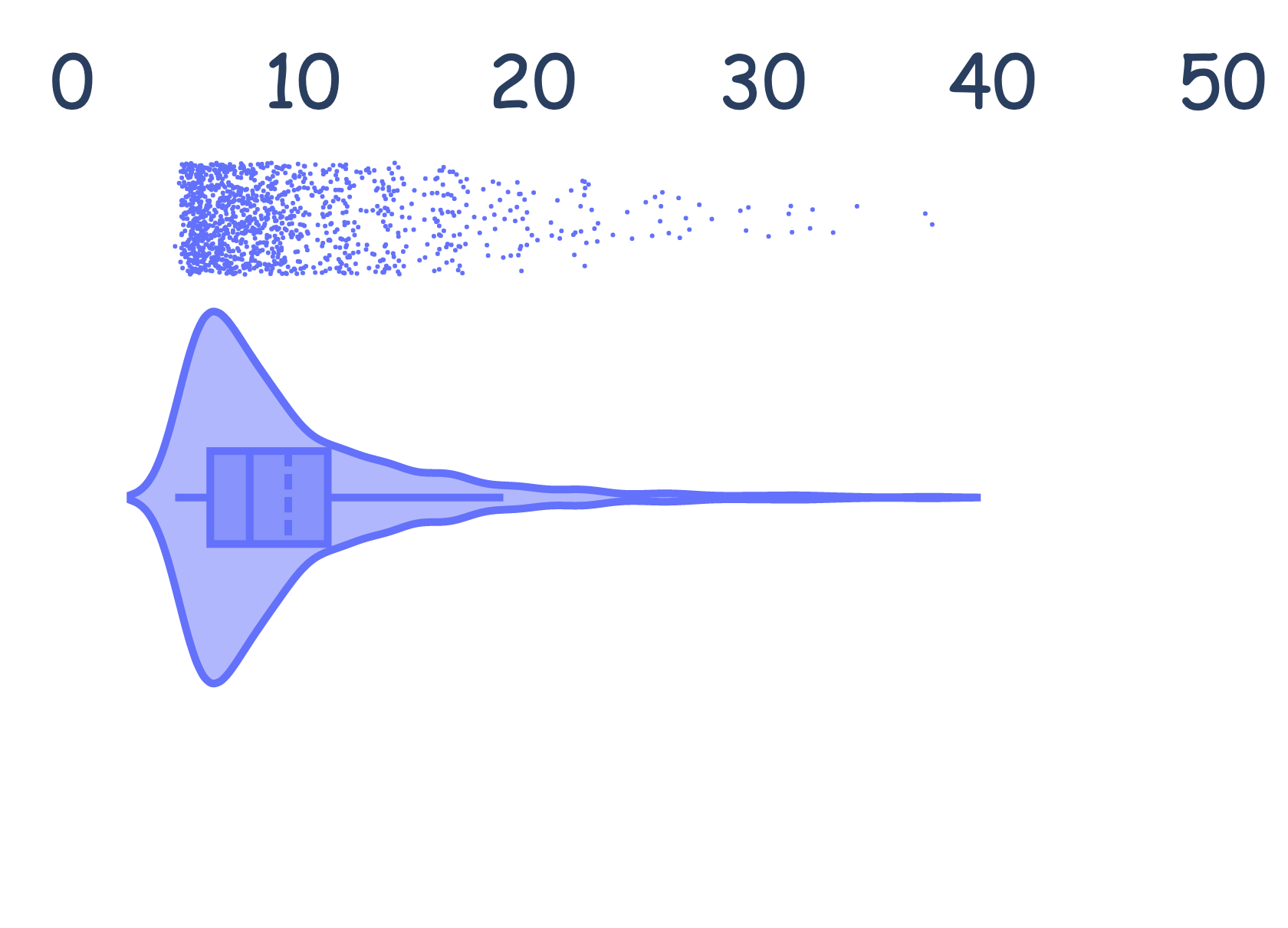}}\end{minipage}
\\
sidewalk & terrain & manmade & vegetation
\\\bottomrule
\end{tabular}
\end{adjustbox}
\label{tab:dataset_nuscenes}
\end{table*}
\begin{table*}[t]
\caption{\textbf{The statistical analysis} of the $19$ semantic classes in the \textit{\textbf{SemanticKITTI}}~\cite{behley2019semanticKITTI} dataset. Statistics are calculated from the \textit{training} split of the dataset. Each violin plot shows the LiDAR point cloud density distribution in a $50$ meters range. Best viewed in colors.}
\vspace{-0.1cm}
\centering
\begin{adjustbox}{width=\textwidth}
\begin{tabular}{c|c|c|c}
\toprule
\multicolumn{4}{c}{\textbf{SemanticKITTI (19 classes)}}
\\\midrule\midrule
\begin{minipage}[b]
{0.49\columnwidth}\centering\raisebox{-.4\height}{\includegraphics[width=\linewidth]{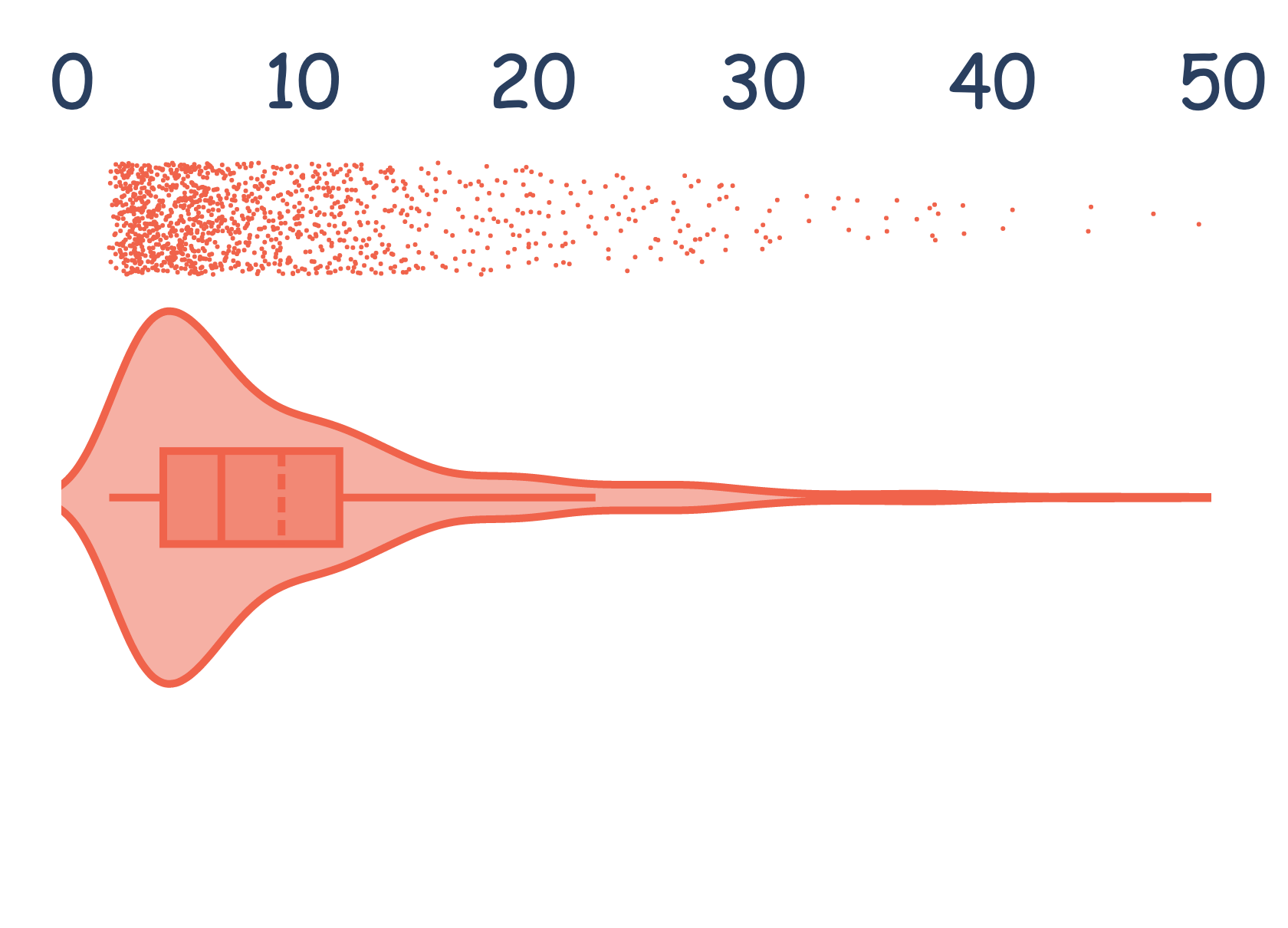}}\end{minipage} & \begin{minipage}[b]
{0.49\columnwidth}\centering\raisebox{-.4\height}{\includegraphics[width=\linewidth]{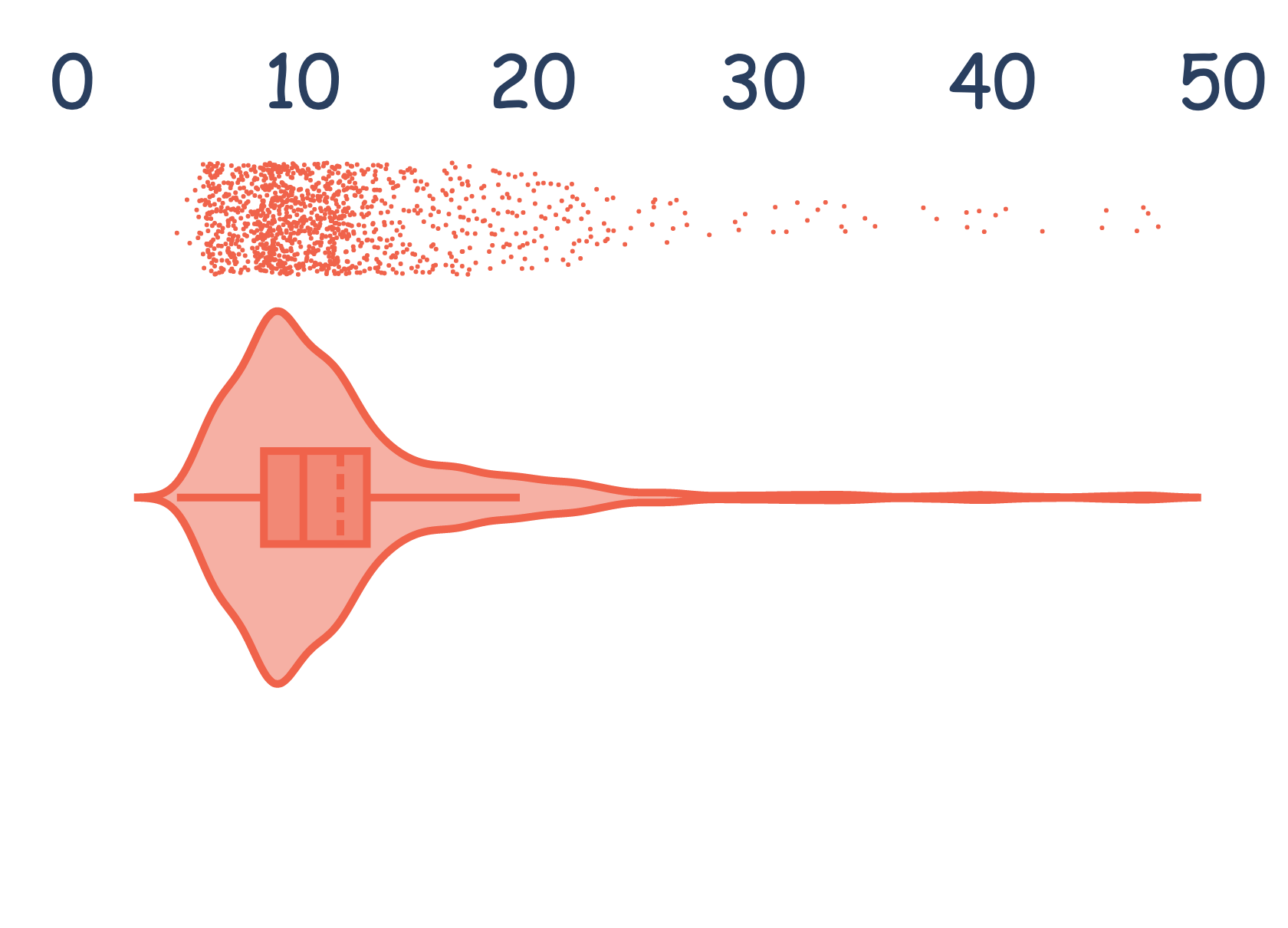}}\end{minipage} & \begin{minipage}[b]
{0.49\columnwidth}\centering\raisebox{-.4\height}{\includegraphics[width=\linewidth]{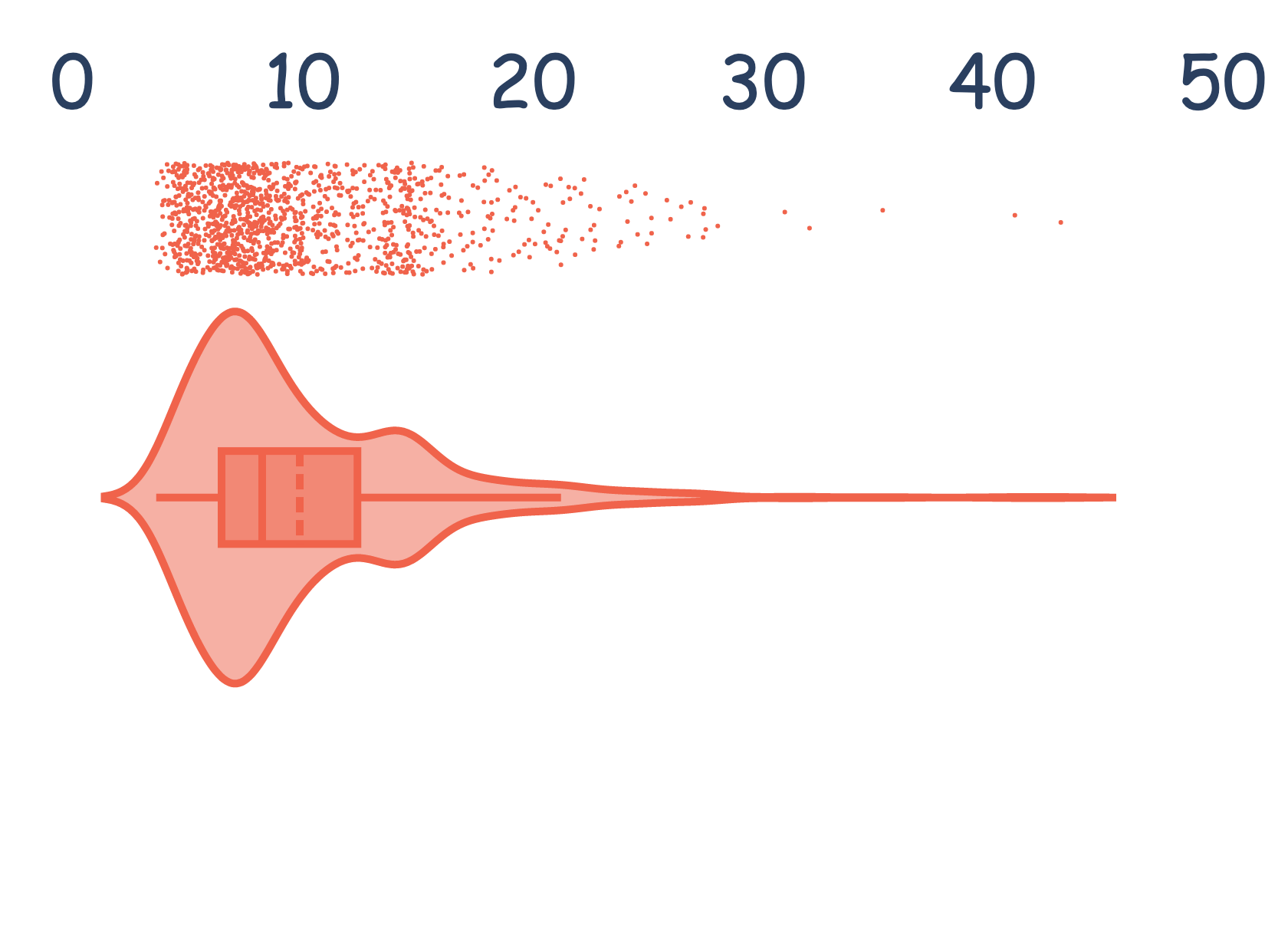}}\end{minipage} & \begin{minipage}[b]
{0.49\columnwidth}\centering\raisebox{-.4\height}{\includegraphics[width=\linewidth]{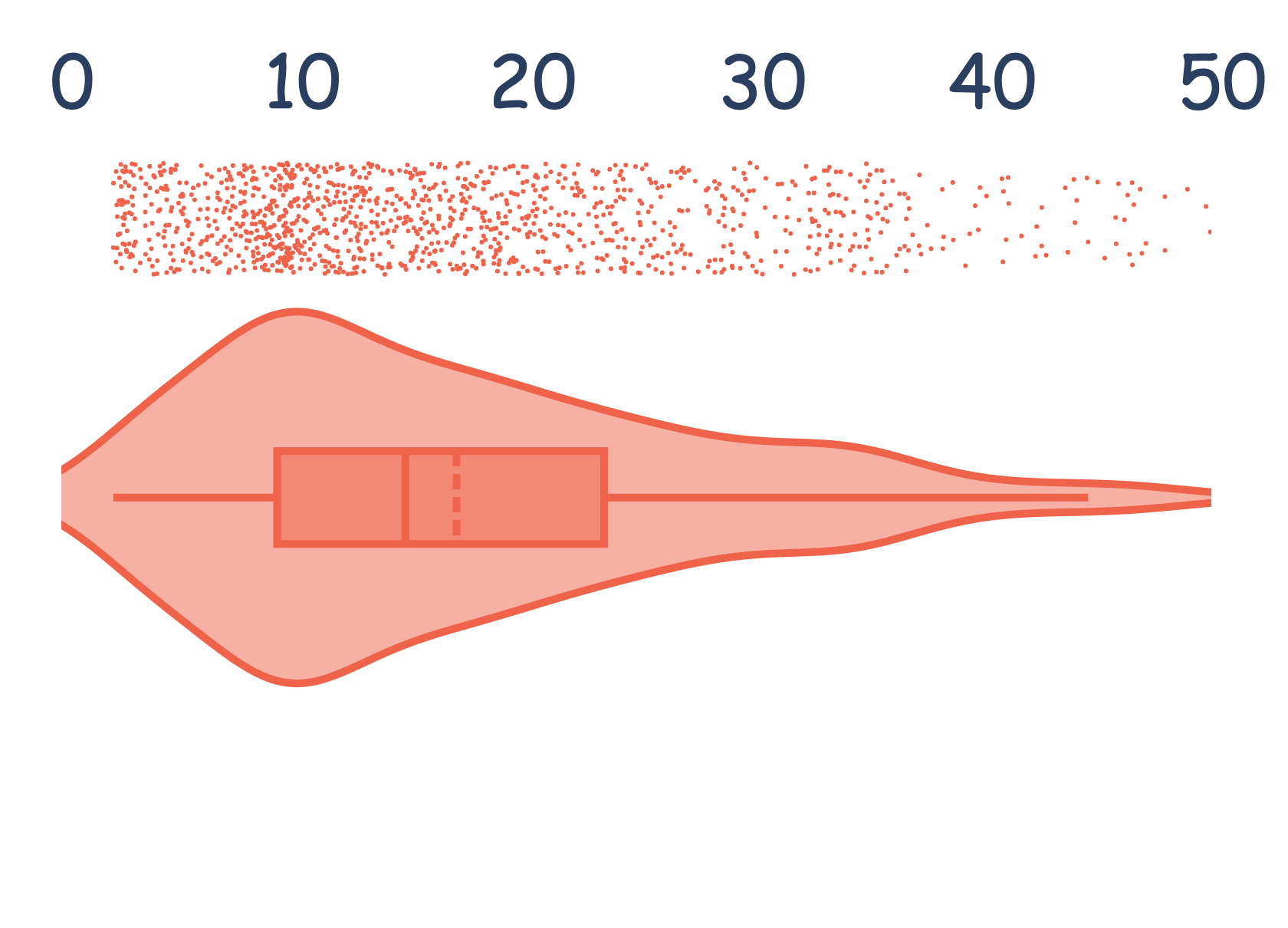}}\end{minipage}
\\
car & bicycle & motorcycle & truck
\\\midrule
\begin{minipage}[b]
{0.49\columnwidth}\centering\raisebox{-.4\height}{\includegraphics[width=\linewidth]{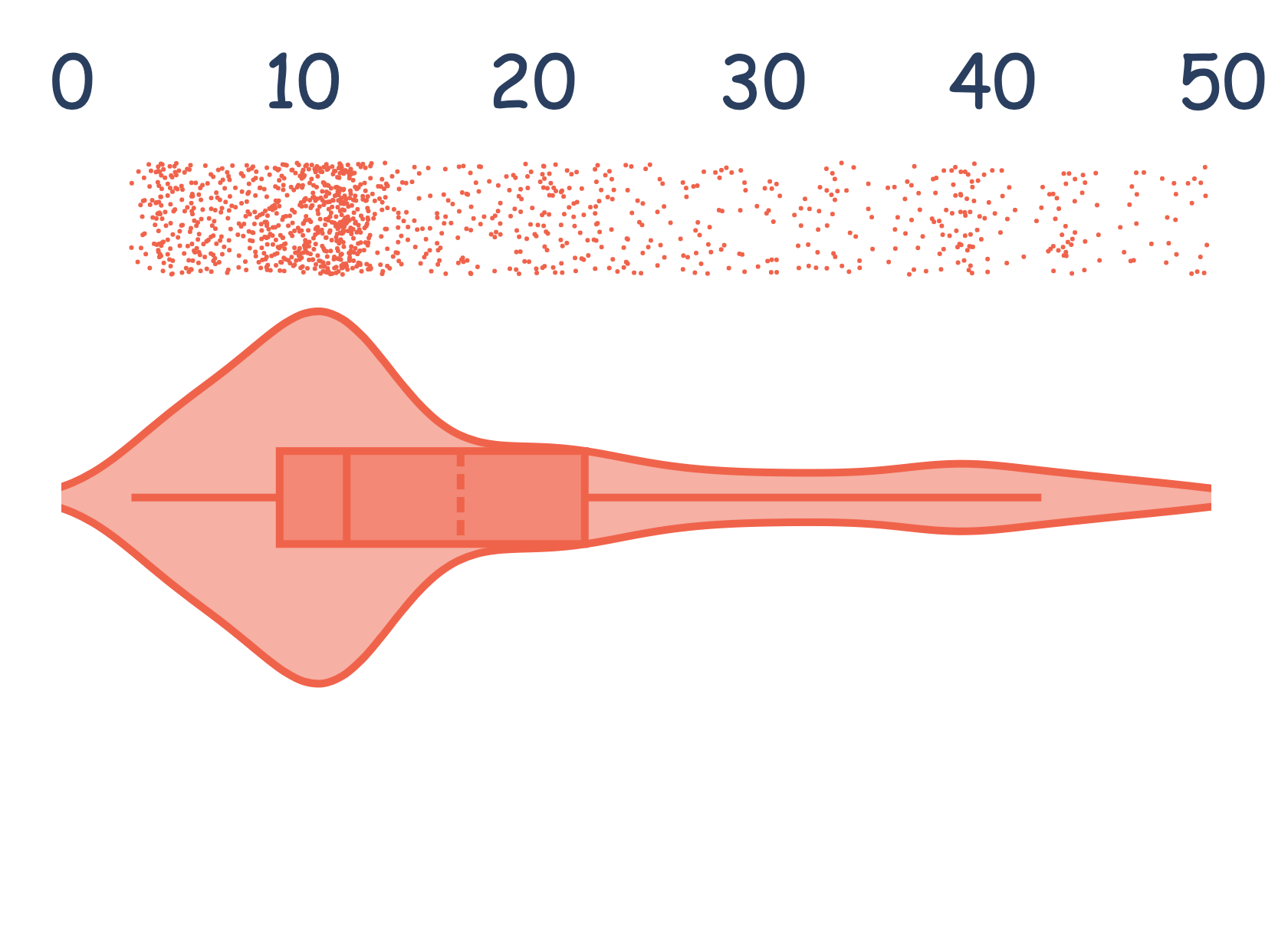}}\end{minipage} & \begin{minipage}[b]
{0.49\columnwidth}\centering\raisebox{-.4\height}{\includegraphics[width=\linewidth]{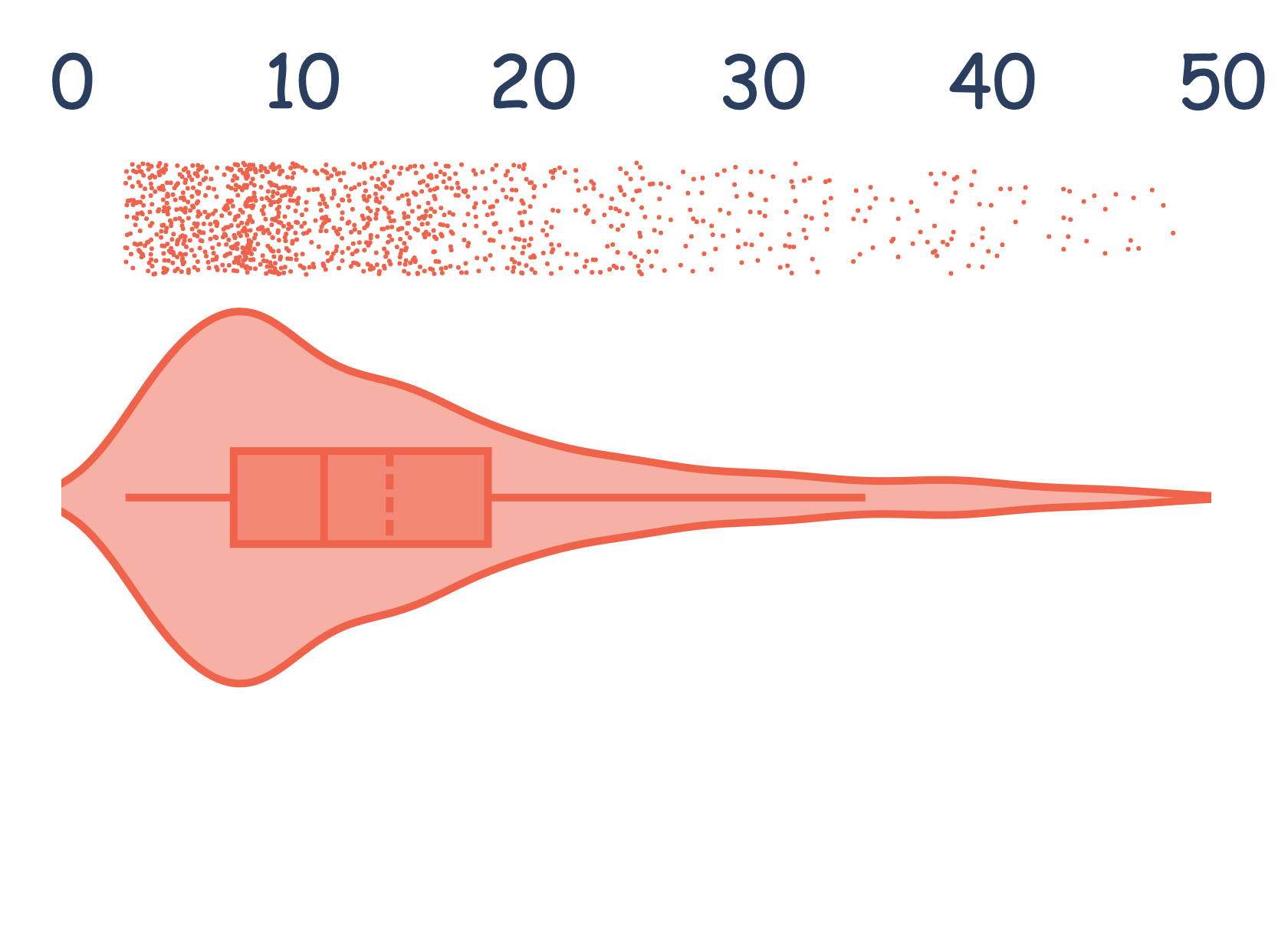}}\end{minipage} & \begin{minipage}[b]
{0.49\columnwidth}\centering\raisebox{-.4\height}{\includegraphics[width=\linewidth]{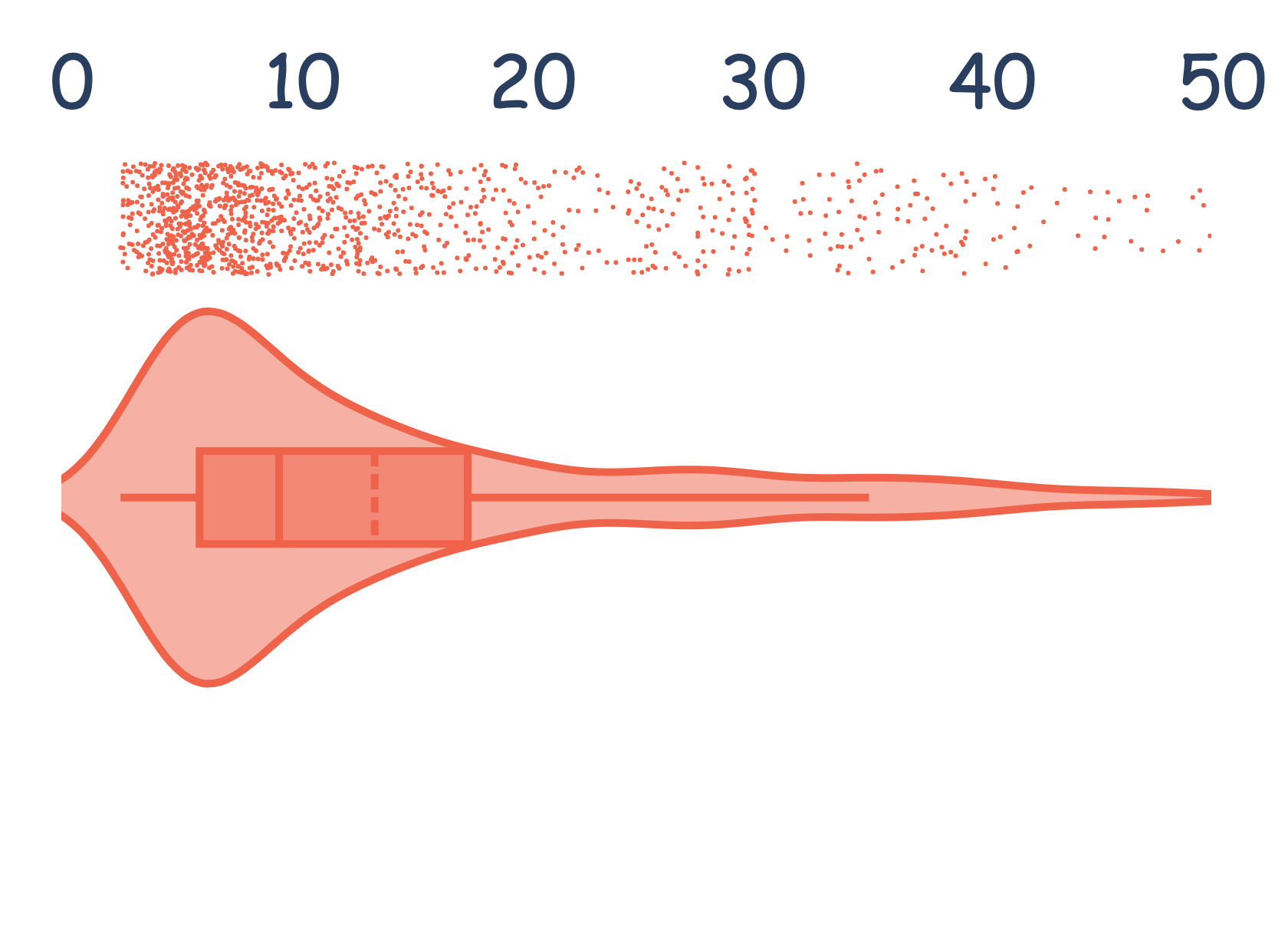}}\end{minipage} & \begin{minipage}[b]
{0.49\columnwidth}\centering\raisebox{-.4\height}{\includegraphics[width=\linewidth]{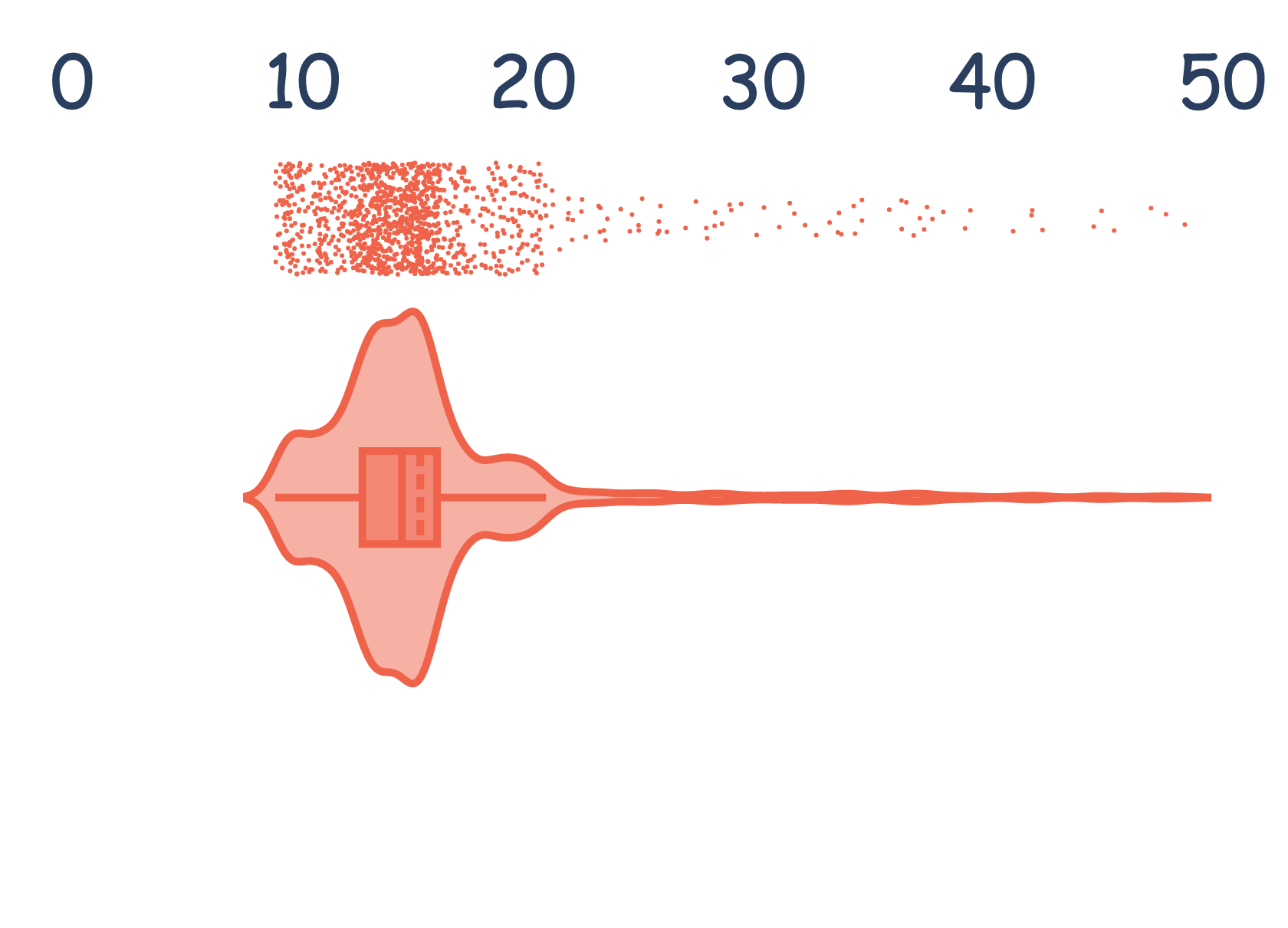}}\end{minipage}
\\
other-vehicle & person & bicyclist & motorcyclist
\\\midrule
\begin{minipage}[b]
{0.49\columnwidth}\centering\raisebox{-.4\height}{\includegraphics[width=\linewidth]{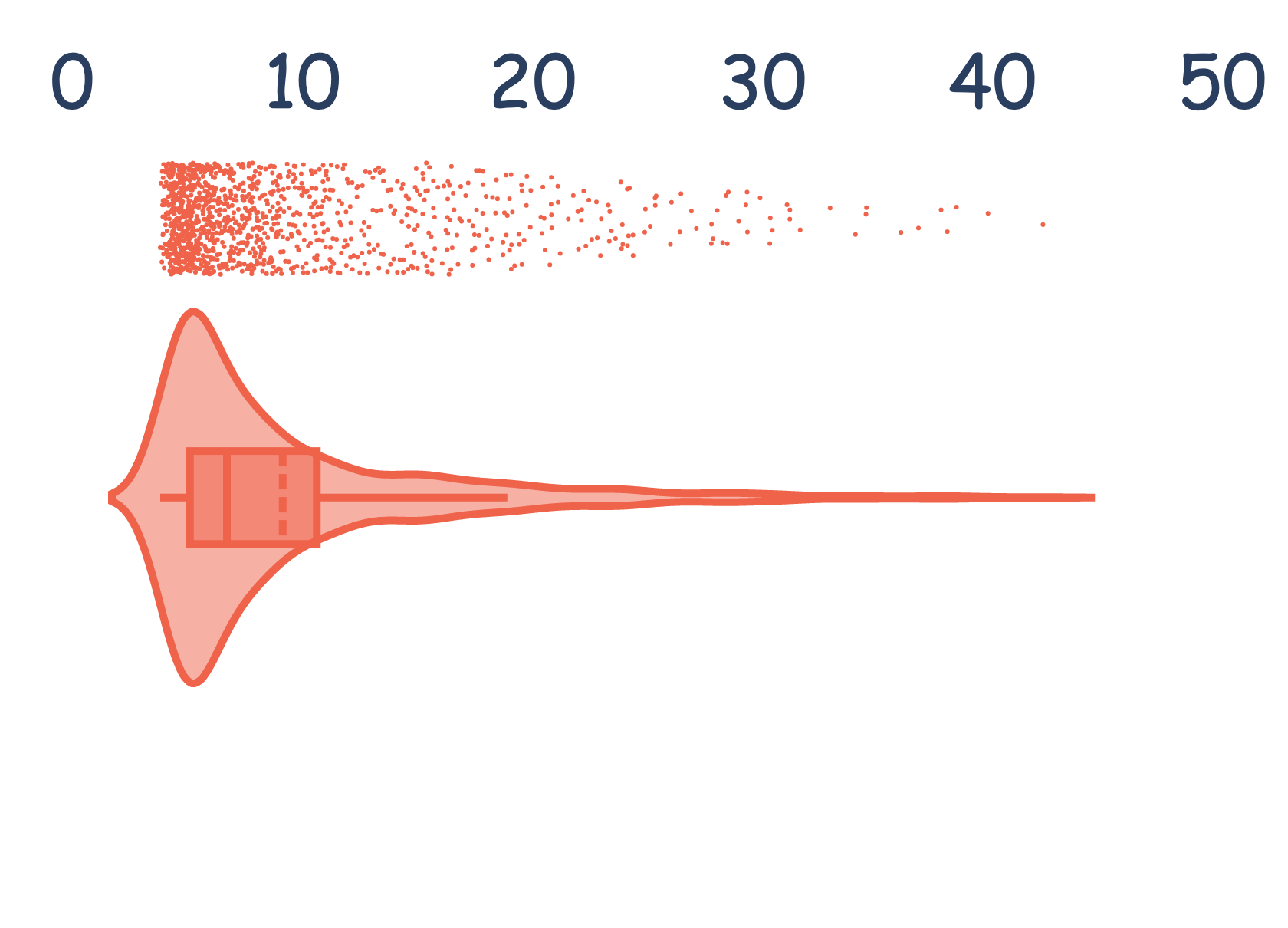}}\end{minipage} & \begin{minipage}[b]
{0.49\columnwidth}\centering\raisebox{-.4\height}{\includegraphics[width=\linewidth]{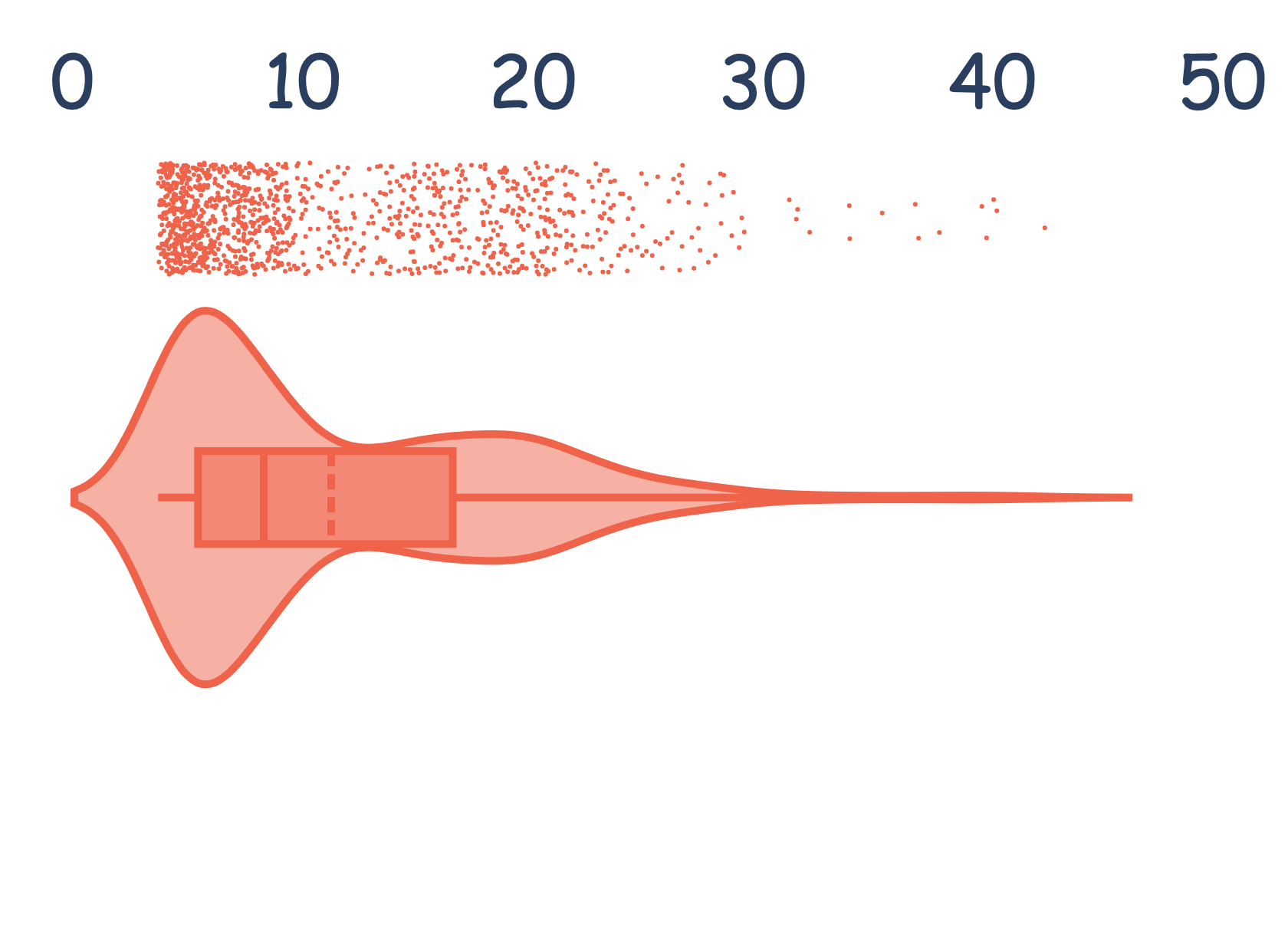}}\end{minipage} & \begin{minipage}[b]
{0.49\columnwidth}\centering\raisebox{-.4\height}{\includegraphics[width=\linewidth]{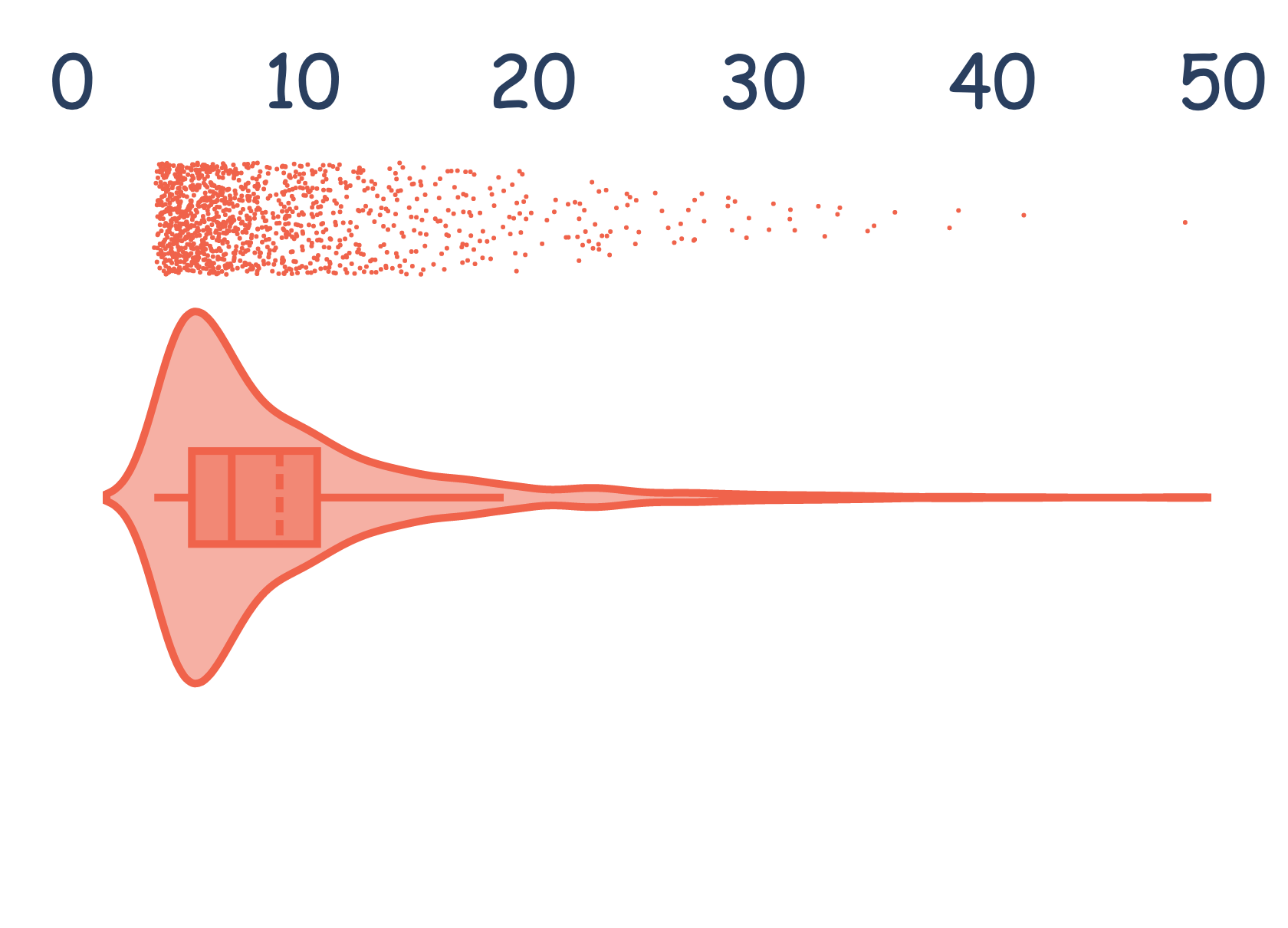}}\end{minipage} & \begin{minipage}[b]
{0.49\columnwidth}\centering\raisebox{-.4\height}{\includegraphics[width=\linewidth]{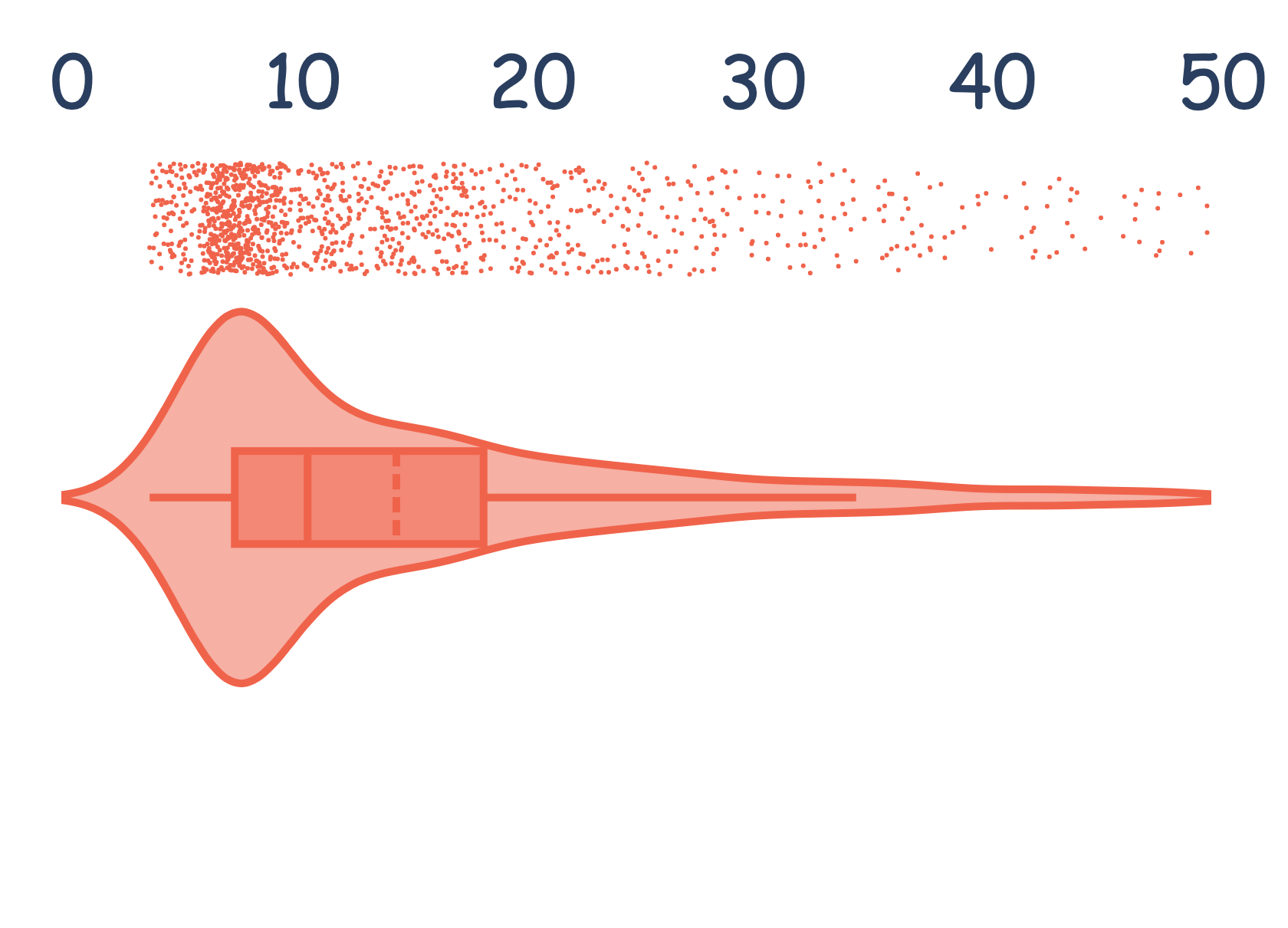}}\end{minipage}
\\
road & parking & sidewalk & other-ground
\\\midrule
\begin{minipage}[b]
{0.49\columnwidth}\centering\raisebox{-.4\height}{\includegraphics[width=\linewidth]{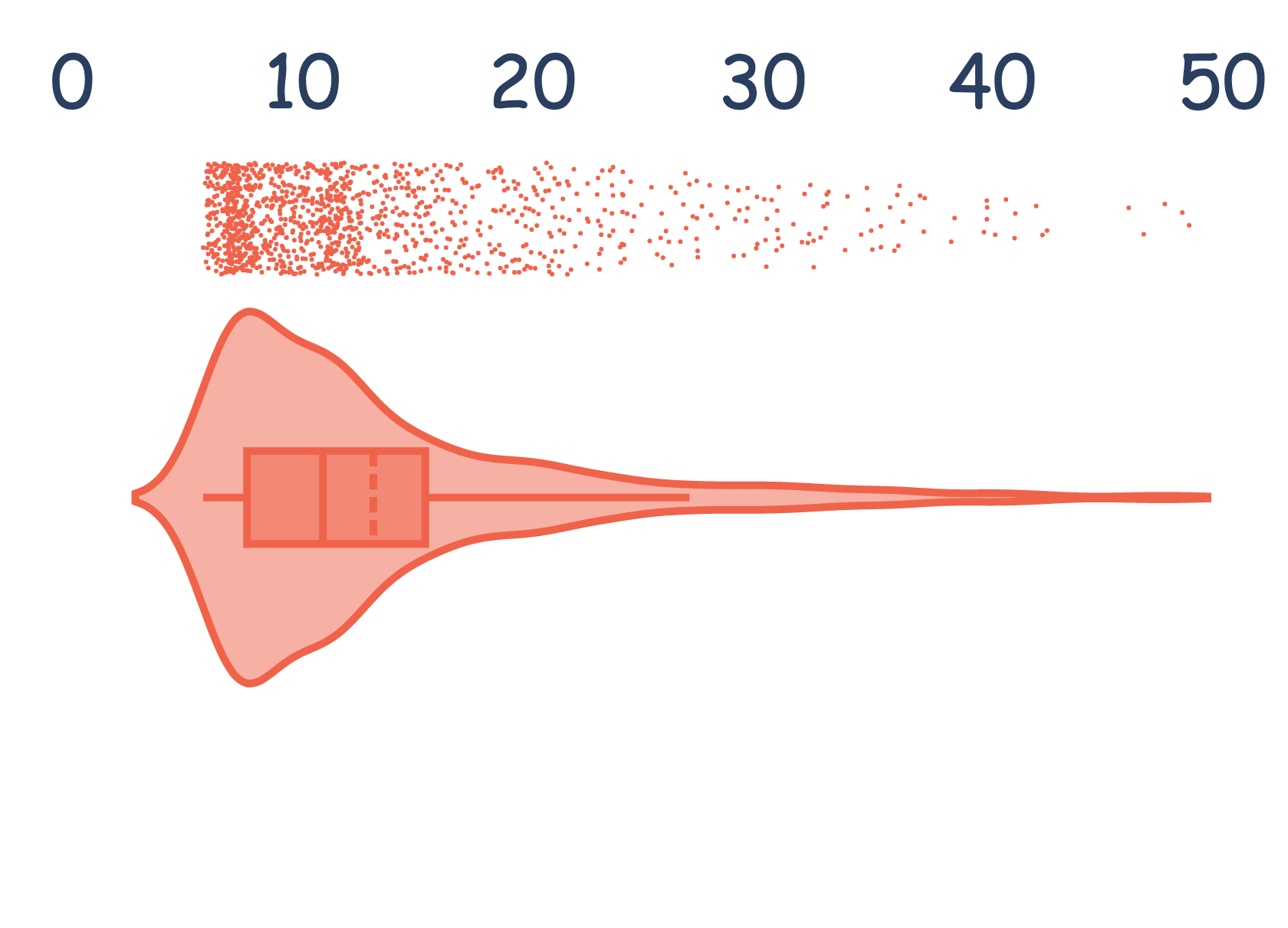}}\end{minipage} & \begin{minipage}[b]
{0.49\columnwidth}\centering\raisebox{-.4\height}{\includegraphics[width=\linewidth]{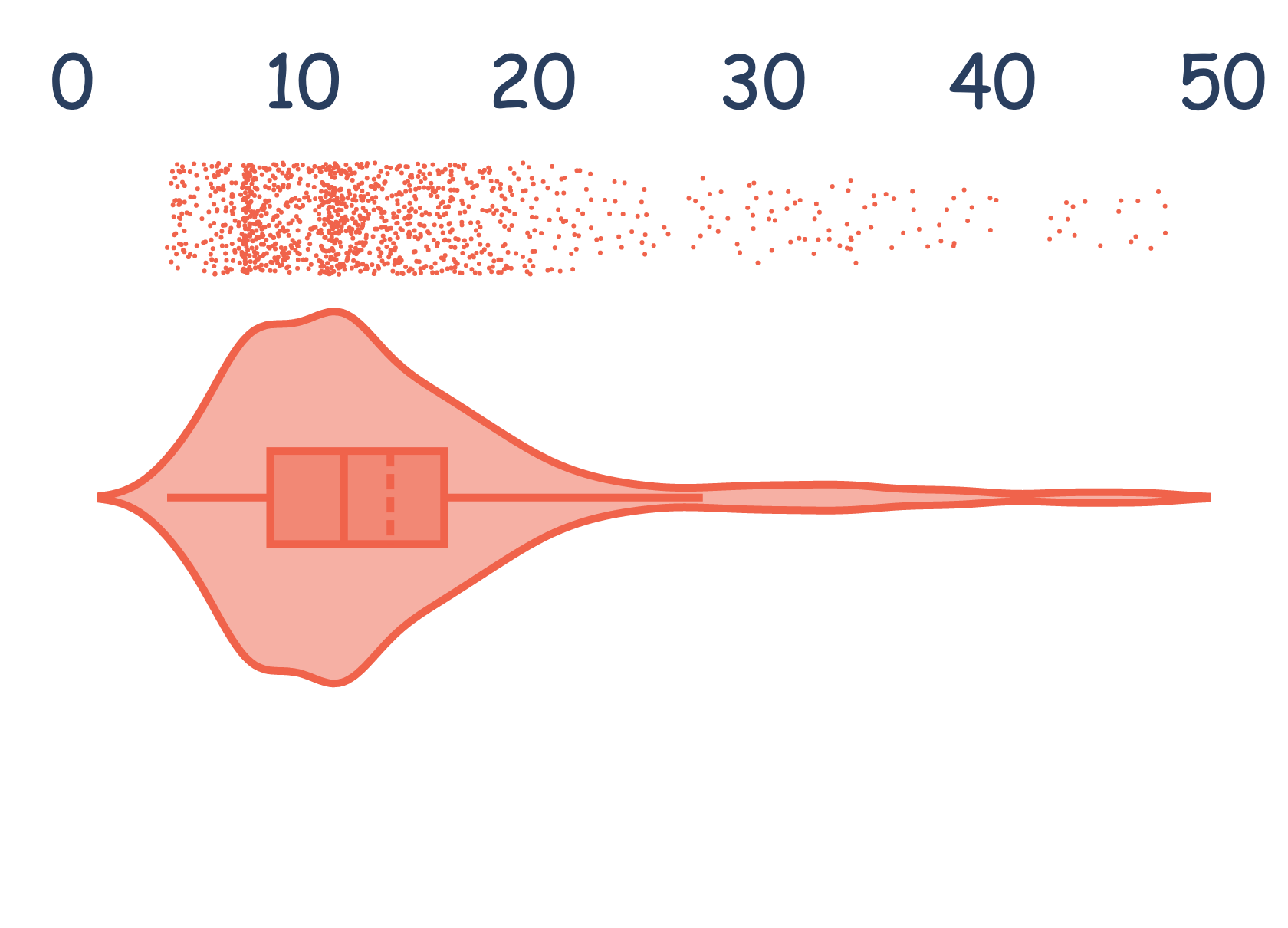}}\end{minipage} & \begin{minipage}[b]
{0.49\columnwidth}\centering\raisebox{-.4\height}{\includegraphics[width=\linewidth]{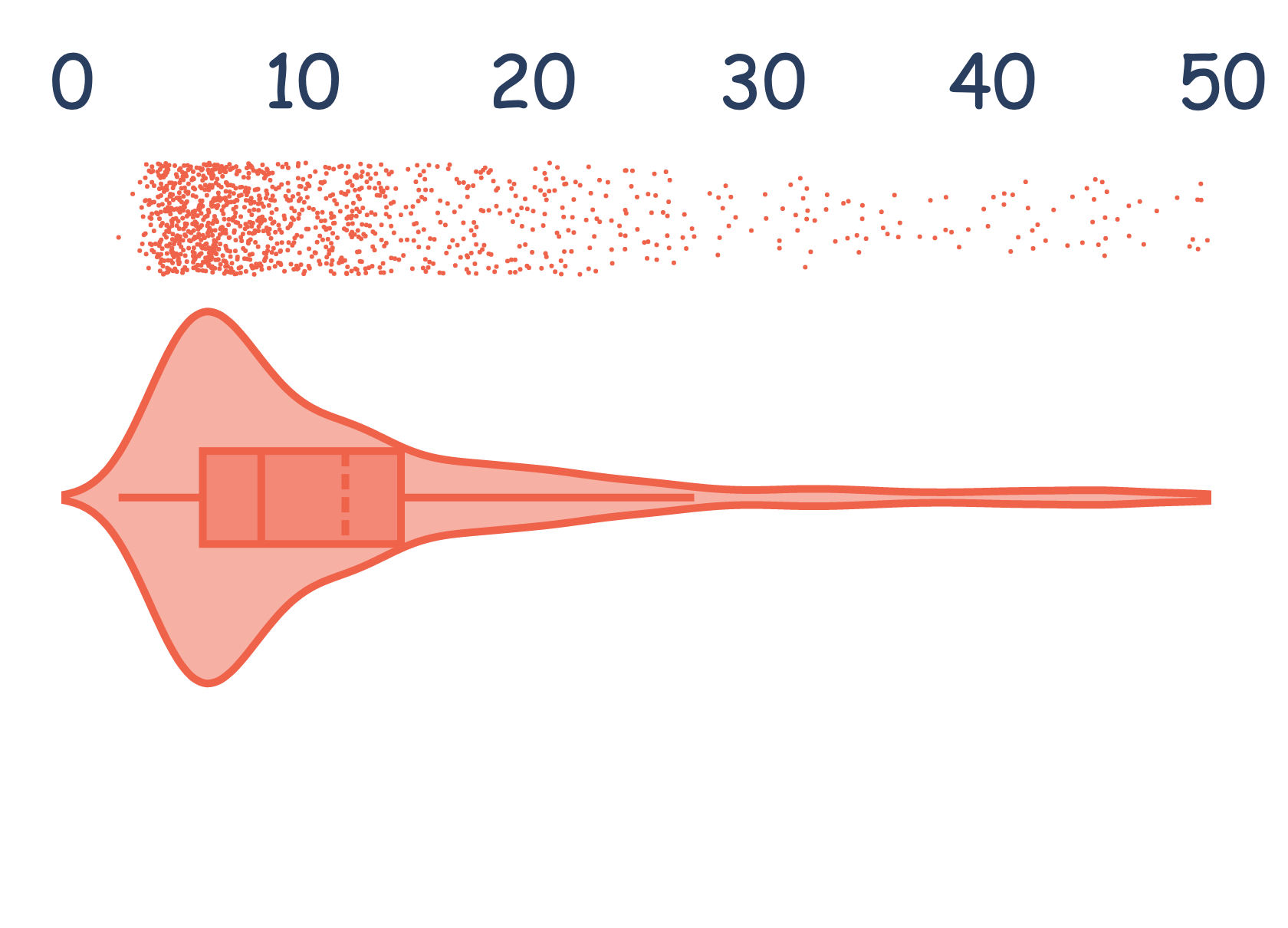}}\end{minipage} & \begin{minipage}[b]
{0.49\columnwidth}\centering\raisebox{-.4\height}{\includegraphics[width=\linewidth]{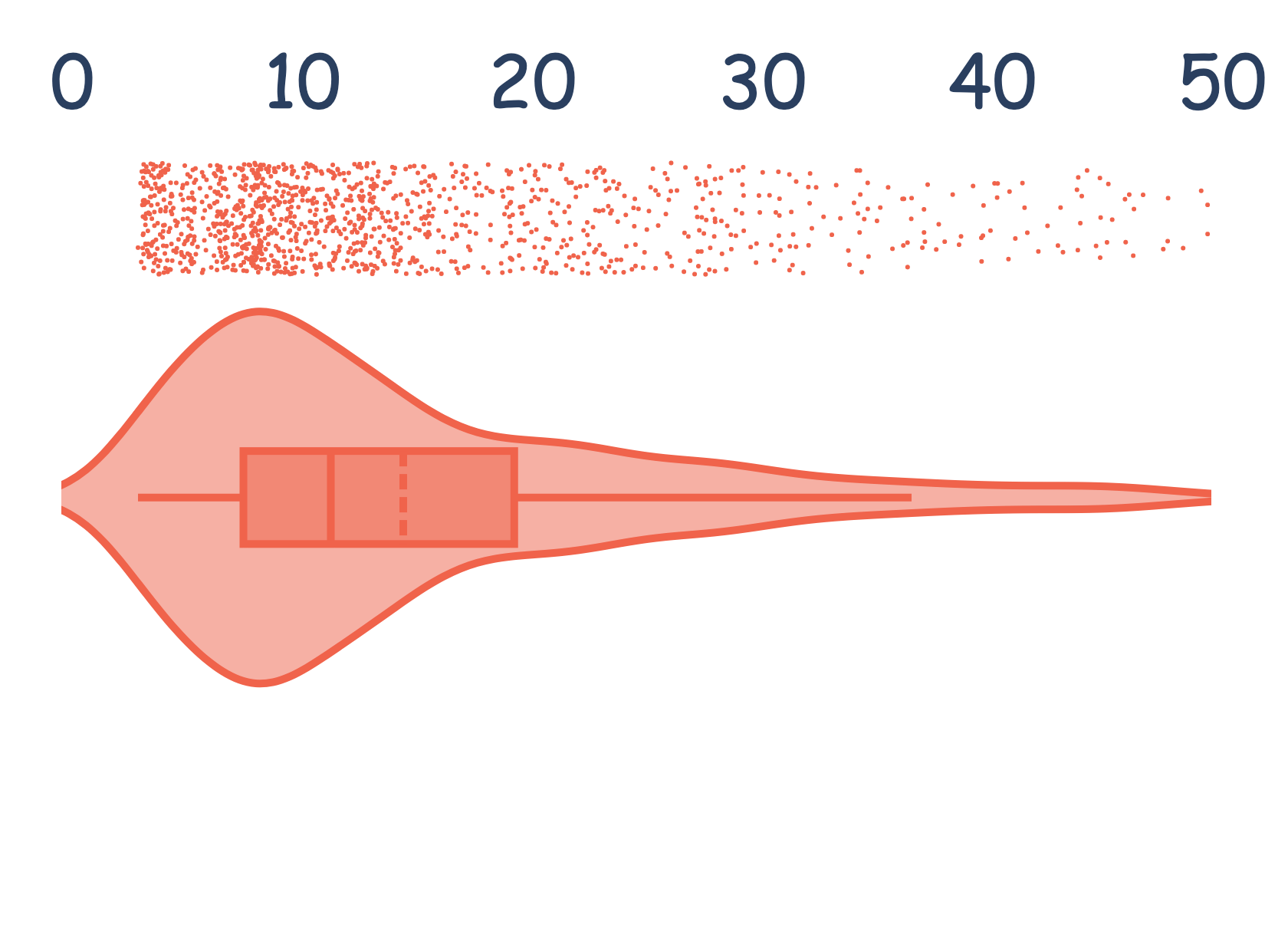}}\end{minipage}
\\
building & fence & vegetation & trunk
\\\midrule
\begin{minipage}[b]
{0.49\columnwidth}\centering\raisebox{-.4\height}{\includegraphics[width=\linewidth]{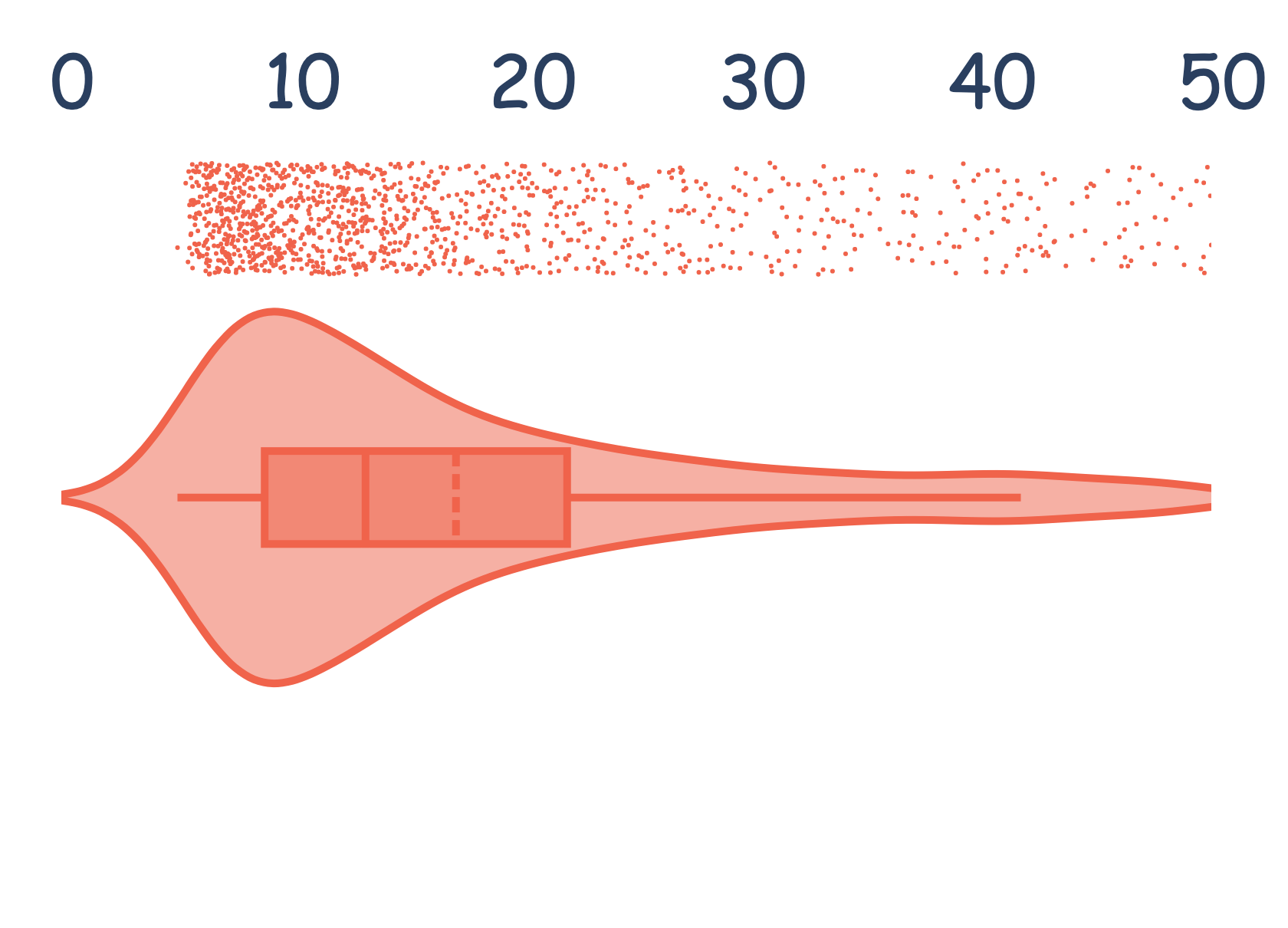}}\end{minipage} & \begin{minipage}[b]
{0.49\columnwidth}\centering\raisebox{-.4\height}{\includegraphics[width=\linewidth]{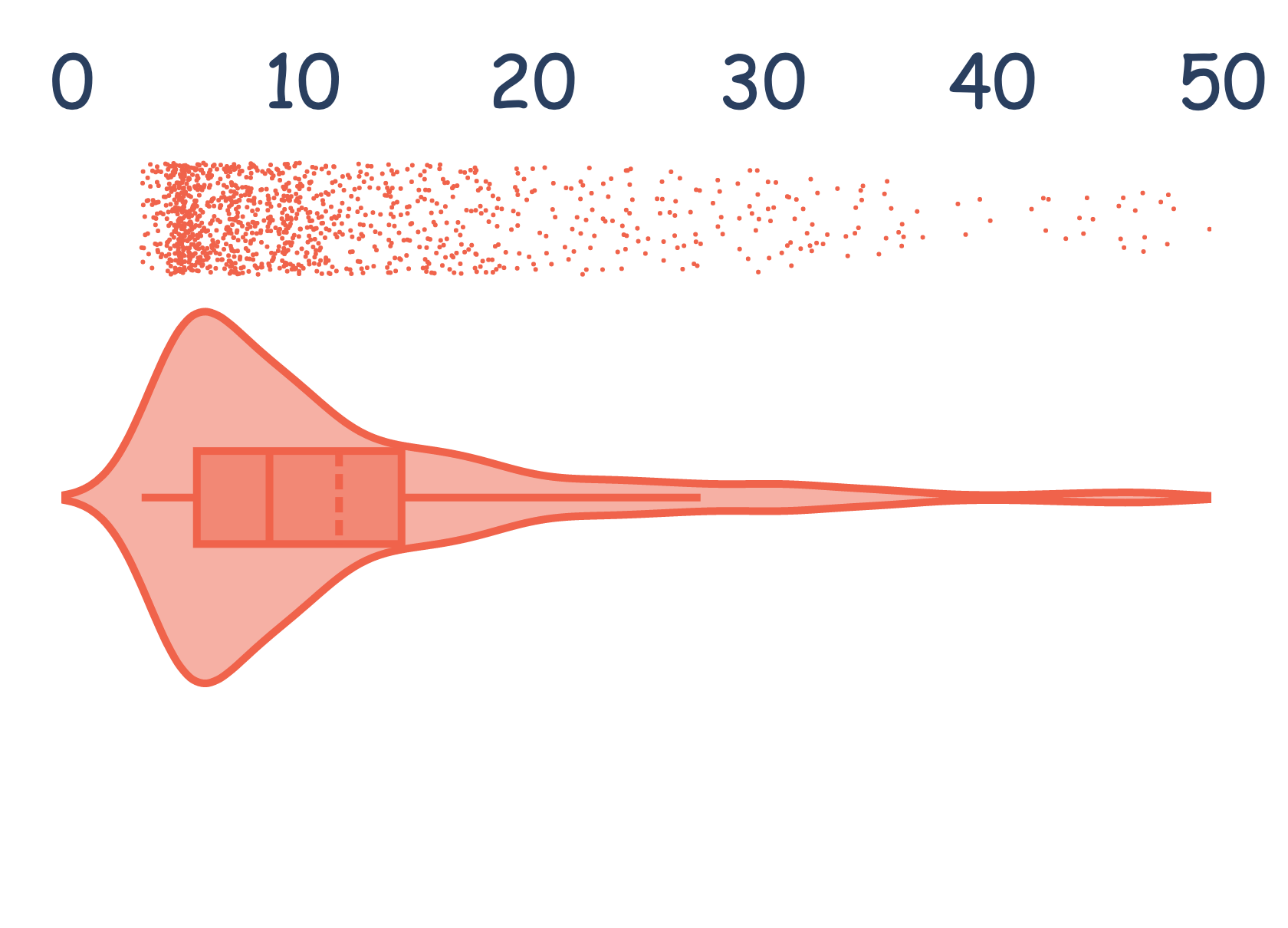}}\end{minipage} & \begin{minipage}[b]
{0.49\columnwidth}\centering\raisebox{-.4\height}{\includegraphics[width=\linewidth]{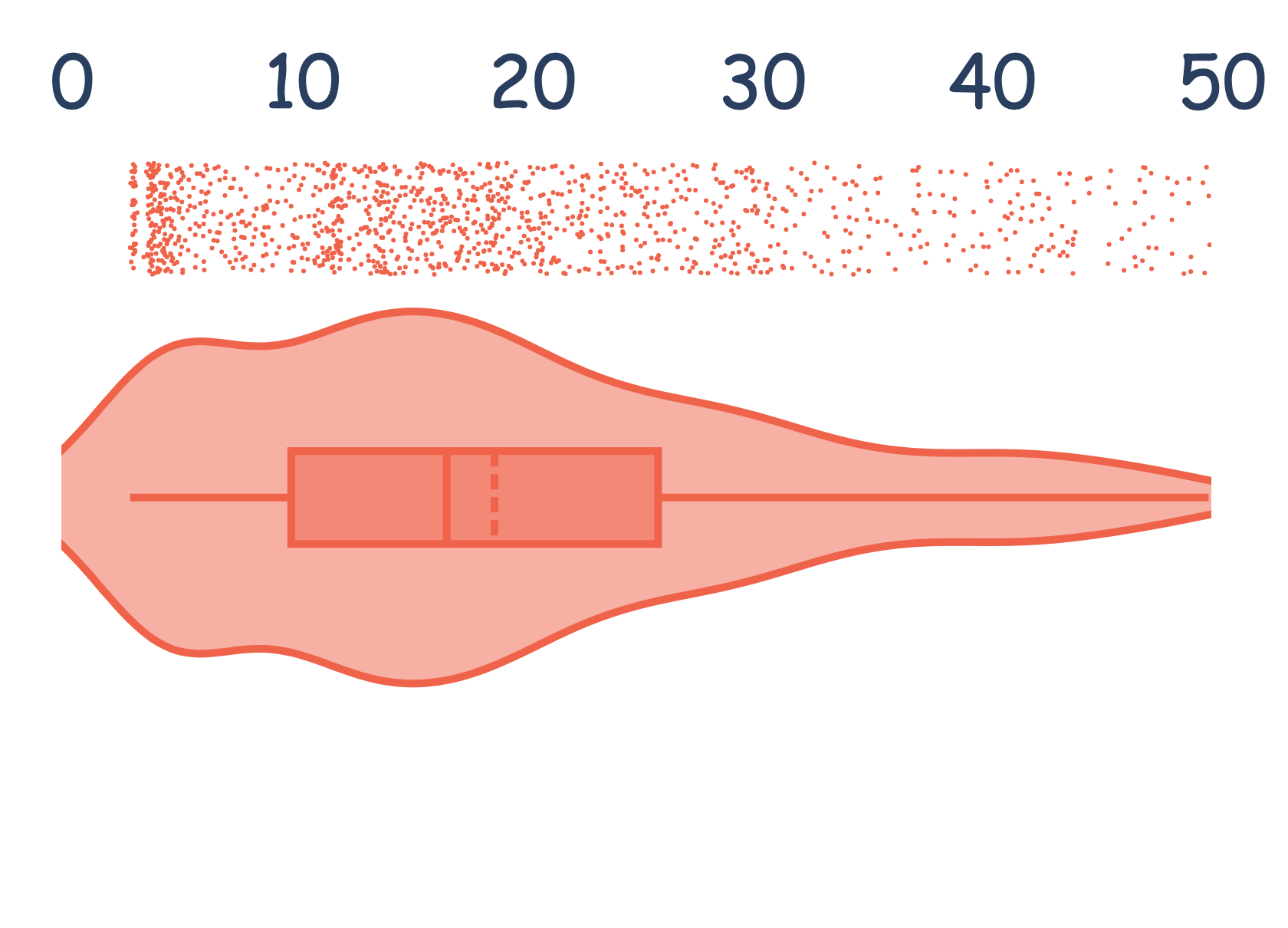}}\end{minipage}
\\
terrain & pole & traffic-sign
\\\bottomrule
\end{tabular}
\end{adjustbox}
\label{tab:dataset_semantickitti}
\end{table*}

Meanwhile, we leverage the \textit{SemanticKITTI-C} and \textit{nuScenes-C} datasets in the Robo3D benchmark \cite{kong2023robo3D} to probe the out-of-training-distribution robustness of 

\begin{itemize}
    \item \textbf{SemanticKITTI-C} is built upon the validation set of the \textit{{SemanticKITTI}} \cite{behley2019semanticKITTI} dataset. It is designed to cover out-of-distribution corruptions that tend to occur in the real world. A total of eight corruption types are benchmarked, including fog, wet ground, snow, motion blur, beam missing, crosstalk, incomplete echo, and cross-sensor cases. For each corruption, three subsets that cover different levels of corruption severity are created, \ie easy, moderate, and hard. The LiDAR segmentation models are expected to be trained on the clean sets while tested on these eight corruption sets. The performance degradation under corruption scenarios is used to measure the model's robustness. Two metrics are designed for such measurements, namely mean Corruption Error (mCE) and mean Resilience Rate (mRR). mCE calculates the relative robustness of a candidate model compared to the baseline model, while mRR computes the absolute performance degradation of a candidate model when it is tested on clean and corruption sets, respectively. In total, there are $97704$ LiDAR scans in \textit{SemanticKITTI-C}, which follow the original dense annotations in \textit{{SemanticKITTI}}. More details of this dataset can be found at \url{https://github.com/ldkong1205/Robo3D}.
    
    \item \textbf{nuScenes-C} shares the same corruption and severity level definitions with \textit{SemanticKITTI-C} and is built upon the validation set of the \textit{nuScenes} \cite{fong2022panoptic-nuScenes} dataset. In total, there are $144456$ LiDAR scans in \textit{nuScenes-C}, which follow the original dense annotations in \textit{nuScenes}. More details of this dataset can be found at \url{https://github.com/ldkong1205/Robo3D}.
\end{itemize}

\subsection{Statistical Analysis}

In this section, we conduct a comprehensive statistical analysis of the \textit{{nuScenes}} \cite{caesar2020nuScenes,fong2022panoptic-nuScenes}, \textit{{SemanticKITTI}} \cite{behley2019semanticKITTI}, and \textit{{Waymo Open}} \cite{sun2020waymoOpen} datasets to showcase the difficulties in merging heterogeneous LiDAR and camera data.

\subsubsection{nuScenes}

The LiDAR point clouds in the \textit{{nuScenes}} \cite{caesar2020nuScenes,fong2022panoptic-nuScenes} dataset are acquired by a Velodyne HDL32E with $32$ beams, $1080$ ($+$/$-10$) points per ring, $20$Hz capture frequency, $360$-degree Horizontal FOV, $+10$-degree to $-30$-degree Vertical FOV, uniform azimuth angles, a $80$m to $100$m range, and up to around $1.39$ million points per second. 
There are a total of $16$ semantic classes in this dataset. The distributions of these classes across a $50$ meters range are shown in \cref{tab:dataset_nuscenes}. As can be seen, most semantic classes distribute within the $20$ meters range. The dynamic classes, such as \texttt{bicycle}, \texttt{motorcycle}, \texttt{bus}, \texttt{car}, and \texttt{pedestrian}, show a high possibility of occurrence at round $5$ meters to $10$ meters. The static classes, on the contrary, are often distributed across a wider range. Typically examples include \texttt{terrain} and \texttt{manmade}. Different semantic classes exhibit unique distribution patterns around the ego-vehicle.

\subsubsection{SemanticKITTI}

The LiDAR point clouds in the \textit{{SemanticKITTI}} \cite{behley2019semanticKITTI} dataset are acquired by a Velodyne HDL-64E with $64$ beams, providing high-resolution data. The Velodyne HDL-64E features a $360$-degree Horizontal Field of View (FOV), a Vertical FOV ranging from $+2$ to $-24.33$ degrees, and an angular resolution of approximately $0.08-0.4$ degrees (vertically) and $0.08-0.35$ degrees (horizontally). The sensor operates at a $10$Hz capture frequency and can detect objects within a range of up to $120$m, delivering densely sampled, detailed point clouds with approximately $1.3$ million points per second. There are a total of $19$ semantic classes in this dataset. The distributions of these classes across a $50$ meters range are shown in \cref{tab:dataset_semantickitti}. It can be seen from these statistical plots that the distributions are distinctly different from each other; points belonging to the \texttt{road} class are intensively distributed in between $5$ meters to $10$ meters around the ego-vehicle, while those dynamic classes like \texttt{bicyclist}, \texttt{motorcyclist}, \texttt{other-vehicle} and \texttt{truck}, tend to appear in a wider range. Similar to the \textit{nuScenes} dataset, the $19$ classes in \textit{SemanticKITTI} also exhibit distinct patterns of occurrence in the driving scenes.

\subsubsection{Waymo Open}

The 3D semantic segmentation subset of the \textit{Waymo Open} \cite{sun2020waymoOpen} dataset features LiDAR point clouds obtained using Waymo's proprietary LiDAR sensors, which include mid-range and short-range LiDARs. There are five LiDARs in total - one mid-range LiDAR (top) and four short-range LiDARs (front, side left, side right, and rear), where the mid-range LiDAR has a non-uniform inclination beam angle pattern. The range of the mid-range LiDAR is truncated to a maximum of $75$ meters. The range of the short-range LiDARs is truncated to a maximum of $20$ meters. The strongest two intensity returns are provided for all five LiDARs. An extrinsic calibration matrix transforms the LiDAR frame to the vehicle frame. The point clouds of each LiDAR in \textit{Waymo Open} are encoded as a range image. Two range images are provided for each LiDAR, one for each of the two strongest returns. There are four channels in the range image, including range, intensity, elongation, and occupancy. The distributions of these classes across a $50$ meters range are shown in \cref{tab:dataset_waymo}. As can be seen. the class distributions of \textit{Waymo Open} are more diverse than those from \textit{nuScenes} and \textit{SemanticKITTI}. Some semantic classes, including \texttt{motorcyclist}, \texttt{pedestrian}, \texttt{construction-cone}, \texttt{vegetation}, and \texttt{tree-trunk}, are distributed across almost the entire driving scenes captured by the five LiDAR sensors.

\begin{table*}[t]
\caption{\textbf{The statistical analysis} of the $22$ semantic classes in the \textit{\textbf{Waymo Open}}~\cite{sun2020waymoOpen} dataset. Statistics are calculated from the \textit{training} split of the dataset. Each violin plot shows the LiDAR point cloud density distribution in a $50$ meters range. Best viewed in colors.}
\vspace{-0.1cm}
\centering
\begin{adjustbox}{width=\textwidth}
\begin{tabular}{c|c|c|c}
\toprule
\multicolumn{4}{c}{\textbf{Waymo Open (22 classes)}}
\\\midrule\midrule
\begin{minipage}[b]
{0.49\columnwidth}\centering\raisebox{-.4\height}{\includegraphics[width=\linewidth]{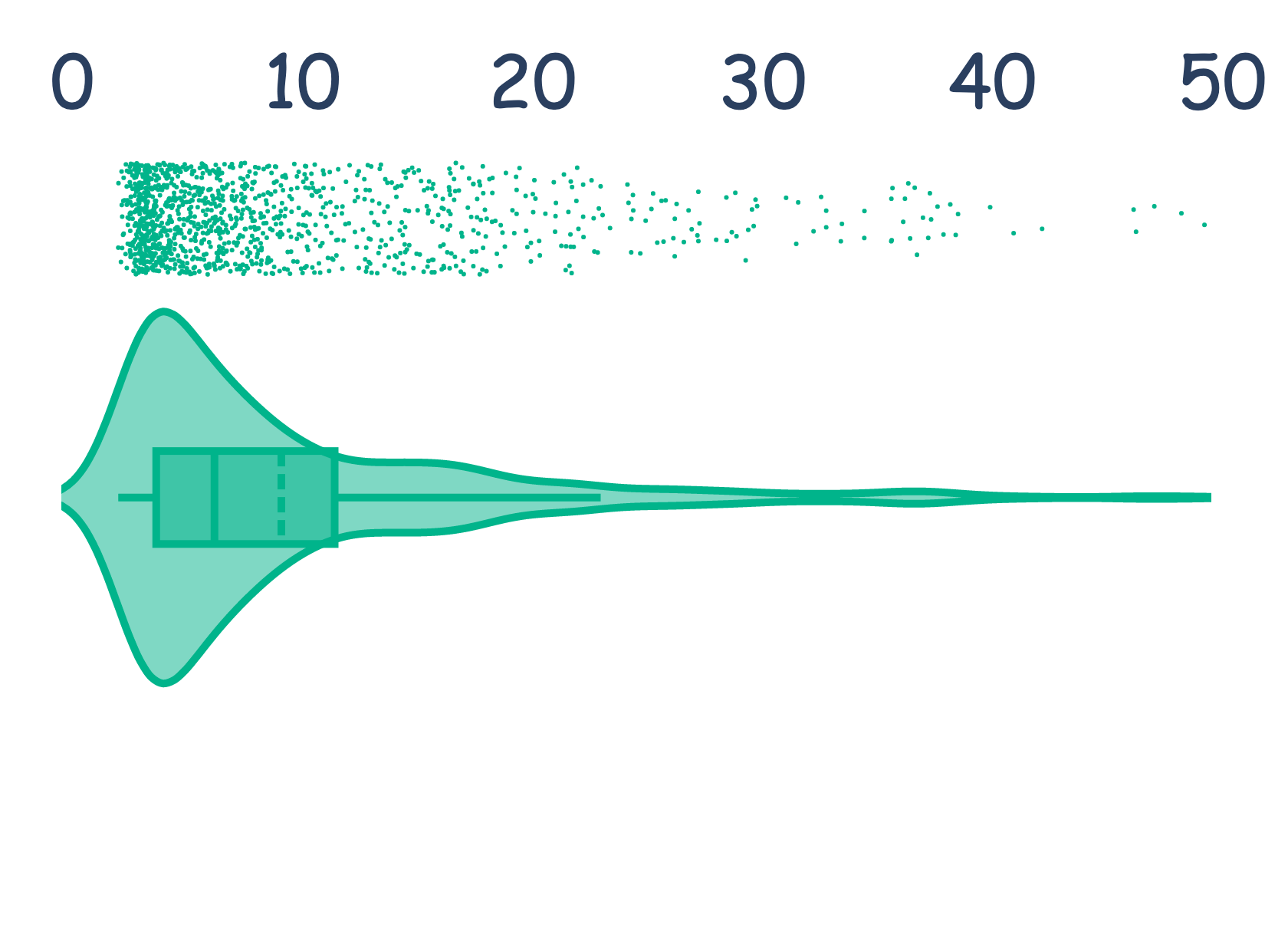}}\end{minipage} & \begin{minipage}[b]
{0.49\columnwidth}\centering\raisebox{-.4\height}{\includegraphics[width=\linewidth]{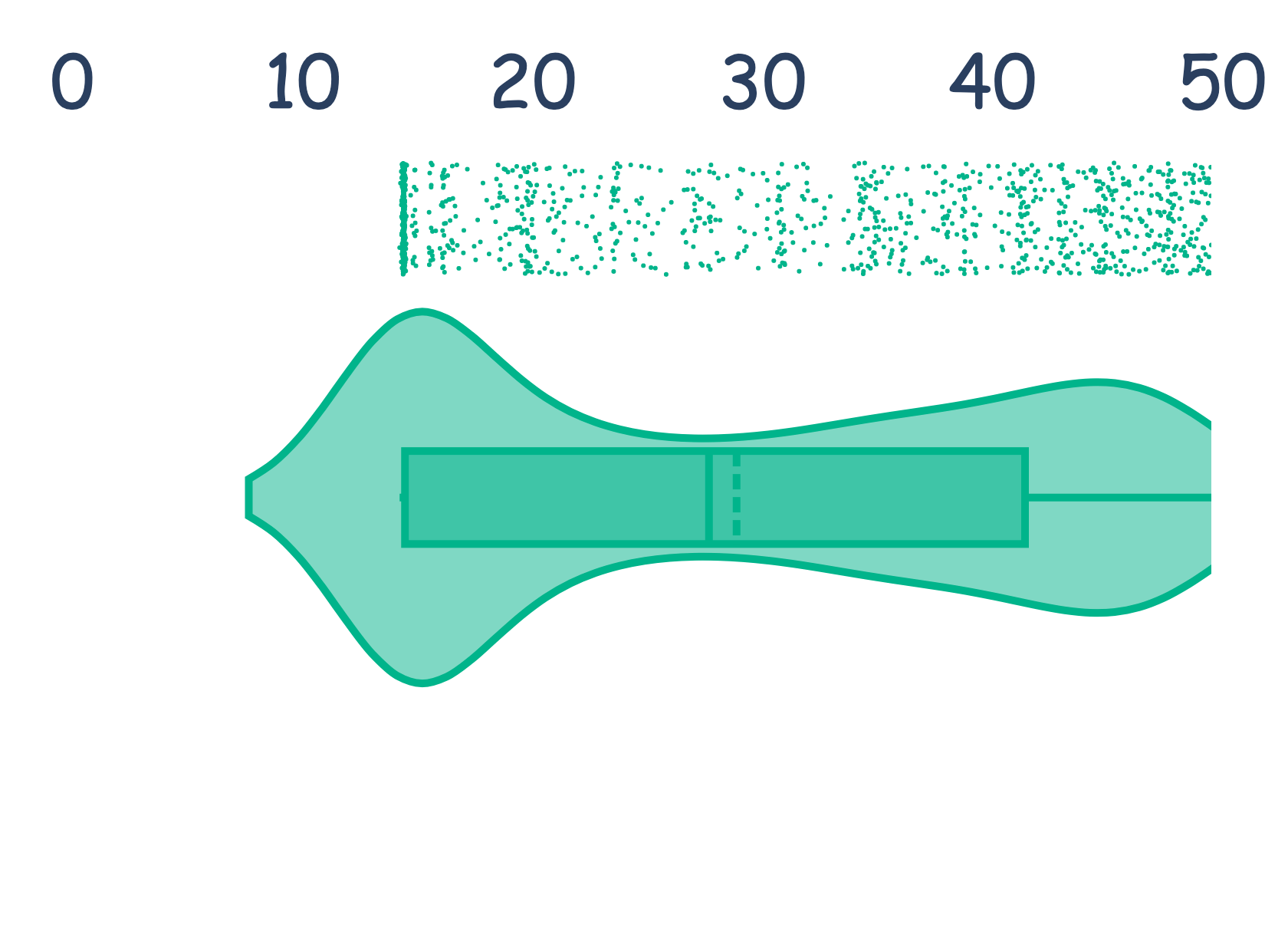}}\end{minipage} & \begin{minipage}[b]
{0.49\columnwidth}\centering\raisebox{-.4\height}{\includegraphics[width=\linewidth]{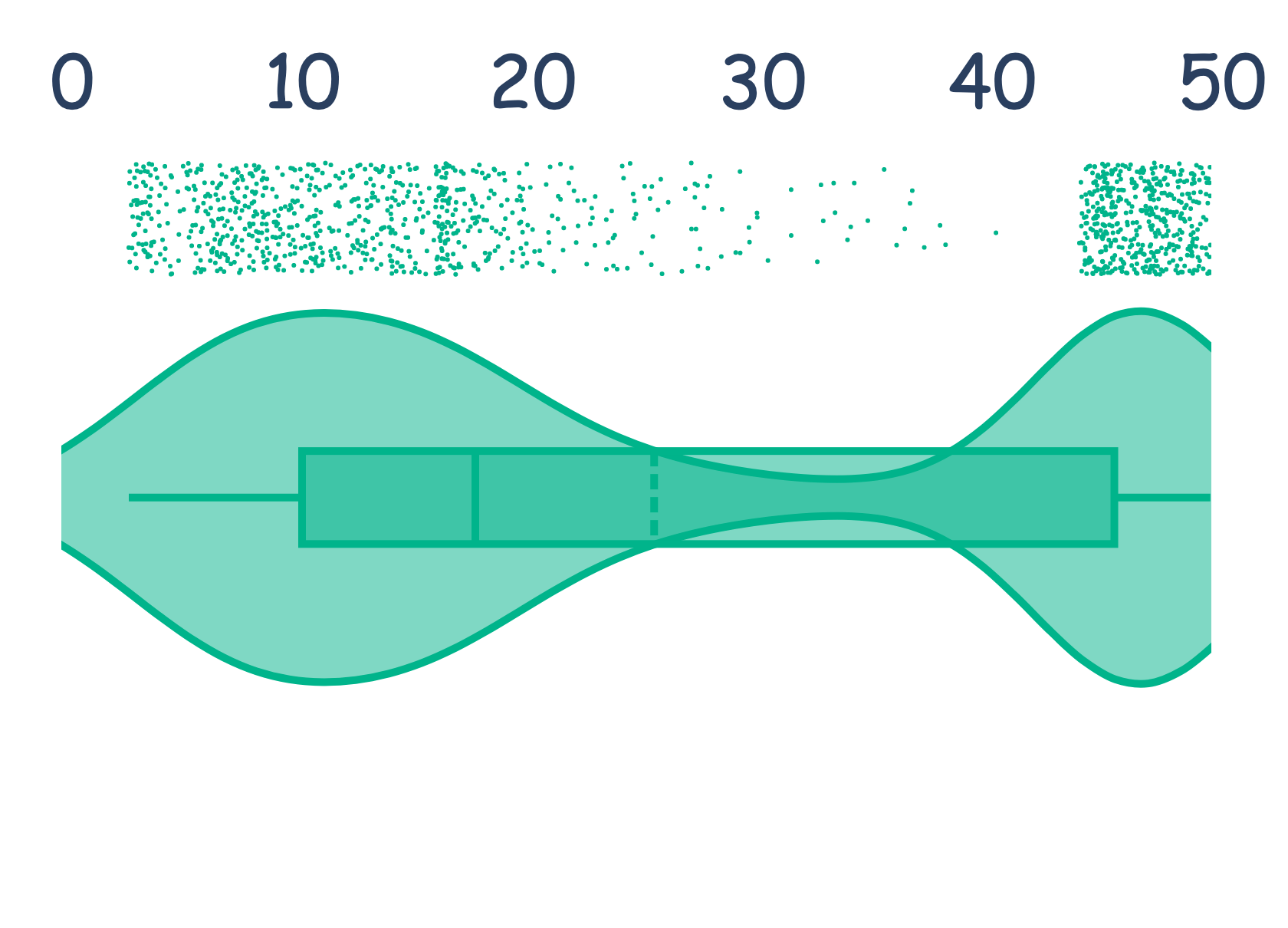}}\end{minipage} & \begin{minipage}[b]
{0.49\columnwidth}\centering\raisebox{-.4\height}{\includegraphics[width=\linewidth]{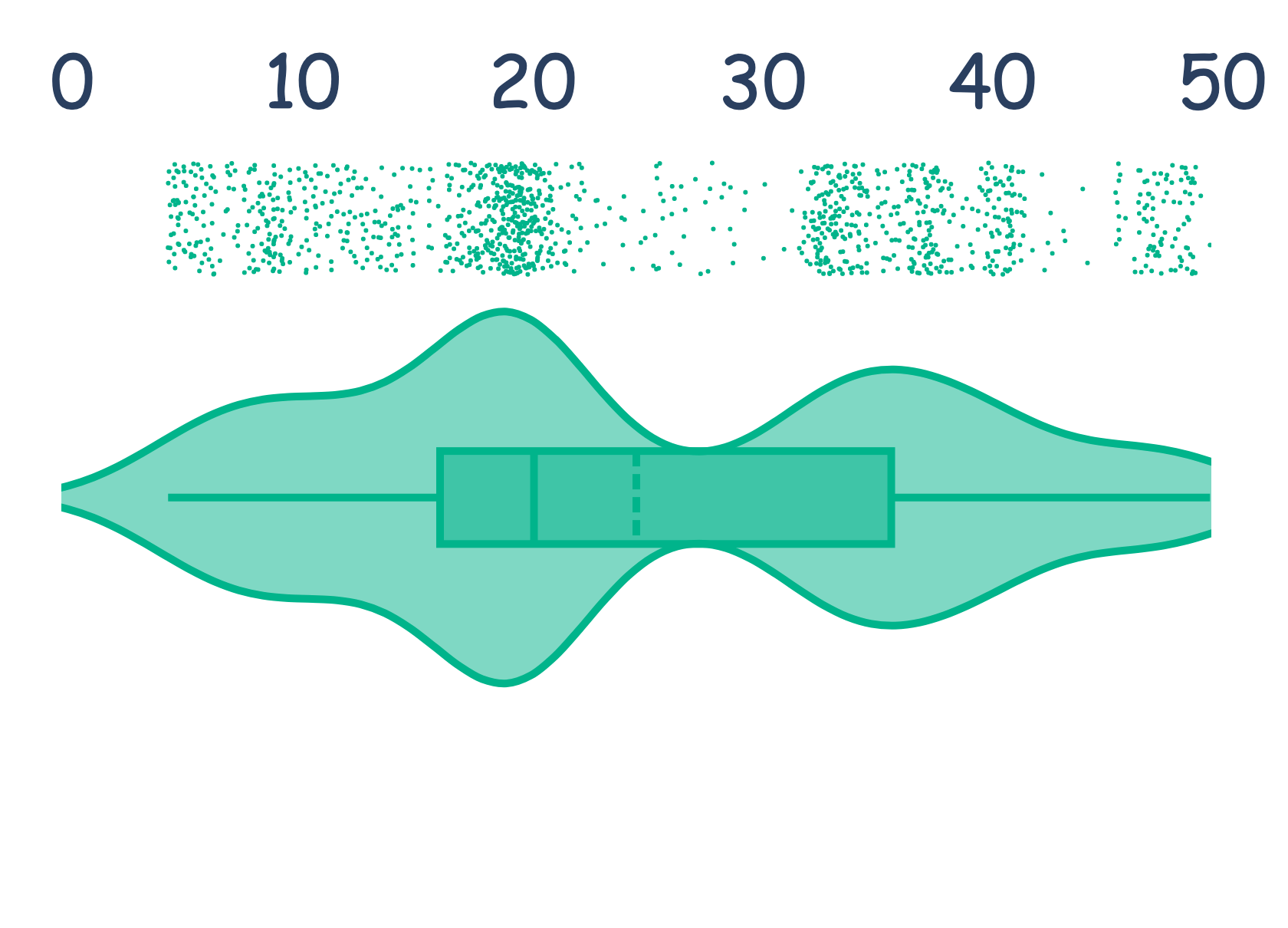}}\end{minipage}
\\
car & truck & bus & other-vehicle
\\\midrule
\begin{minipage}[b]
{0.49\columnwidth}\centering\raisebox{-.4\height}{\includegraphics[width=\linewidth]{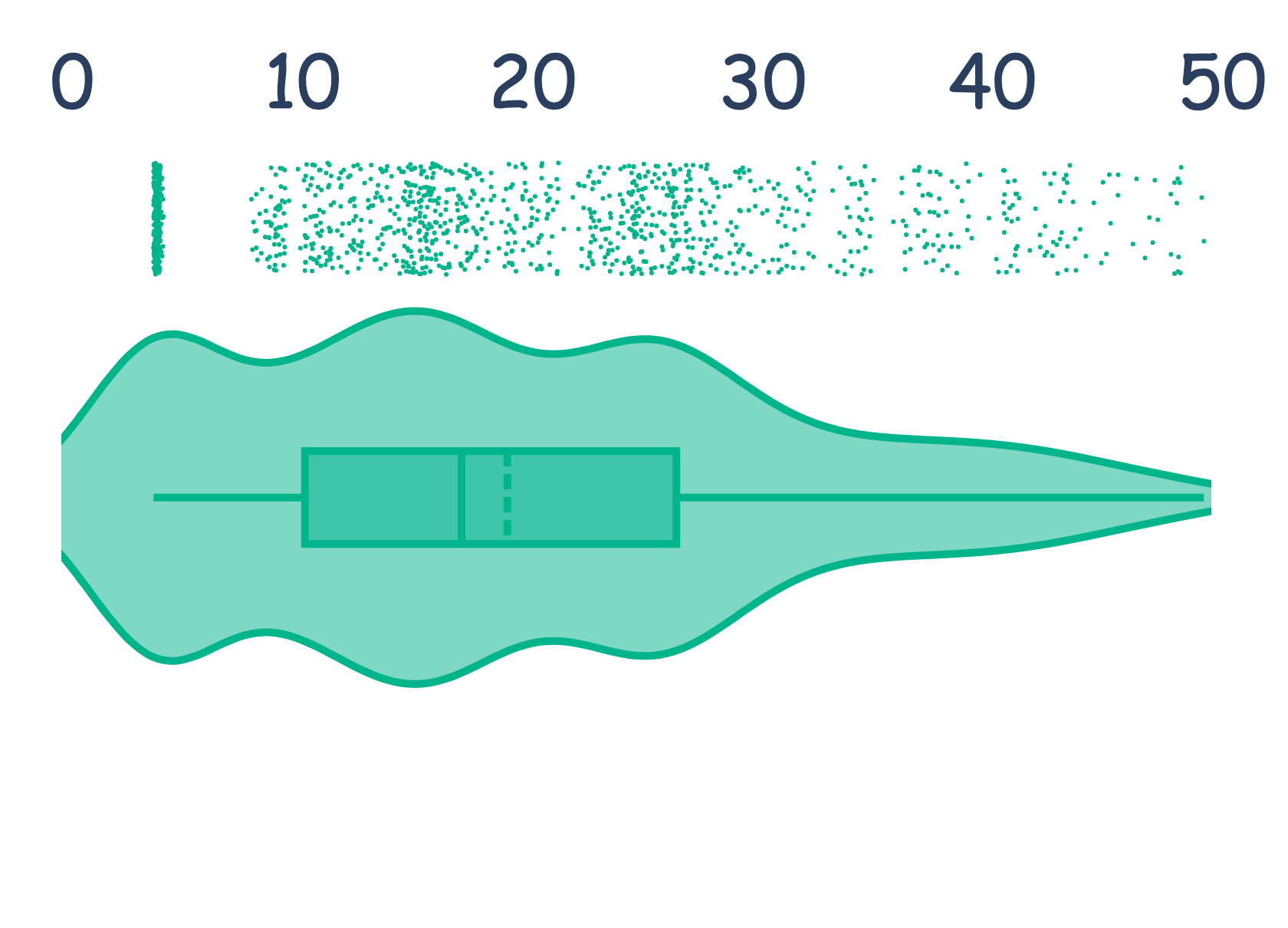}}\end{minipage} & \begin{minipage}[b]
{0.49\columnwidth}\centering\raisebox{-.4\height}{\includegraphics[width=\linewidth]{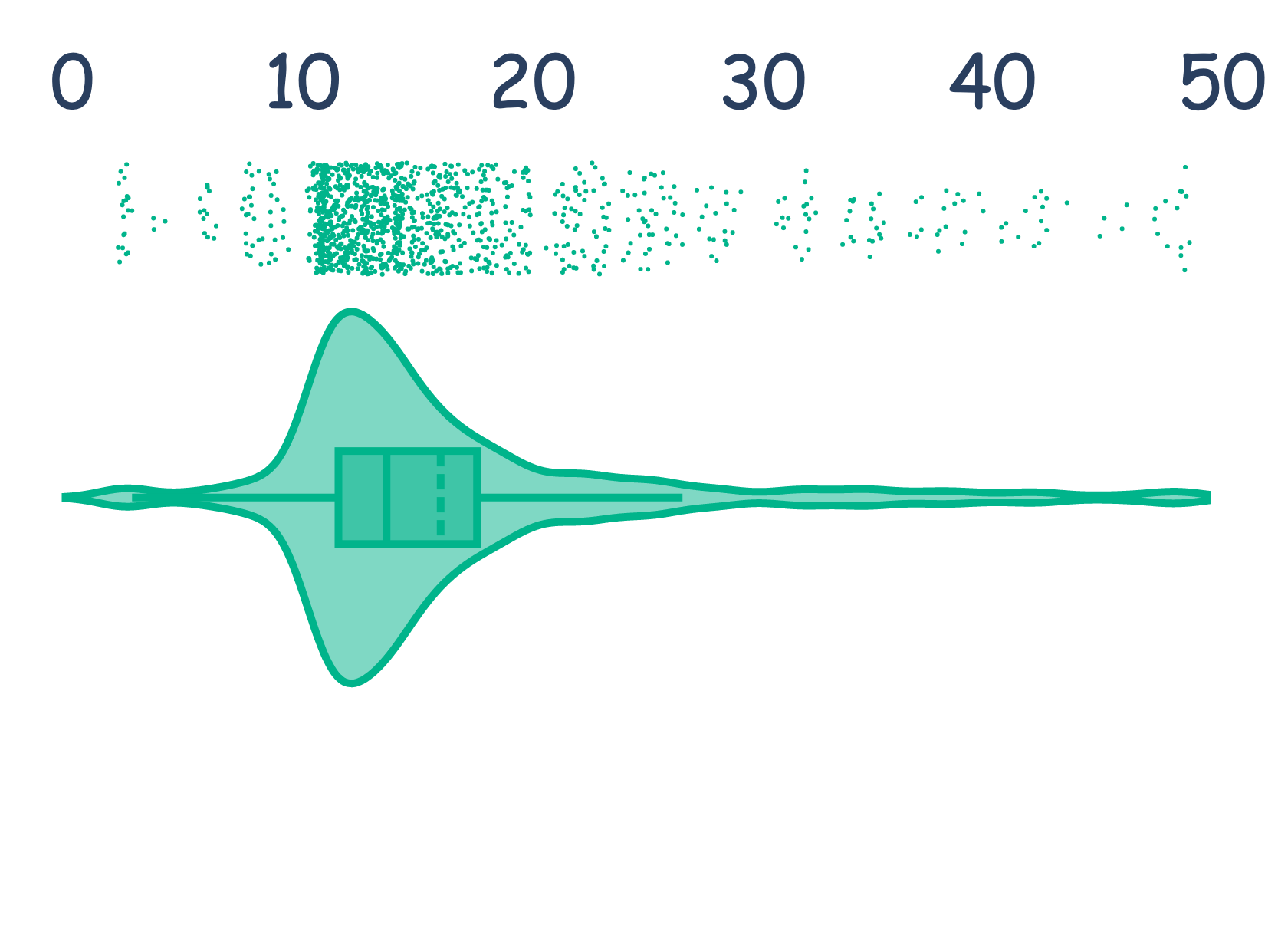}}\end{minipage} & \begin{minipage}[b]
{0.49\columnwidth}\centering\raisebox{-.4\height}{\includegraphics[width=\linewidth]{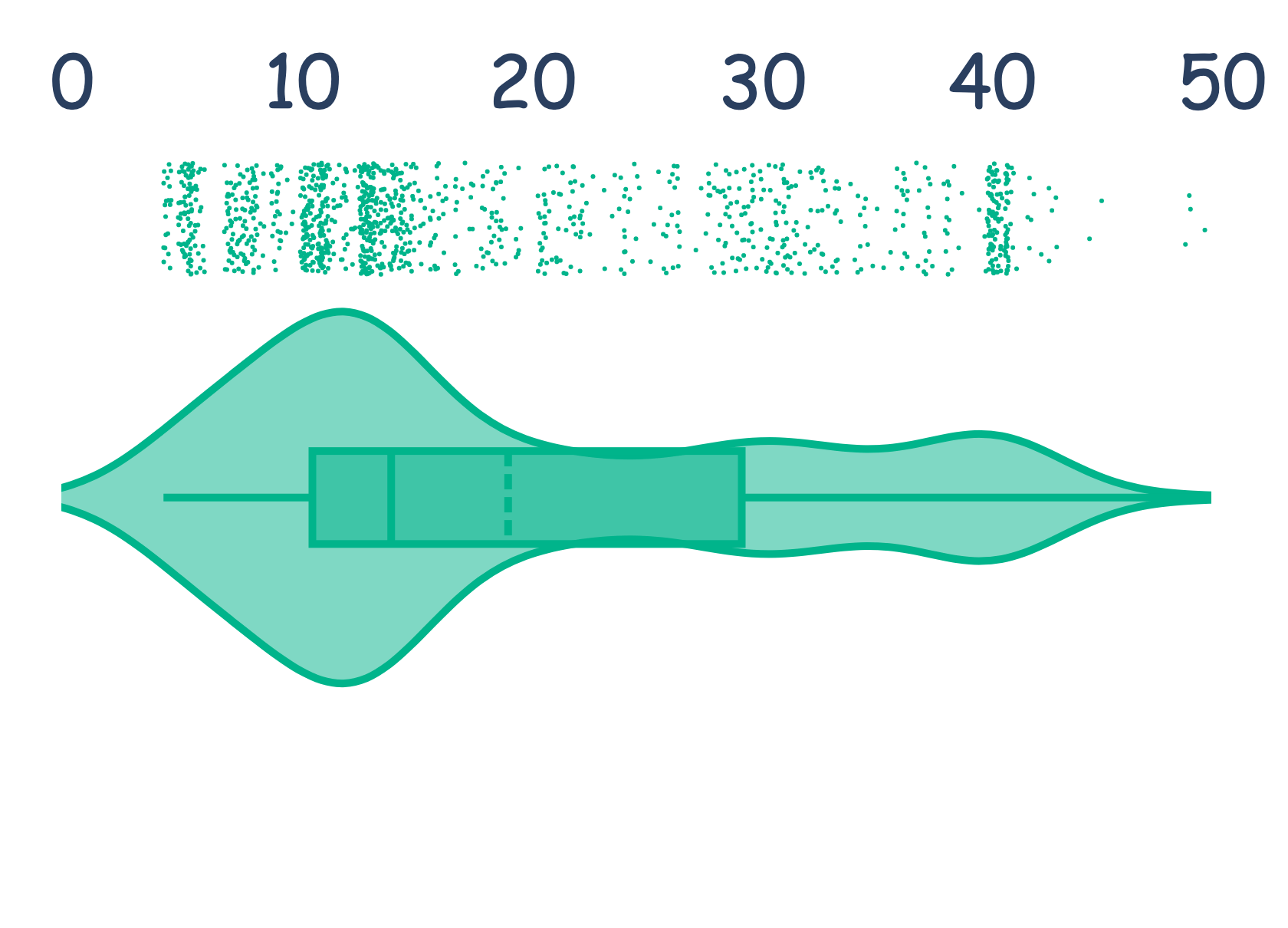}}\end{minipage} & \begin{minipage}[b]
{0.49\columnwidth}\centering\raisebox{-.4\height}{\includegraphics[width=\linewidth]{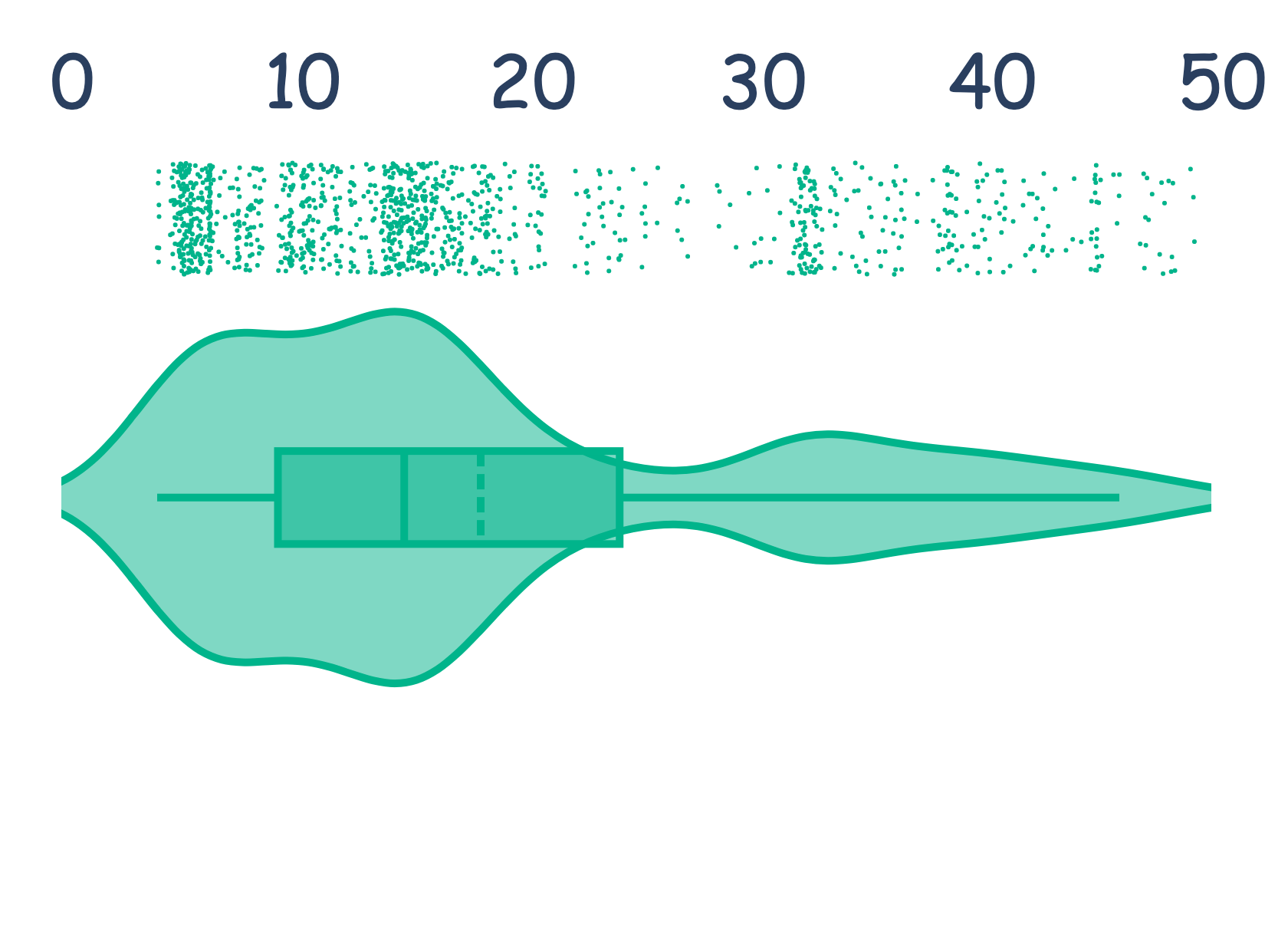}}\end{minipage}
\\
motorcyclist & bicyclist & pedestrian & traffic-sign
\\\midrule
\begin{minipage}[b]
{0.49\columnwidth}\centering\raisebox{-.4\height}{\includegraphics[width=\linewidth]{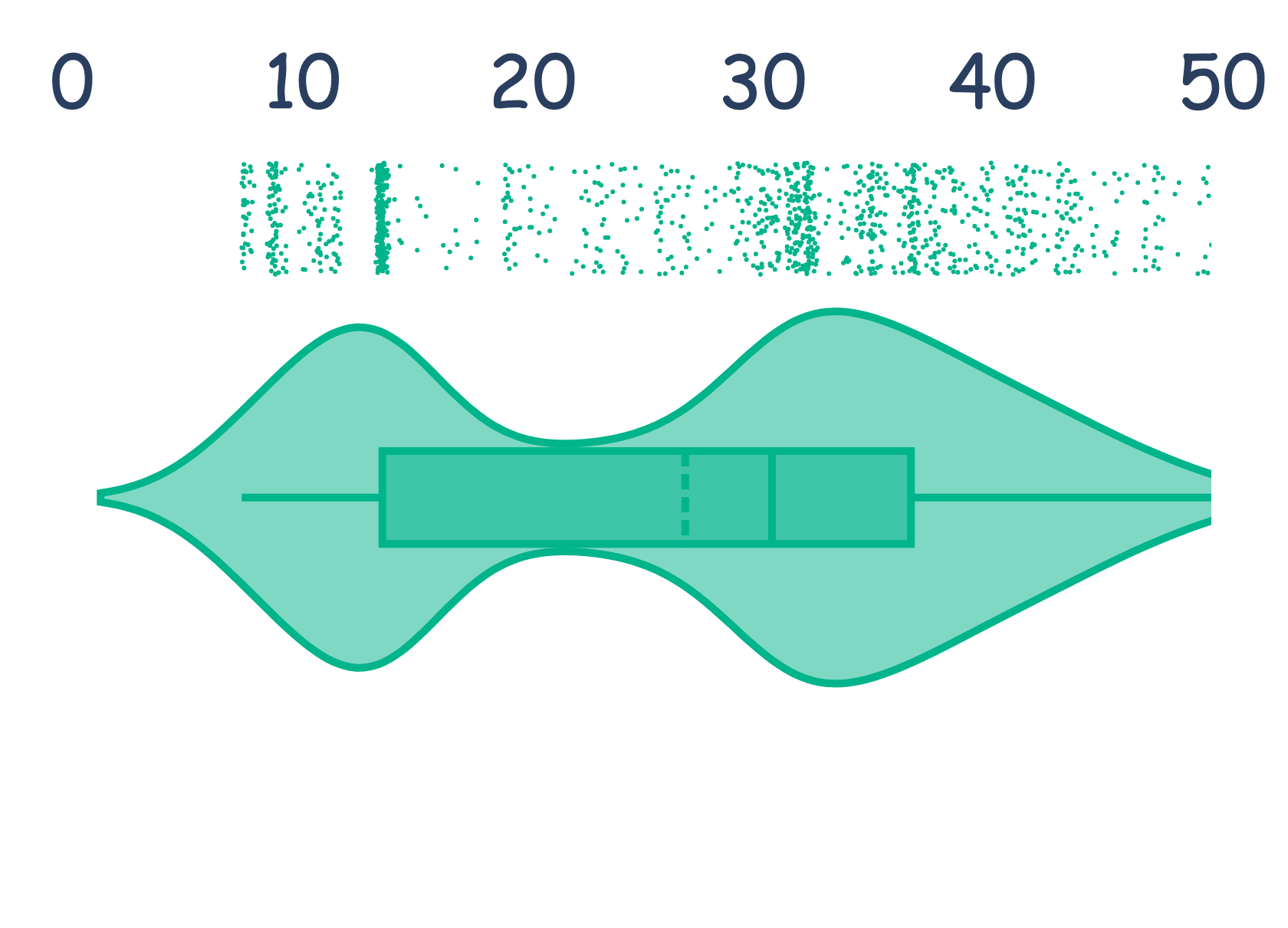}}\end{minipage} & \begin{minipage}[b]
{0.49\columnwidth}\centering\raisebox{-.4\height}{\includegraphics[width=\linewidth]{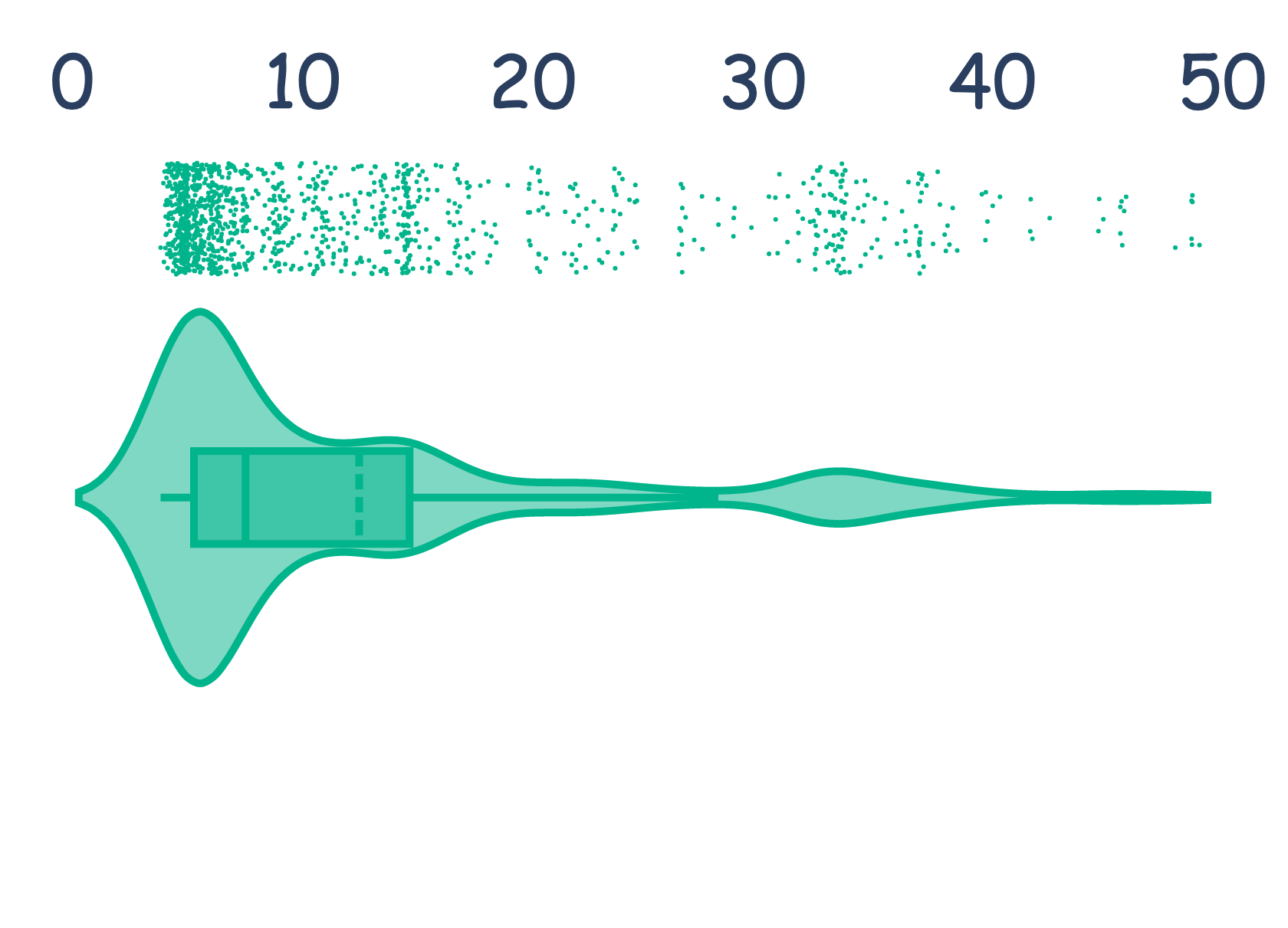}}\end{minipage} & \begin{minipage}[b]
{0.49\columnwidth}\centering\raisebox{-.4\height}{\includegraphics[width=\linewidth]{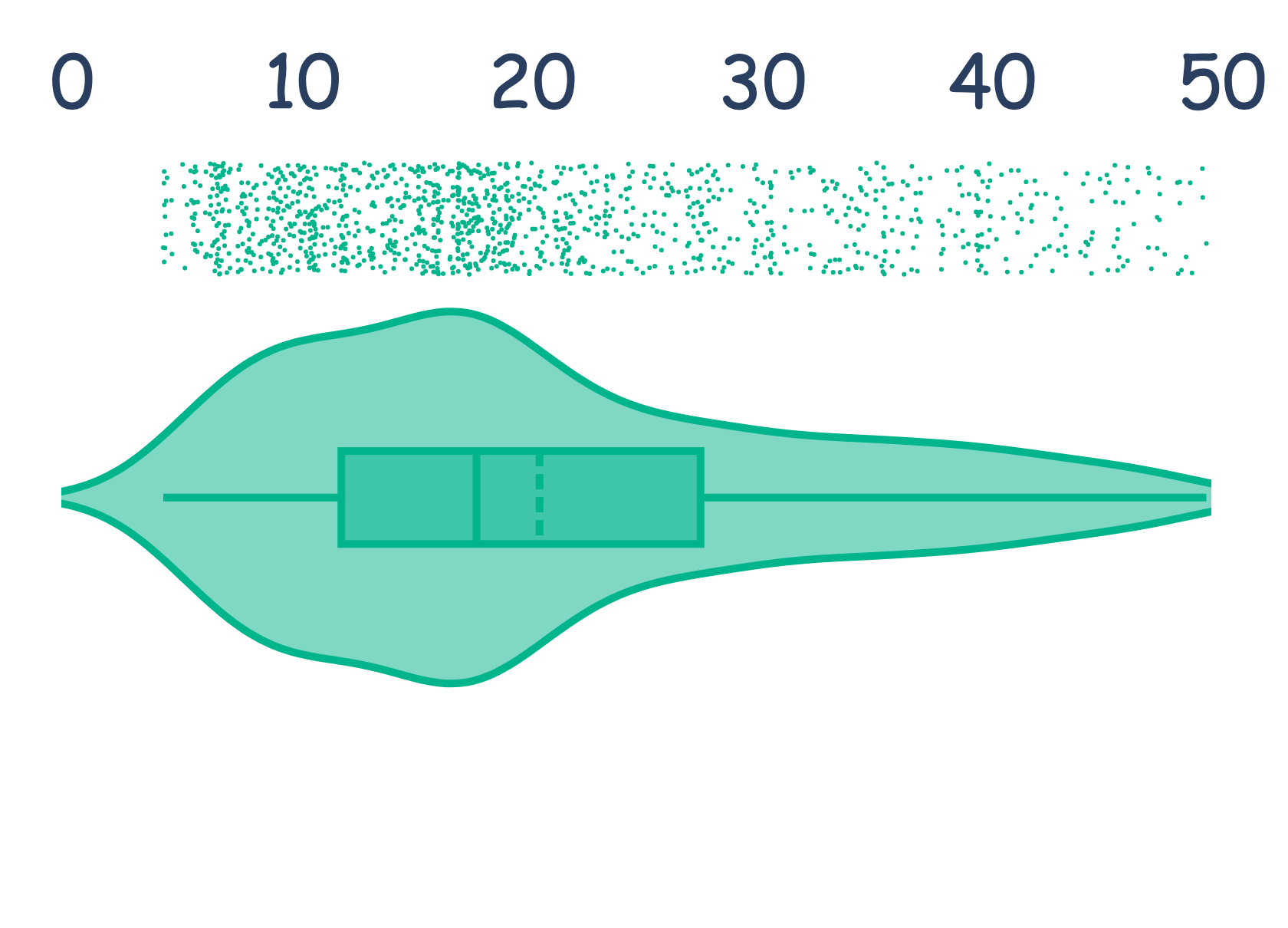}}\end{minipage} & \begin{minipage}[b]
{0.49\columnwidth}\centering\raisebox{-.4\height}{\includegraphics[width=\linewidth]{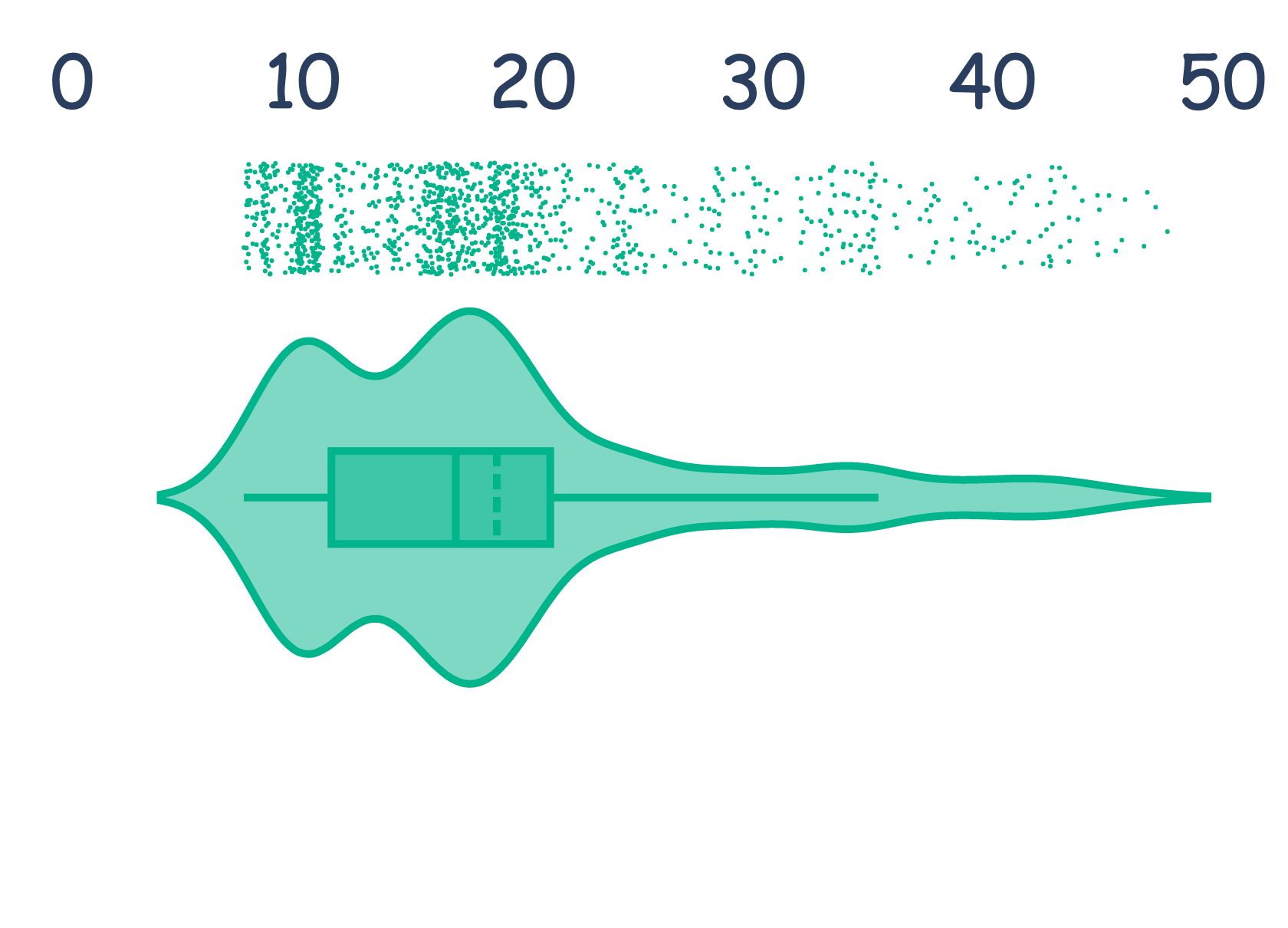}}\end{minipage}
\\
traffic-light & pole & construction-cone & bicycle
\\\midrule
\begin{minipage}[b]
{0.49\columnwidth}\centering\raisebox{-.4\height}{\includegraphics[width=\linewidth]{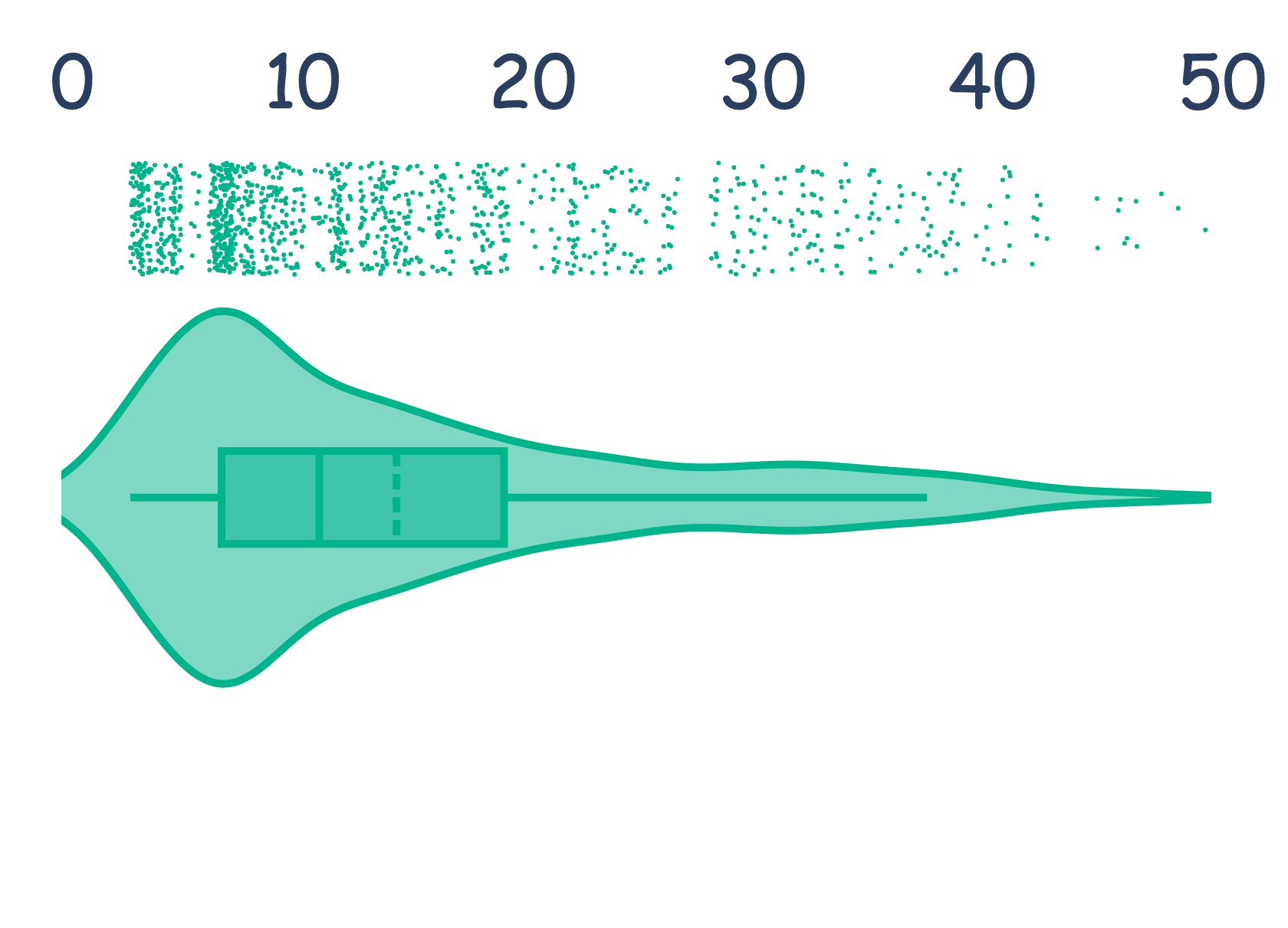}}\end{minipage} & \begin{minipage}[b]
{0.49\columnwidth}\centering\raisebox{-.4\height}{\includegraphics[width=\linewidth]{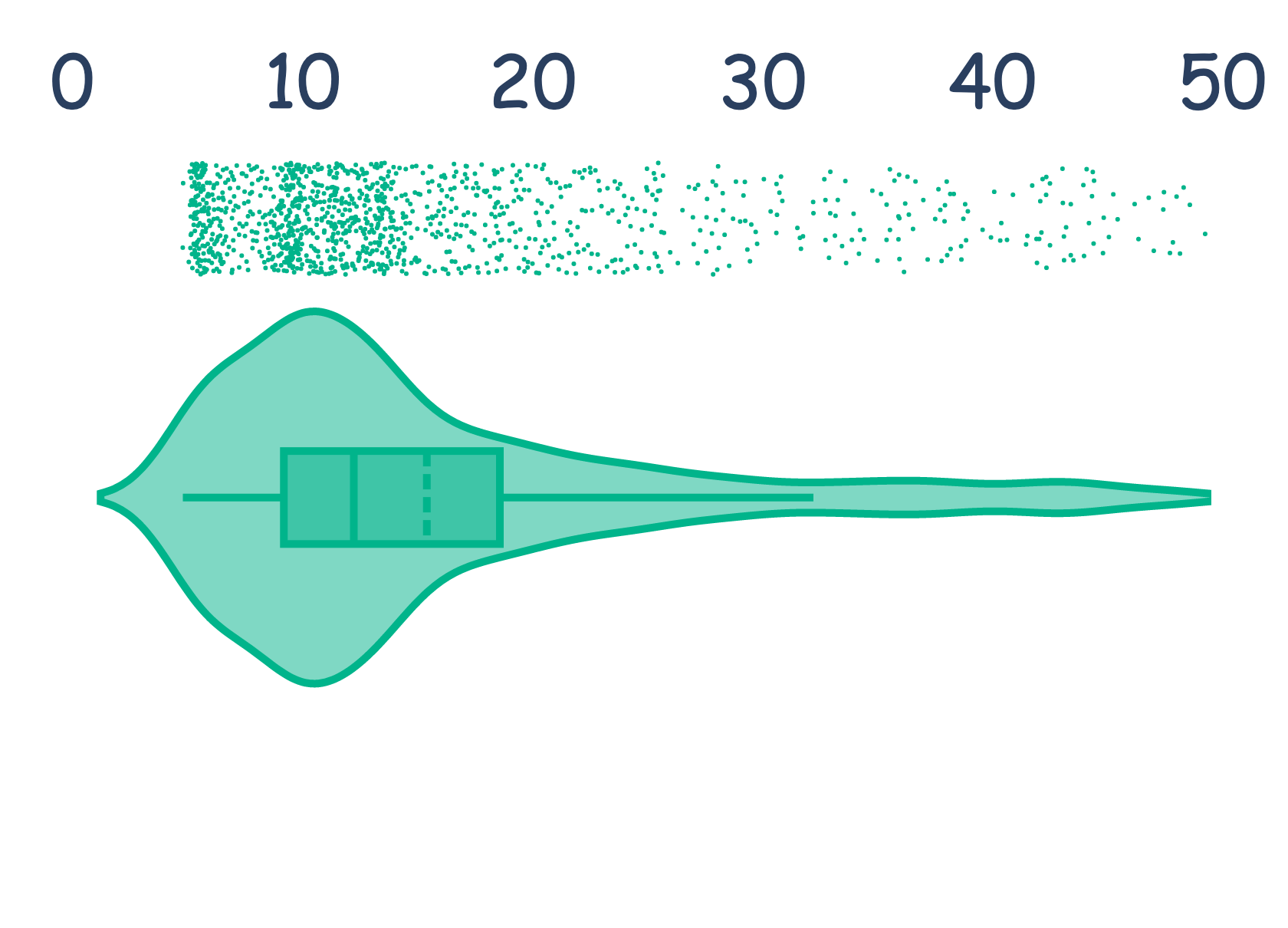}}\end{minipage} & \begin{minipage}[b]
{0.49\columnwidth}\centering\raisebox{-.4\height}{\includegraphics[width=\linewidth]{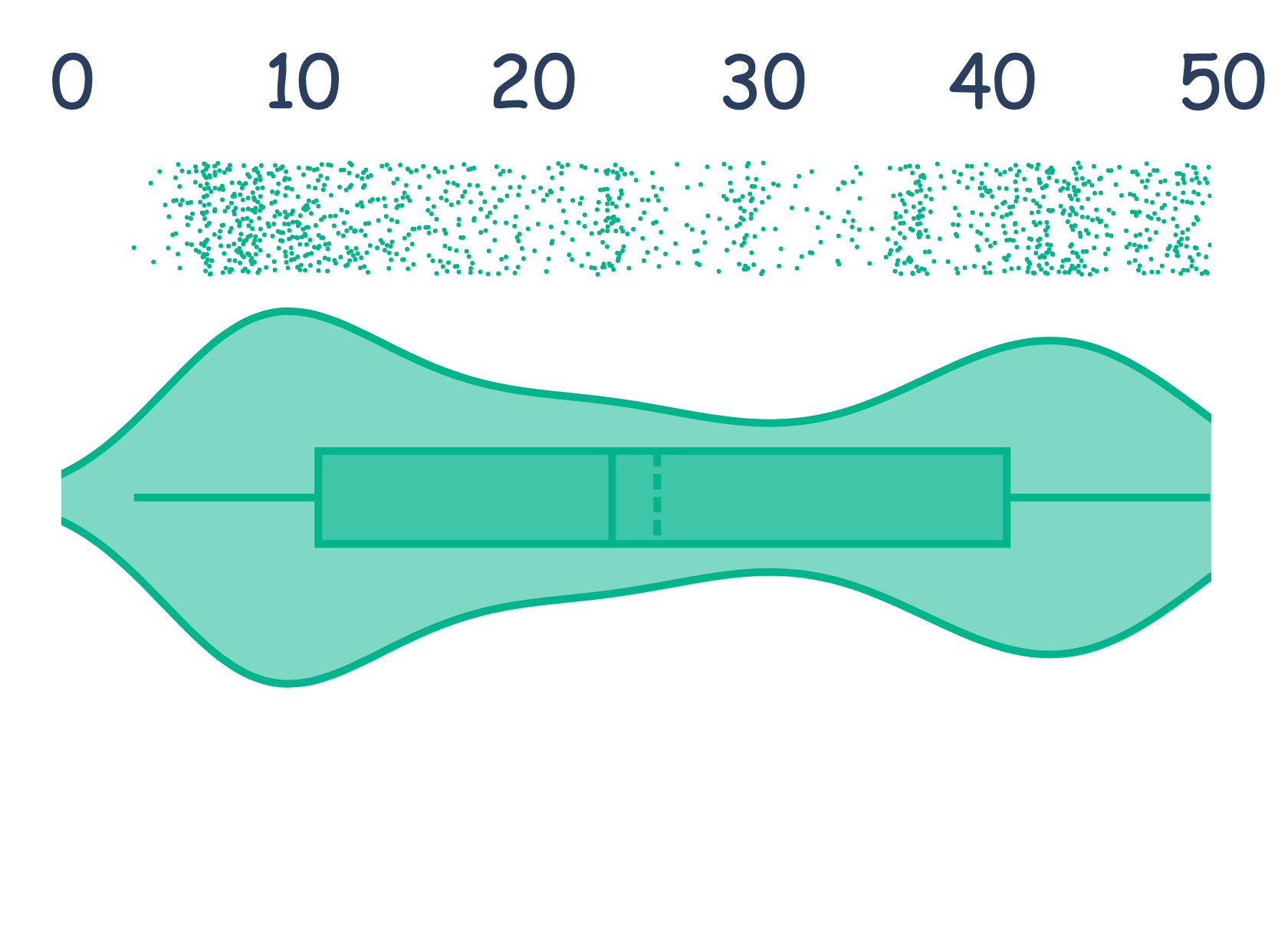}}\end{minipage} & \begin{minipage}[b]
{0.49\columnwidth}\centering\raisebox{-.4\height}{\includegraphics[width=\linewidth]{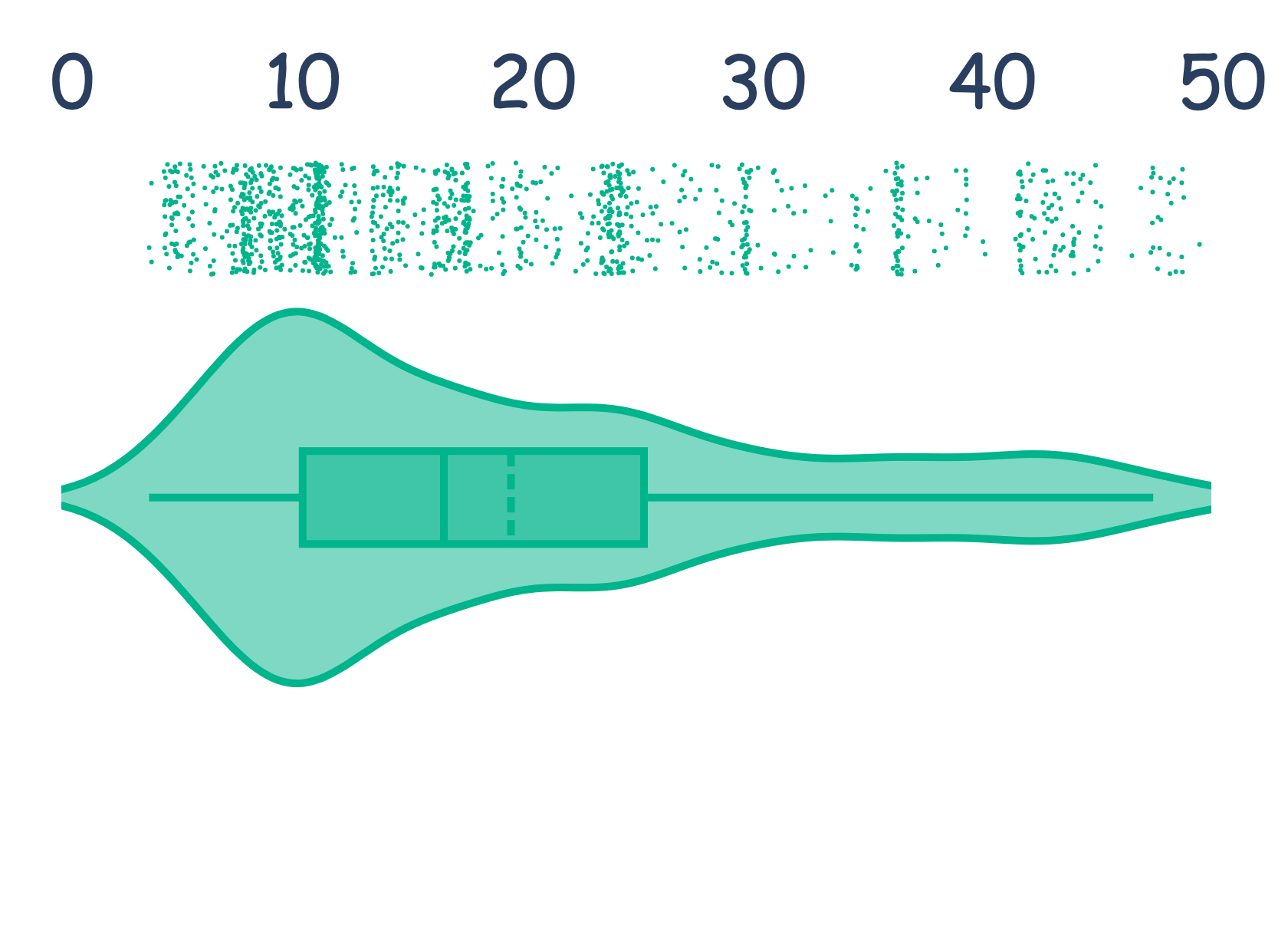}}\end{minipage}
\\
motorcycle & building & vegetation & tree-trunk
\\\midrule
\begin{minipage}[b]
{0.49\columnwidth}\centering\raisebox{-.4\height}{\includegraphics[width=\linewidth]{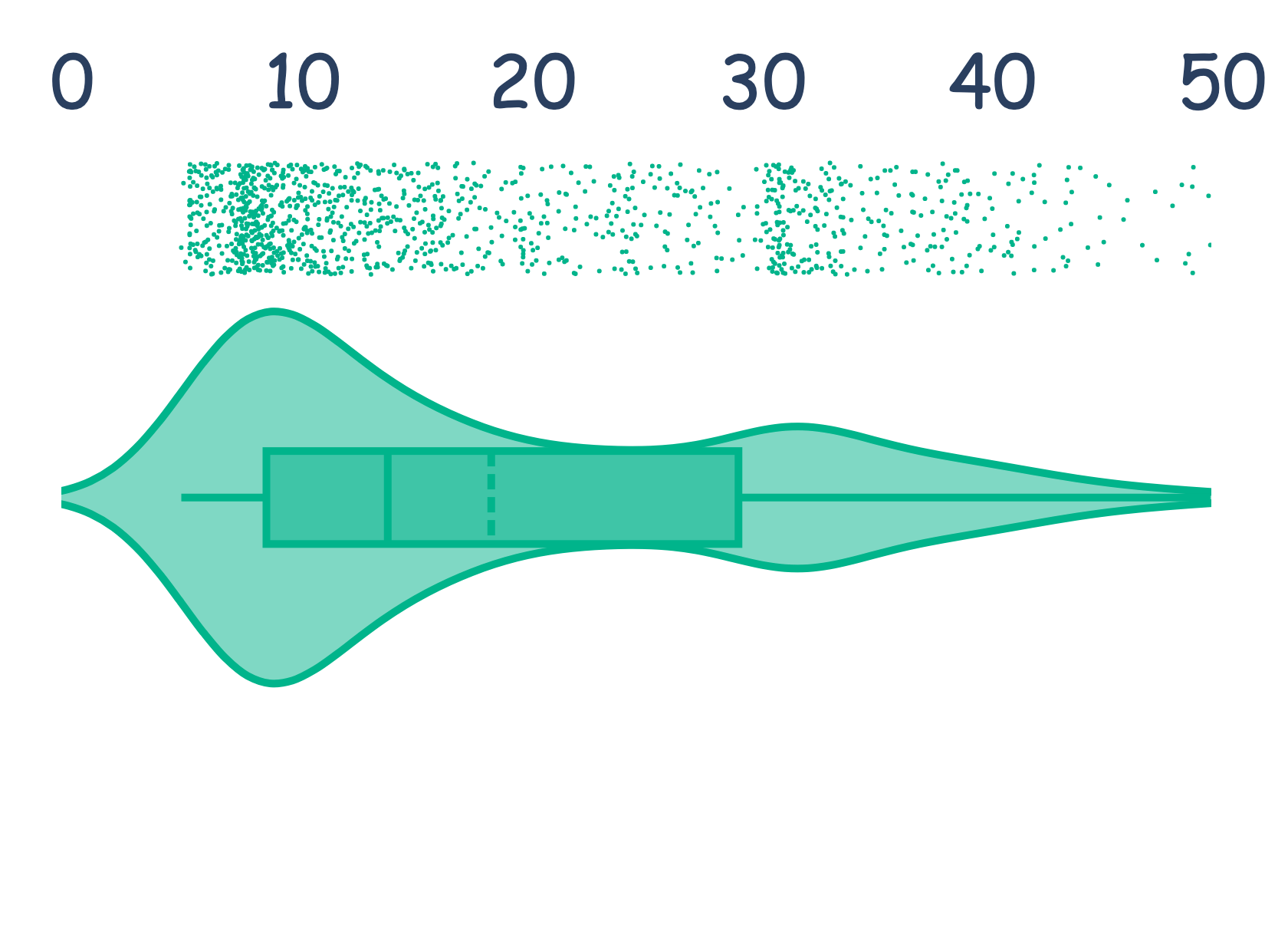}}\end{minipage} & \begin{minipage}[b]
{0.49\columnwidth}\centering\raisebox{-.4\height}{\includegraphics[width=\linewidth]{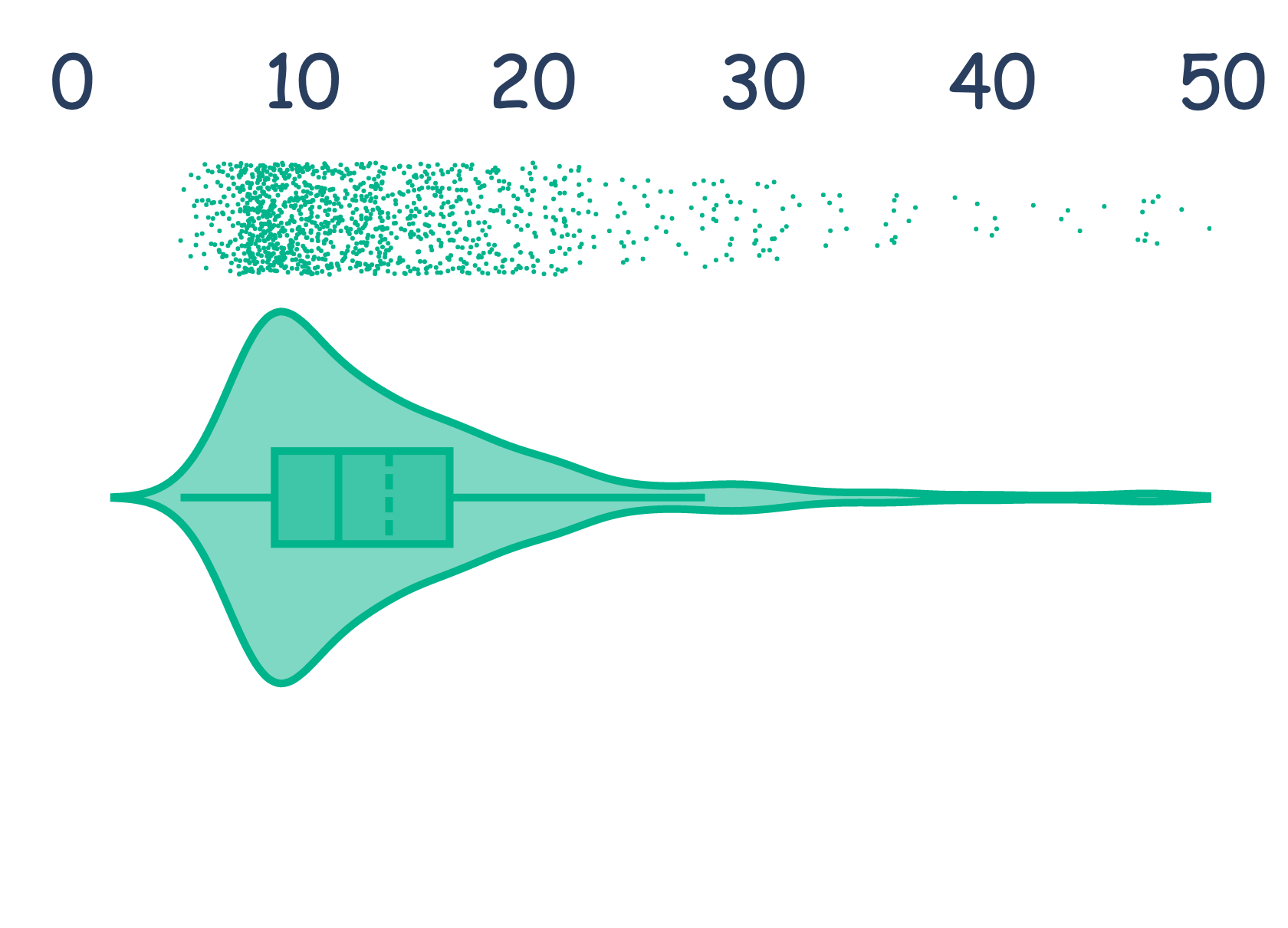}}\end{minipage} & \begin{minipage}[b]
{0.49\columnwidth}\centering\raisebox{-.4\height}{\includegraphics[width=\linewidth]{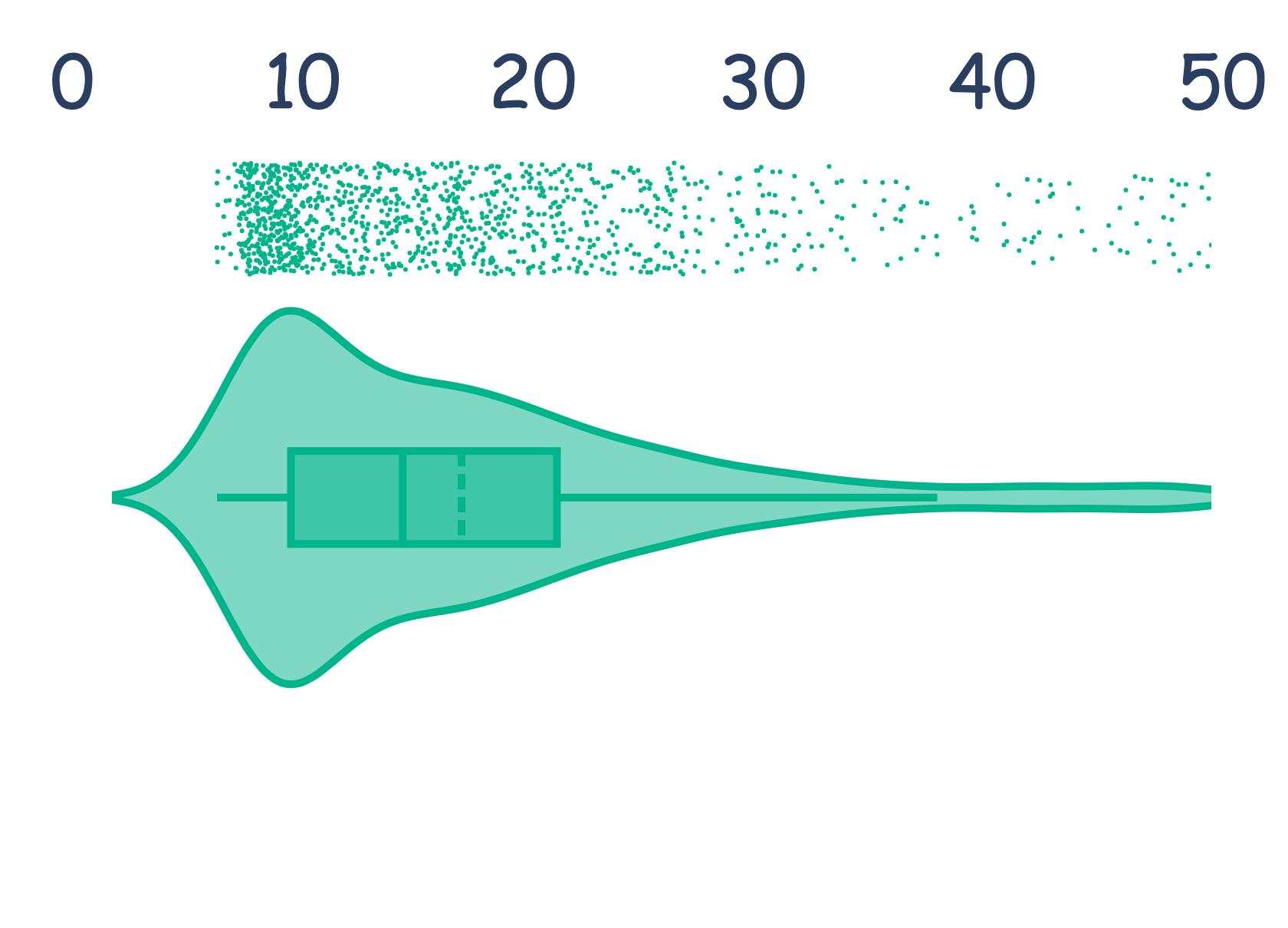}}\end{minipage} & \begin{minipage}[b]
{0.49\columnwidth}\centering\raisebox{-.4\height}{\includegraphics[width=\linewidth]{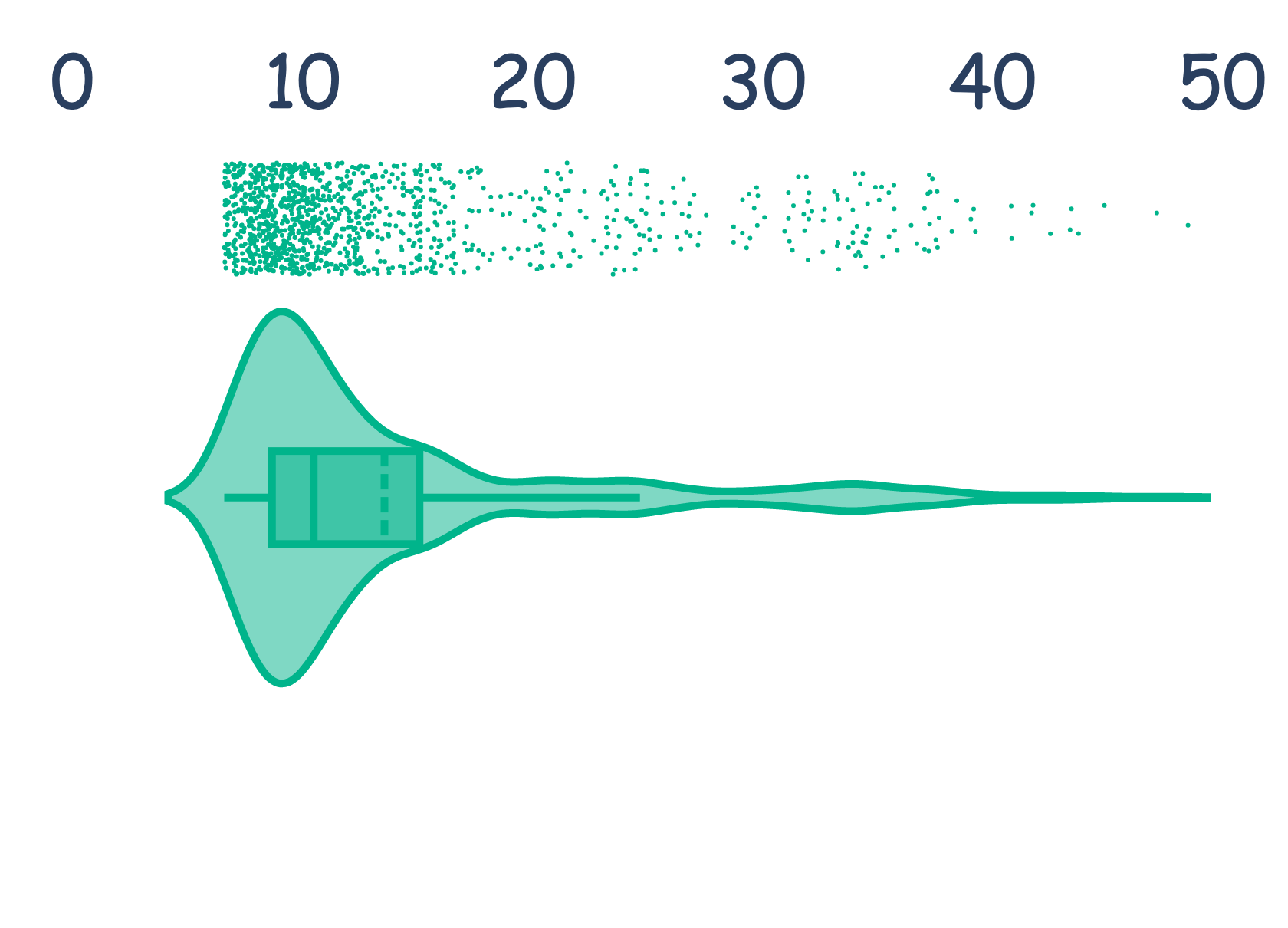}}\end{minipage}
\\
curb & road & lane-marker & other-ground
\\\midrule
\begin{minipage}[b]
{0.49\columnwidth}\centering\raisebox{-.4\height}{\includegraphics[width=\linewidth]{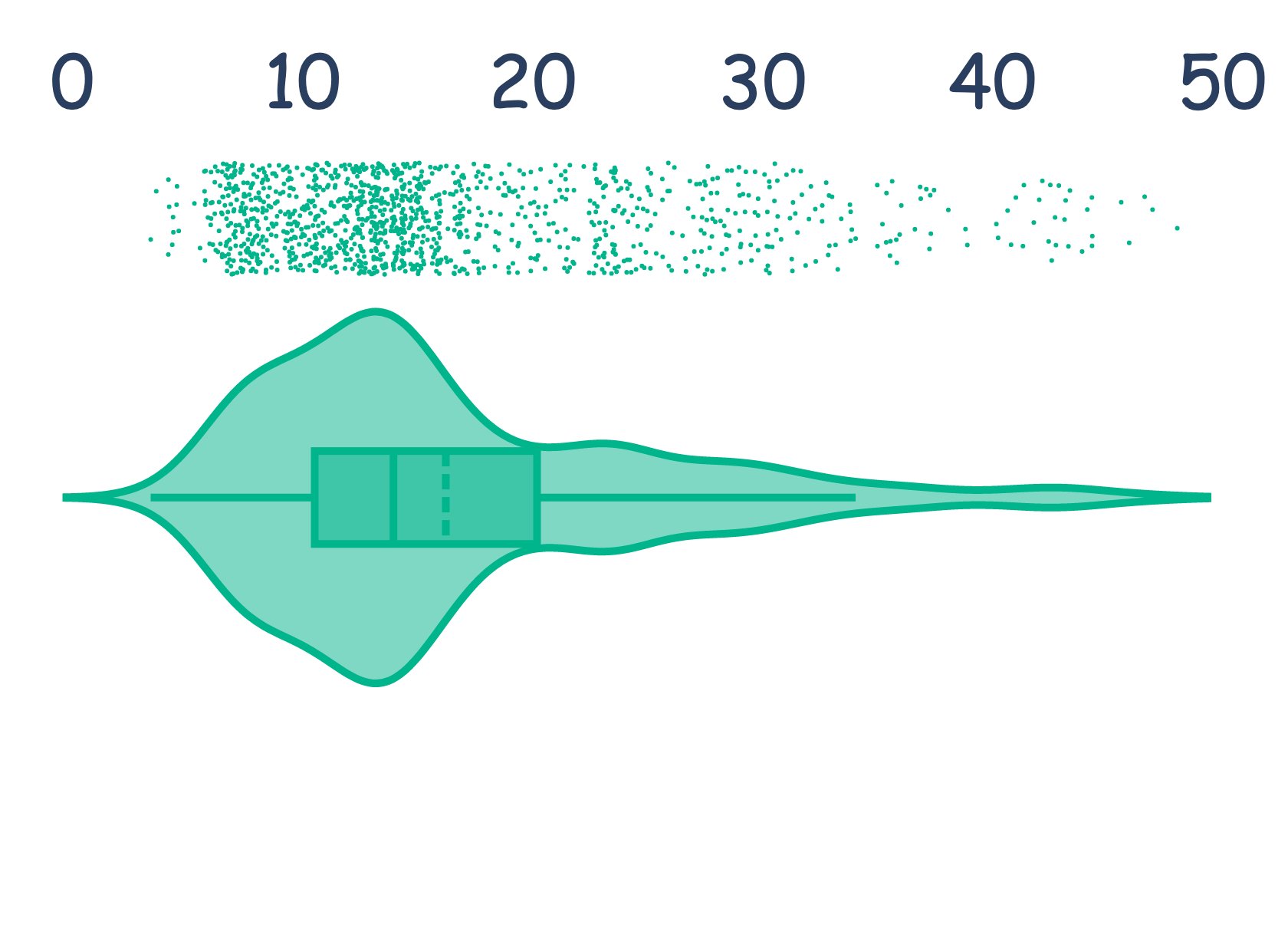}}\end{minipage} & \begin{minipage}[b]
{0.49\columnwidth}\centering\raisebox{-.4\height}{\includegraphics[width=\linewidth]{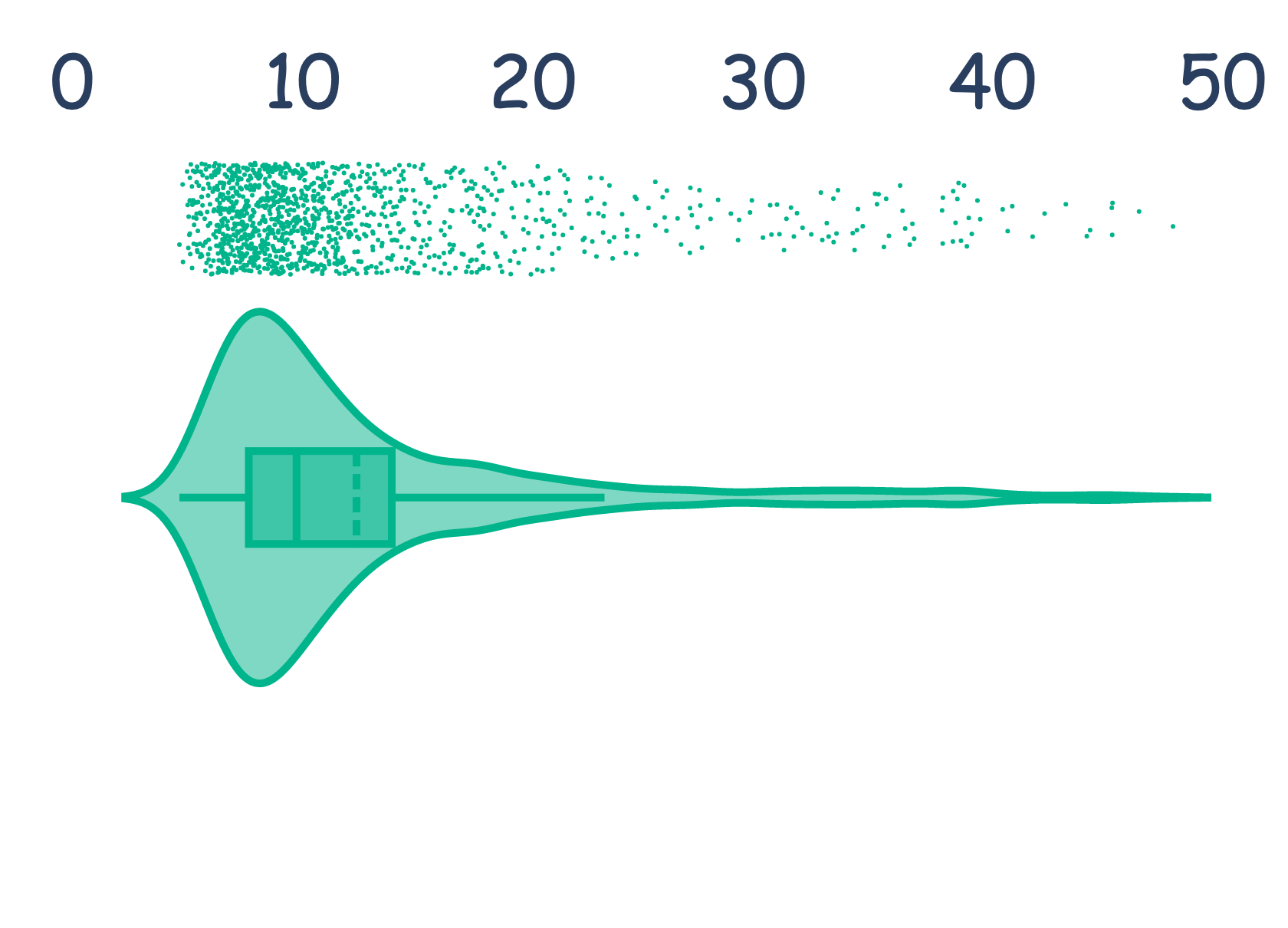}}\end{minipage} & 
\\
walkable & sidewalk & 
\\\bottomrule
\end{tabular}
\end{adjustbox}
\label{tab:dataset_waymo}
\end{table*}

\clearpage

\section{Multi-Task Configuration}
In this section, we supplement more details of our design and implementation toward multi-task (semantic and panoptic) LiDAR segmentation.

\subsection{Overview}
A proper pipeline design could enable the model to generate suitable predictions to fulfill multiple tasks simultaneously. In the context of LiDAR segmentation, we are especially interested in unifying semantic and panoptic segmentation of LiDAR point clouds. Such a holistic way of 3D scene understanding is crucial for the safe perception in autonomous vehicles.

\subsection{Mean Shift}
In this work, we enhance the versatility of our framework in an end-to-end fashion through the integration of a multi-tasking approach. This adaptation involves the modification of the instance extractor on top of the semantic predictions, which enables a dual output for both LiDAR semantic and panoptic segmentation. Specifically, drawing inspiration from DS-Net~\cite{hong2021dsnet,hong20224dDSNet}, our instance extractor comprises an instance head, succeeded by a point clustering step. The instance head encompasses a sequence of multi-layer perceptrons designed to predict the offsets between instance centers. This point clustering step strategically employs semantic predictions to filter out \textit{stuff} points, thereby retaining only those associated with \textit{thing} instances, such as \texttt{pedestrian}, \texttt{car}, and \texttt{bicyclist}. Subsequently, the remaining points undergo mean-shift clustering \cite{marcuzzi2002MeanShift}, utilizing features from the instance head to discern distinct instances. This meticulous process enhances the framework's capacity for accurate instance segmentation. The bandwidth for mean-shift in the \textit{SemanticKITTI} and \textit{Panoptic-nuScenes} datasets is set to $1.2$ and $2.5$, respectively.

\section{Additional Implementation Details}
In this section, we provide additional details to assist the implementation and reproduction of the approach proposed in the main body of this paper.

\subsection{Datasets}
In our multi-dataset training pipeline, we train our M3Net framework on the three most popular large-scale driving datasets, \ie, the \textit{SemanticKITTI} \cite{behley2019semanticKITTI}, \textit{nuScenes} \cite{fong2022panoptic-nuScenes}, and \textit{Waymo Open} \cite{sun2020waymoOpen} datasets. These three datasets consist of $19130$, $29130$, and $23691$ training LiDAR scans, and $4071$, $6019$, and $5976$ validation LiDAR scans, respectively. Besides, we leverage the synchronized camera images from the corresponding datasets as our 2D inputs in the M3Net training pipeline for cross-modality alignments. The \textit{SemanticKITTI}, \textit{nuScenes}, and \textit{Waymo Open} datasets contain $19130$, $174780$, and $118455$ camera images in the train set, respectively, where \textit{SemanticKITTI} has single-camera (front-view) data, \textit{nuScenes} is with a six-camera (three front-view and three back-view) systems, and \textit{Waymo Open} has five camera views in total.

For multi-task experiments on \textit{SemanticKITTI} \cite{behley2019semanticKITTI} and \textit{Panoptic-nuScenes} \cite{fong2022panoptic-nuScenes}, we follow the official data preparation procedures to set up the training and evaluations. Specifically, these two datasets share the same amount of data with their semantic segmentation subsets, \ie, $19130$ and $29130$ training LiDAR scans, and $4071$ and $6019$ validation LiDAR scans, respectively. Each LiDAR scan is associated with a panoptic segmentation map which indicates the instance IDs. For additional details, kindly refer to the original papers.

For the knowledge transfer fine-tuning experiments on the \textit{RELLIS-3D} \cite{jiang2021rellis3D}, \textit{SemanticPOSS} \cite{pan2020semanticPOSS}, \textit{SemanticSTF} \cite{xiao2023semanticSTF}, \textit{SynLiDAR} \cite{xiao2022synLiDAR} and \textit{DAPS-3D} \cite{klokov2023daps3D} datasets, we follow the same procedure as Seal~\cite{liu2023segment} to prepare the training and validation sets. Kindly refer to the original paper for more details on this aspect.

For the out-of-training-distribution generalization experiments on \textit{SemanticKITTI-C} and \textit{nuScenes-C}, we follow the same data preparation procedure in Robo3D \cite{kong2023robo3D}. There are eight different corruption types in each dataset, including fog, wet ground, snow, motion blur, beam missing, crosstalk, incomplete echo, and cross-sensor cases, where each corruption type contains corrupted data from three severity levels. In total, there are $97704$ LiDAR scans in \textit{SemanticKITTI-C} and $144456$ LiDAR scans in \textit{nuScenes-C}. For additional details, kindly refer to the original paper.

\subsection{Text Prompts}
In this work, we adopt the standard templates along with specified class text prompts to generate the CLIP text embedding for the three datasets used in our multi-dataset training pipeline. Specifically, the text prompts associated with the semantic classes in the \textit{{nuScenes}} \cite{fong2022panoptic-nuScenes}, \textit{{SemanticKITTI}} \cite{behley2019semanticKITTI}, and \textit{{Waymo Open}} \cite{sun2020waymoOpen} datasets are shown in \cref{tab:prompts_nuscenes}, \cref{tab:prompts_semantickitti}, and \cref{tab:prompts_waymo}, respectively.

\begin{table*}[t]
    \centering
    \caption{\textbf{Text prompts} defined for the \textit{nuScenes} \cite{fong2022panoptic-nuScenes} dataset ($16$ classes) in our proposed M3Net framework.}
    \vspace{-0.1cm}
    \scalebox{0.9}{\begin{tabular}{c|l|p{11.5cm}}
    \toprule
    \multicolumn{3}{c}{\textbf{nuScenes (16 classes)}}
    \\\midrule\midrule
    \textbf{\#} & \textbf{class} & \textbf{text prompt}
    \\\midrule
    $1$ & \texttt{barrier} & `barrier', `barricade'
    \\\midrule
    $2$ & \texttt{bicycle} & `bicycle'
    \\\midrule
    $3$ & \texttt{bus} & `bus'
    \\\midrule
    $4$ & \texttt{car} & `car'
    \\\midrule
    $5$ & \texttt{construction-vehicle} & `bulldozer', `excavator', `concrete mixer', `crane', `dump truck'
    \\\midrule
    $6$ & \texttt{motorcycle} & `motorcycle'
    \\\midrule
    $7$ & \texttt{pedestrian} & `pedestrian', `person'
    \\\midrule
    $8$ & \texttt{traffic-cone} & `traffic-cone'
    \\\midrule
    $9$ & \texttt{trailer} & `trailer', `semi-trailer', `cargo container', `shipping container', `freight container'
    \\\midrule
    $10$ & \texttt{truck} & `truck'
    \\\midrule
    $11$ & \texttt{driveable-surface} & `road'
    \\\midrule
    $12$ & \texttt{other-flat} & `curb', `traffic island', `traffic median'
    \\\midrule
    $13$ & \texttt{sidewalk} & `sidewalk'
    \\\midrule
    $14$ & \texttt{terrain} & `grass', `grassland', `lawn', `meadow', `turf', `sod'
    \\\midrule
    $15$ & \texttt{manmade} & `building', `wall', `pole', `awning'
    \\\midrule
    $16$ & \texttt{vegetation} & `tree', `trunk', `tree trunk', `bush', `shrub', `plant', `flower', `woods'
    \\\bottomrule
    \end{tabular}}
    \label{tab:prompts_nuscenes}
\end{table*}
\begin{table*}[t]
    \centering
    \caption{\textbf{Text prompts} defined for the \textit{SemanticKITTI} \cite{behley2019semanticKITTI} dataset ($19$ classes) in our proposed M3Net framework.}
    \vspace{-0.1cm}
    \scalebox{0.9}{\begin{tabular}{c|l|p{12.5cm}}
    \toprule
    \multicolumn{3}{c}{\textbf{SemanticKITTI (19 classes)}}
    \\\midrule\midrule
    \textbf{\#} & \textbf{class} & \textbf{text prompt}
    \\\midrule
    $1$ & \texttt{car} & `car'
    \\\midrule
    $2$ & \texttt{bicycle} & `bicycle'
    \\\midrule
    $3$ & \texttt{motorcycle} & `motorcycle'
    \\\midrule
    $4$ & \texttt{truck} & `truck'
    \\\midrule
    $5$ & \texttt{other-vehicle} & `other vehicle', `bulldozer', `excavator', `concrete mixer', `crane', `dump truck', `bus', `trailer', `semi-trailer', `cargo container', `shipping container', `freight container'
    \\\midrule
    $6$ & \texttt{person} & `person'
    \\\midrule
    $7$ & \texttt{bicyclist} & `bicyclist'
    \\\midrule
    $8$ & \texttt{motorcyclist} & `motorcyclist'
    \\\midrule
    $9$ & \texttt{road} & `road'
    \\\midrule
    $10$ & \texttt{parking} & `parking'
    \\\midrule
    $11$ & \texttt{sidewalk} & `sidewalk'
    \\\midrule
    $12$ & \texttt{other-ground} & `other ground', `curb', `traffic island', `traffic median'
    \\\midrule
    $13$ & \texttt{building} & `building'
    \\\midrule
    $14$ & \texttt{fence} & `fence'
    \\\midrule
    $15$ & \texttt{vegetation} & `tree'
    \\\midrule
    $16$ & \texttt{trunk} & `tree trunk', `trunk'
    \\\midrule
    $17$ & \texttt{terrain} & `grass', `grassland', `lawn', `meadow', `turf', `sod'
    \\\midrule
    $18$ & \texttt{pole} & `pole'
    \\\midrule
    $19$ & \texttt{traffic sign} & `traffic sign'
    \\\bottomrule
    \end{tabular}}
    \label{tab:prompts_semantickitti}
\vspace{0.1cm}
\end{table*}
\begin{table*}[t]
    \centering
    \caption{\textbf{Text prompts} defined for the \textit{Waymo Open} \cite{sun2020waymoOpen} dataset ($22$ classes) in our proposed M3Net framework.}
    \vspace{-0.1cm}
    \scalebox{0.9}{\begin{tabular}{c|l|p{11.5cm}}
    \toprule
    \multicolumn{3}{c}{\textbf{Waymo Open (22 classes)}}
    \\\midrule\midrule
    \textbf{\#} & \textbf{class} & \textbf{text prompt}
    \\\midrule
    $1$ & \texttt{car} & `car'
    \\\midrule
    $2$ & \texttt{truck} & `truck'
    \\\midrule
    $3$ & \texttt{bus} & `bus'
    \\\midrule
    $4$ & \texttt{other-vehicle} & `other vehicle', `pedicab', `construction vehicle', `recreational vehicle', `limo', `tram', `trailer', `semi-trailer', `cargo container', `shipping container', `freight container', `bulldozer', `excavator', `concrete mixer', `crane', `dump truck'
    \\\midrule
    $5$ & \texttt{motorcyclist} & `motorcyclist'
    \\\midrule
    $6$ & \texttt{bicyclist} & `bicyclist'
    \\\midrule
    $7$ & \texttt{pedestrian} & `person', `pedestrian'
    \\\midrule
    $8$ & \texttt{traffic-sign} & `traffic sign', `parking sign', `direction sign', `traffic sign without pole', `traffic light box'
    \\\midrule
    $9$ & \texttt{traffic-light} & `traffic light'
    \\\midrule
    $10$ & \texttt{pole} & `lamp post', `traffic sign pole'
    \\\midrule
    $11$ & \texttt{construction-cone} & `construction cone'
    \\\midrule
    $12$ & \texttt{bicycle} & `bicycle'
    \\\midrule
    $13$ & \texttt{motorcycle} & `motorcycle'
    \\\midrule
    $14$ & \texttt{building} & `building'
    \\\midrule
    $15$ & \texttt{vegetation} & `bushes', `tree branches', `tall grasses', `flowers', `grass', `grassland', `lawn', `meadow', `turf', `sod'
    \\\midrule
    $16$ & \texttt{tree-trunk} & `tree trunk', `trunk'
    \\\midrule
    $17$ & \texttt{curb} & `curb'
    \\\midrule
    $18$ & \texttt{road} & `road'
    \\\midrule
    $19$ & \texttt{lane-marker} & `lane marker'
    \\\midrule
    $20$ & \texttt{other-ground} & `other ground', `bumps', `cateyes', `railtracks'
    \\\midrule
    $21$ & \texttt{walkable} & `walkable', `grassy hill', `pedestrian walkway stairs'
    \\\midrule
    $22$ & \texttt{sidewalk} & `sidewalk'
    \\\bottomrule
    \end{tabular}}
    \label{tab:prompts_waymo}
\end{table*}

\subsection{Backbones}
In this work, we adopt two models to serve as the backbone of our proposed M3Net, \ie, the classical MinkUNet \cite{choy2019minkowski} and the more recent PTv2+ \cite{wu2022ptv2}. 

\subsubsection{MinkUNet}
The primary contribution of MinkUNet \cite{choy2019minkowski} is the introduction of a neural network architecture capable of processing 4D spatiotemporal data (3D space + time). This is particularly relevant for applications that involve dynamic environments, like autonomous driving, where understanding the temporal evolution of the scene is crucial. A key feature of the Minkowski convolution, and by extension MinkUNet, is its ability to perform convolutional operations on sparse data. This is achieved through the use of a generalized sparse convolution operation that can handle data in high-dimensional spaces while maintaining computational efficiency. The implementation of Minkowski convolutions is facilitated by the Minkowski Engine, a framework for high-dimensional sparse tensor operations. This engine enables the efficient implementation of the MinkUNet and other similar architectures. In this work, we resort to the Pointcept \cite{pointcept2023} implementation of MinkUNet and adopt the base version as our backbone network in M3Net. More details of this used backbone can be found at \url{https://github.com/Pointcept/Pointcept}.

\subsubsection{PTv2+}

PTv2+ \cite{wu2022ptv2} introduces an effective grouped vector attention (GVA) mechanism. GVA facilitates efficient information exchange both within and among attention groups, significantly enhancing the model's ability to process complex point cloud data. PTv2+ also introduces an improved position encoding scheme. This enhancement allows for better utilization of point cloud coordinates, thereby bolstering the spatial reasoning capabilities of the model. The additional position encoding multiplier strengthens the position information for attention, allowing for more accurate and detailed data processing. Extensive experiments demonstrate that PTv2+ achieves state-of-the-art performance on several challenging 3D point cloud understanding benchmarks. In this work, we resort to the Pointcept \cite{pointcept2023} implementation of PTv2+ implementation of MinkUNet and adopt the base version as our backbone network in M3Net. More details of this used backbone can be found at \url{https://github.com/Pointcept/Pointcept}.

\subsection{Training Configuration}
In this work, we implement the proposed M3Net framework based on Pointcept \cite{pointcept2023} and MMDetection3D \cite{mmdet3d2020}. We trained our baselines and M3Net on four A$100$ GPUs each with $80$ GB memory. We adopt the AdamW optimizer \cite{loshchilov2019adamw} with a weight decay of $0.005$ and a learning rate of $0.002$. The learning rate scheduler utilized is cosine decay and the batch size is set to $6$ for each GPU. 

In the data-specific rasterization process, we rasterize the point clouds with voxel sizes tailored to the dataset characteristics. Specifically, we set the voxel sizes to [$0.05$m, $0.05$m, $0.05$m], [$0.1$m, $0.1$m, $0.1$m], and [$0.05$m, $0.05$m, $0.05$m] for the \textit{SemanticKITTI} \cite{behley2019semanticKITTI}, \textit{nuScenes} \cite{fong2022panoptic-nuScenes}, and \textit{Waymo Open} \cite{sun2020waymoOpen} datasets, respectively. 

For data augmentation, we leverage several techniques, including random flips along the $X$, $Y$, and $XY$ axes, and random jittering within the range of [-$0.02$m, $0.02$m]. Additionally, we incorporate global scaling and rotation, choosing scaling factors and rotation angles randomly from the intervals [$0.9$, $1.1$] and [$0$, $2\pi$], respectively. Furthermore, we integrate Mix3D~\cite{nekrasov2021mix3d} into our augmentation strategy during the training. There also exists some other augmentation techniques, such as LaserMix \cite{kong2022laserMix}, PolarMix \cite{xiao2022polarmix}, RangeMix \cite{kong2023rethinking,kong2023conDA}, and FrustumMix \cite{xu2023frnet}.

For the network backbones, we have opted for MinkUNet~\cite{choy2019minkowski} and PTv2+~\cite{wu2022ptv2}. In the case of MinkUNet, the encoder channels are set as $\{32, 64, 128, 256\}$, and the decoder channels are $\{256, 128, 64, 64\}$, each with a kernel size of 3. Meanwhile, for the PTv2+, the encoder channels are $\{32, 64, 128, 256, 512\}$, and the decoder channels are $\{64, 64, 128, 256\}$. For additional details, kindly refer to the original papers.

For the loss function, we incorporate the conventional cross-entropy loss and the Lovasz-softmax \cite{berman2018lovasz} loss to provide optimization for the LiDAR semantic and panoptic segmentation task. Additionally, we employ the L1 loss to optimize the instance head, aiding in the regression of precise instance offsets. 

\subsection{Evaluation Configuration}
In this work, we follow the conventional reporting and employ the Intersection-over-Union (IoU) for individual classes and the mean Intersection-over-Union (mIoU) across all classes as our evaluation metrics for LiDAR semantic segmentation. Specifically, the IoU score for semantic class $c$ is computed as follows:
\begin{equation}
IoU_{c} = \frac{TP_{c}}{TP_{c} + FP_{c} + FN_{c}}~.
\end{equation}
Here, $TP_{c}$, $FP_{c}$, and $FN_{c}$ represent the true positive, false positive, and false negative of class $c$, respectively. The mIoU score on each dataset is calculated by averaging the IoU scores across every semantic class. Notably, following recent works \cite{yan20202dpass,wu2022ptv2}, we report mIoU with Test Time Augmentation (TTA). For additional details, kindly refer to the original papers.

For panoptic LiDAR segmentation, we follow conventional reporting and utilize the Panoptic Quality (PQ) as our primary metric. The definition and calculation of the Panoptic Quality (PQ), Segmentation Quality (SQ), and Recognition Quality (RQ) scores are given as follows:
\begin{equation}
    \text{PQ} = \underbrace{\frac{\sum_{(i, j)\in TP}\text{IoU}(i, j)}{|TP|}}_\text{SQ} \times \underbrace{\frac{|TP|}{|TP| + \frac{1}{2}|FP| + \frac{1}{2}|FN|}}_\text{RQ}~.
\end{equation}
The three aforementioned metrics are also calculated individually for \textit{things} and \textit{stuff} classes, resulting in PQ$^{th}$, SQ$^{th}$, RQ$^{th}$, and PQ$^{st}$, SQ$^{st}$, RQ$^{st}$. Additionally, we also report the PQ$^\dagger$ score as widely used in many prior works \cite{hong2021dsnet,zhou2021panoptic,li2022panoptic,sirohi2021efficientlps}. This metric is defined by exchanging the PQ of each \textit{stuff} class with its IoU and then averaging across all semantic classes. For additional details, kindly refer to the original papers.

\begin{figure*}[t]
    \vspace{-0.37cm}
    \centering
    \begin{subfigure}[b]{.49\textwidth}
         \centering
         \includegraphics[width=\textwidth]{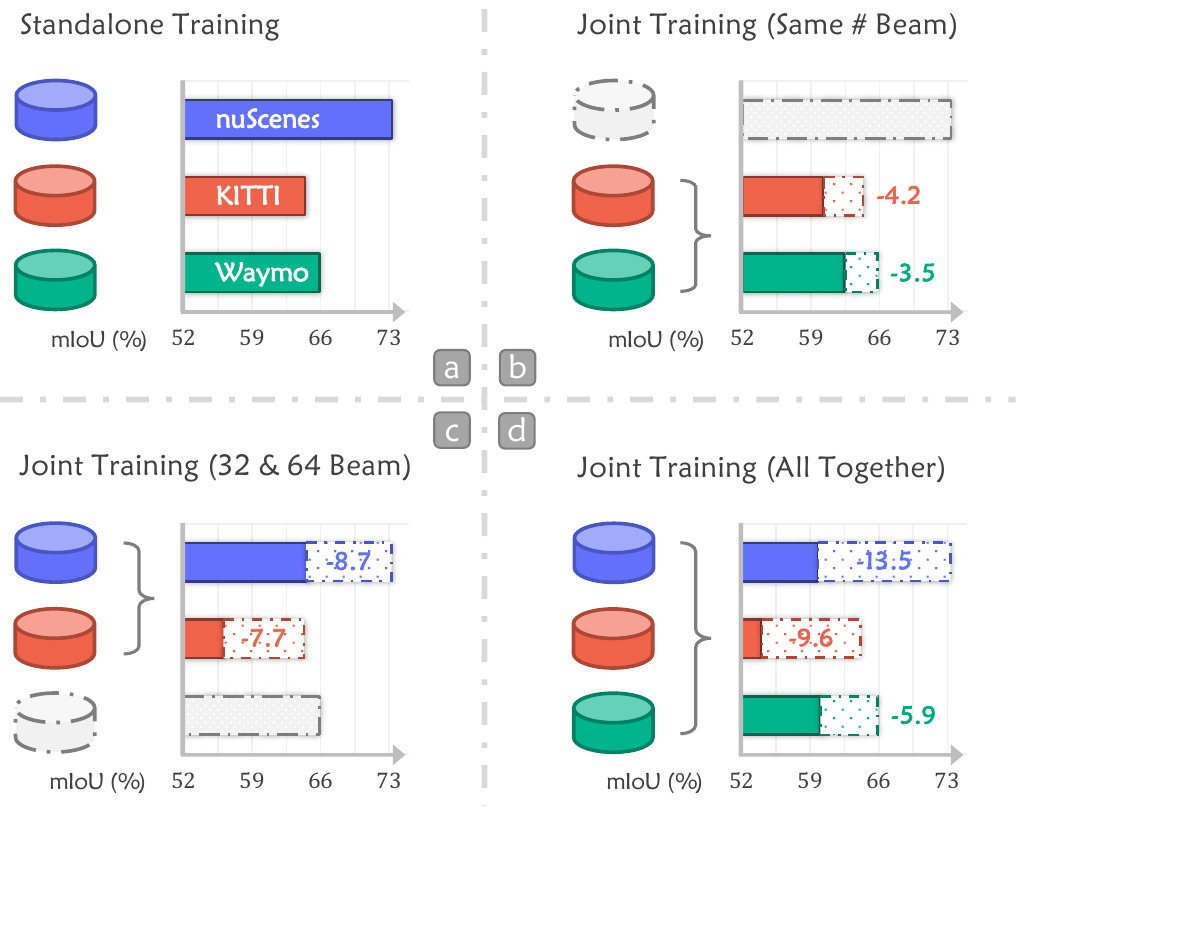}
         \vspace{0.05cm}
         \caption{M3Net (\textit{w/} MinkUNet \cite{choy2019minkowski} backbone)}
         \label{fig:ablation_pilot_study_minkunet}
    \end{subfigure}
    \hfill
    \begin{subfigure}[b]{.49\textwidth}
         \centering
         \includegraphics[width=\textwidth]{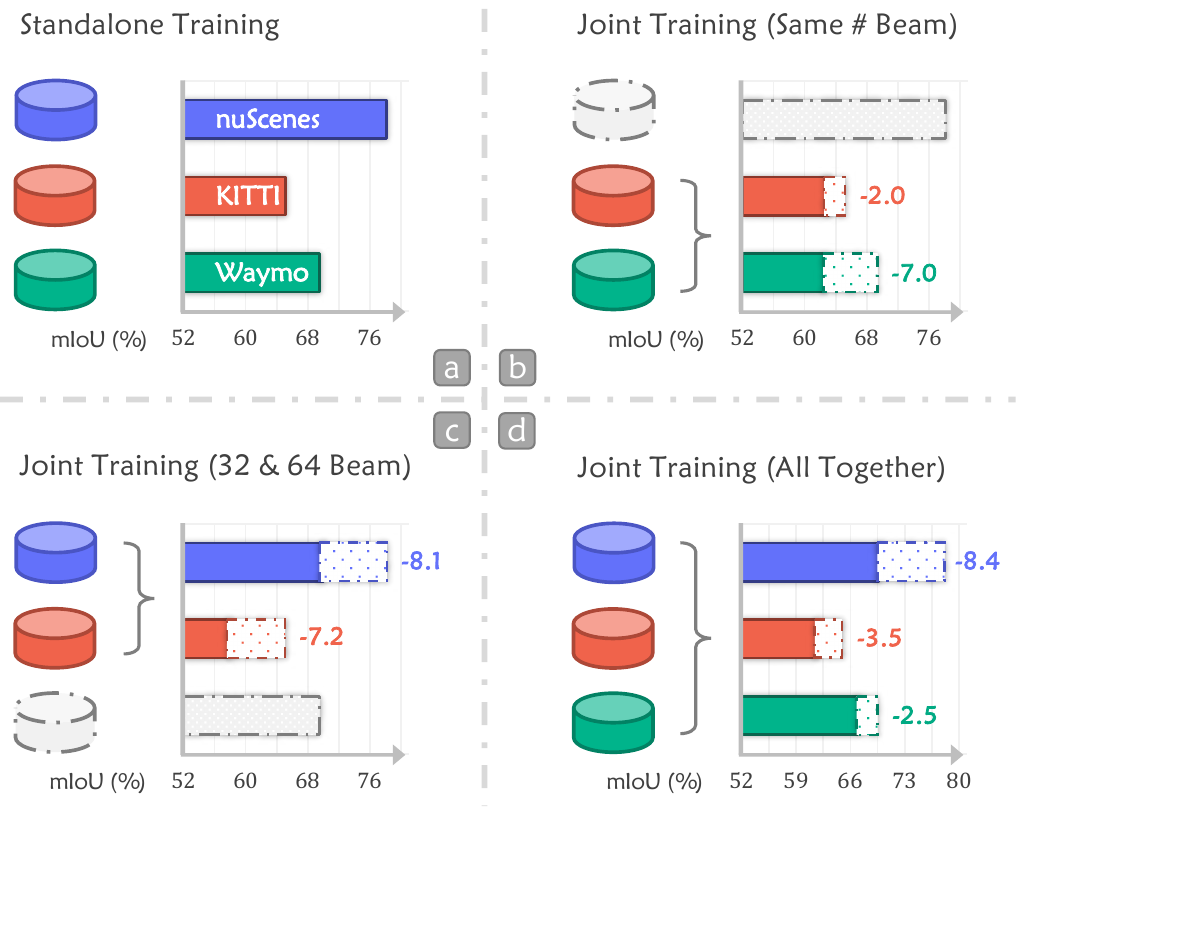}
         \vspace{0.05cm}
         \caption{M3Net (\textit{w/} PTv2+ \cite{wu2022ptv2} backbone)}
         \label{fig:ablation_pilot_study_ptv2}
    \end{subfigure}
    \caption{\textbf{A pilot study} of na\"{\i}vely merging different datasets for training the MinkUNet \cite{choy2019minkowski} model. Compared to the standalone training in \textbf{(a)}, either jointly training with \textbf{(b)} the same, \textbf{(c)} different, or \textbf{(d)} all sensor-acquired data will cause severe degradation. Subfigure (a): M3Net \textit{w/} a MinkUNet \cite{choy2019minkowski} backbone. Subfigure (b): M3Net \textit{w/} a PTv2+ \cite{wu2022ptv2} backbone.}
    \label{fig:ablation_pilot_study}
\end{figure*}

To further assess the capability of a LiDAR segmentation model for out-of-training-distribution generalization, we follow Robo3D \cite{kong2023robo3D} and adopt the corruption error (CE) and resilience rate (RR), as well as the mean corruption error (mCE) and mean resilience rate (mRR) as the evaluation metrics in comparing the robustness. To normalize the severity effects, we chose MinkUNet~\cite{choy2019minkowski} as the baseline model. The CE and mCE scores are calculated as follows:
\begin{equation}
    \text{CE}_k=\frac{\sum^{3}_{l=1}(1 - \text{Acc}_{k,l})}{\sum^{3}_{l=1}(1 - \text{Acc}_{k,l}^{\text{baseline}})}~,~~~
    \text{mCE}=\frac{1}{N}\sum^N_{k=1}\text{CE}_k~,
\end{equation}
where $\text{Acc}_{k,l}$ denotes mIoU scores on corruption type $k$ at severity level $l$. $N=8$ is the total number of corruption types. The mRR serves as the relative robustness indicator for measuring how much accuracy a model can retain when evaluated on the corruption sets. The RR and mRR scores are calculated as follows:
\begin{equation}
    \text{RR}_k=\frac{\sum^{3}_{l=1}\text{Acc}_{k,l}}{3\times \text{Acc}_{\text{clean}}}~,~~~~~
    \text{mRR}=\frac{1}{N}\sum^N_{k=1}\text{RR}_k~,
\end{equation}
where $\text{Acc}_{\text{clean}}$ denotes the mIoU score on the clean validation set of each dataset. Kindly refer to the original paper for additional details.

\section{Additional Experimental Results}
In this section, we present the complete experimental results as a supplement to the findings and conclusions drawn in the main body of this paper.

\subsection{Pilot Study}
In the main body of this paper, we conduct a pilot study to showcase the potential problems in the Single-Dataset Training and Na\"{\i}ve Joint Training pipelines. Specifically, we observe that it is non-trivial to na\"{\i}vely combine heterogeneous data from different driving datasets with large data distribution and sensor configuration gaps to train a universal LiDAR segmentation model.

We show in \cref{fig:ablation_pilot_study} our pilot study with the MinkUNet \cite{choy2019minkowski} backbone in subfigure (a) and the PTv2+ \cite{wu2022ptv2} backbone in subfigure (b), for both standalone and joint training setups. As can be seen, using either the classical MinkUNet or the most recent PTv2+ as the backbone, the brutal combination will undermine the segmentation performance. Due to large discrepancies in aspects like sensor configurations, data acquisitions, label mappings, and domain shifts, the jointly trained representations tend to be disruptive instead of being more general. Such degradation is particularly overt using na\"{\i}vely combining LiDAR data acquired by different sensor setups, such as the direct merge of \textit{nuScenes} \cite{fong2022panoptic-nuScenes} (Velodyne HDL32E with $32$ laser beams) and \textit{SemanticKITTI} \cite{behley2019semanticKITTI} (Velodyne HDL-64E with $64$ laser beams).

\begin{figure*}[t]
    \vspace{-0.37cm}
    \centering
    \begin{subfigure}[b]{.475\textwidth}
         \centering
         \includegraphics[width=\textwidth]{figures/teaser.png}
         \vspace{0.05cm}
         \caption{M3Net (\textit{w/} MinkUNet \cite{choy2019minkowski} backbone)}
         \label{fig:ablation_radar_minkunet}
    \end{subfigure}
    \hfill
    \begin{subfigure}[b]{.475\textwidth}
         \centering
         \includegraphics[width=\textwidth]{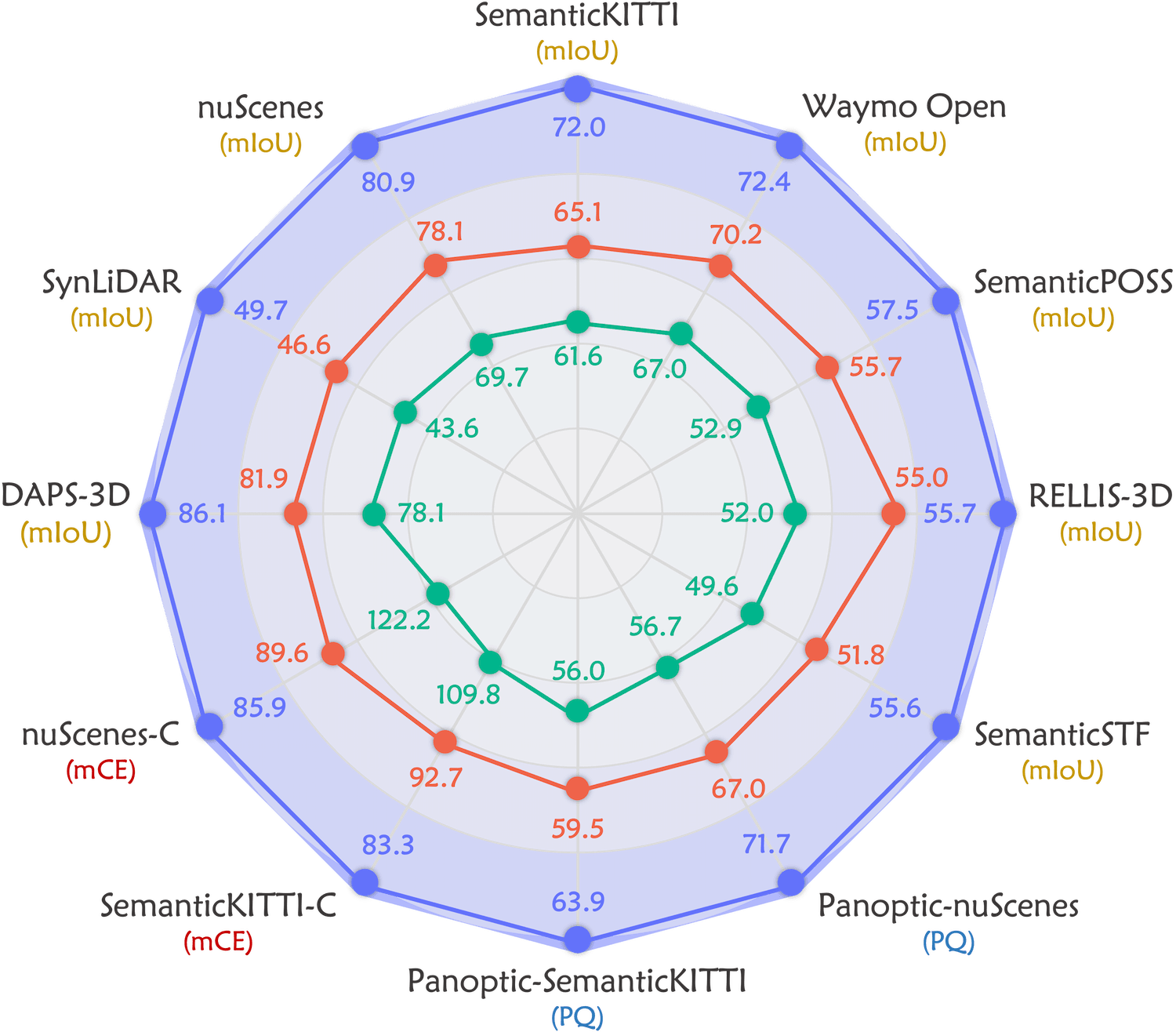}
         \vspace{0.05cm}
         \caption{M3Net (\textit{w/} PTv2+ \cite{wu2022ptv2} backbone)}
         \label{fig:ablation_radar_ptv2}
    \end{subfigure}
    \caption{Performance comparisons among \textbf{M3Net} [\textcolor{m3net_blue}{$\bullet$}], \textit{Single-Dataset Training} [\textcolor{m3net_red}{$\bullet$}], and \textit{Na\"{\i}ve Joint Training} [\textcolor{m3net_green}{$\bullet$}] across \textbf{twelve} LiDAR segmentation datasets. Subfigure (a): M3Net \textit{w/} a MinkUNet \cite{choy2019minkowski} backbone. Subfigure (b): M3Net \textit{w/} a PTv2+ \cite{wu2022ptv2} backbone. For better comparisons, the radius is normalized based on M3Net's scores. The larger the area coverage, the higher the overall performance.}
    \label{fig:ablation_radar}
\end{figure*}

Meanwhile, we also supplement the complete comparison results among the Single-Dataset Training, Na\"{\i}ve Joint Training, and our proposed M3Net pipelines and show the results in \cref{fig:ablation_radar}. As can be seen, compared to the Single-Dataset Training baselines, a na\"{\i}ve merging of heterogeneous LiDAR data will cause severe performance degradation. This observation holds true for both the MinkUNet \cite{choy2019minkowski} backbone as in \cref{fig:ablation_radar_minkunet} and the PTv2+ \cite{wu2022ptv2} backbone as in \cref{fig:ablation_radar_ptv2}, which highlights again the importance of conducting alignments when merging multiple driving datasets for training. Notably, after proper data, feature, and label space alignments, we are able to combine the advantage of leveraging the diverse training data sources and achieve better performance than the Single-Dataset Training baselines. Such improvements are holistic, as shown in the radar charts, our proposed M3Net achieves superior performance gains over the baselines under all the tested scenarios across all twelve LiDAR segmentation datasets.

\subsection{Ablation Study}
\begin{figure*}[t]
    \begin{center}
    \includegraphics[width=1.0\linewidth]{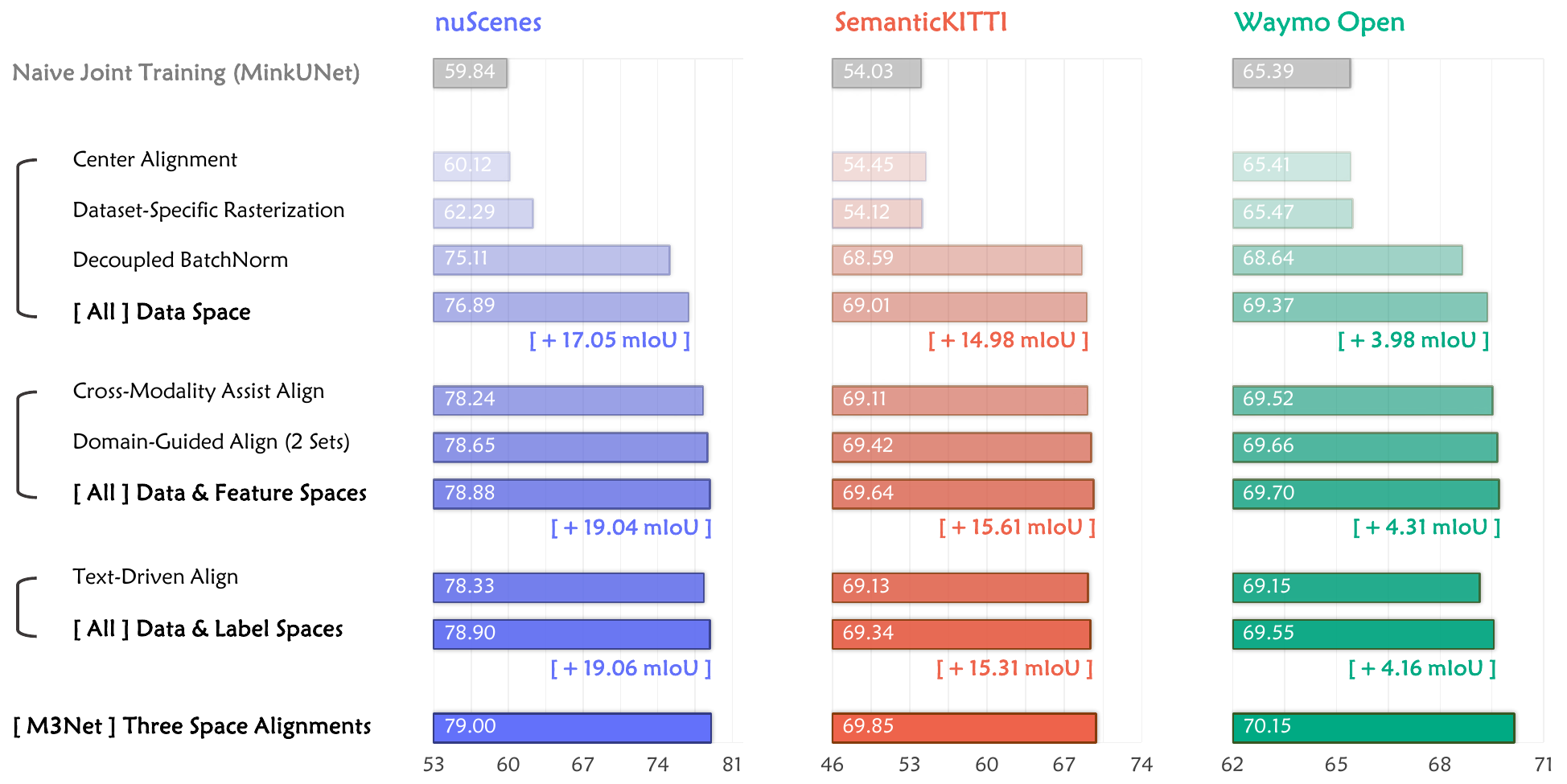}
    \end{center}
    \vspace{-0.4cm}
    \caption{\textbf{Ablation study} of the data, feature, and label space alignments in the proposed M3Net (\textit{w/} MinkUNet \cite{choy2019minkowski} backbone).}
    \label{fig:ablation_minkunet}
    \vspace{0.25cm}
\end{figure*}

\begin{figure*}[t]
    \begin{center}
    \includegraphics[width=1.0\linewidth]{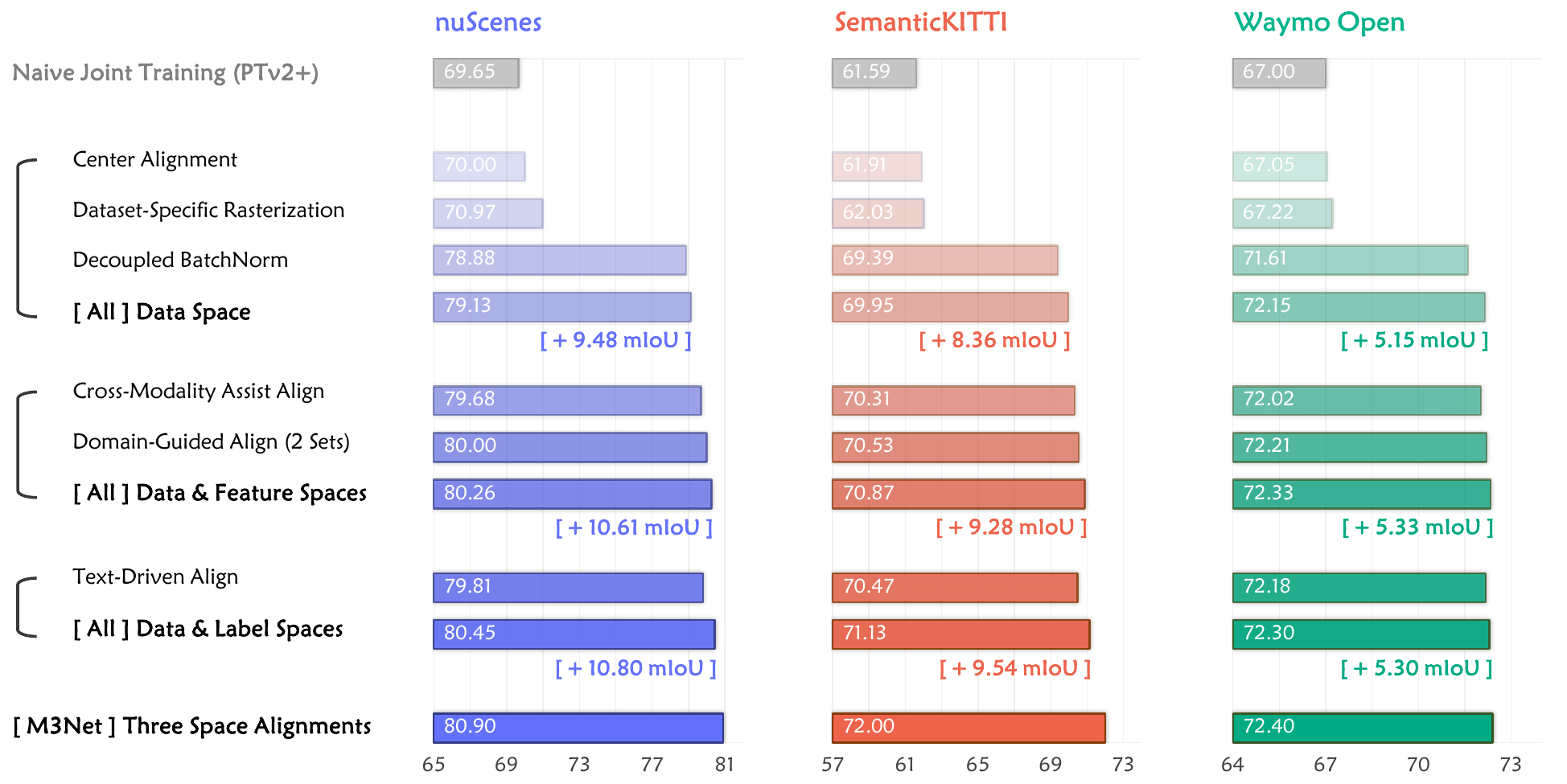}
    \end{center}
    \vspace{-0.4cm}
    \caption{\textbf{Ablation study} of the data, feature, and label space alignments in the proposed M3Net (\textit{w/} PTv2+ \cite{wu2022ptv2} backbone).}
    \label{fig:ablation_ptv2}
    \vspace{0.3cm}
\end{figure*}

In this section, we supplement more fine-grained ablation analysis in the third column of \cref{fig:ablation_minkunet} and \cref{fig:ablation_ptv2} on the \textit{SemanticKITTI} \cite{behley2019semanticKITTI}, \textit{nuScenes} \cite{fong2022panoptic-nuScenes}, and \textit{Waymo Open} \cite{sun2020waymoOpen} datasets. The results verify the effectiveness of each of the three alignments proposed in M3Net.

\subsection{LiDAR Panoptic Segmentation}
In this section, we supplement the PQ, RQ, and SQ scores, as well as their fine-grained scores regarding the \textit{things} and \textit{stuff} classes for our panoptic LiDAR segmentation experiments on the \textit{SemanticKITTI} \cite{behley2019semanticKITTI} and \textit{Panoptic-nuScenes} \cite{fong2022panoptic-nuScenes} datasets.

\subsubsection{Panoptic-SemanticKITTI}
For the detailed PQ, RQ, and SQ scores of our comparative study on the \textit{SemanticKITTI} \cite{behley2019semanticKITTI} dataset, we supplement \cref{tab:panoptic-results-val} to facilitate detailed comparisons with state-of-the-art LiDAR segmentation approaches on the validation set. We observe that the proposed M3Net is capable of achieving new arts on the validation set, especially for the more fine-grained metrics like RQ and SQ. The results verify the effectiveness of the proposed M3Net compared to the singe-dataset training and na\"{\i}ve joint training baselines.

\subsubsection{Panoptic-nuScenes}
For the detailed PQ, RQ, and SQ scores of our comparative study on the \textit{Panoptic-nuScenes} \cite{fong2022panoptic-nuScenes} dataset, we supplement \cref{tab:panoptic-results-val} to facilitate detailed comparisons with state-of-the-art LiDAR segmentation approaches on the validation set. We observe that the proposed M3Net is capable of achieving new arts on the validation set, across almost all the fine-grained metrics. The results verify the effectiveness of the proposed M3Net compared to the singe-dataset training and na\"{\i}ve joint training baselines.

\subsection{Out-of-Distribution Generalization}
In this section, we supplement the class-wise CE and RR scores of the out-of-training-distribution generalization experiments on the \textit{SemanticKITTI-C} and \textit{nuScenes-C} datasets in the Robo3D \cite{kong2023robo3D} benchmark. Specifically, \cref{tab:semantickitti-c} and \cref{tab:nuscenes-c} show the per-corruption IoU scores of prior works, our baselines, and the proposed M3Net on the \textit{SemanticKITTI-C} and \textit{nuScenes-C} datasets, respectively. We observe that M3Net sets up clear superiority over prior arts across almost all eight corruption types. Such robust feature learning is crucial to the safe operation of autonomous vehicles under out-of-training-distribution scenarios, especially in safety-critical areas \cite{kong2023robo3D,xie2023robobev,xie2024robobev}.

\section{Qualitative Assessment}
In this section, we provide a comprehensive qualitative assessment to validate further the effectiveness and superiority of the proposed M3Net framework.

\subsection{Visual Comparisons}
\begin{figure*}[t]
    \begin{center}
    \includegraphics[width=1.0\linewidth]{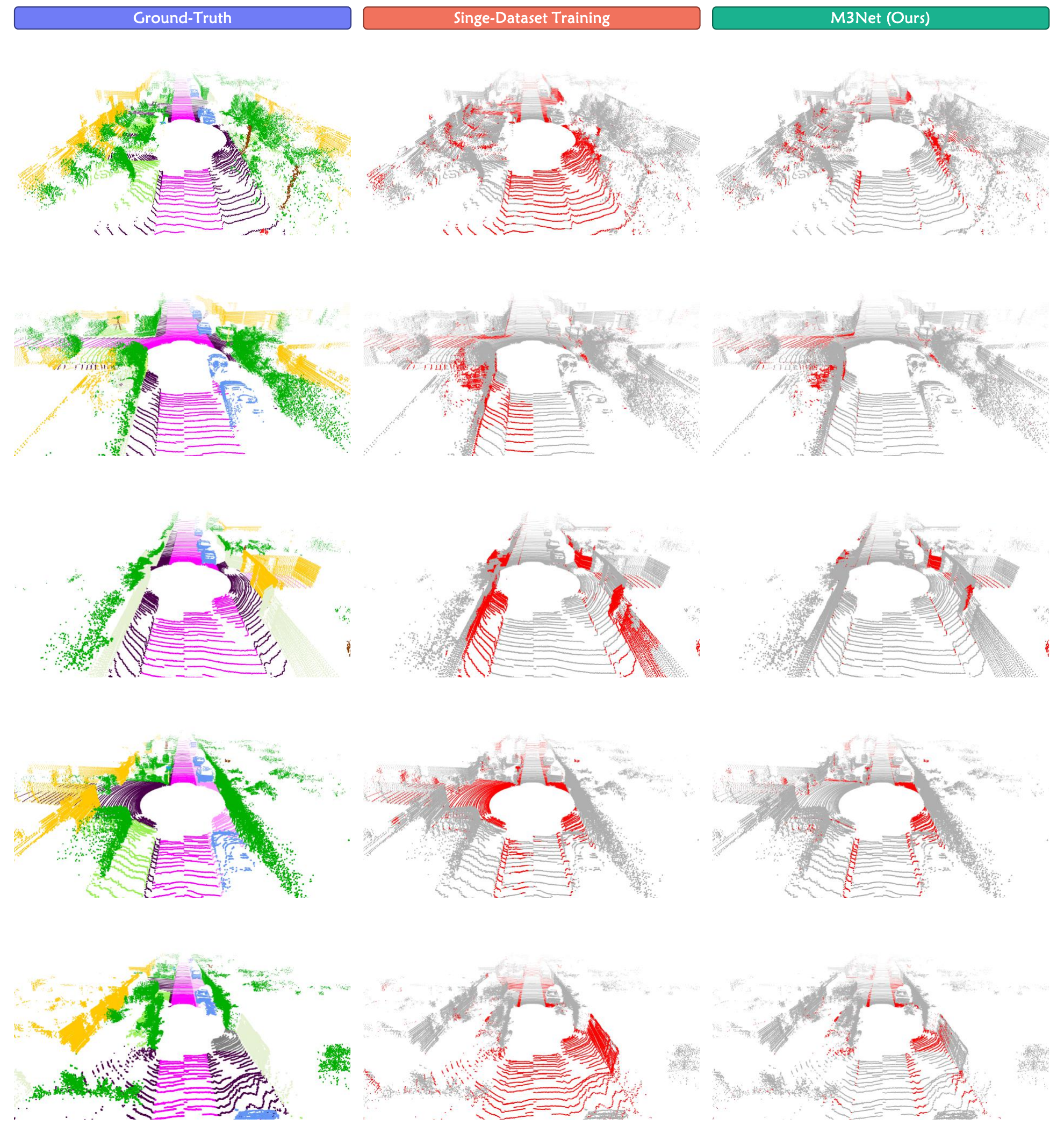}
    \end{center}
    \caption{\textbf{Qualitative comparisons} between the Single-Dataset Training and the proposed M3Net for LiDAR semantic segmentation on the \textit{SemanticKITTI} dataset \cite{behley2019semanticKITTI}. To highlight the differences, the \textbf{\textcolor{correct}{correct}} / \textbf{\textcolor{incorrect}{incorrect}} predictions are painted in \textbf{\textcolor{correct}{gray}} / \textbf{\textcolor{incorrect}{red}}, respectively.}
    \label{fig:vis_semantickitti}
\end{figure*}

\begin{figure*}[t]
    \begin{center}
    \includegraphics[width=1.0\linewidth]{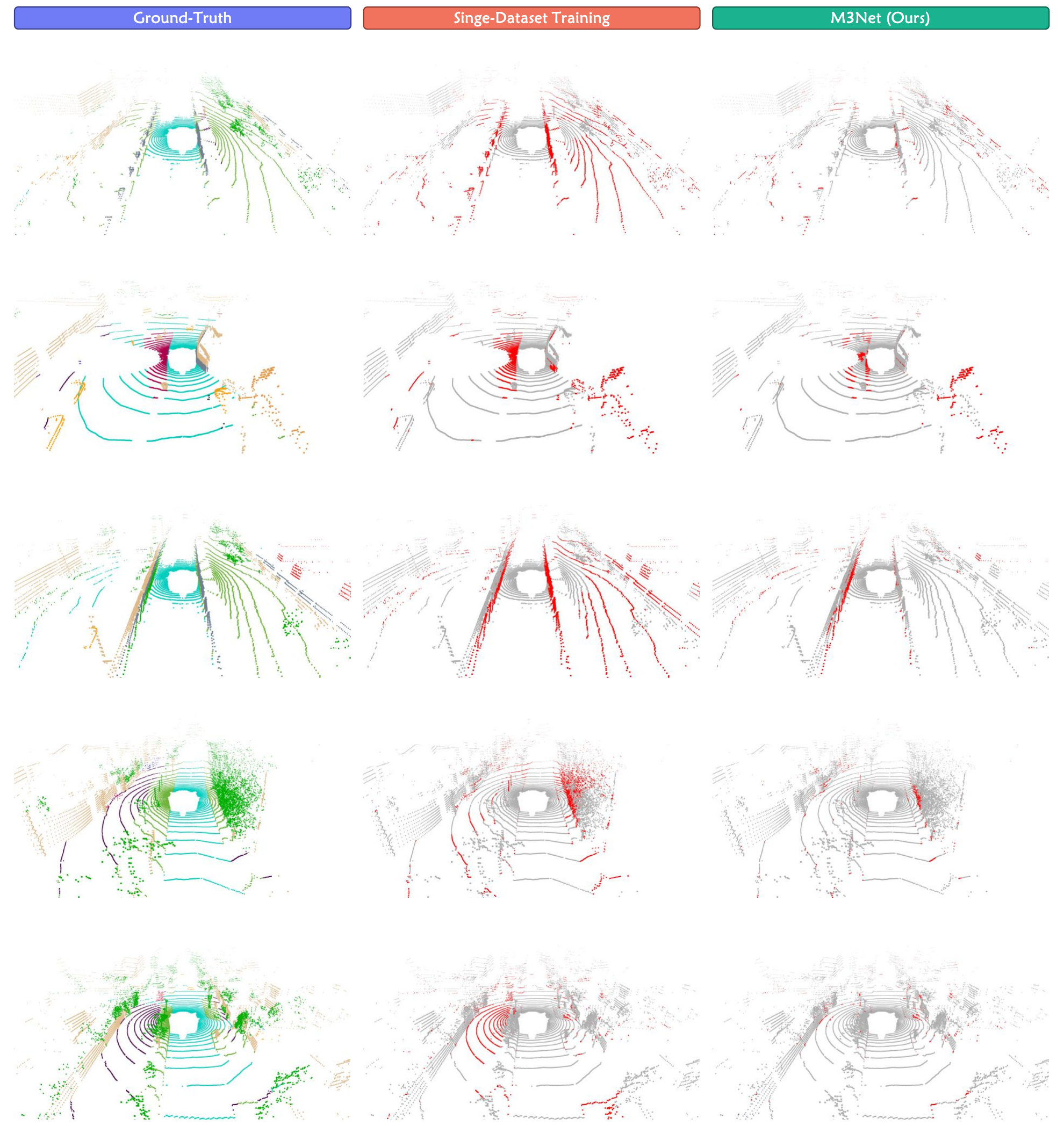}
    \end{center}
    \caption{\textbf{Qualitative comparisons} between the Single-Dataset Training and the proposed M3Net for LiDAR semantic segmentation on the \textit{nuScenes} dataset \cite{fong2022panoptic-nuScenes}. To highlight the differences, the \textbf{\textcolor{correct}{correct}} / \textbf{\textcolor{incorrect}{incorrect}} predictions are painted in \textbf{\textcolor{correct}{gray}} / \textbf{\textcolor{incorrect}{red}}, respectively.}
    \label{fig:vis_nuscenes}
\end{figure*}

\begin{figure*}[t]
    \begin{center}
    \includegraphics[width=1.0\linewidth]{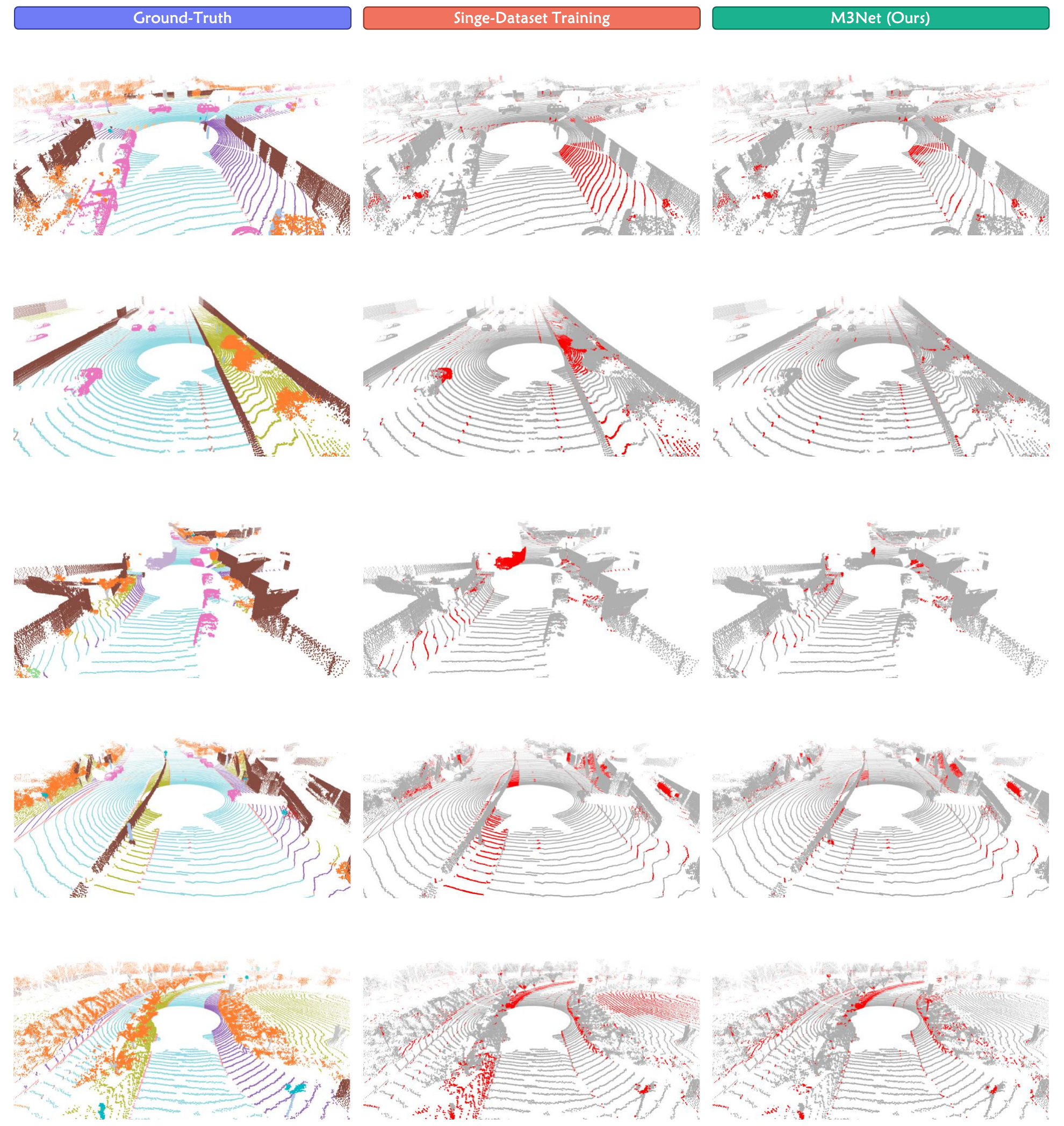}
    \end{center}
    \caption{\textbf{Qualitative comparisons} between the Single-Dataset Training and the proposed M3Net for LiDAR semantic segmentation on the \textit{Waymo Open} dataset \cite{sun2020waymoOpen}. To highlight the differences, the \textbf{\textcolor{correct}{correct}} / \textbf{\textcolor{incorrect}{incorrect}} predictions are painted in \textbf{\textcolor{correct}{gray}} / \textbf{\textcolor{incorrect}{red}}, respectively.}
    \label{fig:vis_waymo}
\end{figure*}

\begin{figure*}[t]
    \begin{center}
    \includegraphics[width=1.0\linewidth]{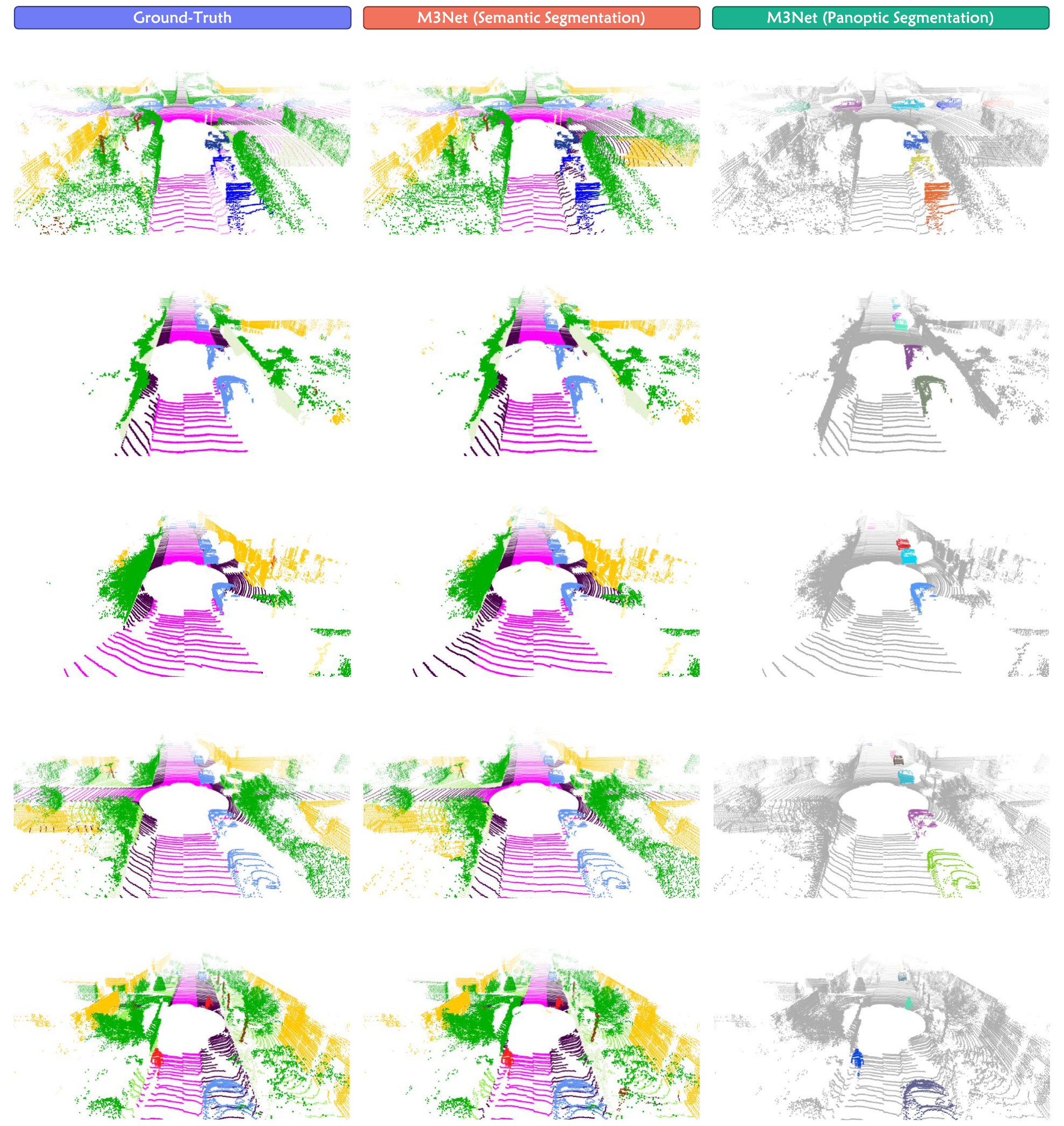}
    \end{center}
    \caption{\textbf{Qualitative comparisons} between the Ground-Truth and the proposed M3Net for LiDAR panoptic segmentation on the \textit{SemanticKITTI} dataset \cite{behley2019semanticKITTI}. To highlight the panoptic segmentation effect, the semantic predictions in the third column are painted in \textbf{\textcolor{correct}{gray}}. For panoptic segmentation predictions, each color-coded cluster represents a distinct instance.}
    \label{fig:vis_panoptic}
\end{figure*}

We supplement several qualitative comparisons of our proposed M3Net over the single-dataset training baseline. Specifically, the visual comparisons across the \textit{SemanticKITTI} \cite{behley2019semanticKITTI}, \textit{nuScenes} \cite{fong2022panoptic-nuScenes}, and \textit{Waymo Open} \cite{sun2020waymoOpen} datasets are shown in~\cref{fig:vis_semantickitti},~\cref{fig:vis_nuscenes}, and ~\cref{fig:vis_waymo}, respectively. As we can see, the proposed M3Net shows superior performance than the baseline under different driving scenarios. Such results highlight the effectiveness of the proposed M3Net in enhancing performance in the multi-task, multi-dataset, multi-modality training setting. Additionally, we present qualitative results in~\cref{fig:vis_panoptic} to showcase the capability of M3Net in tackling both the LiDAR semantic segmentation and panoptic segmentation tasks. As we can see, the proposed M3Net demonstrates effectiveness in making accurate predictions among the complex object and background classes in the driving scenes, underscoring its effectiveness in handling multi-task LiDAR segmentation.

\section{Broader Impact}
In this section, we elaborate on the positive societal influence and potential limitations of our multi-task, multi-dataset, multi-modality LiDAR segmentation framework.

\subsection{Positive Societal Influence}
In this work, we present a versatile LiDAR segmentation framework dubbed M3Net for conducting multi-task, multi-dataset, multi-modality LiDAR segmentation in a unifying pipeline. LiDAR segmentation is crucial for the development of safe and reliable autonomous vehicles. By accurately interpreting the vehicle surroundings, LiDAR helps in obstacle detection, pedestrian safety, and navigation, thereby reducing the likelihood of accidents and enhancing road safety. LiDAR segmentation contributes significantly to societal welfare through its applications in various fields. Its ability to provide accurate, detailed 3D representations of physical environments enables more informed decision-making, enhances safety, and promotes sustainability.

\subsection{Potential Limitation}
Although our proposed M3Net is capable of leveraging multiple heterogeneous driving datasets to train a versatile LiDAR segmentation network and achieve promising universal LiDAR segmentation results, there still exists room for improvement. Firstly, our framework leverages calibrated and synchronized camera data to assist the alignments. Such a requirement might not be met in some older LiDAR segmentation datasets. Secondly, we do not handle the minority classes during multi-dataset learning, especially for some dynamic classes that are uniquely defined by a certain dataset. Thirdly, we do not consider the combination of simulation data with real-world LiDAR point clouds. We believe these aspects are promising for future work to further improve our multi-task, multi-dataset, multi-modality LiDAR segmentation framework.

\section{Public Resources Used}
\label{sec:public-resources-used}
In this section, we acknowledge the use of datasets, models, and codebases, during the course of this work.

\subsection{Public Datasets Used}
We acknowledge the use of the following public datasets, during the course of this work:
\begin{itemize}
    \item nuScenes\footnote{\url{https://www.nuscenes.org/nuscenes}.} \dotfill CC BY-NC-SA 4.0
    \item nuScenes-devkit\footnote{\url{https://github.com/nutonomy/nuscenes-devkit}.} \dotfill Apache License 2.0
    \item SemanticKITTI\footnote{\url{http://semantic-kitti.org}.} \dotfill CC BY-NC-SA 4.0
    \item SemanticKITTI-API\footnote{\url{https://github.com/PRBonn/semantic-kitti-api}.} \dotfill MIT License
    \item Waymo Open Dataset\footnote{\url{https://waymo.com/open}.} \dotfill Waymo Dataset License
    \item RELLIS-3D\footnote{\url{http://www.unmannedlab.org/research/RELLIS-3D}.} \dotfill CC BY-NC-SA 3.0
    \item SemanticPOSS\footnote{\url{http://www.poss.pku.edu.cn/semanticposs.html}.} \dotfill Unknown
    \item SemanticSTF\footnote{\url{https://github.com/xiaoaoran/SemanticSTF}.} \dotfill CC BY-NC-SA 4.0
    \item SynLiDAR\footnote{\url{https://github.com/xiaoaoran/SynLiDAR}.} \dotfill MIT License
    \item DAPS-3D\footnote{\url{https://github.com/subake/DAPS3D}.} \dotfill MIT License
    \item Robo3D\footnote{\url{https://github.com/ldkong1205/Robo3D}.} \dotfill CC BY-NC-SA 4.0
\end{itemize}

\subsection{Public Models Used}
We acknowledge the use of the following public implementations, during the course of this work:
\begin{itemize}
    \item MinkowskiEngine\footnote{\url{https://github.com/NVIDIA/MinkowskiEngine}.} \dotfill MIT License
    \item PointTransformerV2\footnote{\url{https://github.com/Gofinge/PointTransformerV2}.} \dotfill Unknown
    \item spvnas\footnote{\url{https://github.com/mit-han-lab/spvnas}.} \dotfill MIT License
    \item Cylinder3D\footnote{\url{https://github.com/xinge008/Cylinder3D}.}\dotfill Apache License 2.0
    \item SLidR\footnote{\url{https://github.com/valeoai/SLidR}.} \dotfill Apache License 2.0
    \item OpenSeeD\footnote{\url{https://github.com/IDEA-Research/OpenSeeD}.} \dotfill Apache License 2.0
    \item segment-anything\footnote{\url{https://github.com/facebookresearch/segment-anything}.} \dotfill Apache License 2.0
    \item Segment-Any-Point-Cloud\footnote{\url{https://github.com/youquanl/Segment-Any-Point-Cloud}} \dotfill CC BY-NC-SA 4.0
    \item Mix3D\footnote{\url{https://github.com/kumuji/mix3d}.} \dotfill Unknown
    \item LaserMix\footnote{\url{https://github.com/ldkong1205/LaserMix}.} \dotfill CC BY-NC-SA 4.0
\end{itemize}

\subsection{Public Codebases Used}
We acknowledge the use of the following public codebases, during the course of this work:
\begin{itemize}
    \item mmdetection3d\footnote{\url{https://github.com/open-mmlab/mmdetection3d}.} \dotfill Apache License 2.0
    \item Pointcept\footnote{\url{https://github.com/Pointcept/Pointcept}.} \dotfill MIT License
    \item OpenPCSeg\footnote{\url{https://github.com/PJLab-ADG/OpenPCSeg}.} \dotfill Apache License 2.0
\end{itemize}

\clearpage
\begin{table*}[t]
\caption{\textbf{The class-wise panoptic segmentation scores} on the \textit{val} sets of the \textit{Panoptic-SemanticKITTI} \cite{behley2019semanticKITTI} and \textit{Panoptic-nuScenes} \cite{fong2022panoptic-nuScenes} datasets. All scores are given in percentage (\%). For each evaluated metric: \textbf{bold} - best in column; \underline{underline} - second best in column.}
\vspace{-0.1cm}
\centering\scalebox{0.82}{
\begin{tabular}{r|p{25pt}<{\centering}p{25pt}<{\centering}p{25pt}<{\centering}p{25pt}<{\centering}p{25pt}<{\centering}|p{25pt}<{\centering}p{25pt}<{\centering}p{25pt}<{\centering}p{25pt}<{\centering}p{25pt}<{\centering}}
    \toprule
    \multirow{2}{*}{\textcolor{darkgray}{\textbf{Method}}}  & \multicolumn{5}{c|}{\textcolor{darkgray}{\textbf{Panoptic-SemanticKITTI}}} & \multicolumn{5}{c}{\textcolor{darkgray}{\textbf{Panoptic-nuScenes}}}
    \\
     & PQ & PQ$^\dagger$ & RQ & SQ & mIoU & PQ & PQ$^\dagger$ & RQ & SQ & mIoU
    \\\midrule\midrule
    Panoptic-TrackNet \cite{hurtado2020mopt}  & $40.0$ & - & $48.3$ & $73.0$ & $53.8$ & $51.4$ & $56.2$ & $63.3$ & $80.2$ & $58.0$ 
    \\
    Panoptic-PolarNet \cite{zhou2021panoptic} & $59.1$ & $64.1$ & $70.2$ & $78.3$ & $64.5$ & $63.4$ & $67.2$ & $75.3$ & $83.9$ & $66.9$
    \\
    EfficientLPS \cite{sirohi2021efficientlps}  & $59.2$ & $65.1$ & $69.8$ & $75.0$ & $64.9$ & $59.2$ & $62.8$ & $82.9$ & $70.7$ & $69.4$
    \\
    DSNet \cite{hong2021dsnet} & $61.4$ & $65.2$ & $72.7$ & $79.0$ & $69.6$ & $64.7$ & $67.6$ & $76.1$ & $83.5$ & $76.3$
    \\
    Panoptic-PHNet \cite{li2022panoptic}  & \underline{$61.7$} & - & - & - & $65.7$ & $\mathbf{74.7}$ & $\mathbf{77.7}$ & $\mathbf{84.2}$ & $\mathbf{88.2}$ & \underline{$79.7$}
    \\\midrule
    \textcolor{gray}{Na\"{\i}ve Joint (MinkUNet)} & \textcolor{gray}{$47.8$} & \textcolor{gray}{$54.1$} & \textcolor{gray}{$56.9$} & \textcolor{gray}{$71.6$} & \textcolor{gray}{$54.0$} & \textcolor{gray}{$45.0$} & \textcolor{gray}{$50.3$} & \textcolor{gray}{$55.3$} & \textcolor{gray}{$79.4$} & \textcolor{gray}{$59.8$} 
    \\
    Single-Dataset (MinkUNet) & $60.7$ & $65.6$ & $70.6$ & $\mathbf{83.2}$ & $64.4$ & $58.4$ & $62.7$ & \underline{$82.9$} & $69.3$ & $73.3$
    \\
    \textbf{M3Net (MinkUNet)} & $\mathbf{63.9}$ & \underline{$68.5$} & $\mathbf{73.2}$ & $82.3$ & \underline{$69.9$} & $67.9$ & $71.1$ & $78.1$ & $85.9$ & $79.0$
    \\\midrule
    \textcolor{gray}{Na\"{\i}ve Joint (PTv2+)} & \textcolor{gray}{$56.0$} & \textcolor{gray}{$59.6$} & \textcolor{gray}{$65.8$} & \textcolor{gray}{$73.7$} & \textcolor{gray}{$61.6$} & \textcolor{gray}{$56.7$} & \textcolor{gray}{$60.6$} & \textcolor{gray}{$66.8$} & \textcolor{gray}{$83.5$} & \textcolor{gray}{$69.7$}
    \\
    Single-Dataset (PTv2+) & $59.5$ & $63.6$ & $69.5$ & $75.3$ & $65.1$ & $67.0$ & $69.8$ & $77.8$ & $85.0$ & $78.1$
    \\
    \textbf{M3Net (PTv2+)} & $\mathbf{63.9}$ & $\mathbf{68.7}$ & \underline{$73.1$} & \underline{$82.4$} & $\mathbf{72.0}$ & \underline{$71.7$} & \underline{$74.0$} & $82.2$ & \underline{$86.5$} & $\mathbf{80.9}$
    \\\bottomrule
\end{tabular}}
\label{tab:panoptic-results-val}
\end{table*}

\begin{table*}[t]
\caption{\textbf{The class-wise robustness evaluation scores} on the \textit{SemanticKITTI-C} dataset from the Robo3D benchmark \cite{kong2023robo3D}. All scores are given in percentage (\%). For each evaluated metric: \textbf{bold} - best in column; \underline{underline} - second best in column.}

\vspace{-0.1cm}
\centering\scalebox{0.82}{
\begin{tabular}{r|c|c| c c c c c c c c c}
    \toprule
     \textbf{Method}&  \textbf{mCE}~$\downarrow$ & \textbf{mRR}~$\uparrow$ & \rotatebox{90}{fog} & \rotatebox{90}{wet-ground} & \rotatebox{90}{snow} & \rotatebox{90}{motion-blur} & \rotatebox{90}{beam-missing} & \rotatebox{90}{crosstalk} & \rotatebox{90}{incomplete-echo~} & \rotatebox{90}{cross-sensor}
    \\\midrule\midrule
    
    \textcolor{gray}{Na\"{\i}ve Joint (MinkUNet)} & \textcolor{gray}{$113.7$} & \textcolor{gray}{$84.7$} & \textcolor{gray}{$48.5$} & \textcolor{gray}{$54.0$} & \textcolor{gray}{$39.8$} & \textcolor{gray}{$41.1$} & \textcolor{gray}{$49.1$} & \textcolor{gray}{$39.8$} & \textcolor{gray}{$47.7$} & \textcolor{gray}{$46.1$}
    \\
    Single-Dataset (MinkUNet) & $95.1$ & $85.0$ & $50.6$ & $52.3$ & $51.4$ & $54.5$ & $59.3$ & $\mathbf{56.9}$ & $56.2$ & $56.6$
    \\
    \textbf{M3Net (MinkUNet)} & \underline{$86.4$} & \underline{$85.8$} & \underline{$56.7$} & \underline{$63.8$} & $\mathbf{55.1}$ & \underline{$63.3$} & \underline{$64.5$} & $50.6$ & \underline{$60.7$} & $\mathbf{58.1}$
    \\\midrule
    \textcolor{gray}{Na\"{\i}ve Joint (PTv2+)} & \textcolor{gray}{$109.8$} & \textcolor{gray}{$77.5$} & \textcolor{gray}{$49.7$} & \textcolor{gray}{$53.1$} & \textcolor{gray}{$43.6$} & \textcolor{gray}{$45.3$} & \textcolor{gray}{$51.6$} & \textcolor{gray}{$39.7$} & \textcolor{gray}{$50.2$} & \textcolor{gray}{$48.8$}
    \\
    Single-Dataset (PTv2+) & $92.7$ & $\mathbf{85.9}$ & $52.3$ & $53.7$ & $51.8$ & $55.8$ & $60.2$ & \underline{$56.4$} & $59.3$ & $57.6$
    \\
    \textbf{M3Net (PTv2+)} & $\mathbf{83.3}$ & $84.0$ & $\mathbf{60.4}$ & $\mathbf{66.1}$ & \underline{$52.7$} & $\mathbf{63.9}$ & $\mathbf{65.1}$ & $55.1$ & $\mathbf{62.6}$ & \underline{$57.9$}
    \\\bottomrule
\end{tabular}
}
\label{tab:semantickitti-c}
\end{table*}
\begin{table*}[t]
\caption{\textbf{The class-wise robustness evaluation scores} on the \textit{nuScenes-C} dataset from the Robo3D benchmark \cite{kong2023robo3D}. All scores are given in percentage (\%). For each evaluated metric: \textbf{bold} - best in column; \underline{underline} - second best in column.}

\vspace{-0.1cm}
\centering\scalebox{0.82}{
\begin{tabular}{r|c|c|c c c c c c c c c}
\toprule
 \textbf{Method} &  \textbf{mCE}~$\downarrow$ & \textbf{mRR}~$\uparrow$ & \rotatebox{90}{fog} & \rotatebox{90}{wet-ground} & \rotatebox{90}{snow} & \rotatebox{90}{motion-blur} & \rotatebox{90}{beam-missing} & \rotatebox{90}{crosstalk} & \rotatebox{90}{incomplete-echo~} & \rotatebox{90}{cross-sensor}
\\\midrule\midrule

 PPKT \cite{liu2021ppkt} & $105.6$ & $76.1$ & $64.0$ & $72.2$ & \underline{$59.1$} & $57.2$ & $63.9$ & $36.3$ & $60.6$ & $39.6$
\\
 SLidR \cite{sautier2022slidr} & $106.1$ & $76.0$ & \underline{$65.4$} & $72.3$ & $56.0$ & $56.1$ & $62.9$ & $41.9$ & $61.2$ & $38.9$
\\
 Seal \cite{liu2023segment} &$92.6$ & $\mathbf{83.1}$ & $\mathbf{72.7}$ & $74.3$ & $\mathbf{66.2}$ & $66.1$ & $66.0$ & $\mathbf{57.4}$ &  $59.9$ & $39.9$
\\\midrule
\textcolor{gray}{Na\"{\i}ve Joint (MinkUNet)} & \textcolor{gray}{$129.0$} & \textcolor{gray}{\underline{$81.5$}} & \textcolor{gray}{$54.0$} & \textcolor{gray}{$57.3$} & \textcolor{gray}{$50.9$} & \textcolor{gray}{$57.5$} & \textcolor{gray}{$47.3$} & \textcolor{gray}{$42.3$} & \textcolor{gray}{$49.4$} & \textcolor{gray}{$30.9$}
\\
Single-Dataset (MinkUNet) & $99.6$ & $79.1$ & $60.7$ & $74.6$ & $50.8$ & $65.0$ & $67.1$ & $32.4$ & $63.2$ & $50.0$
\\
\textbf{M3Net (MinkUNet)} & \underline{$91.0$} & $79.2$ & $62.5$ & $76.2$ & $49.7$ & \underline{$75.4$} & $66.2$ & $43.3$ & $64.7$ & $52.5$
\\\midrule
\textcolor{gray}{Na\"{\i}ve Joint (PTv2+)} & \textcolor{gray}{$122.2$} & \textcolor{gray}{$73.4$} & \textcolor{gray}{$55.2$} & \textcolor{gray}{$60.0$} & \textcolor{gray}{$51.4$} & \textcolor{gray}{$58.7$} & \textcolor{gray}{$52.7$} & \textcolor{gray}{$43.3$} & \textcolor{gray}{$52.9$} & \textcolor{gray}{$34.7$}
\\
Single-Dataset (PTv2+) & $89.6$ & $79.1$ &$63.1$ & \underline{$76.4$} & $51.6$ & $75.2$ & \underline{$67.9$} & $41.4$ & \underline{$65.4$} & \underline{$53.5$}
\\
\textbf{M3Net (PTv2+)} & $\mathbf{85.9}$ & $78.2$ & $54.4$ & $\mathbf{78.0}$ & $51.2$ & $\mathbf{76.8}$ & $\mathbf{68.0}$ & \underline{$44.3$} & $\mathbf{66.7}$ & $\mathbf{55.9}$
\\\bottomrule
\end{tabular}
}
\label{tab:nuscenes-c}
\end{table*}

\clearpage

\clearpage

\clearpage

\clearpage

\clearpage
\clearpage
{\small
 \bibliographystyle{ieeenat_fullname}
 \bibliography{main}}

\begin{thebibliography}{132}
\providecommand{\natexlab}[1]{#1}
\providecommand{\url}[1]{\texttt{#1}}
\expandafter\ifx\csname urlstyle\endcsname\relax
  \providecommand{\doi}[1]{doi: #1}\else
  \providecommand{\doi}{doi: \begingroup \urlstyle{rm}\Url}\fi

\bibitem[Ando et~al.(2023)Ando, Gidaris, Bursuc, Puy, Boulch, and Marlet]{ando2023rangevit}
Angelika Ando, Spyros Gidaris, Andrei Bursuc, Gilles Puy, Alexandre Boulch, and Renaud Marlet.
\newblock Rangevit: Towards vision transformers for 3d semantic segmentation in autonomous driving.
\newblock In \emph{IEEE/CVF Conference on Computer Vision and Pattern Recognition}, pages 5240--5250, 2023.

\bibitem[Baevski et~al.(2022)Baevski, Hsu, Xu, Babu, Gu, and Auli]{baevski2022data2vec}
Alexei Baevski, Wei-Ning Hsu, Qiantong Xu, Arun Babu, Jiatao Gu, and Michael Auli.
\newblock Data2vec: A general framework for self-supervised learning in speech, vision and language.
\newblock In \emph{International Conference on Machine Learning}, pages 1298--1312, 2022.

\bibitem[Behley et~al.(2019)Behley, Garbade, Milioto, Quenzel, Behnke, Stachniss, and Gall]{behley2019semanticKITTI}
Jens Behley, Martin Garbade, Andres Milioto, Jan Quenzel, Sven Behnke, Cyrill Stachniss, and Juergen Gall.
\newblock Semantickitti: A dataset for semantic scene understanding of lidar sequences.
\newblock In \emph{IEEE/CVF International Conference on Computer Vision}, pages 9297--9307, 2019.

\bibitem[Behley et~al.(2021)Behley, Garbade, Milioto, Quenzel, Behnke, Gall, and Stachniss]{behley2021semanticKITTI}
Jens Behley, Martin Garbade, Andres Milioto, Jan Quenzel, Sven Behnke, Jürgen Gall, and Cyrill Stachniss.
\newblock Towards 3d lidar-based semantic scene understanding of 3d point cloud sequences: The semantickitti dataset.
\newblock \emph{International Journal of Robotics Research}, 40:\penalty0 959--96, 2021.

\bibitem[Berman et~al.(2018)Berman, Triki, and Blaschko]{berman2018lovasz}
Maxim Berman, Amal~Rannen Triki, and Matthew~B Blaschko.
\newblock The lov{\'a}sz-softmax loss: a tractable surrogate for the optimization of the intersection-over-union measure in neural networks.
\newblock In \emph{IEEE/CVF Conference on Computer Vision and Pattern Recognition}, pages 4413--4421, 2018.

\bibitem[Bijelic et~al.(2020)Bijelic, Gruber, Mannan, Kraus, Ritter, Dietmayer, and Heide]{bijelic2020stf}
Mario Bijelic, Tobias Gruber, Fahim Mannan, Florian Kraus, Werner Ritter, Klaus Dietmayer, and Felix Heide.
\newblock Seeing through fog without seeing fog: Deep multimodal sensor fusion in unseen adverse weather.
\newblock In \emph{IEEE/CVF Conference on Computer Vision and Pattern Recognition}, pages 11682--11692, 2020.

\bibitem[Boulch et~al.(2023)Boulch, Sautier, Michele, Puy, and Marlet]{boulch2023also}
Alexandre Boulch, Corentin Sautier, Björn Michele, Gilles Puy, and Renaud Marlet.
\newblock Also: Automotive lidar self-supervision by occupancy estimation.
\newblock In \emph{IEEE/CVF Conference on Computer Vision and Pattern Recognition}, pages 13455--13465, 2023.

\bibitem[Caesar et~al.(2020)Caesar, Bankiti, Lang, Vora, Liong, Xu, Krishnan, Pan, Baldan, and Beijbom]{caesar2020nuScenes}
Holger Caesar, Varun Bankiti, Alex~H Lang, Sourabh Vora, Venice~Erin Liong, Qiang Xu, Anush Krishnan, Yu Pan, Giancarlo Baldan, and Oscar Beijbom.
\newblock nuscenes: A multimodal dataset for autonomous driving.
\newblock In \emph{IEEE/CVF Conference on Computer Vision and Pattern Recognition}, pages 11621--11631, 2020.

\bibitem[Caron et~al.(2021)Caron, Touvron, Misra, Jégou, Mairal, Bojanowski, and Joulin]{caron2021dino}
Mathilde Caron, Hugo Touvron, Ishan Misra, Hervé Jégou, Julien Mairal, Piotr Bojanowski, and Armand Joulin.
\newblock Emerging properties in self-supervised vision transformers.
\newblock In \emph{IEEE/CVF International Conference on Computer Vision}, pages 9650--9660, 2021.

\bibitem[Cen et~al.(2023)Cen, Zhang, Pei, Li, Zheng, Luo, Zhang, and Chen]{cen2023cmdfusion}
Jun Cen, Shiwei Zhang, Yixuan Pei, Kun Li, Hang Zheng, Maochun Luo, Yingya Zhang, and Qifeng Chen.
\newblock Cmdfusion: Bidirectional fusion network with cross-modality knowledge distillation for lidar semantic segmentation.
\newblock \emph{arXiv preprint arXiv:2307.04091}, 2023.

\bibitem[Chen et~al.(2023{\natexlab{a}})Chen, Liu, Kong, Chen, Zhu, Ma, Liu, and Wang.]{chen2023towards}
Runnan Chen, Youquan Liu, Lingdong Kong, Nenglun Chen, Xinge Zhu, Yuexin Ma, Tongliang Liu, and Wenping Wang.
\newblock Towards label-free scene understanding by vision foundation models.
\newblock In \emph{Advances in Neural Information Processing Systems}, 2023{\natexlab{a}}.

\bibitem[Chen et~al.(2023{\natexlab{b}})Chen, Liu, Kong, Zhu, Ma, Li, Hou, Qiao, and Wang]{chen2023clip2Scene}
Runnan Chen, Youquan Liu, Lingdong Kong, Xinge Zhu, Yuexin Ma, Yikang Li, Yuenan Hou, Yu Qiao, and Wenping Wang.
\newblock Clip2scene: Towards label-efficient 3d scene understanding by clip.
\newblock In \emph{IEEE/CVF Conference on Computer Vision and Pattern Recognition}, pages 7020--7030, 2023{\natexlab{b}}.

\bibitem[Chen et~al.(2023{\natexlab{c}})Chen, Zhu, Chen, Li, Ma, Yang, and Wang]{chen2023bridging}
Runnan Chen, Xinge Zhu, Nenglun Chen, Wei Li, Yuexin Ma, Ruigang Yang, and Wenping Wang.
\newblock Bridging language and geometric primitives for zero-shot point cloud segmentation.
\newblock In \emph{ACM International Conference on Multimedia}, pages 5380--5388, 2023{\natexlab{c}}.

\bibitem[Chen et~al.(2020)Chen, An, Zhang, Ma, Wang, Guo, and Zheng]{chen2020improving}
Tian Chen, Shijie An, Yuan Zhang, Chongyang Ma, Huayan Wang, Xiaoyan Guo, and Wen Zheng.
\newblock Improving monocular depth estimation by leveraging structural awareness and complementary datasets.
\newblock In \emph{European Conference on Computer Vision}, pages 90--108, 2020.

\bibitem[Chen et~al.(2023{\natexlab{d}})Chen, Wang, Mittal, Xu, Favaro, Tighe, and Modolo]{chen2023scaledet}
Yanbei Chen, Manchen Wang, Abhay Mittal, Zhenlin Xu, Paolo Favaro, Joseph Tighe, and Davide Modolo.
\newblock Scaledet: A scalable multi-dataset object detector.
\newblock In \emph{IEEE/CVF Conference on Computer Vision and Pattern Recognition}, pages 7288--7297, 2023{\natexlab{d}}.

\bibitem[Cheng et~al.(2021{\natexlab{a}})Cheng, Schwing, and Kirillov]{cheng2021maskformer}
Bowen Cheng, Alex Schwing, and Alexander Kirillov.
\newblock Per-pixel classification is not all you need for semantic segmentation.
\newblock In \emph{Advances in Neural Information Processing Systems}, pages 17864--17875, 2021{\natexlab{a}}.

\bibitem[Cheng et~al.(2022{\natexlab{a}})Cheng, Misra, Schwing, Kirillov, and Girdhar]{cheng2022maskformer}
Bowen Cheng, Ishan Misra, Alexander~G. Schwing, Alexander Kirillov, and Rohit Girdhar.
\newblock Masked-attention mask transformer for universal image segmentation.
\newblock In \emph{IEEE/CVF Conference on Computer Vision and Pattern Recognition}, pages 1290--1299, 2022{\natexlab{a}}.

\bibitem[Cheng et~al.(2022{\natexlab{b}})Cheng, Han, and Xiao]{cheng2022cenet}
Huixian Cheng, Xianfeng Han, and Guoqiang Xiao.
\newblock Cenet: Toward concise and efficient lidar semantic segmentation for autonomous driving.
\newblock In \emph{IEEE International Conference on Multimedia and Expo}, pages 1--6, 2022{\natexlab{b}}.

\bibitem[Cheng et~al.(2021{\natexlab{b}})Cheng, Razani, Taghavi, Li, and Liu]{cheng2021af2S3Net}
Ran Cheng, Ryan Razani, Ehsan Taghavi, Enxu Li, and Bingbing Liu.
\newblock Af2-s3net: Attentive feature fusion with adaptive feature selection for sparse semantic segmentation network.
\newblock In \emph{IEEE/CVF Conference on Computer Vision and Pattern Recognition}, pages 12547--12556, 2021{\natexlab{b}}.

\bibitem[Chowdhery et~al.(2022)Chowdhery, Narang, Devlin, Bosma, Mishra, Roberts, Barham, Chung, Sutton, Gehrmann, Schuh, Shi, Tsvyashchenko, Maynez, Rao, Barnes, Tay, Shazeer, Prabhakaran, Reif, Du, Hutchinson, Pope, Bradbury, Austin, Isard, Gur-Ari, Yin, Duke, Levskaya, Ghemawat, Dev, Michalewski, Garcia, Misra, Robinson, Fedus, Zhou, Ippolito, Luan, Lim, Zoph, Spiridonov, Sepassi, Dohan, Agrawal, Omernick, Dai, Pillai, Pellat, Lewkowycz, Moreira, Child, Polozov, Lee, Zhou, Wang, Saeta, Diaz, Firat, Catasta, Wei, Meier-Hellstern, Eck, Dean, Petrov, and Fiedel]{chowdhery2022palm}
Aakanksha Chowdhery, Sharan Narang, Jacob Devlin, Maarten Bosma, Gaurav Mishra, Adam Roberts, Paul Barham, Hyung~Won Chung, Charles Sutton, Sebastian Gehrmann, Parker Schuh, Kensen Shi, Sasha Tsvyashchenko, Joshua Maynez, Abhishek Rao, Parker Barnes, Yi Tay, Noam Shazeer, Vinodkumar Prabhakaran, Emily Reif, Nan Du, Ben Hutchinson, Reiner Pope, James Bradbury, Jacob Austin, Michael Isard, Guy Gur-Ari, Pengcheng Yin, Toju Duke, Anselm Levskaya, Sanjay Ghemawat, Sunipa Dev, Henryk Michalewski, Xavier Garcia, Vedant Misra, Kevin Robinson, Liam Fedus, Denny Zhou, Daphne Ippolito, David Luan, Hyeontaek Lim, Barret Zoph, Alexander Spiridonov, Ryan Sepassi, David Dohan, Shivani Agrawal, Mark Omernick, Andrew~M. Dai, Thanumalayan~Sankaranarayana Pillai, Marie Pellat, Aitor Lewkowycz, Erica Moreira, Rewon Child, Oleksandr Polozov, Katherine Lee, Zongwei Zhou, Xuezhi Wang, Brennan Saeta, Mark Diaz, Orhan Firat, Michele Catasta, Jason Wei, Kathy Meier-Hellstern, Douglas Eck, Jeff Dean, Slav Petrov, and Noah Fiedel.
\newblock Palm: Scaling language modeling with pathways.
\newblock \emph{arXiv preprint arXiv:2204.02311}, 2022.

\bibitem[Choy et~al.(2019)Choy, Gwak, and Savarese]{choy2019minkowski}
Christopher Choy, JunYoung Gwak, and Silvio Savarese.
\newblock 4d spatio-temporal convnets: Minkowski convolutional neural networks.
\newblock In \emph{IEEE/CVF Conference on Computer Vision and Pattern Recognition}, pages 3075--3084, 2019.

\bibitem[Comaniciu and Meer(2005)]{marcuzzi2002MeanShift}
Dorin Comaniciu and Peter Meer.
\newblock Mean shift: A robust approach toward feature space analysis.
\newblock \emph{IEEE Transactions on Pattern Analysis and Machine Intelligence}, 24\penalty0 (5):\penalty0 603--619, 2005.

\bibitem[Contributors(2020)]{mmdet3d2020}
MMDetection3D Contributors.
\newblock {MMDetection3D: OpenMMLab} next-generation platform for general {3D} object detection.
\newblock \url{https://github.com/open-mmlab/mmdetection3d}, 2020.

\bibitem[Contributors(2023)]{pointcept2023}
Pointcept Contributors.
\newblock Pointcept: A codebase for point cloud perception research.
\newblock \url{https://github.com/Pointcept/Pointcept}, 2023.

\bibitem[Cortinhal et~al.(2020)Cortinhal, Tzelepis, and Aksoy]{cortinhal2020salsanext}
Tiago Cortinhal, George Tzelepis, and Eren~Erdal Aksoy.
\newblock Salsanext: Fast, uncertainty-aware semantic segmentation of lidar point clouds.
\newblock In \emph{International Symposium on Visual Computing}, pages 207--222, 2020.

\bibitem[der Maaten and Hinton(2008)]{maaten2008t-sne}
Laurens~Van der Maaten and Geoffrey Hinton.
\newblock Visualizing data using t-sne.
\newblock \emph{Journal of Machine Learning Research}, 9\penalty0 (11), 2008.

\bibitem[Feichtenhofer et~al.(2022)Feichtenhofer, Li, and He]{feichtenhofer2022mae}
Christoph Feichtenhofer, Yanghao Li, and Kaiming He.
\newblock Masked autoencoders as spatiotemporal learners.
\newblock In \emph{Advances in Neural Information Processing Systems}, 2022.

\bibitem[Fong et~al.(2022)Fong, Mohan, Hurtado, Zhou, Caesar, Beijbom, and Valada]{fong2022panoptic-nuScenes}
Whye~Kit Fong, Rohit Mohan, Juana~Valeria Hurtado, Lubing Zhou, Holger Caesar, Oscar Beijbom, and Abhinav Valada.
\newblock Panoptic nuscenes: A large-scale benchmark for lidar panoptic segmentation and tracking.
\newblock \emph{IEEE Robotics and Automation Letters}, 7:\penalty0 3795--3802, 2022.

\bibitem[Gao et~al.(2021)Gao, Pan, Li, Geng, and Zhao]{gao2021survey}
Biao Gao, Yancheng Pan, Chengkun Li, Sibo Geng, and Huijing Zhao.
\newblock Are we hungry for 3d lidar data for semantic segmentation? a survey of datasets and methods.
\newblock \emph{IEEE Transactions on Intelligent Transportation Systems}, 2021.

\bibitem[Geiger et~al.(2012)Geiger, Lenz, and Urtasun]{geiger2012kitti}
Andreas Geiger, Philip Lenz, and Raquel Urtasun.
\newblock Are we ready for autonomous driving? the kitti vision benchmark suite.
\newblock In \emph{IEEE/CVF Conference on Computer Vision and Pattern Recognition}, pages 3354--3361, 2012.

\bibitem[Gu et~al.(2023)Gu, Cui, Huang, Rashwan, Yang, Zhou, Ghiasi, Kuo, Chen, Chen, and Ross]{gu2023dataseg}
Xiuye Gu, Yin Cui, Jonathan Huang, Abdullah Rashwan, Xuan Yang, Xingyi Zhou, Golnaz Ghiasi, Weicheng Kuo, Huizhong Chen, Liang-Chieh Chen, and David~A. Ross.
\newblock Dataseg: Taming a universal multi-dataset multi-task segmentation model.
\newblock \emph{arXiv preprint arXiv:2306.01736}, 2023.

\bibitem[Gu et~al.(2022)Gu, Huang, Xu, and Kong]{gu2022maskrange}
Yi Gu, Yuming Huang, Chengzhong Xu, and Hui Kong.
\newblock Maskrange: A mask-classification model for range-view based lidar segmentation.
\newblock \emph{arXiv preprint arXiv:2206.12073}, 2022.

\bibitem[He et~al.(2017)He, Gkioxari, Dollár, and Girshick]{he2017mask-rcnn}
Kaiming He, Georgia Gkioxari, Piotr Dollár, and Ross Girshick.
\newblock Mask r-cnn.
\newblock In \emph{IEEE/CVF International Conference on Computer Vision}, pages 2961--2969, 2017.

\bibitem[Hong et~al.(2021)Hong, Zhou, Zhu, Li, and Liu]{hong2021dsnet}
Fangzhou Hong, Hui Zhou, Xinge Zhu, Hongsheng Li, and Ziwei Liu.
\newblock Lidar-based panoptic segmentation via dynamic shifting network.
\newblock In \emph{IEEE/CVF Conference on Computer Vision and Pattern Recognition}, pages 13090--13099, 2021.

\bibitem[Hong et~al.(2024)Hong, Kong, Zhou, Zhu, Li, and Liu]{hong20224dDSNet}
Fangzhou Hong, Lingdong Kong, Hui Zhou, Xinge Zhu, Hongsheng Li, and Ziwei Liu.
\newblock Unified 3d and 4d panoptic segmentation via dynamic shifting networks.
\newblock \emph{IEEE Transactions on Pattern Analysis and Machine Intelligence}, 46\penalty0 (5):\penalty0 3480--3495, 2024.

\bibitem[Hou et~al.(2022)Hou, Zhu, Ma, Loy, and Li]{hou2022pvkd}
Yuenan Hou, Xinge Zhu, Yuexin Ma, Chen~Change Loy, and Yikang Li.
\newblock Point-to-voxel knowledge distillation for lidar semantic segmentation.
\newblock In \emph{IEEE/CVF Conference on Computer Vision and Pattern Recognition}, pages 8479--8488, 2022.

\bibitem[Hu et~al.(2020)Hu, Yang, Xie, Rosa, Guo, Wang, Trigoni, and Markham]{hu2020randla}
Qingyong Hu, Bo Yang, Linhai Xie, Stefano Rosa, Yulan Guo, Zhihua Wang, Niki Trigoni, and Andrew Markham.
\newblock Randla-net: Efficient semantic segmentation of large-scale point clouds.
\newblock In \emph{IEEE/CVF Conference on Computer Vision and Pattern Recognition}, pages 11108--11117, 2020.

\bibitem[Hu et~al.(2021)Hu, Yang, Khalid, Xiao, Trigoni, and Markham]{hu2021sensatUrban}
Qingyong Hu, Bo Yang, Sheikh Khalid, Wen Xiao, Niki Trigoni, and Andrew Markham.
\newblock Towards semantic segmentation of urban-scale 3d point clouds: A dataset, benchmarks and challenges.
\newblock In \emph{IEEE/CVF Conference on Computer Vision and Pattern Recognition}, pages 4977--4987, 2021.

\bibitem[Hurtado et~al.(2020)Hurtado, Mohan, Burgard, and Valada]{hurtado2020mopt}
Juana~Valeria Hurtado, Rohit Mohan, Wolfram Burgard, and Abhinav Valada.
\newblock Mopt: Multi-object panoptic tracking.
\newblock \emph{arXiv preprint arXiv:2004.08189}, 2020.

\bibitem[Jain et~al.(2023)Jain, Li, Chiu, Hassani, Orlov, and Shi]{jain2023oneformer}
Jitesh Jain, Jiachen Li, Mang~Tik Chiu, Ali Hassani, Nikita Orlov, and Humphrey Shi.
\newblock Oneformer: One transformer to rule universal image segmentation.
\newblock In \emph{IEEE/CVF Conference on Computer Vision and Pattern Recognition}, pages 2989--2998, 2023.

\bibitem[Jaritz et~al.(2020)Jaritz, Vu, de~Charette, Wirbel, and Pérez]{jaritz2020xMUDA}
Maximilian Jaritz, Tuan-Hung Vu, Raoul de Charette, Emilie Wirbel, and Patrick Pérez.
\newblock xmuda: Cross-modal unsupervised domain adaptation for 3d semantic segmentation.
\newblock In \emph{IEEE/CVF Conference on Computer Vision and Pattern Recognition}, pages 12605--12614, 2020.

\bibitem[Jaritz et~al.(2023)Jaritz, Vu, Charette, Émilie Wirbel, and Pérez]{jaritz2023xMUDA}
Maximilian Jaritz, Tuan-Hung Vu, Raoul~De Charette, Émilie Wirbel, and Patrick Pérez.
\newblock Cross-modal learning for domain adaptation in 3d semantic segmentation.
\newblock \emph{IEEE Transactions on Pattern Analysis and Machine Intelligence}, 45\penalty0 (2):\penalty0 1533--1544, 2023.

\bibitem[Jiang et~al.(2021)Jiang, Osteen, Wigness, and Saripallig]{jiang2021rellis3D}
Peng Jiang, Philip Osteen, Maggie Wigness, and Srikanth Saripallig.
\newblock Rellis-3d dataset: Data, benchmarks and analysis.
\newblock In \emph{IEEE International Conference on Robotics and Automation}, pages 1110--1116, 2021.

\bibitem[Kalluri et~al.(2019)Kalluri, Varma, Chandraker, and Jawahar]{kalluri2019universal}
Tarun Kalluri, Girish Varma, Manmohan Chandraker, and C.~V. Jawahar.
\newblock Universal semi-supervised semantic segmentation.
\newblock In \emph{IEEE/CVF International Conference on Computer Vision}, pages 5259--5270, 2019.

\bibitem[Kim et~al.(2022)Kim, Tsai, Suh, Faraki, Garg, Chandraker, and Han]{tsai2022learning}
Dongwan Kim, Yi-Hsuan Tsai, Yumin Suh, Masoud Faraki, Sparsh Garg, Manmohan Chandraker, and Bohyung Han.
\newblock Learning semantic segmentation from multiple datasets with label shifts.
\newblock In \emph{European Conference on Computer Vision}, pages 20--36, 2022.

\bibitem[Kirillov et~al.(2023)Kirillov, Mintun, Ravi, Mao, Rolland, Gustafson, Xiao, Whitehead, Berg, Lo, Dollár, and Girshick]{kirillov2023sam}
Alexander Kirillov, Eric Mintun, Nikhila Ravi, Hanzi Mao, Chloe Rolland, Laura Gustafson, Tete Xiao, Spencer Whitehead, Alexander~C. Berg, Wan-Yen Lo, Piotr Dollár, and Ross Girshick.
\newblock Segment anything.
\newblock In \emph{IEEE/CVF International Conference on Computer Vision}, pages 4015--4026, 2023.

\bibitem[Klokov et~al.(2023)Klokov, Pak, Khorin, Yudin, Kochiev, Luchinskiy, and Bezuglyj]{klokov2023daps3D}
Alexey Klokov, Di~Un Pak, Aleksandr Khorin, Dmitry Yudin, Leon Kochiev, Vladimir Luchinskiy, and Vitaly Bezuglyj.
\newblock Daps3d: Domain adaptive projective segmentation of 3d lidar point clouds.
\newblock \emph{IEEE Access}, 11:\penalty0 79341--79356, 2023.

\bibitem[Kong et~al.(2023{\natexlab{a}})Kong, Liu, Chen, Ma, Zhu, Li, Hou, Qiao, and Liu]{kong2023rethinking}
Lingdong Kong, Youquan Liu, Runnan Chen, Yuexin Ma, Xinge Zhu, Yikang Li, Yuenan Hou, Yu Qiao, and Ziwei Liu.
\newblock Rethinking range view representation for lidar segmentation.
\newblock In \emph{IEEE/CVF International Conference on Computer Vision}, pages 228--240, 2023{\natexlab{a}}.

\bibitem[Kong et~al.(2023{\natexlab{b}})Kong, Liu, Li, Chen, Zhang, Ren, Pan, Chen, and Liu]{kong2023robo3D}
Lingdong Kong, Youquan Liu, Xin Li, Runnan Chen, Wenwei Zhang, Jiawei Ren, Liang Pan, Kai Chen, and Ziwei Liu.
\newblock Robo3d: Towards robust and reliable 3d perception against corruptions.
\newblock In \emph{IEEE/CVF International Conference on Computer Vision}, pages 19994--20006, 2023{\natexlab{b}}.

\bibitem[Kong et~al.(2023{\natexlab{c}})Kong, Quader, and Liong]{kong2023conDA}
Lingdong Kong, Niamul Quader, and Venice~Erin Liong.
\newblock Conda: Unsupervised domain adaptation for lidar segmentation via regularized domain concatenation.
\newblock In \emph{IEEE International Conference on Robotics and Automation}, pages 9338--9345, 2023{\natexlab{c}}.

\bibitem[Kong et~al.(2023{\natexlab{d}})Kong, Ren, Pan, and Liu]{kong2022laserMix}
Lingdong Kong, Jiawei Ren, Liang Pan, and Ziwei Liu.
\newblock Lasermix for semi-supervised lidar semantic segmentation.
\newblock In \emph{IEEE/CVF Conference on Computer Vision and Pattern Recognition}, pages 21705--21715, 2023{\natexlab{d}}.

\bibitem[Kong et~al.(2023{\natexlab{e}})Kong, Xie, Hu, Ng, Cottereau, and Ooi]{kong2023robodepth}
Lingdong Kong, Shaoyuan Xie, Hanjiang Hu, Lai~Xing Ng, Benoit~R. Cottereau, and Wei~Tsang Ooi.
\newblock Robodepth: Robust out-of-distribution depth estimation under corruptions.
\newblock In \emph{Advances in Neural Information Processing Systems}, 2023{\natexlab{e}}.

\bibitem[Kong et~al.(2024)Kong, Xu, Cen, Zhang, Pan, Chen, and Liu]{kong2024calib3d}
Lingdong Kong, Xiang Xu, Jun Cen, Wenwei Zhang, Liang Pan, Kai Chen, and Ziwei Liu.
\newblock Calib3d: Calibrating model preferences for reliable 3d scene understanding.
\newblock \emph{arXiv preprint arXiv:2403.17010}, 2024.

\bibitem[Lai et~al.(2023)Lai, Chen, Lu, Liu, and Jia]{lai2023sphereformer}
Xin Lai, Yukang Chen, Fanbin Lu, Jianhui Liu, and Jiaya Jia.
\newblock Spherical transformer for lidar-based 3d recognition.
\newblock In \emph{IEEE/CVF Conference on Computer Vision and Pattern Recognition}, pages 17545--17555, 2023.

\bibitem[Lambert et~al.(2020)Lambert, Liu, Sener, Hays, and Koltun]{lambert2020mseg}
John Lambert, Zhuang Liu, Ozan Sener, James Hays, and Vladlen Koltun.
\newblock Mseg: A composite dataset for multi-domain semantic segmentation.
\newblock In \emph{IEEE/CVF Conference on Computer Vision and Pattern Recognition}, pages 2879--2888, 2020.

\bibitem[Li et~al.(2023{\natexlab{a}})Li, Kang, Wang, Wei, and Yang]{li2023adversarially}
Guangrui Li, Guoliang Kang, Xiaohan Wang, Yunchao Wei, and Yi Yang.
\newblock Adversarially masking synthetic to mimic real: Adaptive noise injection for point cloud segmentation adaptation.
\newblock In \emph{IEEE/CVF Conference on Computer Vision and Pattern Recognition}, pages 20464--20474, 2023{\natexlab{a}}.

\bibitem[Li et~al.(2022{\natexlab{a}})Li, He, Wen, Gao, Cheng, and Zhang]{li2022panoptic}
Jinke Li, Xiao He, Yang Wen, Yuan Gao, Xiaoqiang Cheng, and Dan Zhang.
\newblock Panoptic-phnet: Towards real-time and high-precision lidar panoptic segmentation via clustering pseudo heatmap.
\newblock In \emph{IEEE/CVF Conference on Computer Vision and Pattern Recognition}, pages 11809--11818, 2022{\natexlab{a}}.

\bibitem[Li et~al.(2023{\natexlab{b}})Li, Shum, and Breckon]{li2023lim3d}
Li Li, Hubert~PH Shum, and Toby~P. Breckon.
\newblock Less is more: Reducing task and model complexity for 3d point cloud semantic segmentation.
\newblock In \emph{IEEE/CVF Conference on Computer Vision and Pattern Recognition}, pages 9361--9371, 2023{\natexlab{b}}.

\bibitem[Li et~al.(2022{\natexlab{b}})Li, de~Charette, and Anh-Quan]{li2022coarse3D}
Rong Li, Raoul de Charette, and C.~A.~O. Anh-Quan.
\newblock Coarse3d: Class-prototypes for contrastive learning in weakly-supervised 3d point cloud segmentation.
\newblock In \emph{British Machine Vision Conference}, 2022{\natexlab{b}}.

\bibitem[Li et~al.(2022{\natexlab{c}})Li, Shi, Hou, Wu, Ma, Li, and He]{li2022homogeneous}
Xin Li, Botian Shi, Yuenan Hou, Xingjiao Wu, Tianlong Ma, Yikang Li, and Liang He.
\newblock Homogeneous multi-modal feature fusion and interaction for 3d object detection.
\newblock In \emph{European Conference on Computer Vision}, pages 691--707. Springer, 2022{\natexlab{c}}.

\bibitem[Li et~al.(2022{\natexlab{d}})Li, Zhang, Pang, Chen, Cheng, Tong, and Loy]{li2022video-k-net}
Xiangtai Li, Wenwei Zhang, Jiangmiao Pang, Kai Chen, Guangliang Cheng, Yunhai Tong, and Chen~Change Loy.
\newblock Video k-net: A simple, strong, and unified baseline for video segmentation.
\newblock In \emph{IEEE/CVF Conference on Computer Vision and Pattern Recognition}, pages 18847--18857, 2022{\natexlab{d}}.

\bibitem[Li et~al.(2023{\natexlab{c}})Li, Ma, Hou, Shi, Yang, Liu, Wu, Chen, Li, Qiao, et~al.]{li2023logonet}
Xin Li, Tao Ma, Yuenan Hou, Botian Shi, Yuchen Yang, Youquan Liu, Xingjiao Wu, Qin Chen, Yikang Li, Yu Qiao, et~al.
\newblock Logonet: Towards accurate 3d object detection with local-to-global cross-modal fusion.
\newblock In \emph{IEEE/CVF Conference on Computer Vision and Pattern Recognition}, pages 17524--17534, 2023{\natexlab{c}}.

\bibitem[Li et~al.(2024)Li, Kong, Hu, Xu, and Huang]{li2024place3d}
Ye Li, Lingdong Kong, Hanjiang Hu, Xiaohao Xu, and Xiaonan Huang.
\newblock Optimizing lidar placements for robust driving perception in adverse conditions.
\newblock \emph{arXiv preprint arXiv:2403.17009}, 2024.

\bibitem[Liong et~al.(2020)Liong, Nguyen, Widjaja, Sharma, and Chong]{liong2020amvNet}
Venice~Erin Liong, Thi Ngoc~Tho Nguyen, Sergi Widjaja, Dhananjai Sharma, and Zhuang~Jie Chong.
\newblock Amvnet: Assertion-based multi-view fusion network for lidar semantic segmentation.
\newblock \emph{arXiv preprint arXiv:2012.04934}, 2020.

\bibitem[Liu et~al.(2022)Liu, Zhou, Qi, Gong, Su, and Anguelov]{liu2022less}
Minghua Liu, Yin Zhou, Charles~R. Qi, Boqing Gong, Hao Su, and Dragomir Anguelov.
\newblock Less: Label-efficient semantic segmentation for lidar point clouds.
\newblock In \emph{European Conference on Computer Vision}, pages 70--89, 2022.

\bibitem[Liu et~al.(2023{\natexlab{a}})Liu, Chen, Li, Kong, Yang, Xia, Bai, Zhu, Ma, Li, Qiao, and Hou]{liu2023uniseg}
Youquan Liu, Runnan Chen, Xin Li, Lingdong Kong, Yuchen Yang, Zhaoyang Xia, Yeqi Bai, Xinge Zhu, Yuexin Ma, Yikang Li, Yu Qiao, and Yuenan Hou.
\newblock Uniseg: A unified multi-modal lidar segmentation network and the openpcseg codebase.
\newblock In \emph{IEEE/CVF International Conference on Computer Vision}, pages 21662--21673, 2023{\natexlab{a}}.

\bibitem[Liu et~al.(2023{\natexlab{b}})Liu, Kong, Cen, Chen, Zhang, Pan, Chen, and Liu]{liu2023segment}
Youquan Liu, Lingdong Kong, Jun Cen, Runnan Chen, Wenwei Zhang, Liang Pan, Kai Chen, and Ziwei Liu.
\newblock Segment any point cloud sequences by distilling vision foundation models.
\newblock In \emph{Advances in Neural Information Processing Systems}, 2023{\natexlab{b}}.

\bibitem[Liu et~al.(2021)Liu, Huang, Chiang, Su, Liu, Chen, Tseng, and Hsu]{liu2021ppkt}
Yueh-Cheng Liu, Yu-Kai Huang, Hung-Yueh Chiang, Hung-Ting Su, Zhe-Yu Liu, Chin-Tang Chen, Ching-Yu Tseng, and Winston~H. Hsu.
\newblock Learning from 2d: Contrastive pixel-to-point knowledge transfer for 3d pretraining.
\newblock \emph{arXiv preprint arXiv:2104.0468}, 2021.

\bibitem[Liu et~al.(2023{\natexlab{c}})Liu, Haotian~Tang, Yang, Mao, Rus, and Han]{liu2023bevfusion}
Zhijian Liu, Alexander~Amini Haotian~Tang, Xinyu Yang, Huizi Mao, Daniela~L. Rus, and Song Han.
\newblock Bevfusion: Multi-task multi-sensor fusion with unified bird's-eye view representation.
\newblock In \emph{IEEE International Conference on Robotics and Automation}, pages 2774--2781, 2023{\natexlab{c}}.

\bibitem[Loshchilov and Hutter(2019)]{loshchilov2019adamw}
Ilya Loshchilov and Frank Hutter.
\newblock Decoupled weight decay regularization.
\newblock In \emph{International Conference on Learning Representations}, 2019.

\bibitem[Lu et~al.(2023)Lu, Jiang, Chen, Hou, Zhu, and Ma]{lu2023see}
Yuhang Lu, Qi Jiang, Runnan Chen, Yuenan Hou, Xinge Zhu, and Yuexin Ma.
\newblock See more and know more: Zero-shot point cloud segmentation via multi-modal visual data.
\newblock In \emph{IEEE/CVF International Conference on Computer Vision}, pages 21674--21684, 2023.

\bibitem[Mao et~al.(2023)Mao, Shi, Wang, and Li]{mao2023survey}
Jiageng Mao, Shaoshuai Shi, Xiaogang Wang, and Hongsheng Li.
\newblock 3d object detection for autonomous driving: A comprehensive survey.
\newblock \emph{International Journal of Computer Vision}, 2023.

\bibitem[Marcuzzi et~al.(2023)Marcuzzi, Nunes, Wiesmann, Behley, and Stachniss]{marcuzzi2023maskpls}
Rodrigo Marcuzzi, Lucas Nunes, Louis Wiesmann, Jens Behley, and Cyrill Stachniss.
\newblock Mask-based panoptic lidar segmentation for autonomous driving.
\newblock \emph{IEEE Robotics and Automation Letters}, 8\penalty0 (2):\penalty0 1141--1148, 2023.

\bibitem[Meletis and Dubbelman(2018)]{meletis2018training}
Panagiotis Meletis and Gijs Dubbelman.
\newblock Training of convolutional networks on multiple heterogeneous datasets for street scene semantic segmentation.
\newblock In \emph{IEEE Intelligent Vehicles Symposium}, pages 1045--1050, 2018.

\bibitem[Michele et~al.(2023)Michele, Boulch, Puy, Vu, Marlet, and Courty]{michele2023saluda}
Björn Michele, Alexandre Boulch, Gilles Puy, Tuan-Hung Vu, Renaud Marlet, and Nicolas Courty.
\newblock Saluda: Surface-based automotive lidar unsupervised domain adaptation.
\newblock \emph{arXiv preprint arXiv:2304.03251}, 2023.

\bibitem[Milioto et~al.(2019)Milioto, Vizzo, Behley, and Stachniss]{milioto2019rangenet++}
Andres Milioto, Ignacio Vizzo, Jens Behley, and Cyrill Stachniss.
\newblock Rangenet++: Fast and accurate lidar semantic segmentation.
\newblock In \emph{IEEE/RSJ International Conference on Intelligent Robots and Systems}, pages 4213--4220, 2019.

\bibitem[Nekrasov et~al.(2021)Nekrasov, Schult, Litany, Leibe, and Engelmann]{nekrasov2021mix3d}
Alexey Nekrasov, Jonas Schult, Or Litany, Bastian Leibe, and Francis Engelmann.
\newblock Mix3d: Out-of-context data augmentation for 3d scenes.
\newblock In \emph{International Conference on 3D Vision}, pages 116--125, 2021.

\bibitem[Oquab et~al.(2023)Oquab, Darcet, Moutakanni, Vo, Szafraniec, Khalidov, Fernandez, Haziza, Massa, El-Nouby, Assran, Ballas, Galuba, Howes, Huang, Li, Misra, Rabbat, Sharma, Synnaeve, Xu, Jegou, Mairal, Labatut, Joulin, and Bojanowski]{oquab2023dinov2}
Maxime Oquab, Timothée Darcet, Théo Moutakanni, Huy Vo, Marc Szafraniec, Vasil Khalidov, Pierre Fernandez, Daniel Haziza, Francisco Massa, Alaaeldin El-Nouby, Mahmoud Assran, Nicolas Ballas, Wojciech Galuba, Russell Howes, Po-Yao Huang, Shang-Wen Li, Ishan Misra, Michael Rabbat, Vasu Sharma, Gabriel Synnaeve, Hu Xu, Hervé Jegou, Julien Mairal, Patrick Labatut, Armand Joulin, and Piotr Bojanowski.
\newblock Dinov2: Learning robust visual features without supervision.
\newblock \emph{arXiv preprint arXiv:2304.07193}, 2023.

\bibitem[Pan et~al.(2020)Pan, Gao, Mei, Geng, Li, and Zhao]{pan2020semanticPOSS}
Yancheng Pan, Biao Gao, Jilin Mei, Sibo Geng, Chengkun Li, and Huijing Zhao.
\newblock Semanticposs: A point cloud dataset with large quantity of dynamic instances.
\newblock In \emph{IEEE Intelligent Vehicles Symposium}, pages 687--693, 2020.

\bibitem[Peng et~al.(2023{\natexlab{a}})Peng, Genova, Jiang, Tagliasacchi, Pollefeys, and Funkhouser]{peng2023openscene}
Songyou Peng, Kyle Genova, Chiyu Jiang, Andrea Tagliasacchi, Marc Pollefeys, and Thomas Funkhouser.
\newblock Openscene: 3d scene understanding with open vocabularies.
\newblock In \emph{IEEE/CVF Conference on Computer Vision and Pattern Recognition}, pages 815--824, 2023{\natexlab{a}}.

\bibitem[Peng et~al.(2023{\natexlab{b}})Peng, Chen, Qiao, Kong, Liu, Wang, Zhu, and Ma]{peng2023sam}
Xidong Peng, Runnan Chen, Feng Qiao, Lingdong Kong, Youquan Liu, Tai Wang, Xinge Zhu, and Yuexin Ma.
\newblock Sam-guided unsupervised domain adaptation for 3d segmentation.
\newblock \emph{arXiv preprint arXiv:2310.08820}, 2023{\natexlab{b}}.

\bibitem[Puy et~al.(2023)Puy, Boulch, and Marlet]{puy23waffleiron}
Gilles Puy, Alexandre Boulch, and Renaud Marlet.
\newblock Using a waffle iron for automotive point cloud semantic segmentation.
\newblock In \emph{IEEE/CVF International Conference on Computer Vision}, pages 3379--3389, 2023.

\bibitem[Radford et~al.(2021)Radford, Kim, Hallacy, Ramesh, Goh, Agarwal, Sastry, Askell, Mishkin, Clark, Krueger, and Sutskever]{radford2021clip}
Alec Radford, Jong~Wook Kim, Chris Hallacy, Aditya Ramesh, Gabriel Goh, Sandhini Agarwal, Girish Sastry, Amanda Askell, Pamela Mishkin, Jack Clark, Gretchen Krueger, and Ilya Sutskever.
\newblock Learning transferable visual models from natural language supervision.
\newblock In \emph{International Conference on Machine Learning}, pages 8748--8763, 2021.

\bibitem[Ranftl et~al.(2020)Ranftl, Lasinger, Hafner, Schindler, and Koltun]{rene2020towards}
René Ranftl, Katrin Lasinger, David Hafner, Konrad Schindler, and Vladlen Koltun.
\newblock Towards robust monocular depth estimation: Mixing datasets for zero-shot cross-dataset transfer.
\newblock \emph{IEEE Transactions on Pattern Analysis and Machine Intelligence}, 44\penalty0 (3):\penalty0 1623--1637, 2020.

\bibitem[Rizzoli et~al.(2022)Rizzoli, Barbato, and Zanuttigh]{rizzoli2022survey}
Giulia Rizzoli, Francesco Barbato, and Pietro Zanuttigh.
\newblock Multimodal semantic segmentation in autonomous driving: A review of current approaches and future perspectives.
\newblock \emph{Technologies}, 10\penalty0 (4), 2022.

\bibitem[Saltori et~al.(2022)Saltori, Galasso, Fiameni, Sebe, Ricci, and Poiesi]{saltori2022cosmix}
Cristiano Saltori, Fabio Galasso, Giuseppe Fiameni, Nicu Sebe, Elisa Ricci, and Fabio Poiesi.
\newblock Cosmix: Compositional semantic mix for domain adaptation in 3d lidar segmentation.
\newblock In \emph{European Conference on Computer Vision}, pages 586--602, 2022.

\bibitem[Sanchez et~al.(2023)Sanchez, Deschaud, and Goulette]{sanchez2022cola}
Jules Sanchez, Jean-Emmanuel Deschaud, and François Goulette.
\newblock Cola: Coarse-label multi-source lidar semantic segmentation for autonomous driving.
\newblock \emph{arXiv preprint arXiv:2311.03017}, 2023.

\bibitem[Sautier et~al.(2022)Sautier, Puy, Gidaris, Boulch, Bursuc, and Marlet]{sautier2022slidr}
Corentin Sautier, Gilles Puy, Spyros Gidaris, Alexandre Boulch, Andrei Bursuc, and Renaud Marlet.
\newblock Image-to-lidar self-supervised distillation for autonomous driving data.
\newblock In \emph{IEEE/CVF Conference on Computer Vision and Pattern Recognition}, pages 9891--9901, 2022.

\bibitem[Seppänen et~al.(2022)Seppänen, Ojala, and Tammi]{seppanen2022snowyKITTI}
Alvari Seppänen, Risto Ojala, and Kari Tammi.
\newblock 4denoisenet: Adverse weather denoising from adjacent point clouds.
\newblock \emph{IEEE Robotics and Automation Letters}, 8:\penalty0 456--463, 2022.

\bibitem[Sirohi et~al.(2021)Sirohi, Mohan, B{\"u}scher, Burgard, and Valada]{sirohi2021efficientlps}
Kshitij Sirohi, Rohit Mohan, Daniel B{\"u}scher, Wolfram Burgard, and Abhinav Valada.
\newblock Efficientlps: Efficient lidar panoptic segmentation.
\newblock \emph{IEEE Transactions on Robotics}, 2021.

\bibitem[Soum-Fontez et~al.(2023)Soum-Fontez, Deschaud, and Goulette]{fontez2023mdt3d}
Louis Soum-Fontez, Jean-Emmanuel Deschaud, and François Goulette.
\newblock Mdt3d: Multi-dataset training for lidar 3d object detection generalization.
\newblock \emph{arXiv preprint arXiv:2308.01000}, 2023.

\bibitem[Sun et~al.(2020)Sun, Kretzschmar, Dotiwalla, Chouard, Patnaik, Tsui, Guo, Zhou, Chai, Caine, Vasudevan, Han, Ngiam, Zhao, Timofeev, Ettinger, Krivokon, Gao, Joshi, Zhang, Shlens, Chen, and Anguelov]{sun2020waymoOpen}
Pei Sun, Henrik Kretzschmar, Xerxes Dotiwalla, Aurelien Chouard, Vijaysai Patnaik, Paul Tsui, James Guo, Yin Zhou, Yuning Chai, Benjamin Caine, Vijay Vasudevan, Wei Han, Jiquan Ngiam, Hang Zhao, Aleksei Timofeev, Scott Ettinger, Maxim Krivokon, Amy Gao, Aditya Joshi, Yu Zhang, Jonathon Shlens, Zhifeng Chen, and Dragomir Anguelov.
\newblock Scalability in perception for autonomous driving: Waymo open dataset.
\newblock In \emph{IEEE/CVF Conference on Computer Vision and Pattern Recognition}, pages 2446--2454, 2020.

\bibitem[Tang et~al.(2020)Tang, Liu, Zhao, Lin, Lin, Wang, and Han]{tang2020searching}
Haotian Tang, Zhijian Liu, Shengyu Zhao, Yujun Lin, Ji Lin, Hanrui Wang, and Song Han.
\newblock Searching efficient 3d architectures with sparse point-voxel convolution.
\newblock In \emph{European Conference on Computer Vision}, pages 685--702, 2020.

\bibitem[Thomas et~al.(2019)Thomas, Qi, Deschaud, Marcotegui, Goulette, and Guibas]{thomas2019kpconv}
Hugues Thomas, Charles~R Qi, Jean-Emmanuel Deschaud, Beatriz Marcotegui, Fran{\c{c}}ois Goulette, and Leonidas~J Guibas.
\newblock Kpconv: Flexible and deformable convolution for point clouds.
\newblock In \emph{IEEE/CVF International Conference on Computer Vision}, pages 6411--6420, 2019.

\bibitem[Triess et~al.(2020)Triess, Peter, Rist, and Z{\"o}llner]{triess2020scan}
Larissa~T Triess, David Peter, Christoph~B Rist, and J~Marius Z{\"o}llner.
\newblock Scan-based semantic segmentation of lidar point clouds: An experimental study.
\newblock In \emph{IEEE Intelligent Vehicles Symposium}, pages 1116--1121, 2020.

\bibitem[Triess et~al.(2021)Triess, Dreissig, Rist, and Zöllner]{triess2021survey}
Larissa~T. Triess, Mariella Dreissig, Christoph~B. Rist, and J.~Marius Zöllner.
\newblock A survey on deep domain adaptation for lidar perception.
\newblock In \emph{IEEE Intelligent Vehicles Symposium Workshop}, pages 350--357, 2021.

\bibitem[Tsai et~al.(2023{\natexlab{a}})Tsai, Berrio, Shan, Nebot, and Worrall]{tsai2023ms3d}
Darren Tsai, Julie~Stephany Berrio, Mao Shan, Eduardo Nebot, and Stewart Worrall.
\newblock Ms3d: Leveraging multiple detectors for unsupervised domain adaptation in 3d object detection.
\newblock \emph{arXiv preprint arXiv:2304.02431}, 2023{\natexlab{a}}.

\bibitem[Tsai et~al.(2023{\natexlab{b}})Tsai, Berrio, Shan, Nebot, and Worrall]{tsai2023ms3d++}
Darren Tsai, Julie~Stephany Berrio, Mao Shan, Eduardo Nebot, and Stewart Worrall.
\newblock Ms3d++: Ensemble of experts for multi-source unsupervised domain adaption in 3d object detection.
\newblock \emph{arXiv preprint arXiv:2308.05988}, 2023{\natexlab{b}}.

\bibitem[Unal et~al.(2022)Unal, Dai, and Gool]{unal2022scribbleKITTI}
Ozan Unal, Dengxin Dai, and Luc~Van Gool.
\newblock Scribble-supervised lidar semantic segmentation.
\newblock In \emph{IEEE/CVF Conference on Computer Vision and Pattern Recognition}, pages 2697--2707, 2022.

\bibitem[Wang et~al.(2021)Wang, Zhu, Adam, Yuille, and Chen]{wang2021max-deeplab}
Huiyu Wang, Yukun Zhu, Hartwig Adam, Alan Yuille, and Liang-Chieh Chen.
\newblock Max-deeplab: End-to-end panoptic segmentation with mask transformers.
\newblock In \emph{IEEE/CVF Conference on Computer Vision and Pattern Recognition}, pages 5463--5474, 2021.

\bibitem[Wang et~al.(2019)Wang, Cai, Gao, and Vasconcelos]{wang2019towards}
Xudong Wang, Zhaowei Cai, Dashan Gao, and Nuno Vasconcelos.
\newblock Towards universal object detection by domain attention.
\newblock In \emph{IEEE/CVF Conference on Computer Vision and Pattern Recognition}, pages 7289--7298, 2019.

\bibitem[Wang et~al.(2023{\natexlab{a}})Wang, Li, Kallidromitis, Kato, Kozuka, and Darrell]{wang2023hipie}
Xudong Wang, Shufan Li, Konstantinos Kallidromitis, Yusuke Kato, Kazuki Kozuka, and Trevor Darrell.
\newblock Hierarchical open-vocabulary universal image segmentation.
\newblock In \emph{Advances in Neural Information Processing Systems}, 2023{\natexlab{a}}.

\bibitem[Wang et~al.(2023{\natexlab{b}})Wang, Zhang, Cao, Wang, Shen, and Huang]{wang2023segGPT}
Xinlong Wang, Xiaosong Zhang, Yue Cao, Wen Wang, Chunhua Shen, and Tiejun Huang.
\newblock Seggpt: Segmenting everything in context.
\newblock \emph{arXiv preprint arXiv:2304.03284}, 2023{\natexlab{b}}.

\bibitem[Wei et~al.(2022)Wei, Wei, Rao, Li, Zhou, and Lu]{wei2022lidar}
Yi Wei, Zibu Wei, Yongming Rao, Jiaxin Li, Jie Zhou, and Jiwen Lu.
\newblock Lidar distillation: Bridging the beam-induced domain gap for 3d object detection.
\newblock In \emph{European Conference on Computer Vision}, pages 179--195. Springer, 2022.

\bibitem[Wu et~al.(2022)Wu, Lao, Jiang, Liu, and Zhao]{wu2022ptv2}
Xiaoyang Wu, Yixing Lao, Li Jiang, Xihui Liu, and Hengshuang Zhao.
\newblock Point transformer v2: Grouped vector attention and partition-based pooling.
\newblock In \emph{Advances in Neural Information Processing Systems}, 2022.

\bibitem[Wu et~al.(2023)Wu, Tian, Wen, Peng, Liu, Yu, and Zhao]{wu2023ppt}
Xiaoyang Wu, Zhuotao Tian, Xin Wen, Bohao Peng, Xihui Liu, Kaicheng Yu, and Hengshuang Zhao.
\newblock Towards large-scale 3d representation learning with multi-dataset point prompt training.
\newblock \emph{arXiv preprint arXiv:2308.09718}, 2023.

\bibitem[Xiao et~al.(2022{\natexlab{a}})Xiao, Huang, Guan, Cui, Lu, and Shao]{xiao2022polarmix}
Aoran Xiao, Jiaxing Huang, Dayan Guan, Kaiwen Cui, Shijian Lu, and Ling Shao.
\newblock Polarmix: A general data augmentation technique for lidar point clouds.
\newblock In \emph{Advances in Neural Information Processing Systems}, pages 11035--11048, 2022{\natexlab{a}}.

\bibitem[Xiao et~al.(2022{\natexlab{b}})Xiao, Huang, Guan, Zhan, and Lu]{xiao2022synLiDAR}
Aoran Xiao, Jiaxing Huang, Dayan Guan, Fangneng Zhan, and Shijian Lu.
\newblock Transfer learning from synthetic to real lidar point cloud for semantic segmentation.
\newblock In \emph{AAAI Conference on Artificial Intelligence}, pages 2795--2803, 2022{\natexlab{b}}.

\bibitem[Xiao et~al.(2023{\natexlab{a}})Xiao, Huang, Xuan, Ren, Liu, Guan, Saddik, Lu, and Xing]{xiao2023semanticSTF}
Aoran Xiao, Jiaxing Huang, Weihao Xuan, Ruijie Ren, Kangcheng Liu, Dayan Guan, Abdulmotaleb~El Saddik, Shijian Lu, and Eric Xing.
\newblock 3d semantic segmentation in the wild: Learning generalized models for adverse-condition point clouds.
\newblock In \emph{IEEE/CVF Conference on Computer Vision and Pattern Recognition}, pages 9382--9392, 2023{\natexlab{a}}.

\bibitem[Xiao et~al.(2023{\natexlab{b}})Xiao, Zhang, Wang, Loy, Lin, and Pang]{xiao2023p3former}
Zeqi Xiao, Wenwei Zhang, Tai Wang, Chen~Change Loy, Dahua Lin, and Jiangmiao Pang.
\newblock Position-guided point cloud panoptic segmentation transformer.
\newblock \emph{arXiv preprint arXiv:2303.13509}, 2023{\natexlab{b}}.

\bibitem[Xie et~al.(2023{\natexlab{a}})Xie, Kong, Zhang, Ren, Pan, Chen, and Liu]{xie2023robobev}
Shaoyuan Xie, Lingdong Kong, Wenwei Zhang, Jiawei Ren, Liang Pan, Kai Chen, and Ziwei Liu.
\newblock Robobev: Towards robust bird's eye view perception under corruptions.
\newblock \emph{arXiv preprint arXiv:2304.06719}, 2023{\natexlab{a}}.

\bibitem[Xie et~al.(2023{\natexlab{b}})Xie, Kong, Zhang, Ren, Pan, Chen, and Liu]{xie2024robobev}
Shaoyuan Xie, Lingdong Kong, Wenwei Zhang, Jiawei Ren, Liang Pan, Kai Chen, and Ziwei Liu.
\newblock Benchmarking and analyzing bird's eye view perception robustness to corruptions.
\newblock \emph{Preprint}, 2023{\natexlab{b}}.

\bibitem[Xu et~al.(2020)Xu, Wu, Wang, Zhan, Vajda, Keutzer, and Tomizuka]{xu2020squeezesegv3}
Chenfeng Xu, Bichen Wu, Zining Wang, Wei Zhan, Peter Vajda, Kurt Keutzer, and Masayoshi Tomizuka.
\newblock Squeezesegv3: Spatially-adaptive convolution for efficient point-cloud segmentation.
\newblock In \emph{European Conference on Computer Vision}, pages 1--19, 2020.

\bibitem[Xu et~al.(2021)Xu, Zhang, Dou, Zhu, Sun, and Pu]{xu2021rpvnet}
Jianyun Xu, Ruixiang Zhang, Jian Dou, Yushi Zhu, Jie Sun, and Shiliang Pu.
\newblock Rpvnet: A deep and efficient range-point-voxel fusion network for lidar point cloud segmentation.
\newblock In \emph{IEEE/CVF International Conference on Computer Vision}, pages 16024--16033, 2021.

\bibitem[Xu et~al.(2024)Xu, Yang, Kong, Liu, Zhang, Zhou, and Fei]{xu2024visual}
Jingyi Xu, Weidong Yang, Lingdong Kong, Youquan Liu, Rui Zhang, Qingyuan Zhou, and Ben Fei.
\newblock Visual foundation models boost cross-modal unsupervised domain adaptation for 3d semantic segmentation.
\newblock \emph{arXiv preprint arXiv:2403.10001}, 2024.

\bibitem[Xu et~al.(2023)Xu, Kong, Shuai, and Liu]{xu2023frnet}
Xiang Xu, Lingdong Kong, Hui Shuai, and Qingshan Liu.
\newblock Frnet: Frustum-range networks for scalable lidar segmentation.
\newblock \emph{arXiv preprint arXiv:2312.04484}, 2023.

\bibitem[Yan et~al.(2022)Yan, Gao, Zheng, Zheng, Zhang, Cui, and Li]{yan20202dpass}
Xu Yan, Jiantao Gao, Chaoda Zheng, Chao Zheng, Ruimao Zhang, Shuguang Cui, and Zhen Li.
\newblock 2dpass: 2d priors assisted semantic segmentation on lidar point clouds.
\newblock In \emph{European Conference on Computer Vision}, pages 677--695, 2022.

\bibitem[Yang et~al.(2021)Yang, Shi, Wang, Li, and Qi]{yang2021st3d}
Jihan Yang, Shaoshuai Shi, Zhe Wang, Hongsheng Li, and Xiaojuan Qi.
\newblock St3d: Self-training for unsupervised domain adaptation on 3d object detection.
\newblock In \emph{IEEE/CVF Conference on Computer Vision and Pattern Recognition}, pages 10368--10378, 2021.

\bibitem[Ye et~al.(2023)Ye, Zhou, Chen, Xie, Wang, Wang, and Foroosh]{ye2023lidarmultinet}
Dongqiangzi Ye, Zixiang Zhou, Weijia Chen, Yufei Xie, Yu Wang, Panqu Wang, and Hassan Foroosh.
\newblock Lidarmultinet: Towards a unified multi-task network for lidar perception.
\newblock In \emph{AAAI Conference on Artificial Intelligence}, pages 3231--3240, 2023.

\bibitem[Zhang et~al.(2023{\natexlab{a}})Zhang, Yuan, Shi, Chen, Li, and Qiao]{zhang2023uni3d}
Bo Zhang, Jiakang Yuan, Botian Shi, Tao Chen, Yikang Li, and Yu Qiao.
\newblock Uni3d: A unified baseline for multi-dataset 3d object detection.
\newblock In \emph{IEEE/CVF Conference on Computer Vision and Pattern Recognition}, pages 9253--9262, 2023{\natexlab{a}}.

\bibitem[Zhang et~al.(2023{\natexlab{b}})Zhang, Li, Zou, Liu, Li, Gao, Yang, and Zhang]{zhang2023openSeeD}
Hao Zhang, Feng Li, Xueyan Zou, Shilong Liu, Chunyuan Li, Jianfeng Gao, Jianwei Yang, and Lei Zhang.
\newblock A simple framework for open-vocabulary segmentation and detection.
\newblock \emph{arXiv preprint arXiv:2303.08131}, 2023{\natexlab{b}}.

\bibitem[Zhang et~al.(2021)Zhang, Pang, Chen, and Loy]{zhang2021k-net}
Wenwei Zhang, Jiangmiao Pang, Kai Chen, and Chen~Change Loy.
\newblock K-net: Towards unified image segmentation.
\newblock In \emph{Advances in Neural Information Processing Systems}, pages 10326--10338, 2021.

\bibitem[Zhang et~al.(2020)Zhang, Zhou, David, Yue, Xi, Gong, and Foroosh]{zhou2020polarNet}
Yang Zhang, Zixiang Zhou, Philip David, Xiangyu Yue, Zerong Xi, Boqing Gong, and Hassan Foroosh.
\newblock Polarnet: An improved grid representation for online lidar point clouds semantic segmentation.
\newblock In \emph{IEEE/CVF Conference on Computer Vision and Pattern Recognition}, pages 9601--9610, 2020.

\bibitem[Zhang et~al.(2023{\natexlab{c}})Zhang, Yang, Wang, and Li]{zhang2023growsp}
Zihui Zhang, Bo Yang, Bing Wang, and Bo Li.
\newblock Growsp: Unsupervised semantic segmentation of 3d point clouds.
\newblock In \emph{IEEE/CVF Conference on Computer Vision and Pattern Recognition}, pages 17619--17629, 2023{\natexlab{c}}.

\bibitem[Zhao et~al.(2021)Zhao, Bai, and Huang]{zhao2021fidnet}
Yiming Zhao, Lin Bai, and Xinming Huang.
\newblock Fidnet: Lidar point cloud semantic segmentation with fully interpolation decoding.
\newblock In \emph{IEEE/RSJ International Conference on Intelligent Robots and Systems}, pages 4453--4458, 2021.

\bibitem[Zhou et~al.(2022{\natexlab{a}})Zhou, Liu, Yu, Li, Wang, and Wang]{zhou2022lmseg}
Qiang Zhou, Yuang Liu, Chaohui Yu, Jingliang Li, Zhibin Wang, and Fan Wang.
\newblock Lmseg: Language-guided multi-dataset segmentation.
\newblock In \emph{International Conference on Learning Representations}, 2022{\natexlab{a}}.

\bibitem[Zhou et~al.(2022{\natexlab{b}})Zhou, Girdhar, Joulin, Krähenbühl, and Misra]{zhou2022detecting}
Xingyi Zhou, Rohit Girdhar, Armand Joulin, Philipp Krähenbühl, and Ishan Misra.
\newblock Detecting twenty-thousand classes using image-level supervision.
\newblock In \emph{European Conference on Computer Vision}, pages 350--368, 2022{\natexlab{b}}.

\bibitem[Zhou et~al.(2022{\natexlab{c}})Zhou, Koltun, and Krähenbühl]{zhou2022simple}
Xingyi Zhou, Vladlen Koltun, and Philipp Krähenbühl.
\newblock Simple multi-dataset detection.
\newblock In \emph{IEEE/CVF Conference on Computer Vision and Pattern Recognition}, pages 7571--7580, 2022{\natexlab{c}}.

\bibitem[Zhou et~al.(2021)Zhou, Zhang, and Foroosh]{zhou2021panoptic}
Zixiang Zhou, Yang Zhang, and Hassan Foroosh.
\newblock Panoptic-polarnet: Proposal-free lidar point cloud panoptic segmentation.
\newblock In \emph{IEEE/CVF Conference on Computer Vision and Pattern Recognition}, pages 13194--13203, 2021.

\bibitem[Zhu et~al.(2021)Zhu, Zhou, Wang, Hong, Ma, Li, Li, and Lin]{zhu2021cylindrical}
Xinge Zhu, Hui Zhou, Tai Wang, Fangzhou Hong, Yuexin Ma, Wei Li, Hongsheng Li, and Dahua Lin.
\newblock Cylindrical and asymmetrical 3d convolution networks for lidar segmentation.
\newblock In \emph{IEEE/CVF Conference on Computer Vision and Pattern Recognition}, pages 9939--9948, 2021.

\bibitem[Zhuang et~al.(2021)Zhuang, Li, Jia, Wang, Li, and Tan]{zhuang2021pmf}
Zhuangwei Zhuang, Rong Li, Kui Jia, Qicheng Wang, Yuanqing Li, and Mingkui Tan.
\newblock Perception-aware multi-sensor fusion for 3d lidar semantic segmentation.
\newblock In \emph{IEEE/CVF International Conference on Computer Vision}, pages 16280--16290, 2021.

\bibitem[Zou et~al.(2023)Zou, Yang, Zhang, Li, Li, Gao, and Lee]{zou2023seem}
Xueyan Zou, Jianwei Yang, Hao Zhang, Feng Li, Linjie Li, Jianfeng Gao, and Yong~Jae Lee.
\newblock Segment everything everywhere all at once.
\newblock In \emph{Advances in Neural Information Processing Systems}, 2023.

\end{thebibliography}

\end{document}